\documentclass{article} 

\usepackage[margin=3cm]{geometry}
\usepackage{mathtools}
\usepackage[utf8]{inputenc}
\usepackage[T1]{fontenc}
\usepackage{authblk}
\usepackage{relsize}
\usepackage{times}
\usepackage{amsmath}
\usepackage{amsthm}
\usepackage{faktor}
\usepackage{amssymb}
\usepackage{amscd}
\usepackage{graphicx}
\usepackage{caption}
\usepackage{subcaption}
\usepackage{empheq}
\usepackage{cancel}
\usepackage{color}
\usepackage{url}
\usepackage{appendix}
\usepackage{amsfonts}
\usepackage{float}
\usepackage{comment}
\usepackage{hyperref}
\usepackage{xcolor}
\usepackage[ruled,vlined]{algorithm2e}

\usepackage{longtable}
\usepackage{booktabs} 
\usepackage{makecell}
\usepackage{array}
\usepackage{bbm}
\usepackage{siunitx} 
\usepackage{multirow}
\usepackage{ifthen}

\newcommand{\R}{\mathbb{R}}
\newcommand{\Xml}{X^{(m)}_{\ell}}
\newcommand{\tXml}{\tilde{X}^{(m)}_{\ell}}
\newcommand{\Vml}{V^{(m)}_{\ell}}

\newcommand{\Rml}{R^{(m)}_{\ell}}
\newcommand{\Hml}{H^{(m)}_{\ell}}
\newcommand{\Aml}{A^{(m)}_{\ell}}

\newcommand{\Cml}{C^{(m)}_{\ell}}

\newcommand{\Psiml}{\Psi^{(m)}_{\ell}}
\newcommand{\RdN}{\R^{dN}}
\newcommand{\Hf}{\mathcal{H}_{f}}
\newcommand{\Hphi}{\mathcal{H}_{\phi}}
\newcommand{\nf}{n_{f}}
\newcommand{\nphi}{n_{\phi}}

\DeclareMathOperator*{\argmin}{arg\,min}
\DeclarePairedDelimiterX{\inp}[2]{\langle}{\rangle}{#1, #2}
\DeclarePairedDelimiter{\norm}{\lVert}{\rVert}

\newcommand{\etaml}[1][]{%
  \ifthenelse{\equal{#1}{}}%
    {\eta_{\ell,m}}%
    {\eta^{#1}_{\ell,m}}%
}

\begin{document}
\title{Simultaneous inference of environmental and interaction forces in collective dynamics}

\author{Nipuni de Silva$^{1}$\quad Ming Zhong$^{2}$\quad James M. Greene$^{1,*}$}

\date{}
\maketitle

\footnotetext[1]{Department of Mathematics, Clarkson University, Potsdam, NY, United States}
\footnotetext[2]{Department of Mathematics, University of Houston, Houston, TX, United States}

\begingroup
\renewcommand{\thefootnote}{*}
\footnotetext{Corresponding author, \href{mailto:jgreene@clarkson.edu}{jgreene@clarkson.edu}}
\endgroup

\begin{abstract}
Collective dynamics arise in a wide range of physical, biological, and engineering applications. Examples include cell migration, swarm robotics, social dynamics, and animal behavior. A defining characteristic of these systems is the emergence of large-scale coordination from local interactions among agents; a fundamental scientific question is thus to understand the local interactions that give rise to the observed emergent dynamics.  We are interested in methods for learning interactions generally, which can describe a wide class of physical systems exhibiting collective dynamics defined by an interaction kernel, without making any a priori assumptions regarding the analytical form of this kernel (i.e. it is nonparametric). The advantage of this kernel-based approach is that it incorporates the underlying physics of the model (i.e. collective dynamics), which more general equation-learning approaches may ignore, potentially limiting their effectiveness with regards to model accuracy and predictions.  In this work, we extend existing variational learning approaches to collective systems with both interaction kernels and environmental/intra-agent forces. The proposed framework simultaneously infers the interaction kernel non-parametrically while learning the environmental force using either semi-parametric or fully nonparametric representations. The proposed methodology is validated on several benchmark models exhibiting synchronization, alignment, attraction-repulsion, and external environmental forces.  In addition, we introduce a model-selection procedure based on our nonparametric learning framework to identify models that optimally explain a given set of trajectory observations. By exploiting the feature-identification capability of the learned models, the proposed procedure can distinguish among different collective dynamics frameworks and recover mechanistic interaction mechanisms directly from trajectory data.
 
\end{abstract}

\section{Introduction}
\label{sec:introduction}
An important problem in the study of collective dynamics of interacting agents is how one may quantify and predict their behavior via mathematical models.  Such models may be constructed from first principles, utilizing (for example) Newton's equations of motion and other well-known physical theories, or from phenomenological considerations based on experimental observations; we note also that such models may be constructed by combining aspects of both approaches.  Many such models of collective dynamics exist in a wide variety of disciplines, including physics, biology, ecology, neurobiology, social sciences, and economics \cite{carrillo2017review,kolokolnikov2011stability,vicsek2012collective,shoham2008multiagent}.  Furthermore, once a mathematical model is known and validated, the emergence of complex behavior may be analyzed and understood theoretically and precisely; again, such models provide mechanistic insight into the causality of emergence, which may be both scientifically and practically important. Indeed, inter-agent dynamics have had applications across biology, engineering, and the social sciences, providing mathematical frameworks for understanding collective cell migration, tissue morphogenesis and tumor invasion \cite{friedl2009collective, rorth2009collective}, microbial swarming \cite{czirok2000collective, sokolov2007concentration, zhang2010collective}, animal flocking \cite{cucker2007emergent,cucker2010avoiding,cucker2011general,cucker2013conditional,vicsek1995novel, albi2014stability,chuang2016swarming}, pedestrian and traffic flow \cite{moussaid2011simple}, opinion formation \cite{krause2000discrete, couzin2005effective}, swarm robotics \cite{li2025reverse, chen2024synchronous}, and the control of distributed autonomous systems \cite{ren2008distributed, olfati2007consensus}.

Such mechanistic insight will thus be challenging without prior knowledge of the governing equations.  A learning-based approach that first discovers the governing equations from observations of collective dynamics can address these challenges in predictive modeling, since predictions are then generated from the learned mechanistic model rather than from direct extrapolation of dynamical data~\cite{zhong2020data}, the latter of which generally fails in predictive capacity outside of training data \cite{sahoo2018learning,brunton2016discovering}.

Research on the data-driven discovery of governing equations of dynamical systems can be traced back to the early works of Lagrange, Laplace, and Gauss \cite{stigler1990history}.  The work presented here focuses on discovering the governing equations of collective dynamics base on observed temporal agent trajectory data, which is often available with modern experimental techniques~\cite{moussaid2011simple, becco2006experimental, pham2024watching, bialek2012statistical, de2025data, yang2024learning}. We note that all models considered in this work are defined via ordinary differential equations (ODEs), and as such take the general form
\begin{align}
\dot{x} = F(x), \quad x(t_0)=x_0 \in \R^D, \quad t\in [t_0,T],
\label{eq:autonomous_ode}
\end{align}
for some $D \in \mathbb{N}$.  For the general ODE system~\eqref{eq:autonomous_ode}, the discovery of the governing equations is thus represented as follows: infer $F$ from observations of $x(t)$ at various times $t$.  Many system identification methods have been developed to solve this problem, including classical regression-inspired techniques~\cite{ jeliazkov2013nonparametric,devore2006approximation,binev2005universal}, sparse-identification methods \cite{schaeffer2013sparse,brunton2016discovering,tran2017exact,messenger2022learning,messenger2022learning2}, and statistical mechanics formulations~\cite{bialek2012statistical}. Physics-informed neural networks (PINNs)~\cite{raissi2019physics}, neural ODEs~\cite{chen2018neural}, physics-informed neural vector fields~\cite{djeumou2022neural,yu2024learning} have also been utilized. 

Several challenges arise when attempting to learn governing equations for collective systems from trajectory data. If one treats the dynamics as an arbitrary vector field on the full state space, then the resulting regression problem is high-dimensional and computationally demanding, with a dimension that grows with the number of agents. This is particularly problematic in biological collective dynamics, where systems may consist of many organisms or cells, each evolving in $\mathbb{R}^{2}$ or $\mathbb{R}^{3}$. In addition, the relevant functional forms of the governing mechanisms are often not known a priori, making it difficult to construct an appropriate candidate library. Library-based methods may therefore be susceptible to model misspecification when important interaction, environmental, or self-propulsion terms are omitted~\cite{brunton2022data,champion2019data}. For a more detailed discussion of these issues in the context of learning collective dynamics, we refer the interested reader to~\cite{lu2019nonparametric,gelss2019multidimensional}.

The previously discussed methods are applicable for general dynamical systems of the form~\eqref{eq:autonomous_ode} (we note that extensions to partial differential equations and stochastic differential equations also exist~\cite{messenger2021weak,rudy2017data,messenger2022learning2, boninsegna2018sparse, huang2022sparse}).  However, these methods generally do not take into account the specific form of the vector field $F$, or any symmetry that may be a priori known in the application of interest.  For agents interacting collectively, $F$ typically assumes a structured form, and one can use this known structure to improve inference.  Indeed, the authors in~\cite{lu2019nonparametric,bongini2017inferring,lu2021learning,zhong2020data,feng2024learning} have designed variational-based learning methods for collective systems that can be applied to a diverse range of fields; these methods avoid the problems faced by regression-based dynamical methods, and are the basis of the work presented here. More specifically, this general framework assumes that the generally high-dimensional dynamical system~\eqref{eq:autonomous_ode} is defined by a low-dimensional interaction kernel $\phi : \R_{\geq 0} \to \R$, which quantifies the effect of local pairwise interactions between agents.  The simplest example of such a system is that the interactions directly influence dynamics via pairwise averaging:
\begin{align}
\dot{x}_{i} &= \frac{1}{N}\sum_{\substack{j=1 \\ j \neq i}}^{N}\phi(|x_{j}-x_{i}|)(x_{j}-x_{i}).
\label{eq:first_order_kernel_only_framework}
\end{align}
Here, $x_{i}$ denotes the state in $\R^{d}$ of the $i^{\text{th}}$ agent, and $\phi(|x_{j}-x_{i}|)$ denotes the strength of the interaction of agent $j$ on agent $i$:  $\phi > 0$ models attraction, while $\phi < 0$ models repulsion.  The function $\phi = \phi(r)$ is \textit{assumed} to be a function of only the distance between agents, where $|\cdot|$ denotes the standard Euclidean distance in $\R^{d}$.  Many models in science and engineering take the form~\eqref{eq:first_order_kernel_only_framework}, including~\cite{krause2000discrete, cucker2007emergent}.  Note that the problem of inferring $F:\R^{D} \to \R^{D}$ is thus reduced to inferring $\phi: \R_{\geq 0} \to \R$, as we are utilizing physical knowledge of the phenomena to reduce the dimensionality of the inverse problem.  Of course, other collective systems exist beyond those of the form~\eqref{eq:first_order_kernel_only_framework} (e.g. second-order systems), and we emphasize that the learning methods generally apply; indeed, such variational methods will perform effectively for systems when a significant degree of dimensionality reduction in the candidate vector fields exists and is known a priori. 

We note that system~\eqref{eq:first_order_kernel_only_framework} is defined entirely by interaction forces between agents; there are no intra-agent forces assumed.  However, in many applications, agents experience both intra- and inter-agent dynamics, the former of which includes self-propulsion, friction, and other environmental forces. Thus there remains a need for a unified learning framework that simultaneously identifies both the interactions between agents ($\phi$) and the intra-agent environmental forces via trajectory data.  It is the goal of this work to introduce an extended mechanistic learning approach to capture both interactions and the environmental forces simultaneously by generalizing the variational framework~\cite{lu2019nonparametric}.  Mathematically, we are thus generalizing the approach for system~\eqref{eq:first_order_kernel_only_framework} to (for example) first-order equations of the form
\begin{align}
\dot{x}_{i} &= f(x_{i}) + \frac{1}{N}\sum_{\substack{j=1 \ j \neq i}}^{N}\phi(|x_{j}-x_{i}|)(x_{j}-x_{i}),
\label{eq:kernel_and_f_framework}
\end{align}
where the environmental forces are represented by the function  $f: \R^{d} \to \R^{d}$; extensions to more general collective systems will be discussed in the manuscript.  The extension~\eqref{eq:kernel_and_f_framework} thus includes both the interaction-driven dynamics ($\phi$) with agent-level intrinsic effects ($f$) while preserving the pairwise interaction structure.  Note that we are now interested in inferring both $f:\R^{d} \to \R^{d}$ and $\phi:\R_{\geq 0} \to \R$, which still represents a significant reduction in dimension when compared to~\eqref{eq:autonomous_ode}, where $F:\R^{D} \to \R^{D}$ with $D=dN$ in the case of collective dynamics, as generally $ d \ll dN$. Equations such as~\eqref{eq:kernel_and_f_framework} can be utilized to describe a broad class of phenomena, including attraction, repulsion, milling, flocking, swarming, clustering, synchronization, and alignment in agent-based systems, and are observed in a diverse range of processes such as the phototaxis of bacteria~\cite{ha2009particle}, self-propelled particles and the coordinated movement of fish schools ~\cite{d2006self},  swarm robotics~\cite{olfati2006flocking, fax2004information}, chemotaxis with social interactions~\cite{burger2007aggregation}, crowd dynamics with destination guidance~\cite{helbing2000simulating}, and firefly synchronization~\cite{kuramoto1975international}.   

The main contributions of this work are as follows. We introduce two approaches (semi-parametric and fully non-parametric) and demonstrate that they can both be applied within a unified first- and second-order model framework to simultaneously recover the environmental force $f$ and the interaction kernel $\phi$.  We note that the parametric formulation may be utilized in the case when a functional form for the intra-agent dynamics is known, and can be naturally combined to non-parametrically identify the interaction kernel $\phi$. We next quantify the performance of the learning methods by comparing the mechanistic estimates for the forces utilizing a dynamics-driven weighted $L^2$-distance, as well as by predicting trajectories from both trained and untrained initial conditions. We validate the performance of the proposed approaches through quantifying the dependence on training data using using statistics computed over different training sets, evaluating the sample-complexity (the amount of training data required to learn the model), and robustness analyses under observational noise on a number of benchmark models of collective dynamics.  Numerical evidence further suggests that these estimators scale favorably with respect to prediction on a significantly larger number of agents than trained on, which is experimentally relevant. Lastly, we demonstrate how the generalized framework can be used for model selection to identify the ``active" forces in a given dataset so as to determine the mechanistic equations that likely describe the observed dynamics.

The structure of the manuscript is as follows. In Section~\ref{sec:learn_methods}, we describe in detail the unified learning approaches which capture both the environmental intra-agent dynamics as well as the interaction kernel under the semi-parametric and fully non-parametric learning approaches. In Section~\ref{sec:evaluation_methodology}, we define performance measures for the learning approaches via a number of metrics, including feature recovery, governing-equation consistency, dynamical accuracy, statistical reliability, sample complexity, and robustness to observational noise.  Section~\ref{sec:model_selection} introduces a framework which can be applied to sampled trajectory data to determine the most likely mechanistic model of collective dynamics; we note that this identifies the prescence of intra-agent forces $f$.  Section~\ref{sec:results} presents a detailed study of three fundamental dynamical systems that vary across order, features, and agent characteristics for both learning approaches, together with an extended study of model selection using an additional three dynamical systems that vary across feature complexity. Lastly, we conclude the paper with a discussion of limitations and possible future directions.
\subsection{Related Methods}
\label{subsec:related_methods}
We briefly discuss results in the variational approach on systems~\eqref{eq:first_order_kernel_only_framework}.  In~\cite{lu2019nonparametric} and~\cite{bongini2017inferring}, the non-parametric variational approach is proposed to learn the interaction kernel $\phi$ of the framework~\eqref{eq:first_order_kernel_only_framework}.  Specifically,~\cite{bongini2017inferring} considers a first-order model of homogeneous agents and studies the convergence to its mean-field limit, together with the inference of the mean-field interaction kernel from observations of trajectories of systems with a finite, yet increasing, number of agents.  In~\cite{lu2019nonparametric}, the authors extend the approach to the setting where the number of agents is fixed, but the number of observations increases, showing that the non-parametric estimators for the interaction kernel converge at the near-optimal rate for one-dimensional regression, independent of the dimension $d$ of the state space. It further generalizes the estimators to first- and second-order heterogeneous agent systems with one-dimensional interaction kernels based on pairwise distances, providing substantial numerical evidence of the performance of these generalizations. The work of~\cite{lu2021learning} analyzes in detail the estimators for first-order heterogeneous agent-based systems, generalizing the theoretical results of~\cite{lu2019nonparametric} to that setting while sharpening some of the constructions. The work of~\cite{zhong2020data} extends the application of these approaches to rather general classes of agent-based systems driven by first- and second-order dynamics, with interaction kernels depending not only on pairwise distances but also on other pairwise quantities that depend on the states of the agents. The overview in~\cite{zhong2020data} provides a summary of these learning methods under different scenarios, together with possible future extensions and novel approaches related to the framework introduced in~\cite{lu2019nonparametric}.

\section{Methods for learning collective systems}
\label{sec:learn_methods}

Here we discuss the methods utilized to simultaneously infer both the interaction and environmental dynamics in general models of collective dynamics.   For notational simplicity, we restrict our attention to first-order systems; an analogous description of the methods for second-order systems is provided in Appendix~\ref{sec:app:variational_learning_NP_second_order}.  Section~\ref{subsec:variational_learning_NP_first_order} provides a detailed description of the fully non-parametric procedure, while the semi-parametric approach is described in Section~\ref{subsec:variational_learning_P}.  
\subsection{Variational methods for learning collective and environmental forces:  fully non-parametric approach (first-order)}
\label{subsec:variational_learning_NP_first_order}

In this section, we describe our algorithm for inferring both the environmental and interaction forces in first-order models of collective dynamics from observed trajectory data.  Specifically, we assume that the dynamics of a system of $N$ particles are described by the following system of first-order ordinary differential equations (ODEs), for $i=1,2,\ldots , N$:
\begin{align}
    \dot{x}_{i} &= f(x_{i}) + \frac{1}{N}\sum_{\substack{j=1 \\ j \neq i}}^{N}\phi(|x_{j}-x_{i}|)(x_{j}-x_{i}).
    \label{eq:first_order_system_methods}
\end{align}
Here $x_{i}=x_{i}(t) \in \R^{d}$ denotes the state of the $i^{\text{th}}$ agent at time $t$.  The function $f:\R^{d} \to \R^{d}$ models the environmental forces on the agents, which for simplicity we assume to be identical for all agents in the system; typical examples include frictional forces and self-propulsion, which may be due to both the external environment and/or intrinsic agent dynamics of the specific model considered.  We also note that in certain applications, the domain of $f$ may be a proper subset of $\Omega \subseteq \R^{d}$, which will not change the proposed algorithm in any significant manner.  The function $\phi: \R_{\geq 0} \to \R$ encodes the interaction strength between agents, which we assume, again for simplicity of presentation, to be a function only of the distance $|\cdot |$ between agents.  As discussed previously and presented in Section~\ref{sec:introduction}, the general form of system~\eqref{eq:first_order_system_methods} has been utilized in many physical, biological, engineering, and social science contexts.  We note that this is a modeling assumption, and that the proposed algorithm can be readily adapted to more general interaction kernels.   Here $|\cdot |$ denotes the standard Euclidean norm on $\R^{d}$, but other norms or metrics may be utilized as appropriate for the scientific application considered.

Our goal is to obtain estimators for $f$ and $\phi$ in~\eqref{eq:first_order_system_methods} from observed trajectory data.  In the following subsections, we describe both the form of the observed data which we use to obtain the estimators, as well as the estimation algorithm.  We note that the subsequent descriptions are for systems of the form~\eqref{eq:first_order_system_methods}; the analogous procedure for second-order systems can be found in Appendix~\ref{sec:app:variational_learning_NP_second_order}

\subsubsection{Trajectory data}
\label{subsubsec:trajectory_data}

In this subsection, we formalize the form and notation for the trajectory data which will be utilized to construct the estimators for $f$ and $\phi$ in~\eqref{eq:first_order_system_methods}.  We begin by rewriting~\eqref{eq:first_order_system_methods} as an explicit dynamical system in $\RdN$ ($N$ agents each taking values in $\R^{d}$).  Denote the full state of the system at time $t$ via
\begin{align}
    x(t) &:= \begin{pmatrix}
        x_{1}(t) \\
        x_{2}(t) \\
        \vdots \\
        x_{N}(t) 
    \end{pmatrix} \in \RdN.
    \label{eq:x_state_def}
\end{align}
Thus, by defining
\begin{align}
    F_{f}(x) &:= \begin{pmatrix}
        f(x_{1}) \\
        f(x_{2}) \\
        \vdots \\
        f(x_{N})
    \end{pmatrix} \in \RdN \quad \quad
    F_{\phi}(x) := \frac{1}{N} \begin{pmatrix}
        \sum\limits_{\substack{j=1 \\ j \neq 1}}^{N}\phi(|x_{j}-x_{1}|)(x_{j}-x_{1}) \\
        \sum\limits_{\substack{j=1 \\ j \neq 2}}^{N}\phi(|x_{j}-x_{2}|)(x_{j}-x_{2}) \\
        \vdots \\
        \sum\limits_{\substack{j=1 \\ j \neq N}}^{N}\phi(|x_{j}-x_{N}|)(x_{j}-x_{N}),
    \end{pmatrix} \in \RdN
    \label{eq:f_phi_stack}
\end{align}
we see that we can realize~\eqref{eq:first_order_system_methods} in the form
\begin{align}
    \dot{x} &= F_{f}(x) + F_{\phi}(x).
    \label{eq:first_order_stacked}
\end{align}
We denote $x_{i} \in \R^{d}$ as the $i^{\text{th}}$ component of $x \in \R^{d}$, and will utilize this notation to precisely define the form of the observation data and estimation algorithm below.

Assume that $M$ replicates of initial conditions are independently and identically (i.i.d.) sampled from a fixed but generally unknown probability distribution $\mu_{0}$ on $\RdN$; note that each replicate corresponds to a set of initial conditions in $\R^{d}$ for the $N$ agents.  Denote these initial conditions via  
\begin{align}
    X^{(m)}_{0} \in \RdN,
    \label{eq:X_m_0}
\end{align}
for $m=1,2,\ldots , M$.  For each such $m$, denote the solution at time $t$ of the corresponding initial-value problem (IVP)
\begin{align}
    \begin{cases}
    \dot{x} &= F_{f}(x) + F_{\phi}(x) \\
    x(0) &= X^{(m)}_{0}
    \end{cases}
    \label{eq:IVP}
\end{align}
as $x^{(m)}(t)$.  We assume that the experimental data consists of observations at discrete time points $0=t_{1} < t_{2} <  \cdots < t_{L}=T$, which we denote as
\begin{align}
    \Xml &:= x^{(m)}(t_{\ell})
    \label{eq:X_m_l}
\end{align}
As before, $\Xml \in \RdN$, and thus the set $(\Xml)_{m,\ell=1}^{M,L}$ is generally the set of observed trajectory data which will be utilized to obtain the estimators for $f$ and $\phi$.  Note that by construction,
\begin{align}
(\Xml)_{i} &= x^{(m)}_{i}(t_{\ell}) \in \R^{d},
\label{eq:i_component_Xml}
\end{align}
i.e. that the $i^{\text{th}}$ component (with respect to the natural decomposition of $\Xml$ via~\eqref{eq:x_state_def}) of $\Xml$ is the state of agent $i$ at time $t_{\ell}$ of the $m^{\text{th}}$ replicate.

The learning algorithm introduced in Section~\ref{subsubsec:estimation_algorithm} will also require observations of the velocity $v:=\dot{x}$ in~\eqref{eq:first_order_stacked}.  Specifically, we define
\begin{align}
    \Vml &:= \dot{x}^{(m)}(t_{\ell}),
    \label{eq:V_m_l}
\end{align}
which may be approximated from $(\Xml)_{m,\ell=1}^{M,L}$.  In practice, we may obtain this approximation by first applying a low-pass filter $\mathcal{H}$ to the trajectory data, and thus estimate the necessary velocities via forward differences of the filtered data:
\begin{align}
    \Vml &\approx \frac{1}{\Delta t_{\ell}}(\mathcal{H}(X^{(m)}_{\ell+1}) - \mathcal{H}(\Xml)), 
    \label{eq:Vml_approx}
\end{align}
where $\Delta t_{\ell}:=t_{\ell+1}-t_{\ell}$ for $\ell=1,2,\ldots, L$; see also the discussion in Section~\ref{subsec:noise_robustness}.  Taken together, the observed data utilized to estimate $\phi$ and $f$ takes the form $(\Xml, \Vml)_{m,\ell=1}^{M,L}$, where $\Xml,\Vml \in \RdN.$  Since the interaction kernel $\phi$ depends on on pairwise distances $r$ between agents, it is natural to also introduce notation for the observed pairwise distances from the trajectory data.  Thus, we define $\Rml \in \R^{N \times N}$ via
\begin{align}
    (\Rml)_{ij} := | (\Xml)_i -(\Xml)_j |\in \R,
    \label{eq:Rml}
\end{align}
i.e. $(\Rml)_{ij}$ measures the distance between agents $i$ and $j$ at time $t_{\ell}$ of replicate (initial condition) $m$.

Prior to formalizing the estimation procedure, we note that the formulation~\eqref{eq:first_order_stacked} induces a natural inner product (and thus norm) on $\RdN$ via the standard Euclidean norm $|\cdot |$ on $\R^{d}$ and the $N$-agent symmetry of~\eqref{eq:first_order_system_methods}.  Specifically, we define, for $X, Y \in \RdN,$
\begin{align}
\begin{split}
\inp*{X}{Y}_{\RdN} &:= \frac{1}{N}\sum_{i=1}^{N}\inp*{X_{i}}{Y_{i}} =\frac{1}{N}X^{T}Y,
\end{split}
\label{eq:inp_RdN}
\end{align}
where $X^{T}$ denotes the transpose of $X$, $X = (X_{1},X_{2},\ldots , X_{N}), Y= (Y_{1},Y_{2},\ldots , Y_{N})$ and $X_{i},Y_{i} \in \R^{d}$ for $i=1,2,\ldots, N$, and $\inp*{\cdot}{\cdot}$ denotes the standard Euclidean inner product on $\R^{d}$.  We then define the associated norm on $\RdN$ as
\begin{align}
    \norm*{\cdot}_{\RdN} &:= \sqrt{\inp*{\cdot}{\cdot}_{\RdN}}.
    \label{eq:norm_RdN}
\end{align}
This norm and inner product will be utilized in the error functional defined in Section~\ref{subsubsec:estimation_algorithm}.

\subsubsection{Estimation algorithm}
\label{subsubsec:estimation_algorithm}

The central idea in the learning algorithm is that we assume the unknown functions $f$ and $\phi$ can be well-approximated within finite-dimensional hypothesis spaces; our goal then is to calculate these approximations, which will then serve as our desired estimators.  Specifically, we fix two finite-dimensional functional vector spaces $\Hf\subseteq \mathcal{F}(\R^{d},\R^{d})$ and $\Hphi \subseteq \mathcal{F}(\R_{\geq 0},\R)$, where $\mathcal{F}(X,Y)$ denotes the set of all functions between sets $X$ and $Y$; we refer to these spaces as hypothesis spaces.  A discussion of hypothesis space selection is provided in Section~\ref{subsubsec:hypothesis_space_selection}.  We then define the following error functional $\mathcal{E}_{\Hf,\Hphi}$ over the hypothesis spaces $\Hf$ and $\Hphi$ as
\begin{align}
    \mathcal{E}_{\Hf,\Hphi}(\tilde{f},\tilde{\phi}) &:= \frac{1}{ML}\sum_{m,\ell=1}^{M,L}\norm*{\Vml - \Big(F_{\tilde{f}}(\Xml)+F_{\tilde{\phi}}(\Xml)\Big)}_{\RdN}^{2},
    \label{eq:error_functional}
\end{align}
where $(\Xml,\Vml)_{m,\ell=1}^{M,L}$ denotes the (fixed) trajectory data as introduced in Section~\ref{subsubsec:trajectory_data}, and $\tilde{f}\in \Hf, \tilde{\phi}\in \Hphi$, with $F_{\tilde{f}}$ and $F_{\tilde{\phi}}$ defined as in~\eqref{eq:f_phi_stack} and~\eqref{eq:first_order_stacked}. Note that the notational choice of $\tilde{f}$ and $\tilde{\phi}$ reflects our assumption that $f$ and $\phi$ in~\eqref{eq:first_order_system_methods} represent the \textit{true} environmental and inter-agent forces, respectively, and our goal is to construct estimators $\hat{f}$ and $\hat{\phi}$ via the error functional $\mathcal{E}_{\Hf,\Hphi}.$  Intuitively, $\mathcal{E}_{\Hf,\Hphi}(\tilde{f},\tilde{\phi})$ measures the average error (over time and replicates) of the sum-of-squares error of observing the data $(\Xml, \Vml)_{m,\ell=1}^{M,L}$ when predicting with the vector fields $\tilde{f}$ and $\tilde{\phi}$ in~\eqref{eq:first_order_stacked}.  Indeed, by the law of large numbers, $\mathcal{E}_{\Hf,\Hphi}(\tilde{f},\tilde{\phi})$ is an estimator for a time-discretized version of
\begin{align}
    \mathbb{E}_{X(0) \sim \mu_{0}} \left[\frac{1}{T}\int_{0}^{T} \norm{\dot{X}(t) - \left(F_{\tilde{f}}(X(t)) + F_{\tilde{\phi}}(X(t))\right)}_{\RdN}^{2} \, \mathrm{d}t \right].
    \label{eq:continuous_error_funcitonal}
\end{align}
As in standard likelihood approaches, our goal is to minimize~\eqref{eq:error_functional} over $\Hf$ and $\Hphi$.

As $\Hf$ and $\Hphi$ are finite dimensional, let $(h_{k})_{k=1}^{\nf}$ and $(\psi_{k})_{k=1}^{\nphi}$ be respective ordered bases.  Thus, for any $\tilde{f} \in \Hf$ and $\tilde{\phi} \in \Hphi$, there exists $\alpha = (\alpha_{1},\alpha_{2},\ldots , \alpha_{\nf}) \in \R^{\nf}$ and $\beta =(\beta_{1},\beta_{2},\ldots , \beta_{\nphi}) \in \R^{\nphi}$ such that
\begin{align}
    \begin{split}
        \tilde{f} &= \sum_{k=1}^{\nf}\alpha_{k}h_{k} \quad \text{and} \quad \tilde{\phi} = \sum_{k=1}^{\nphi}\beta_{k}\psi_{k}.
    \end{split}
    \label{eq:basis_coefficients}
\end{align}
As in~\eqref{eq:f_phi_stack}, for each $1 \leq k \leq \nf$, writing $x=(x_{1},x_{2},\ldots , x_{N}) \in \R^{dN}$ with $x_{i} \in \R^{d}$, we define
\begin{align}
    h_{k}(x) &:= \begin{pmatrix}
        h_{k}(x_{1}) \\
        h_{k}(x_{2}) \\
        \cdots \\
        h_{k}(x_{N})
    \end{pmatrix}
    \label{eq:hk_stacked}
\end{align}
and the matrix $\Hml \in \R^{dN  \times \nf}$
\begin{align}
    \Hml &:= \Big( \, h_{1}(\Xml) \quad h_{2}(\Xml) \quad \cdots \quad h_{\nf}(\Xml) \, \Big).
    \label{eq:Hml}
\end{align}
We note that for $1 \leq m \leq M$ and $1 \leq \ell \leq L$, each matrix $\Hml$ is constant in the estimation algorithm, as a function only of the trajectory data $(\Xml)_{m,\ell=1}^{M,L}$ and the selected basis $(h_{k})_{k=1}^{\nf}$ for $\Hf$.  Thus, expanding $\tilde{f} \in \Hf$ as in~\eqref{eq:basis_coefficients}, $F_{\tilde{f}}(\Xml)$ in~\eqref{eq:error_functional} takes the form of matrix-vector multiplication:
\begin{align}
    F_{\tilde{f}}(\Xml) &= \Hml \alpha.
    \label{eq:F_f_ml_simple}
\end{align}
Similarly, we define
\begin{align}
    \psi_{k}(x) &:= \begin{pmatrix}
        \psi_{k}(x_{1}) \\
        \psi_{k}(x_{2}) \\
        \cdots \\
        \psi_{k}(x_{N})
    \end{pmatrix}
    \label{eq:psik_stacked}
\end{align}
and
\begin{align}
   F_{\psi_{k}}(x) := \frac{1}{N} \begin{pmatrix}
        \sum\limits_{\substack{j=1 \\ j \neq 1}}^{N}\psi_{k}(|x_{j}-x_{1}|)(x_{j}-x_{1}) \\
        \sum\limits_{\substack{j=1 \\ j \neq 2}}^{N}\psi_{k}(|x_{j}-x_{2}|)(x_{j}-x_{2}) \\
        \vdots \\
        \sum\limits_{\substack{j=1 \\ j \neq N}}^{N}\psi_{k}(|x_{j}-x_{N}|)(x_{j}-x_{N}),
    \end{pmatrix}
    \label{eq:F_psik}
\end{align}
for each $1 \leq k \leq \nphi$ as in~\eqref{eq:f_phi_stack}, from which we further define, for $1 \leq m \leq M$ and $1 \leq \ell \leq L$ the matrix $\Psiml \in \R^{dN \times {\nphi}}$
\begin{align}
    \Psiml &:= \Big( \, F_{\psi_{1}}(\Xml) \quad F_{\psi_{2}}(\Xml) \quad \cdots \quad F_{\psi_{\nphi}}(\Xml) \, \Big).
    \label{eq:Psiml}
\end{align}
Recall again that for $1 \leq m \leq M$ and $1 \leq \ell \leq L$, each matrix $\Psiml$ is constant in the estimation algorithm, as a function only of the trajectory data $(\Xml)_{m,\ell=1}^{M,L}$ and the selected basis $(\psi_{k})_{k=1}^{\nf}$ for $\Hphi$. Expanding $\tilde{\phi} \in \Hphi$ as in~\eqref{eq:basis_coefficients}, the term $F_{\tilde{\phi}}(\Xml)$ in~\eqref{eq:error_functional} thus also takes the form of matrix-vector multiplication:
\begin{align}
    F_{\tilde{\phi}}(\Xml) &= \Psiml \beta.
    \label{eq:F_phi_ml_simple}
\end{align}

Using~\eqref{eq:F_f_ml_simple} and~\eqref{eq:F_phi_ml_simple}, we observe that minimizing~\eqref{eq:error_functional} over $\Hf$ and $\Hphi$ is equivalent to minimizing
\begin{align}
    \mathcal{E}(\alpha,\beta) &:= \frac{1}{ML}\sum_{m,\ell=1}^{M,L}\norm*{\Vml - \Big( \Hml \alpha + \Psiml \beta \Big)}_{\RdN}^{2},
    \label{eq:error_functional_simple}
\end{align}
over $(\alpha,\beta) \in \R^{\nf +\nphi}$.  As~\eqref{eq:error_functional_simple} is quadratic in $(\alpha,\beta)$ (recall that the norm $\norm*{\cdot}_{\RdN}$ is defined in~\eqref{eq:inp_RdN}~-~\eqref{eq:norm_RdN} via the standard Euclidean inner product on $\R^{d}$), a solution $(\hat{\alpha},\hat{\beta})$ always exists:
\begin{align}
    (\hat{\alpha},\hat{\beta}) &:= \argmin_{\alpha \in \R^{\nf}, \, \beta \in \R^{\nphi}}\mathcal{E}(\alpha,\beta).
    \label{eq:alpha_hat_beta_hat_def}
\end{align}

\subsubsection{Normal equations}
\label{subsubsec:normal}

We provide the system of linear equations (i.e. the normal equations) which characterize solutions of~\eqref{eq:alpha_hat_beta_hat_def}.  For simplicity, define $\theta \in \R^{\nf + \nphi}$ as ordered set of all coefficients to be determined,
\begin{align}
    \theta &:= \begin{pmatrix}
        \alpha \\
        \beta
    \end{pmatrix},
    \label{eq:theta}
\end{align}
and the matrices $\Cml \in \R^{dN \times {(\nf + \nphi)}}$
\begin{align}
    \Cml &:= ( \,\Hml \quad \Psiml \,)
    \label{eq:Aml}
\end{align}
for $m=1,2,\ldots , M$ and $\ell=1,2,\ldots , L$.  With respect to this notation, equation~\eqref{eq:error_functional_simple} takes the form
\begin{align}
    \mathcal{E}(\theta) &:= \frac{1}{ML}\sum_{m,\ell=1}^{M,L}\norm*{\Vml - \Cml \theta}_{\RdN}^{2}.
    \label{eq:error_functional_theta}
\end{align}
Thus, we minimize~\eqref{eq:error_functional_theta} with respect to $\theta = (\alpha,\beta) \in  \R^{\nf + \nphi}$, and obtain the estimator coefficients
\begin{align}
     (\hat{\alpha},\hat{\beta}) &= \hat{\theta} := \argmin_{\theta \in \R^{\nf + \nphi}}\mathcal{E}(\theta).
     \label{eq:theta_hat_def}
\end{align}
Since $\norm*{\cdot}_{\RdN}$ is derived from the standard inner product on $\RdN$ via~\eqref{eq:inp_RdN}, solving the minimization problem~\eqref{eq:theta_hat_def} corresponds precisely to a least-squares problem with multiple observations (one observation for each $m=1,2,\ldots, M$ and $\ell=1,2,\ldots, L$).  Thus, as a minimizing $\hat{\theta}$ necessarily satisfies
\begin{align}
    \nabla_{\theta}\mathcal{E}(\hat{\theta}) &= 0,
    \label{eq:nec_condition_theta_hat}
\end{align}
and as
\begin{align}
    \nabla_{\theta}\inp*{\Vml - \Cml \theta}{\Vml - \Cml \theta}_{\RdN} &= \frac{2}{N}\Big((\Cml)^{T}\Cml \theta - (\Cml)^{T}\Vml \Big),
    \label{eq:gradient_inner_product}
\end{align}
a minimizing $\hat{\theta}$ must satisfy the linear system
\begin{align}
    \left(\sum_{m,\ell=1}^{M,L}(\Cml)^{T}\Cml \right) \hat{\theta} &= \sum_{m,\ell=1}^{M,L}(\Cml)^{T}\Vml.
    \label{eq:normal_eq_v1}
\end{align}
Defining
\begin{align}
    \begin{split}
    A &:= \sum_{m,\ell=1}^{M,L}(\Cml)^{T}\Cml \in \R^{(\nf+\nphi)\times(\nf+\nphi)} \quad \text{and} \quad b := \sum_{m,\ell=1}^{M,L}(\Cml)^{T}\Vml \in \R^{\nf + \nphi},
    \end{split}
    \label{eq:normal_eq_mat_vec}
\end{align}
we see that~\eqref{eq:normal_eq_v1} is equivalent to solving
\begin{align}
    A \hat{\theta} &= b.
    \label{eq:normal_eq_v2}
\end{align}
We lastly note that since $\Cml$ takes the form~\eqref{eq:Aml}, the components of $(\Cml)^{T} \Cml$ and $(\Cml)^{T}\Vml$ take the form
\begin{align}
    \begin{split}
       (\Cml)^{T} \Cml = \begin{pmatrix}
           (\Hml)^{T} \Hml & (\Hml)^{T} \Psiml \\
           (\Psiml)^{T} \Hml & (\Psiml)^{T} \Psiml
       \end{pmatrix} \quad \text{and} \quad (\Cml)^{T} \Vml = \begin{pmatrix}
           (\Hml)^{T} \Vml \\
           (\Psiml)^{T} \Vml
       \end{pmatrix}.
    \end{split}
    \label{eq:matrices_explicit}
\end{align}

\subsubsection{Hypothesis space selection}
\label{subsubsec:hypothesis_space_selection}

As discussed in Section~\ref{subsubsec:estimation_algorithm}, estimators are determined within fixed hypothesis spaces $\Hphi$ and $\Hf$ by minimizing the error functional~\eqref{eq:error_functional}.  If the hypothesis spaces are finite-dimensional, minimizing~\eqref{eq:error_functional} is equivalent to minimizing~\eqref{eq:error_functional_simple}, i.e. to solving a least-squares problem, and in this section, we describe typical constructions for the spaces $\Hphi$ and $\Hf$.  Motivated by the Weierstrass approximation theorem, we seek relatively simple basis functions with localized support; intuitively, altering a single coefficient in the estimation will thus only affect the approximation locally, making the estimation procedure robust to both outlier data and noise.

Consider first $\Hphi$, the hypothesis space approximating the (true) interaction kernel $\phi$.   Define
\begin{align*}
    R_{\min}:=\min_{m,\ell, 1\leq i < j \leq N} (\Rml)_{ij} \quad \text{and} \quad R_{\max}:=\max_{m,\ell, 1\leq i < j \leq N} (\Rml)_{ij}
\end{align*} 
to be the minimum and maximum, respectively, of pairwise distances obtained from the observed trajectory data as discussed in Section~\ref{subsubsec:trajectory_data}.  Thus, all observed pairwise distances are within the interval $[R_{\min},R_{\max}]$, which serves as the effective domain on which $\phi$ can be learned from the data.  We then partition $[R_{\min},R_{\max}]$ into sub-intervals based on the observed data; this is analogous to selecting bins in a histogram or choosing
knots for a spline approximation: the partition should be fine enough to resolve variation in $\phi$, but coarse enough to ensure that each interval contains sufficient observations.  In practice, this partition may be chosen using standard data-dependent
histogram/bin-width rules, such as Scott's rule or the Freedman-Diaconis rule, or by placing breakpoints at quantiles of the empirical pairwise distance distribution $\hat{\rho}_R$ discussed below in Section~\ref{subsec:measure_on_data}~\cite{scott1979optimal,freedman1981histogram,wand1997statistical}.  For each element of the partition of $[R_{\min},R_{\max}]$, we then define a set of functions whose support is precisely that element of the partition; standard choices are indicator functions for the partition, or localized low-degree polynomials.  For example, if the partition element is of the form $I_{p}:=[r_{p},r_{p+1}) \subseteq [R_{\min},R_{\max}]$, then we may define the basis of localized polynomials to $I_{p}$ of degree at most two as $\{\mathbf{1}_{I_{p}}(r),r\mathbf{1}_{I_{p}}(r), r^{2}\mathbf{1}_{I_{p}}(r)\}$.  We then extend this basis naturally to all partitions of $[R_{\min},R_{\max}]$ to obtain the basis $(\psi_{k})_{k=1}^{\nphi}$, with $\Hphi$ defined as the span of this set of functions.  Note that if there are $P$ partitions of $[R_{\min},R_{\max}]$ and $n_{P}$ number of localized basis functions, then $\nphi=P\times n_{P}$.  As localized polynomials are in general discontinuous, the resulting approximation is also discontinuous, as a linear combination of discontinuous functions (see~\eqref{eq:basis_coefficients}).  If it known that the interaction kernel is continuous, it may be natural to select a set of continuous basis functions; an example of such a choice is B-splines, which are utilized frequently in this work.  We refer the interested reader to the supplementary material of~\cite{lu2019nonparametric} for a detailed discussion regarding the choice of basis functions.

The same localization principle applies to the construction of the hypothesis space $\Hf$ for variables associated with the environmental forcing term $f$. For the first-order system~\eqref{eq:first_order_system_methods}, $f=f(x)$, where $x \in \R^{d}$; note the same principle applies for second-order systems, but in general in this case $f = f(x,v)$, i.e. $(x,v) \in \R^{d} \times \R^{d}$. Specifically, for first-order systems, the data of observed states $(\Xml)_{ij} \in \R^{d}$ are partitioned into $d$-dimensional hyper-rectangles.  For example, when $d=2$, the state data is partitioned into rectangles of the form $[x_{1,p},x_{1,p+1}) \times [x_{2,p},x_{2,p+1})$.  Localized basis functions are constructed on these hyper-rectangular partition elements, with each function defining a specific $h_{k}$, and the hypothesis space $\Hf$ is thus defined as the span of this collection of localized basis functions.  As $f \in \mathcal{F}(\R^{d},\R^{d})$, each basis function $h_k \in \mathcal{F}(\R^{d},\R^{d})$ of the hypothesis space $\Hf$ is $\R^{d}$ valued, i.e. is a function of the form $h_{k}=(h^{1}_{k},h^{2}_{k},\ldots , h^{d}_{k})$, where $h^{j}_{k}:\R^{d} \to \R$.  Thus, each $h^{b}_{k}$ is a localized scalar-valued function defined on a partition element of $\R^{d}$; again, we may take localized multivariate (with $d$ being the number of variates) polynomials of small degree, or the tensor product of B-splines.  In practice, we thus construct a localized scalar basis, which we assume identical for each partition, and then take the (pairwise) tensor product of this basis with the standard basis of $\R^{d}$ to extend this to a set of functions in $\mathcal{F}(\R^{d},\R^{d})$, which defines $(h_{k})_{k=1}^{\nf}$ and thus $\Hf$ as the span of these functions.   

\subsubsection{Algorithm}
\label{subsubsec:algorithm}

\begin{algorithm}[H]
\DontPrintSemicolon
\SetAlgoLined
\KwIn{$( \Xml, \Vml   )_{m, \ell=1}^{M,L}$}
\KwOut{Estimators: $\hat{\phi}$, $\hat{f}$}

\BlankLine
Construct pairwise quantities and estimate the observed supports\;
\Indp Generate interaction distances $( \Rml )_{m,\ell=1}^{M,L}$
\BlankLine
Find the maximum and minimum interaction radii $R_{\max}, R_{\min} \in \R$\;
Find the component-wrise maximum and minimum values of state variable $X_{\max}, X_{\min} \in \R^d$\;
Observed supports$ [ R_{\min}, R_{\max}], [X_{\min}, X_{\max}]$\;
\Indm
\BlankLine
Construct localized basis functions \;
\Indp Construct the kernel basis $(\psi_{k})_{k=1}^{\nphi}$\;
    Construct $(h_{k})_{k=1}^{\nf}$ \;
\Indm
Create $\Cml$ by column-wise concatenating $( \,\Hml \quad \Psiml \,)$ as in ~\eqref{eq:Aml}\;
Assemble $A\hat{\theta}=b$ (in parallel for $\ell, m$), using~\eqref{eq:normal_eq_v1},~\eqref{eq:normal_eq_mat_vec},~\eqref{eq:normal_eq_v2},~\eqref{eq:theta}\;
Solve for $\hat{\theta}$ \;
\BlankLine
Assemble $\hat{f}$ and $\hat{\phi}$ as in ~\eqref{eq:basis_coefficients}\;

\caption{Algorithm for learning $\phi$ (interaction) and $f$ (environmental) forces from trajectory data of first-order systems of the form~\eqref{eq:first_order_system_methods}. }
\label{alg:1st_order_variational_algorithm_for_kernel_and_fenv_NP}
\end{algorithm}

\subsubsection{Computational complexity}
\label{subsubsec:computational_complexity}

The proposed learning framework, described in Sections~\ref{subsubsec:estimation_algorithm} and~\ref{subsubsec:normal}, is naturally parallelizable across both replicates and observation times.  In general, the data utilized in training is $\mathcal{O}(MLD)$, where $D=dN$ for first-order systems and $D=2dN$ for second-order systems, as there exist $M$ replicates (initial conditions) of $L$ time observations, with each observation an element of $\R^{D}$. However, the assembly of the least-squares system~\eqref{eq:normal_eq_mat_vec} is performed independently across both replicates and times (i.e. each $\Cml$ may be constructed independently across different computational cores).  For each replicate $m$ and observation time $t_{\ell}$, all pairwise interaction distances $\Rml$ must be computed to construct $\Psiml$, which requires $\mathcal{O}(N^2)$ operations, where $N$ is the number of agents, and thus requires (across all $M$ replicates and $L$ observation times) $\mathcal{O}(MLN^{2})$ operations.  Let $n:= n_\phi+n_f$ be the number of basis functions (corresponding to $\Hphi$ and $\Hf$).  Distances $\Rml$ and state $\Xml$ must be evaluated on the $n$ basis functions, which yields an additional total complexity of $\mathcal{O}(MLn)$.  Note that we are assuming that the complexity of evaluation of each basis function is $O(1)$; regardless, this cost is fixed when a set of basis functions for $\Hphi$ and $\Hf$ is selected.    Thus, the construction of $A$ and $b$ in~\eqref{eq:normal_eq_mat_vec} requires  $\mathcal{O}(MLN^2 + MLn)$ operations, with the pairwise-distance computations dominating in practice (i.e. $n \ll N^{2}$), so that we assume $\mathcal{O}(MLN^{2} + MLn) = \mathcal{O}(MLN^{2})$.  Solving the resulting linear system via $QR$ factorization or singular-value decomposition requires $\mathcal{O}(n^3)$ operations. Therefore, the overall computational complexity of the learning algorithm is $\mathcal{O}(MLN^2+n^3)$. Note that since the basis dimension $n$ is typically several orders of magnitude smaller than the number of observations, the dominant computational cost arises from matrix construction rather than solving the resulting linear system $A \hat{\theta}=b$. New trajectories may also be incorporated incrementally by updating the accumulated matrix/vector in the normal equations, thus making the framework naturally amenable to online-learning scenarios.


\subsection{Variational methods for learning collective and environmental forces:  semi-parametric approach (first-order)}
\label{subsec:variational_learning_P}
In contrast to the fully non-parametric learning approach described in Section~\ref{subsec:variational_learning_NP_first_order}, in certain physical scenarios we may have knowledge of the functional form of the environmental forces acting on the collective system.  More precisely,   in this case we assume that there exists a prescribed functional form for the environmental force term \( f \) in~\eqref{eq:first_order_system_methods}, which we generally write
\begin{align}
f= f(x;p),
\end{align}
where $p \in \R^{n_{p}}$ is a vector of parameters which are needed to fully specify $f$.  The task of estimating $f$ thus reduces to estimating the parameter vector $p$ from trajectory data.  We continue to estimate the interaction kernel $\phi$ non-parametrically as described in Section~\ref{subsec:variational_learning_NP_first_order}, and thus we term this approach \textit{semi-parametric.}

We assume the same training data notation as described in Section~\ref{subsubsec:trajectory_data}, and all results presented comparing the fully non-parametric and semi-parametric approaches will utilize identical samples from initial conditions.  Analogously to the algorithm presented in Section~\ref{subsubsec:estimation_algorithm}, we define an error functional of the form
\begin{align}
    \mathcal{E}_{\Hphi}(p,\tilde{\phi}) &:= \frac{1}{ML}\sum_{m,\ell=1}^{M,L}\norm*{\Vml - \Big(F_{f}(\Xml;p)+F_{\tilde{\phi}}(\Xml)\Big)}_{\RdN}^{2},
    \label{eq:error_functional_parametric}
\end{align}
where $F_{f}(x;p):=(f(x_{1};p),f(x_{2};p),\ldots , f(x_{N};p)) \in \RdN$ for $x \in \RdN$ as in~\eqref{eq:f_phi_stack}.  As before, $\Hphi$ is a fixed hypothesis space, and once a basis $(\psi_{k})_{k=1}^{\nphi}$ is chosen, the error functional~\eqref{eq:error_functional_parametric} takes the form
\begin{align}
    \mathcal{E}(p,\beta) &:= \frac{1}{ML}\sum_{m,\ell=1}^{M,L}\norm*{\Vml - \Big( F_{f}(\Xml;p)+ \Psiml \beta \Big)}_{\RdN}^{2},
    \label{eq:error_functional_simple_parametric}
\end{align}
where $\tilde{\phi}=\sum_{k=1}^{\nphi}\beta_{k}\psi_{k}$ and $\Psiml$ is given by~\eqref{eq:Psiml}.  We then minimize $\mathcal{E}(p,\beta)$ with respect to $(p,\beta) \in \R^{n_{p}} \times \R^{\nphi}$ (note that if certain constraints are known for the parameters $p$, then we minimize over a subset of $\R^{n_{p}}$).  In general, this is non-convex optimization problem, as the vector field $f$ may depend nonlinearly on $p$ (e.g $f(x;p) = x/(x+p)$), and existence, uniqueness, and computation of minimizers (global and local) may be non-trivial~\cite{tarantola2005inverse,nocedal2006numerical}.   We note that the structure of~\eqref{eq:error_functional_simple_parametric} can be exploited computationally.  Although the objective is generally
non-quadratic in $p$, it is quadratic in the coefficients $\beta$, and a ``variable projection'' or
separable nonlinear least-squares formulation reduces the dimension of the
nonlinear optimization problem and is often preferable to optimizing over
$(p,\beta)$ simultaneously~\cite{golub1973differentiation,golub2003separable,o2013variable}.  The reduced problem may then be solved using standard methods for nonlinear least squares, such as Gauss--Newton, Levenberg--Marquardt, or trust-region
methods, typically with multiple initial guesses when non-convexity is
expected~\cite{more2006levenberg,nocedal2006numerical,bjorck2024numerical}.  We note that if $f$ is linear in the parameters $p$ (i.e. $f=\sum_{k=1}^{n_{p}}\eta_{k}p_{k}$, where $\eta_{k}:\R^{d} \to \R^{d}$ are known and fixed), then~\eqref{eq:error_functional_simple_parametric} is indeed quadratic in $(p,\beta)$, and we can construct $\Hml$ analogously as in~\eqref{eq:Hml} to obtain a set of normal equations~\eqref{eq:normal_eq_v2} for the estimators $(\hat{p},\hat{\beta})$.  We note that all the systems considered in this work are linear in parameters, so that we obtain that we obtain a linear least-squares problem even in the semi-parametric case, and hence do not discuss issues related to non-convex optimization.  See Sections~\ref{subsec:results_nonparametric_vs_parametric_learning} and~\ref{subsec:results_model_selection}for details on the models considered in this manuscript.

\section{Evaluation of estimators}
\label{sec:evaluation_methodology}

In this section we describe various methods to evaluate the effectiveness of the learning methods presented in Sections~\ref{sec:learn_methods} and Appendix~\ref{sec:app:variational_learning_NP_second_order}.  Our goal is to assess the estimated model with respect to both model features (i.e. interaction and environmental mechanisms) and trajectory predictions. The proposed methodology quantifies the performance of the inference in various capacities, including feature recovery, consistency of governing equations, trajectory accuracy, statistical variability with respect to observation data, scalability with respect to the number of agents, and robustness to observational noise.  Collectively, these metrics provide a comprehensive assessment of the identifiability, predictive capacity, consistency, and practical applicability of the proposed variational learning framework.

Throughout this section, we define metrics for the first-order system~\eqref{eq:first_order_system_methods}, and discuss natural extensions to the second-order system~\eqref{eq:second_order_system_methods}.  We will denote by $f$ and $\phi$ the true (but unknown) environmental and interaction forces, and by $\tilde{f}$ and $\tilde{\phi}$ their respective estimators.  Note that we define following evaluation metrics with respect to \textit{any} pair of functions $\tilde{f}$ and $\tilde{\phi}$, but results in Section~\ref{sec:results} will generally be shown for the estimators defined by the variational approaches, as discussed in Sections~\ref{subsec:variational_learning_NP_first_order},~\ref{subsec:variational_learning_P}, and Appendix~\ref{sec:app:variational_learning_NP_second_order}; e.g. for $\hat{f}$ and $\hat{\phi}$ defined by
\begin{align}
\begin{split}
    \hat{f} &= \sum_{k=1}^{\nf}\hat{\alpha}_{k}h_{k} \quad \text{and} \quad \hat{\phi} = \sum_{k=1}^{\nphi}\hat{\beta}_{k} \psi_{k},
\end{split}
\label{eq:hat_h_hat_phi}
\end{align}
where $(\hat{\alpha},\hat{\beta})$ are given by~\eqref{eq:alpha_hat_beta_hat_def} for the fully non-parametric approach, or by minimization of~\eqref{eq:error_functional_simple_parametric} for the semi-parametric approach. 

\subsection{Induced measures on data}
\label{subsec:measure_on_data}

The randomness $\mu_{0}$ in the initial conditions of~\eqref{eq:IVP} implies that the observed trajectory data (states $x$ and velocities $\dot{x}$) are random variables, and thus that there exists distributions both on the state space $x$ and thus also the pairwise distances $r$ induced by the solutions of the (generally nonlinear) IVP~\eqref{eq:IVP}.  As our goal is to estimate $f = f(x)$ and $\phi = \phi(r)$, it is natural to measure the accuracy of these estimators with respect to these distributions.  Indeed, one can generally only obtain accurate estimates on regions of state and pairwise distance space that are explored by the solutions of~\eqref{eq:IVP}; regions that are highly explored are thus prioritized when determining accuracy as they contain a larger number of observations.  Hence, following the constructions in~\cite{lu2019nonparametric}, we consider the expected empirical measures for continuous-time observations and their counterparts for discrete-time observations, the latter of which are utilized numerically.

Consider the IVP
\begin{align}
\begin{split}
\dot{x} &= F_{f}(x) + F_{\phi}(x), \quad x(0) = x_{0},
\end{split}
\label{eq:ivp_random_ics}
\end{align}
where $x_{0} \sim \mu_{0}$ and $\mu_{0}$ is a fixed distribution on $\RdN$.  Note that this is the same notation as introduced previously in~\eqref{eq:IVP}, but here are emphasizing the randomness in the solution $x=x(t,x_{0}) = x(t,x_{0}(\omega))$ induced by sampling $x_{0}$ from $\mu_{0}$; the notation $x_{0}(\omega)$ denotes the dependence of $x_{0}$ on given sample $\omega$.  Assume that the solution exists on an interval $[0,T]$.  Since the solution $x$ consists of $N$ components in $\R^{d}$ (one for each agent; see~\eqref{eq:x_state_def}), we obtain $\binom{N}{2}$ unique pairwise distance trajectories,
\begin{align}
    r_{ij}(t,\omega) &:= |x_{j}(t,x_{0}(\omega)) - x_{i}(t,x_{0}(\omega))|,
    \label{eq:r_ij_omega}
\end{align}
where $1 \leq i,j \leq N$.  The distribution of $(r_{ij})_{i,j=1}^{N}$ naturally induces a measure on $\R_{\geq 0}$ of average pairwise distances:
\begin{align}
\rho_R(A):=
\frac{1}{\binom{N}{2}\,T}
\int_0^T
\mathbb{E}_{x_0\sim\mu_0}
\left[
\sum_{1\le i<j\le N}
\mathbf{1}_{A}(r_{ij}(t,\omega))
\right] \, \mathrm{d}t
\label{eq:measure_on_pairwise_interactions}
\end{align}
for any Borel set $A \subseteq \R_{\geq 0}$, where $\mathbf{1}_{A}$ denotes the indicator function of $A$.  Intuitively, $\rho_{R}(A)$ measures the time-averaged expected fraction of observed trajectories whose pairwise distance lies in $A$.  

Assume now that we have measurement data as discussed in Section~\ref{subsubsec:trajectory_data}, i.e. $M$ samples of initial conditions $m=1,2,\ldots , M$ evaluated on time interval $[0,T]$ at points $t_{\ell}$ for $\ell=1,2,\ldots , L$.  Simultaneously discretizing both the integration interval $[0,T]$ (via $\Delta t := T/L$) and utilizing the law of large numbers to estimate the expectation over $x_{0} \sim \mu_{0}$ in~\eqref{eq:measure_on_pairwise_interactions}, we obtain the empirical version of~\eqref{eq:measure_on_pairwise_interactions}:
\begin{align}
    \hat{\rho}_R(A)&:=\frac{1}{\binom{N}{2}\,ML}\sum_{m, \ell=1}^{M,L}
                        \left(
                        \sum_{1\le i<j\le N} \mathbf{1}_{A}((\Rml)_{ij})
                        \right).
    \label{eq:empirical_measure_on_pairwise_interactions}
\end{align}
Recall from~\eqref{eq:Rml} that $(\Rml)_{ij}$ denotes the observed distance between agents $i$ and $j$ of sample (replicate) $m$ at time $t_{\ell}$.  As in~\eqref{eq:measure_on_pairwise_interactions}, $A$ denotes a Borel subset of $\R_{\geq 0}$.  Note that in typical examples, the true measure $\rho_{R}$ is unobservable, so that for all computations demonstrated in the manuscript, the approximation $\hat{\rho}_{R}$ will be employed.

We analogously define an induced measure on the state variable for determining the accuracy of the environmental force $f$ in~\eqref{eq:ivp_random_ics}; note that $f=f(x)$.  Recalling the notation $x(t,x_{0}(\omega))=(x_{i}(t,x_{0}(\omega)))_{i=1}^{N}$ for a solution of~\eqref{eq:ivp_random_ics}, we define the induced average state measure
\begin{align}
    \rho_{X}(B)&:=\frac{1}{TN}\int_0^T\mathbb{E}_{x_{0} \sim \mu_0}
    \left[
    \sum_{i=1}^{N}
    \mathbf{1}_{B}(x_{i}(t,x_{0}(\omega)))
    \right] \, \mathrm{d}t,
    \label{eq:measure_on_state}
\end{align}
where $B$ is a Borel set in $\R^{d}$, and its empirical approximation
\begin{align}
    \hat{\rho}_X(B)&:=\frac{1}{MLN}\sum_{m,\ell,i=1}^{M,L,N}
    \mathbf{1}_{B}((\Xml)_{i}).
    \label{eq:empirical_measure_on_state}
\end{align}
As with $\rho_{R}(A)$, $\rho_{X}(B)$ is time-averaged expected fraction of observed trajectories whose state lies in $B$. Here $(\Xml)_{i} \in \R^{d}$ denotes the observation of agent $i$ at time $t_{\ell}$ of sample (replicate) $m$; see equation~\eqref{eq:i_component_Xml}.  For second-order systems (as discussed in Appendix~\ref{sec:app:variational_learning_NP_second_order}), the environmental forcing $f$ may depend on both position and velocity ($f=f(x,v)$; see~\eqref{eq:second_order_system_methods}), in which case we write $z:=(x,v)$ , and we extend the measure $\rho_{Z}$ to Borel sets in $\R^{d} \times \R^{d}$ in the natural way.  As for $\rho_{R}$, in practice computations are performed  via the empirical approximation $\hat{\rho}_{X}$ (or $\hat{\rho}_{Z}$, for second-order systems), but for notational simplicity, we typically define expressions involving induced measures via $\rho_{R}$ and $\rho_{X}$.  Specifically, as there does not exist a universal criterion for how large $M$ needs to be for accurate estimation of $\rho_{R}$ and $\rho_{X}$ (via $\hat{\rho}_{R}$ and $\hat{\rho}_{X}$, respectively), we approximate $\rho_R$ and $\rho_X$ using datasets generated with large values of $M$ (denoted by $M_{\rho}$), and compare them with the empirical estimators $\hat{\rho}_R$ and $\hat{\rho}_X$ obtained from the training data (denoted by $M_{\text{train}}$); see Table~\ref{tab:app:data_generation_parameters} for parameter values utilized for the various example systems studied in this manuscript.  

The approximation errors of the learned estimators are then evaluated in the weighted least-squares spaces $L^2(\rho_R)$ and $L^2(\rho_X)$ via $\norm{\tilde{\phi}-\phi}_{L^{2}(\rho_R)}$, where
\begin{align}
    \begin{split}
    \norm{\psi}_{L^{2}(\rho_R)} &:= \left(\int_{\R \geq 0}|\psi(r)|^{2}r^{2} \, \mathrm{d} \rho_{R}(r)\right)^{1/2} \quad \text{and} \quad \norm{h}_{L^{2}(\rho_X)} := \left(\int_{\R^{d}}|h(x)|^{2} \, \mathrm{d} \rho_{X}(x)\right)^{1/2}
    \end{split}
    \label{eq:L2_weighted_def}
\end{align}
Note the factor of $r^{2}$ in $\norm{\cdot}_{L^{2}(\rho_{R})}$, which arises due to the assumed form of the collective force in~\eqref{eq:first_order_system_methods}.  An extended discussion of methods for quantifying errors is also provided in Section~\ref{subsec:feature_recovery_metrics}.  As previously discussed, the weighted $L^{2}$ norms in~\eqref{eq:L2_weighted_def} are estimated via the empirical measures $\hat{\rho}_{R}$ and $\hat{\rho}_{X}$, and for all calculations, these approximations are utilized.  In Appendix~\ref{app:sec:system_profiles} (see, for example, Figures~\ref{fig:Kuramoto_observation_measures},~\ref{fig:Phototaxis_observation_measures}, and~\ref{fig:SPP_observation_measures}), the observed distributions obtained from the training data closely align with the empirical approximations of $\rho_R$ and $\rho_X$, which suggests that the training data adequately capture the regions of the domain explored by the dynamics, making the use of $L^2(\hat{\rho}_R)$ and $L^2(\hat{\rho}_X)$ reasonable for assessing estimation errors in simulations.

\subsection{Feature recovery metrics}
\label{subsec:feature_recovery_metrics}
Feature recovery metrics quantify the accuracy with which the underlying interaction kernels and environmental forces are recovered by the variational (non-parametric and semi-parametric) estimators as discussed in Section~\ref{sec:learn_methods} and Appendix~\ref{sec:app:variational_learning_NP_second_order}. As these quantities are known only in synthetic experiments, the metrics in this section are primarily used to validate the  ability of the learning algorithms to recover the true interaction and environmental forces, and cannot generally be utilized in experimental settings.  Nevertheless, they provide evidence that the introduced algorithms are able to mechanistically infer dynamics in collective systems. 

As discussed in Section~\ref{subsec:measure_on_data}, we bias the accuracy of estimation to regions highly explored by the dynamics of systems~\eqref{eq:first_order_system_methods} or~\eqref{eq:second_order_system_methods} via the metrics $\rho_{R}$ (see~\eqref{eq:measure_on_pairwise_interactions}) and $\rho_{X}$ (see~\eqref{eq:empirical_measure_on_state}). We thus define the interaction force and environmental kernel errors as
\begin{align}
    E_{\phi}(\tilde{\phi}) := \norm*{\tilde{\phi}-\phi}_{L^{2}(\rho_{R})} \label{eq:kernel_error} \\
    E_{f}(\tilde{f}) := \norm*{\tilde{f}-f}_{L^{2}(\rho_{X})},\label{eq:environmental_forces_error}
\end{align}
respectively.  To facilitate comparisons across systems with different scales, we also report relative versions of the above errors, defined as
\begin{align}
    \begin{split}
    E_{\phi}^{\mathrm{rel}}(\tilde{\phi}) &:=\frac{E_{\phi}(\tilde{\phi})}{\norm*{\phi}_{L^2(\rho_R)}} \quad \text{and} \quad E_f^{\mathrm{rel}}(\tilde{f}):=\frac{E_f(\tilde{f})}{\norm*{f}_{L^2(\rho_X)}}.
    \end{split}
    \label{eq:feature_functions_relative_errors}
\end{align}
Note that this error is theoretically independent of the observed data, but in practice is estimated via $\hat{\rho}_{R}$ and $\hat{\rho}_{X}$.

\subsection{Residual error}
\label{subsec:residual_error}
While feature recovery metrics in Section~\ref{subsec:feature_recovery_metrics} assess recovery of the underlying interaction mechanisms, residual errors evaluate how well the estimator satisfies the observed data, i.e. it measures the discrepancy between the derivatives derived from the observation data (velocity for a first-order system, acceleration for a second-order system) and the predicted derivatives obtained from estimators (e.g. $\tilde{\phi}$ and $\tilde{f}$).  Unlike the previous mechanistic errors, residual errors are a function of the observed trajectory data.

For notational simplicity, consider the first-order system~\eqref{eq:first_order_system_methods}.  As discussed in Section~\ref{subsubsec:trajectory_data}, the observed position and velocity data takes the form $(\Xml,\Vml)_{m,\ell=1}^{M,L}$.  For estimators $\tilde{\phi}$ and $\tilde{f}$, the residual error is defined by
\begin{align}
    E_{\mathrm{res}}(\tilde{f},\tilde{\phi}):=\left(\frac{1}{ML}
                            \sum_{m,\ell=1}^{M,L}\norm*{\Vml - \left(F_{\tilde{f}}(\Xml) + F_{\tilde{\phi}}(\Xml)\right)}_{\RdN}^{2}
                    \right)^{1/2},
    \label{eq:residue_error}
\end{align}
where $\norm*{\cdot}_{\RdN}$ denotes the induced Euclidean norm on $\RdN$ as defined by~\eqref{eq:norm_RdN}. Note the close correspondence between the residual error~\eqref{eq:residue_error} and the error functional $\mathcal{E}_{\Hf, \Hphi}$ defined in~\eqref{eq:error_functional}; indeed, the goal of the variational approach is precisely to minimize the residual error over the hypothesis spaces $\Hf$ and $\Hphi$.  For second-order systems~\eqref{eq:second_order_system_methods}, the observed trajectory data takes the form of position, velocity, and acceleration $(\Xml,\Vml,\Aml)_{m,\ell=1}^{M,L}$ as discussed in Section~\ref{subsec:app:trajectory_data_second_order}, and the goal is to minimize the error functional~\eqref{eq:second_order_error_functional} over hypothesis spaces $\Hf$ and $\Hphi$.  Thus, we define the corresponding residual error (and relative residual error) analogously, where (for example) we replace velocities $\Vml$ with accelerations $\Aml$ and $F_{\tilde{f}}(\Xml)$ with $F_{\tilde{f}}(\Xml,\Vml)$ in~\eqref{eq:residue_error}.  

As with the mechanistic errors of Section~\ref{subsec:feature_recovery_metrics}, we also define a relative residual error measuring how well the observed trajectory data are approximated by the candidate functions $\tilde{\phi}$ and $\tilde{f}$.  Since the magnitude of the residual depends on the natural scale of the observed velocities, we normalize by a characteristic velocity scale. Specifically, we define the residual scale $S_{\mathrm{res}}$ as the root mean square (RMS) of the observed velocity data $(\Vml)_{m,\ell=1}^{M,L}$:
\begin{align}
 S_{\mathrm{res}}
 &=
 \left(
 \frac{1}{ML} \sum_{m=1}^{M}\sum_{\ell=1}^{L}\norm*{\Vml}_{\RdN}^{2}
 \right)^{1/2}.
\label{eq:residual_scale}
\end{align}
This normalization makes the residual error dimensionless and allows errors to be compared across data sets, parameter regimes, and candidate models with different velocity scales.  In particular, the normalized residual measures the
typical discrepancy between the observed and predicted velocities relative to the typical magnitude of the observed velocities.  We therefore define
\begin{align}
    E^{\mathrm{rel}}_{\mathrm{res}}(\tilde{f},\tilde{\phi}) &:= \frac{E_{\mathrm{res}}(\tilde{f},\tilde{\phi})}{S_{\mathrm{res}}}.
    \label{eq:relative_residue_error}
\end{align}
As discussed previously, for second-order systems $S_{\mathrm{res}}$ is defined with respect to the observed acceleration data ($\Aml)_{m,\ell=1}^{M,L}$

In Section~\ref{sec:model_selection}, we utilize a regularized form of the relative residual error in~\eqref{eq:relative_residue_error}, where a small constant $\epsilon=10^{-12}$ is included in the denominator of~\eqref{eq:relative_residue_error} to improve numerical stability when scales of the observed velocities (accelerations for second-order systems) are small:
\begin{align}
E_{\mathrm{res},\epsilon}^{\mathrm{rel}}(\tilde{f},\tilde{\phi})
=
\frac{E_{\mathrm{res}}(\widetilde{f},\widetilde{\phi})}
{S_{\mathrm{res}}+\epsilon}.
\label{eq:ms_relative_residual_error}
\end{align}

\subsection{Trajectory metrics}
\label{subsec:trajectory_Based_metrics}

A small residual error implies that estimators predict the observed velocity (first-order) or acceleration (second-order) well. However, small residual errors do not necessarily guarantee accurate trajectory estimation. In this section we introduce trajectory-based metrics to assess accuracy of trajectory data utilized in both training and predictions.  Specifically, if we have observation data on an interval $[0,T_{f}]$, we obtain estimators as described in Sections~\ref{sec:learn_methods} and Appendix~\ref{sec:app:variational_learning_NP_second_order} utilizing data restricted to $[0,T] \subseteq [0,T_{f}]$ (training data), while the remaining observations in $(T,T_f]$ are reserved for trajectory extrapolation (testing data) via predictions from the obtained estimators.

As in Section~\ref{subsubsec:trajectory_data}, we define the observed trajectory data from replicate $m$ at time $t_{\ell}$ as $\Xml \in \RdN$.  For estimators $\tilde{f}$ and $\tilde{\phi}$, define $\tilde{x}=(\tilde{x}_{1},\tilde{x}_{2},\ldots , \tilde{x}_{N}) \in \RdN$ as the corresponding solution of the IVP
\begin{align}
    \begin{cases}
    \dot{x} &= F_{\tilde{f}}(x) + F_{\tilde{\phi}}(x) \\
    x(0) &= X^{(m)}_{0}.
    \end{cases}
    \label{eq:IVP_estimator}
\end{align}
Define $\tXml \in \RdN$ as the solution $\tilde{x}$ evaluated at time $t_{\ell}$ of each replicate $m$:
\begin{align}
    (\tXml)_{i} &:= \tilde{x}_{i}^{(m)}(t_{\ell}).
    \label{eq:tXml}
\end{align}
We then define the pointwise agent-averaged trajectory error of replicate (initial condition) $m = 1,2,\ldots, M$ at time $t_{\ell}$ as
\begin{align}
    e^{(m)}_{\ell}(\tilde{f},\tilde{\phi}) &:= \norm*{\Xml - \tXml}_{\RdN}.
    \label{eq:point_wise_trajectory_error}
\end{align}
Note that for each sample $m$ we obtain an averaged (over agents) time series of trajectory errors.

To quantify goodness-of-fit for trajectory data over an entire time interval, we define the following trajectory errors:
\begin{align}
    E_{\mathrm{recon}}^{(m)}(\tilde{f},\tilde{\phi}) &:= \max_{t_{\ell} \in [0,T]} e^{(m)}_{\ell}(\tilde{f},\tilde{\phi})\label{eq:m_th_traj_recon_error} \\
    E_{\mathrm{pred}}^{(m)}(\tilde{f},\tilde{\phi}) &:= \max_{t_{\ell} \in (T,T_f]} e^{(m)}_{\ell} (\tilde{f},\tilde{\phi})\label{eq:m_th_traj_pred__error} \\
    E_{\mathrm{traj}}^{(m)}(\tilde{f},\tilde{\phi}) &:= \max_{t_{\ell} \in [0,T_f]} e^{(m)}_{\ell}(\tilde{f},\tilde{\phi})\label{eq:m_th_traj_full_error}
\end{align}
That is, $E_{\mathrm{recon}}^{(m)}, E_{\mathrm{pred}}^{(m)}$, and $E_{\mathrm{traj}}^{(m)}$ measure trajectory accuracy with respect to the training data $[0,T]$, the testing (prediction) data $(T,T_{f}]$, and the entire trajectory $[0,T_{f}]$, respectively.  As $E_{\mathrm{recon}}^{(m)}$ measures the ability of the estimators to reconstruct the training data, we refer to this as the reconstruction error of sample $m$.  The trajectory error $E_{\mathrm{traj}}^{(m)}$ combines reconstruction and prediction performance into a single quantity, and will serve as the primary trajectory-based metric used in the model selection procedure discussed in Section~\ref{sec:model_selection}.

We have defined trajectory-based metrics for each sampled initial condition $m$.  To assess performance over an entire observed trajectory data set, we aggregate these quantities across all $M$ realizations.  Specifically, we define the sample means
\begin{align}
\begin{split}
    \bar{E}_{\mathrm{recon}}(\tilde{f},\tilde{\phi}) &:= \frac{1}{M}\sum_{m=1}^{M} E_{\mathrm{recon}}^{(m)}(\tilde{f},\tilde{\phi}) \\
    \bar{E}_{\mathrm{pred}}(\tilde{f},\tilde{\phi}) &:= \frac{1}{M} \sum_{m=1}^{M} E_{\mathrm{pred}}^{(m)}(\tilde{f},\tilde{\phi}) \\
    \bar{E}_{\mathrm{traj}}(\tilde{f},\tilde{\phi}) &:= \frac{1}{M}\sum_{m=1}^{M} E_{\mathrm{traj}}^{(m)}(\tilde{f},\tilde{\phi}),
\end{split}
    \label{eq:mean_trajectory_errors}
\end{align}
and the corresponding sample standard deviations
\begin{align}
\begin{split}
    \sigma_{E_{\mathrm{recon}}}(\tilde{f},\tilde{\phi}) &:= \left( \frac{1}{M - 1}\sum_{m=1}^{M} \left(E_{\mathrm{recon}}^{(m)}(\tilde{f},\tilde{\phi}) -\bar{E}_{\mathrm{recon}}(\tilde{f},\tilde{\phi}) \right)^2\right)^{1/2} \\
    \sigma_{E_{\mathrm{pred}}}(\tilde{f},\tilde{\phi}) &:=\left(\frac{1}{M-1}\sum_{m=1}^{M} \left(E_{\mathrm{pred}}^{(m)}(\tilde{f},\tilde{\phi}) - \bar{E}_{\mathrm{pred}}(\tilde{f},\tilde{\phi}) \right)^2 \right)^{1/2} \\
    \sigma_{E_{\mathrm{traj}}}(\tilde{f},\tilde{\phi}) &:=\left(\frac{1}{M-1}\sum_{m=1}^{M} \left(E_{\mathrm{traj}}^{(m)} (\tilde{f},\tilde{\phi}) - \bar{E}_{\mathrm{traj}}(\tilde{f},\tilde{\phi}) \right)^2\right)^{1/2}.
\end{split}
    \label{eq:std_trajectory_error}
\end{align}
The averaged errors summarize the sampled-averaged accuracy of the estimated mechanisms, while the corresponding standard deviations quantify variability across trajectory realizations. When evaluated on prediction data (e.g. on observation times in $(T,T_{f}]$), these quantities additionally provide an assessment of predictive performance.

Additionally, as opposed to the maximum pointwise trajectory error, we also compute the RMS error, which characterizes an average measure of fidelity of trajectory data. Compared with the maximum errors~\eqref{eq:mean_trajectory_errors}, this metric is less dominated by outliers, and is utilized primarily in Section~\ref{sec:model_selection}, where the corresponding relative RMS errors (defined in~\eqref{eq:relative_reconstruction_error} below) is utilized to determine model frameworks which are incompatible with the observed trajectory data.  We define trajectory RMS error as
\begin{align}
E_{*}^{\mathrm{RMS}} (\tilde{f},\tilde{\phi}) :=\left(\frac{1}{ML} \sum_{m,\ell=1}^{M,L}e_{\ell}^{(m)} (\tilde{f},\tilde{\phi})\right)^{1/2},
\label{eq:reconstruction_rmse}
\end{align}
where the subscript ``${*}$'' indicates the time interval that the error is evaluated, i.e. ${*} \in \{ \mathrm{recon}, \mathrm{pred}, \mathrm{traj}\}$ as discussed in~\eqref{eq:m_th_traj_recon_error}~-~\eqref{eq:m_th_traj_full_error}.  As in~\eqref{eq:residual_scale}, the trajectory scale $S_{X}$ is defined as the RMS of the trajectory data,
\begin{align}
S_{X}:= \left( \frac{1}{ML} \sum_{m, \ell=1}^{M, L} \norm*{\Xml}_{\RdN}^{2} \right)^{1/2},
\label{eq:observed_trajectories_scale}
\end{align}
and the corresponding relative reconstruction error is defined as
\begin{align}
E_{*}^{\mathrm{rel}} :=\frac{E_{*}^{\mathrm{RMS}}(\tilde{f},\tilde{\phi})}{ S_{X}+\epsilon},
\label{eq:relative_reconstruction_error}
\end{align}
where again ${*} \in \{ \mathrm{recon}, \mathrm{pred}, \mathrm{traj}\}$.  Note that as opposed to the case for residual errors, the definitions for trajectory errors for first- and second-order systems are identical.

\subsection{Metrics for consistency}
\label{subsec:reliability_assessment} 
Beyond accuracy, we quantify the consistency of the proposed variational approach with respect to the trajectory data utilized to estimate environmental and interaction forces in models of collective dynamics. We assess consistency via uncertainty quantification on observation (training) trajectory data, scalability with respect to the number of agents ($N$), accuracy with respect to the number of sample ($M$) and time ($L$) measurements, and robustness to observational noise.

\subsubsection{Uncertainty quantification}
\label{subsubsec:repeated_trial_runs}

To evaluate variation with respect to sample data $(\Xml, \Vml)_{m,\ell=1}^{M,L}$, the estimation algorithm is repeated independently a total of $T_r$ times using $M$ samples generated from the same underlying distribution $\mu_0$.  That is, we consider $T_{r}$ replicates of $M$ samples, and for each trial in $T_r$, all evaluation metrics introduced in Sections~\ref{subsec:feature_recovery_metrics}~-~\ref{subsec:trajectory_Based_metrics} are computed, and trial means and trial standard deviations of the error metrics are subsequently reported.  Small trial-to-trial variability indicates that the learning procedure is relatively insensitive to sampling fluctuations in observed trajectory data, and therefore produces reliable estimates of the underlying environmental and interaction mechanisms.

\subsubsection{Scalability}
\label{subsubsec:scalability}
We evaluate the estimators ability to scale from a small to large number of agents by obtaining estimators for systems containing $N$ agents, and subsequently evaluating the performance of those estimators on the same system with $4N$ agents.  The objective of this experiment is to determine whether the recovered interaction mechanisms generalize across system sizes and continue to reproduce the collective dynamics of larger populations without the need to re-estimate.  Scientifically this is also useful, as it is often more experimentally tractable to measure trajectories on small samples.  Note that in this numerical experiment, we continue to report reconstruction error metrics for trajectories (e.g $\bar{E}_{\mathrm{recon}}$), but emphasize that we do not train on the $4N$-agent system on any time interval; this is a purely predictive experiment, and in this case such errors simply measure the predictive trajectory error on the early time window $[0,T]$.

\subsubsection{Dependence on training data}
\label{subsubsec:sample_complexity_experiments}
We investigate how estimation accuracy improves for the mechanisms $f$ and $\phi$ as a function of the amount of training data.  The following two types of experiments are considered:
\begin{enumerate}
    \item Increase the number of samples $M$.
    \item Increase the temporal resolution $L$ of the observations.
\end{enumerate}
For each experiment, the training data is expanded while all other experimental settings remain fixed, and we compute how the relative feature recovery metrics~\eqref{eq:feature_functions_relative_errors} vary with $M$ and $L$.  This resulting convergence analysis provide empirical evidence regarding the amount of data required to accurately recover the underlying interaction mechanisms.

\subsubsection{Robustness to noise in observation data}
\label{subsec:noise_robustness}
Robustness refers to the ability of the learned model to maintain accuracy in the presence of noisy observations. To assess robustness, controlled perturbations are introduced into the observed trajectories $(\Xml, \Vml)_{m,\ell=1}^{M,L}$ for first-order systems~\eqref{eq:first_order_system_methods}; for second-order systems~\eqref{eq:second_order_system_methods}, perturbations are also added to the acceleration data $(\Aml)_{m,\ell=1}^{M,L}$.
 
In this manuscript, we consider multiplicative noise.  In the simplest case, for data utilized in the estimation algorithm (i.e. for $t_{\ell} \in [0,T]$ as described in Section~\ref{subsec:trajectory_Based_metrics}), the observation data takes the perturbed form
\begin{align}
    \begin{split}
        x_{i}^{(m), \text{noisy}} &:= x_{i}^{(m)}(t_{\ell})\left(1 + \eta_{i,\ell}^{(m)} \right) \quad \text{and} \quad v_{i}^{(m), \text{noisy}} := v_{i}^{(m)}(t_{\ell})\left(1 + \bar{\eta}_{i,\ell}^{(m)} \right) 
    \end{split}
    \label{eq:noisy_data}
\end{align}
where
\begin{align}
    \eta_{i,\ell}^{(m)}, \bar{\eta}_{i,\ell}^{(m)} &\stackrel{\mathrm{i.i.d.}}{\sim} \text{Unif}[-\zeta, \zeta]
    \label{eq:noisy_dist}
\end{align}
and $((\Xml)_{i}, (\Vml)_{i})= (x_{i}^{(m)}(t_{\ell}), v_{i}^{(m)}(t_{\ell}))$.  That is, noise is applied independently from a uniform distribution to all observed state variables; thus we assume in~\eqref{eq:noisy_data} that noise affects position and velocity independently.  Numerical experiments are performed over the range
\begin{align}
    \zeta \in \{0,\;0.01,\;0.03,\;0.05,\;0.10\},
    \label{eq:noise_levels}
\end{align}
and for each noise level, we evaluate  $E_{\phi}$, $E_f$, and also the trajectory-based metrics of performance.

The above assumption of independent noise for position, velocity, and acceleration data is generally unrealistic, as experimentally only position measurements are typically available, from which velocity and acceleration must be constructed. Since the proposed learning framework requires velocity and, for second-order systems, acceleration observations, these quantities must be estimated from measured positions. A straightforward estimation approach is to apply finite-difference schemes; however, numerical differentiation amplifies measurement noise and often produces poor derivative estimates~\cite{van2020numerical}.  Several smoothing and differentiation techniques have been proposed for trajectory pre-processing, including moving-average windows~\cite{smith1997scientist}, Butterworth filters~\cite{butterworth1930theory}, Savitzky--Golay filters~\cite{savitzky1964smoothing}, Whittaker smoothers~\cite{eilers2003perfect}, and Kalman filtering with Rauch-Tung-Striebel (RTS) smoothing~\cite{rauch1965maximum, jiang2024new}. We note that these techniques have been applied to trajectory data from animal movement, unmanned aerial vehicle (UAV) tracking, biomechanics, ecological monitoring, and other systems which require analysis of motion. Each method has its own advantages and limitations depending on the characteristics of the data.  However, since noise filtering is not the primary focus of this work, we do not attempt a comprehensive comparison of these approaches.
Instead, we evaluate two practical approaches for pre-processing noisy trajectory data: a moving-average smoother and a Kalman filter with an RTS smoother. Both methods reduce the effects of measurement noise before applying the learning algorithm, but they accomplish this in different ways. The moving-average filter smooths the position measurements over a local time window, after which velocities and accelerations are obtained through numerical differentiation. In contrast, the Kalman smoother estimates positions, velocities, and, when appropriate, accelerations simultaneously through a state-space model, thereby eliminating the need for numerical differentiation.  Both methods will be applied to estimate velocity data for noisy trajectory position data for a first-order system in Section~\ref{subsec:results_noise_robustness}.

\section{Model selection}
\label{sec:model_selection}

The framework introduced in Section~\ref{subsec:variational_learning_NP_first_order} and Appendix~\ref{sec:app:variational_learning_NP_second_order} assumes that the precise form of the dynamical system is known a priori, e.g. that there exists both an interaction kernel $\phi$ and an environmental force $f$.  In practice however, the exact form of the governing dynamics is often unknown.  For example, a system may involve interaction forces, alignment effects, environmental influences, and/or combinations of these mechanisms. Therefore, it is natural to consider a more general class of models of collective dynamics and determine, directly from the trajectory data, precisely which mechanistic forces are influencing the dynamics.  Ideally, this selection procedure should be data-driven: rather than manually choosing the model structure in advance, the algorithm should identify a parsimonious candidate model whose inferred forces accurately reproduce the observed dynamics.  It is the goal of this section to outline such a selection procedure.

For first-order systems, we consider the generalized framework
\begin{align}
    \dot{x}_i &=  f(x_i) +\frac{1}{N} \sum_{\substack{j=1 \\ j \neq i}}^{N} \phi^{E}(|x_j - x_i|)(x_j - x_i),  \quad i = 1, \dots, N
    \label{eq:model_selection_general_framework_1order}
\end{align}
where the environmental force $f$ and interaction kernel $\phi^E$ may be identically zero.  Here we label the interaction kernel $\phi=\phi^{E}$ to be consistent with~\eqref{eq:model_selection_general_framework-2order}.  Analogously, for second-order systems, we consider the generalized framework (for $i= 1, \dots, N$)
\begin{equation}
\label{eq:model_selection_general_framework-2order}
\begin{cases}
    \dot{x}_i &=  v_i\\
    \dot{v}_i &= f(x_i, v_i) + \frac{1}{N} \sum_{\substack{j=1 \\ j \neq i}}^{N} \phi^E(|x_j - x_i|)(x_j - x_i) + \frac{1}{N} \sum_{\substack{j=1 \\ j \neq i}}^{N} \phi^A(|x_j - x_i|)(v_j - v_i), 
\end{cases}
\end{equation}
where the environmental force $f$, the energy-based interaction kernel $\phi^E$, and the alignment interaction kernel $\phi^A$ may be identically zero.  The previously introduced variational learning procedure naturally extends to these generalized formulations. Under the assumption that the data is generated from dynamics corresponding to either the full generalized model or to one of its sub-models (e.g. if $\phi^{A} \equiv 0$ in~\eqref{eq:model_selection_general_framework-2order}), the learned coefficients of the estimators with respect to fixed hypothesis spaces provide information about the likelihood of specific interaction mechanisms from the observed data.  Relatedly, we are also interested in determining whether the data supports a first-order or second-order model description.  In this section, we discuss a systematic framework for determining which mechanisms are likely present in data generated from an inter-agent particle system.  We further note that although similar methodologies can be applied to extended systems of collective dynamics (or more generally to systems exhibiting a high degree of symmetry), we restrict our attention to the above frameworks throughout this manuscript.

We begin by enumerating a collection of candidate frameworks, referred to as \emph{candidates} and denoted by the label $c$, each corresponding to a specific combination of both order and mechanisms.  As described below, each candidate model is utilized to construct estimators, from which each are evaluated and ranked according to predefined selection criteria. The framework with the strongest support from the data (i.e. highest ranked) is selected as the most likely framework which describes the available data.  We assign to each candidate a complexity rank based on its dynamical order and the number of active features, as summarized in Table~\ref{tab:model_selection_candidates}; generally first-order models are less complex when compared to second-order models, as are models with fewer terms comprising their vector fields.  Thus, complexity is based on the notion of parsimony, i.e. our goal is to select the simplest model that is able to describe the data.  For every case, the corresponding model yields estimators as described in Sections~\ref{subsec:variational_learning_NP_first_order} (first-order) and Appendix~\ref{sec:app:variational_learning_NP_second_order} (second-order), and its performance on the available trajectory data is evaluated utilizing the metrics described in Section~\ref{subsec:trajectory_Based_metrics}.

\begin{table}
\centering
\begin{tabular}{clc}
\toprule
Candidate ($c)$ & Candidate Model & Complexity \\
\midrule
$F_1$ & $\dot{x}=F_{\phi^E}$ & 1 \\

$F_2$ & $\dot{x}=F_{f}+F_{\phi^E}$ & 2 \\

$S_1$ & $\ddot{x}=F_{\phi^E}$ & 3 \\

$S_2$ & $\ddot{x}=F_{\phi^A}$ & 3 \\

$S_3$ & $\ddot{x}=F_{f}+F_{\phi^E}$ & 4 \\

$S_4$ & $\ddot{x}=F_{f}+F_{\phi^A}$ & 4 \\

$S_5$ & $\ddot{x}=F_{\phi^E}+F_{\phi^A}$ & 5 \\

$S_6$ & $\ddot{x}=F_{f}+F_{\phi^E}+F_{\phi^A}$ & 6 \\
\bottomrule
\end{tabular}
\caption{Hierarchy of candidate models used for model selection. In the Candidate column, $F$ denote a first-order framework, while $S$ denote the second-order framework.  The notation utilized for vector fields in $\RdN$ is as described in Sections~\ref{subsubsec:trajectory_data} and Appendix~\ref{subsec:app:trajectory_data_second_order} (e.g. equation~\ref{eq:first_order_stacked}).  We also suppress $\dot{x}_{i}=v_{i}$ for second-order systems for notational simplicity.}
\label{tab:model_selection_candidates}
\end{table}

Before selecting the most plausible framework, each candidate model must first demonstrate that it is capable of explaining the observed data. As is standard in machine-learning contexts, we partition the available observation trajectory data into training, validation, and testing sets.  That is, if $M$ samples are available, we partition them into three subsets, where
\begin{align}
    M &= M_{\mathrm{train}} + M_{\mathrm{val}} + M_{\mathrm{test}},
    \label{eq:M_split}
\end{align}
each of whose respective purposes are summarized in Table~\ref{tab:train_validate_test_split}.  Candidate models are fit to the trajectory data using the training subset and subsequently compared via the validation subset. Once the candidate that best represents the observed dynamics has been selected, the test set is used only for its final evaluation by comparing the predicted trajectories with previously unseen test trajectories; thus the test subset does not participate in either model fitting or model selection and is not utilized in the model selection algorithm.  Testing data is however utilized when we fit a given model to data, as in Section~\ref{subsec:results_nonparametric_vs_parametric_learning} (see for example Figure~\ref{fig:Kuramoto_True_vs_Learned_Trajectories}).  In general, the language for partitioning data is as follows:  training and testing are utilized when discussing a \textit{pre-determined} model framework, while training and validating apply for model selection.

Candidate frameworks that cannot adequately reproduce the observed dynamics are removed prior to final model selection. In particular, poor candidates often exhibit large residual errors, inaccurate reconstruction of the training trajectories, or even fail to generate numerically integrable trajectories. Motivated by these observations, we introduce a framework compatibility gate that filters out inadmissible candidates prior to ranking.  Specifically, on the $M_{\mathrm{train}}$ training replicates, define a trajectory reconstruction tolerance $\tau_{\mathrm{recon}}$ and a residual tolerance $\tau_{\mathrm{res}}$.  A candidate framework $c$ is then considered \textit{admissible} if both of the following conditions are satisfied:
\begin{align}
\begin{split}
E_{\mathrm{recon}}^{\mathrm{rel}}(c) &<\tau_{\mathrm{recon}} \quad \text{and} \quad E_{\mathrm{res},\epsilon}^{\mathrm{rel}}(c) < \tau_{\mathrm{res}}.
\end{split}
\label{eq:compatible_reconstruction}
\end{align}
where $E_{\mathrm{recon}}^{\mathrm{rel}}$ is defined in~\eqref{eq:relative_reconstruction_error} ($* = \mathrm{recon}$), and $E_{\mathrm{res},\epsilon}^{\mathrm{rel}}$ is defined in~\eqref{eq:ms_relative_residual_error}.  Intuitively, this requires that the candidate model $c$ must approximate well both the training trajectory and velocity (first-order) or acceleration (second-order) data.  In numerical simulations, the above thresholds are fixed at $\tau_{\mathrm{recon}}= \tau_{\mathrm{res}}=0.25$, and if at least one condition in~\eqref{eq:compatible_reconstruction} is violated by candidate $c$, it is deemed \textit{inadmissible} and not considered further in the model selection procedure.  In addition, a candidate framework is rejected if its learned dynamics cannot be integrated successfully or if the condition number of the associated regression matrix $A$ in~\eqref{eq:normal_eq_mat_vec} exceeds a prescribed numerical stability threshold ($10^{12}$). The corresponding evaluation metrics are reported as \textit{N/A}. Denote by $\mathcal{C}$ the set of admissible candidate frameworks:
\begin{align}
    \mathcal{C} &:= \{c \, | \, E_{\mathrm{recon}}^{\mathrm{rel}}(c) <\tau_{\mathrm{recon}}, \, E_{\mathrm{res},\epsilon}^{\mathrm{rel}}(c) < \tau_{\mathrm{res}} \}
    \label{eq:admissible_C}
\end{align}
If $\mathcal{C}=\varnothing$, then no first- or second-order models of the form~\eqref{eq:model_selection_general_framework_1order} or~\eqref{eq:model_selection_general_framework-2order} are selected to describe the trajectory data.

\begin{table}
\centering
\begin{tabular}{ll}
\toprule
Data set & Function \\
\midrule
Training ($M_{\mathrm{train}}$ samples) & Obtain estimators for mechanistic forces for each model \\

Validation ($M_{\mathrm{val}}$ samples) & Compare candidate models and tune hyperparameters to select candidate model \\

Testing ($M_{\mathrm{test}}$ samples) & Measure final evaluation of models and report performance\\
\bottomrule
\end{tabular}
\caption{Description of training, validation, and testing data sets for model selection algorithm.  Note that the full set of initial condition replicates $M$ is partitioned between the three classifications.}
\label{tab:train_validate_test_split}
\end{table}

By construction, all candidates in $\mathcal{C}$ possess small trajectory and residual error with respect to the training replicates.  The primary metric utilized to further evaluate admissible candidates is validation on trajectory data.  That is, candidate frameworks $c \in \mathcal{C}$ are further selected if their mean trajectory error on the validation replicates is below a threshold.  More precisely, define
\begin{align}
    \begin{split}
        E_{\min} &:= \min_{c\in\mathcal{C}} \bar{E}_{\mathrm{traj}}(c) \quad \text{and} \quad \tau_{\mathrm{traj}} := \max \left\{ E_{\min}+\varepsilon_{\mathrm{abs}}, E_{\min}(1+\varepsilon_{\mathrm{rel}}) \right\}
    \end{split}
    \label{eq:model_selection_admissible}
\end{align}
Error $E_{\min}$ thus denotes the minimum average trajectory error (defined by~\eqref{eq:mean_trajectory_errors}) over all admissible candidates on the validation replicates (recall that there are $M_{\mathrm{val}}$ of them), and $\tau_{\mathrm{traj}}$ measures both an absolute and a relative tolerance with respect to $E_{\min}$.  In numerical experiments, we fix $\varepsilon_{\mathrm{abs}}=10^{-3}$ and $\varepsilon_{\mathrm{rel}} =0.05$.  Denote that $\mathcal{C}_{\mathrm{val}}$ the subset of admissible candidates that posses a mean trajectory error (on validation replicates) below $\tau_{\mathrm{traj}}$:
\begin{align}
     \mathcal{C}_{\mathrm{val}} &:= \left\{ c\in\mathcal{C} \, | \, \bar{E}_{\mathrm{traj}}(c) \leq \tau_{\mathrm{traj}}\right\}
    \label{eq:tie_candidate_treshhold}
\end{align}

A model framework is selected from $\mathcal{C}_{\mathrm{val}}$ as follows.  If $|\mathcal{C}_{\mathrm{val}}|= 1$, then the unique $c \in \mathcal{C}_{\mathrm{val}}$ is selected as the most likely model framework to describe the trajectory data.  If $|\mathcal{C}_{\mathrm{val}}|> 1$, i.e. multiple models exhibit statistically insignificant differences in mean validation trajectory error, then preference is given to the framework with lower complexity as defined in Table~\ref{tab:model_selection_candidates}, following the principle of parsimony~\cite{gori2023machine}.  If multiple models in $\mathcal{C}_{\mathrm{val}}$ exhibit the same complexity, the framework with the smallest residual error $E_{\text{res}}$~\eqref{eq:residue_error} on the $M_{\mathrm{train}}$ replicates is selected.

For reference, Table~\ref{tab:evaluation_metrics_summary} summarizes the metrics and partitions of the trajectory data utilized throughout the selection procedure. The complete algorithm is provided in Algorithm~\ref{alg:model_selection}.

\begin{table}[H]
\centering
\begin{tabular}{llll}
\toprule
Metric & Data & Time horizon & Function \\
\midrule

$E_{\mathrm{recon}}^{\mathrm{rel}}$
& $M_{\mathrm{train}}$
& $[0,T]$
& Obtain admissible frameworks (trajectory data) \\

$E_{\mathrm{res},\epsilon}^{\mathrm{rel}}$
& $M_{\mathrm{train}}$
& $[0,T]$
& Obtain admissible frameworks (residual data) \\

$\bar E_{\mathrm{traj}}$
& $M_{\mathrm{val}}$
& $[0,T_f]$
& Primary validation metric (trajectories) \\

$E_{\mathrm{res}}$
& $M_{\mathrm{train}}$
& $[0,T]$
& Residual consistency \\
\bottomrule
\end{tabular}
\caption{Summary of the evaluation metrics, the datasets on which they are computed, their associated time horizons, and their intended purposes.}
\label{tab:evaluation_metrics_summary}
\end{table}

\begin{algorithm}[H]
\DontPrintSemicolon
\SetAlgoLined

\KwIn{
Training trajectories $M_{\mathrm{train}}$ on $[0,T]$,
validation trajectories $M_{\mathrm{val}}$ on $(T,T_f]$,
test trajectories $M_{\mathrm{test}}$ on $[0,T_f]$
}

\KwOut{
Selected candidate model $c^*$ and its learned interaction laws,
or a declaration that the dataset is incompatible with the proposed framework
}

Using $M_{\mathrm{train}}$ on $[t_0,T]$:\;

\Indp
Construct pairwise quantities and estimate the observed supports\;

\Indp
Interaction distances $(\Rml)_{m,\ell=1}^{M,L}$

Observed supports
$[R_{\min},R_{\max}]$
and
$[Z_{\min},Z_{\max}]$,
where
$Z=X$ for first-order models and
$Z= V $ or $ X,V$ for second-order models\;
\Indm

Construct localized basis functions
$(\psi_k^{E})_{k=1}^{n_{\phi}}$,
$(\psi_k^{A})_{k=1}^{n_{\phi}}$,
and
$(h_k)_{k=1}^{n_f}$\;
\Indm

\ForEach{candidate $c\in\mathcal{C}$}{

    Assemble the least-squares system $ A_c\hat{\theta}_c=\vec{b}_c.$ Solve for
    $\hat{\theta}_c$\;

    Recover the learned interaction laws: $\hat\phi^E, \hat\phi^A, \hat{f}$
    associated with $c$\;

    Compute
    $E_{\mathrm{res}}$ ~\eqref{eq:residue_error}
    and
    $\bar{E}_{\mathrm{recon}}$ 
    using
    $M_{\mathrm{train}}$~\eqref{eq:mean_trajectory_errors}\;

    Compute
    $\bar{E}_{\mathrm{traj}}$
    using
    $M_{\mathrm{val}}$~\eqref{eq:mean_trajectory_errors}\;

    Assign the candidate status according to
    compatibility and numerical checks\;
}
Reject inadmissible candidates using
$E_{\mathrm{res},\epsilon}^{\mathrm{rel}}$ ~\eqref{eq:ms_relative_residual_error}
and
$E_{\mathrm{recon}}^{\mathrm{rel}}$ ~\eqref{eq:relative_reconstruction_error} and inability to integrate numerically\;

\eIf{at least one admissible candidate remains}{

    Identify $E_{\min}$ via ~\eqref{eq:model_selection_admissible}

    Form the set of tied candidates $\mathcal{C}_{\mathrm{val}}$ as in ~\eqref{eq:tie_candidate_treshhold}

    Rank candidates in
    $\mathcal{C}_{\mathrm{val}}$
    according to
    \begin{enumerate}
        \item lower model complexity
        (Table~\ref{tab:model_selection_candidates}),
        \item lower residual error
        $E_{\mathrm{res}}$.
    \end{enumerate}

    Select the highest-ranked candidate
    $c^*$\;

}{
    Declare that no candidate within the proposed framework adequately explains the observed dataset\;
}

\caption{Algorithm for identifying model framework from trajectory data}
\label{alg:model_selection}
\end{algorithm}


\section{Results}
\label{sec:results}

\subsection{Implementation details}
\label{subsec:numerical_setup}
In this section, we describe the implementation details common to the numerical experiments presented throughout the manuscript. The models considered here span a range of collective behaviors, including clustering, flocking, milling, and synchronization. We study both first- and second-order systems, as well as models with and without environmental forcing. The examples are organized by the types of mechanisms present in the dynamics, with the presentation progressing generally from simpler to more complex model systems.

For each system, we generate replicates via the distribution $\mu_{0}$ of initial conditions for training $(M_{\mathrm{train}})$, testing $(M_{\mathrm{test}})$, and validation $(M_{\mathrm{val}})$; recall that validation is only utilized for results relating to model selection as discussed in Section~\ref{sec:model_selection}. Initial conditions are sampled independently from the prescribed distribution $\mu_0$, and the governing equations are numerically solved from $0$ to $T_f$.\footnote{Simulation data are generated using the Python \texttt{solve\_ivp} function from the \texttt{scipy.integrate} package with the BDF and Radau methods. Default relative and absolute tolerances are used throughout.}  To obtain approximations for the induced empirical measures on interaction distances and state ($\hat{\rho}_{R}$ and $\hat{\rho}_{X}$, respectively), we also perform $M_{\rho}$ independent samples of $\mu_{0}$, as discussed in Section~\ref{subsec:measure_on_data}.  Estimators are obtained on the $M_{\mathrm{train}}$ samples restricted to the observation interval ($[0,T]$). The same observation data is also used to construct the induced probability measures $\hat{\rho}_{R}$ and $\hat{\rho}_{X}$, which approximate the underlying empirical probability measures $\rho_{R}$ and $\rho_{X}$; again recall that the latter are constructed utilizing the $M_{\rho}$ samples.

To quantify the variation  of the estimators that arises via the distribution $\mu_{0}$, we generate $T_r=10$ independent training replicates, with each training replicate containing $M_{\mathrm{train}}$ samples (so a total of $M_{\mathrm{train}} \times T_{r}$ samples), together with a test sample (of size $M_{\mathrm{test}}$), a validation sample (of size $M_{\mathrm{val}}$), and an induced measure-estimating sample (of size $M_{\rho}$).  Means and standard deviations will be constructed from the $T_{r}$ training replicates. The general trajectory data takes the form $(\Xml, \Vml, \Aml)_{m,\ell=1}^{M,L}$ as described in Section ~\ref{subsubsec:trajectory_data} and Appendix ~\ref{subsec:app:trajectory_data_second_order}. From this data, we construct the interaction-distance data $(\Rml)_{m,\ell=1}^{M,L}$, which contains the relative pairwise information required for estimating interaction kernels.  Initial condition distributions utilized are provided in Table~\ref{tab:app:system-initial-conditions} in Appendix~\ref{app:subsec:system_initial_conditions}.

The hypothesis spaces are constructed directly from the observed trajectory data as discussed in Section~\ref{subsubsec:hypothesis_space_selection}. The interaction kernel $\phi$ is estimated on the interval $[R_{\min},R_{\max}]$, while environmental forces are learned on hyper-rectangles derived from component-wise ranges $[X_{\min},X_{\max}]$ and $[V_{\min},V_{\max}]$ obtained from $(\Xml, \Vml)_{m,\ell=1}^{M,L}$.  Throughout the manuscript, we employ piecewise-linear B-spline basis functions; the number of basis functions and other hyper-parameters are reported in Appendix ~\ref{app:subsec:learning_algorithm_parameters}. For all model systems, we report the evaluation metrics introduced in Section~\ref{sec:evaluation_methodology}, including feature recovery errors (Section~\ref{subsec:feature_recovery_metrics}), residual errors (Section~\ref{subsec:residual_error}), trajectory errors (Section~\ref{subsec:trajectory_Based_metrics}), and consistency measures (Section~\ref{subsec:reliability_assessment}).  The presentation of results is organized as follows.  We begin by introducing each mathematical model discuss its relevance to the proposed learning framework presented in Sections~\ref{subsec:variational_learning_NP_first_order} and Appendix~\ref{sec:app:variational_learning_NP_second_order}. We then present the estimated interaction kernels and environmental forces, together with trajectory comparisons and evaluation metrics. For the model selection experiments, we present the a summary of candidate frameworks, tables demonstrating model selection, and supporting visualizations.  Further details relating to data generation may be found in Table~\ref{tab:app:data_generation_parameters}, algorithm parameters in Table~\ref{tab:app:learning_parameters}, and estimates from the semi-parametric approach in Table~\ref{tab:app:param-recovery}.

\subsection{Semi-parametric versus fully nonparametric variational learning}
\label{subsec:results_nonparametric_vs_parametric_learning}
In this section, we compare estimated interaction kernels $\hat{\phi}$ and environmental forces $\hat{f}$ together with generated trajectory dynamics from both the semi-parametric ($SP$) and fully nonparametric ($NP$) variants of the proposed learning framework. Comparisons are conducted through three representative systems: the Kuramoto model of synchronization (Section~\ref{subsubsec:results_kuramoto}), a model self-propelled particles (SPP) (Section~\ref{subsubsec:results_SPP}), and a model of phototaxis (Section~\ref{subsubsec:results_phototaxis}).  For each system, we compare the ability of the two methods to recover the mechanistic forces, their consistency with respect to the error functional utilized in estimation, and the predicted trajectory accuracy obtained by the two learning approaches. The primary objective is to assess the advantages and limitations of incorporating prior parametric knowledge into the environmental force model relative to the fully nonparametric formulation.

\subsubsection{Kuramoto model}
\label{subsubsec:results_kuramoto}
The Kuramoto model, initially introduced in~\cite{kuramoto1975international}, describes synchronization phenomena in large populations of coupled oscillators. The model was initially motivated by collective behavior observed in chemical and
biological oscillators~\cite{acebron2005kuramoto} and has since found applications in neuroscience~\cite{breakspear2010generative}, electrical power-grid dynamics~\cite{filatrella2008analysis,dorfler2013synchronization}, and oscillatory combustion and flame systems~\cite{susanto2011variational,kiss2002emerging}.  For a review of applications of the Kuramoto model to computation in swarming and synchronization, we refer the interested reader to~\cite{o2019review}, while a historical review of the development of synchronization models can be found in~\cite{strogatz2000kuramoto}. 

The Kuramoto model is a first-order interacting particle system whose state variable is phase $\theta_i \in \R$.  The governing equations are given by
\begin{align}
    \dot{\theta}_i = \omega_i + \frac{1}{N} \sum_{j=1}^{N} K_{ij} \sin (\theta_j - \theta_i),
    \label{eq:Kuramoto-ODE}
    \end{align}
where $N$ denotes the number of oscillators, $\theta_i$ represents the phase of the $i^{\mathrm{th}}$ oscillator,  $\omega_i = \omega_{i}(\theta_{i})$ denotes its intrinsic (generally state dependent) frequency, and $K_{ij}$ is the coupling strength between oscillators $i$ and $j$. In numerical simulations presented below, the coupling strength is assumed to be homogeneous between agents, so that $K_{ij}=K$, and all oscillators share the same functional form of intrinsic frequency $\omega_{i}=\omega(\theta_{i})$. Under these assumptions, the model can be written in the form of a first-order system~\eqref{eq:first_order_system_methods}, where $x_{i}=\theta_{i}, f(x)= \omega(x)$, and  $\phi(r) = K\sin(r)/r$.  For all experiments reported in this section, $K_{ij} =2$ and $\omega(x) = x/100$.   For the semi-parametric approach, we note that the environmental force term possesses a single parameter $p = 1/100 = 0.01$.

\begin{figure}
    \centering
    \begin{subfigure}{0.48\textwidth}
      \centering
      \includegraphics[width=\textwidth]{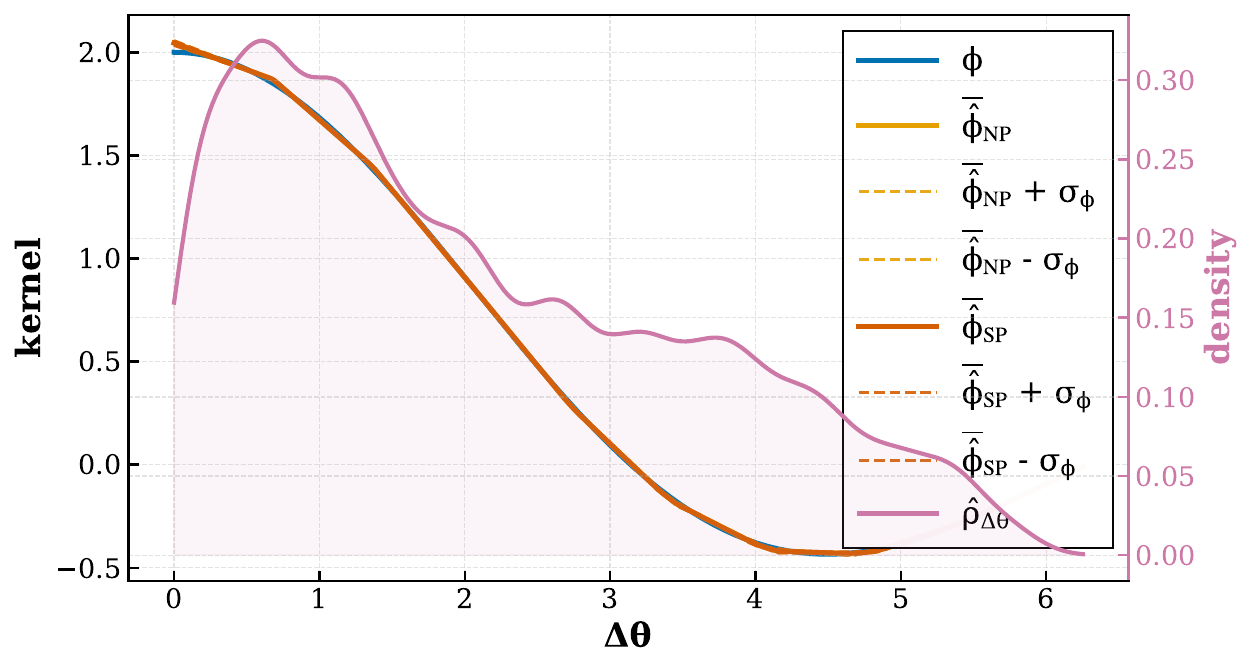}
      \caption{Estimation of interaction kernel $\phi$}
    \label{subfig:Kuramoto-interaction-kernel-true-vs-learned-with-KDE_10_trials}
    \end{subfigure}
    \begin{subfigure}{0.48\textwidth}
      \centering
      \includegraphics[width=\textwidth]{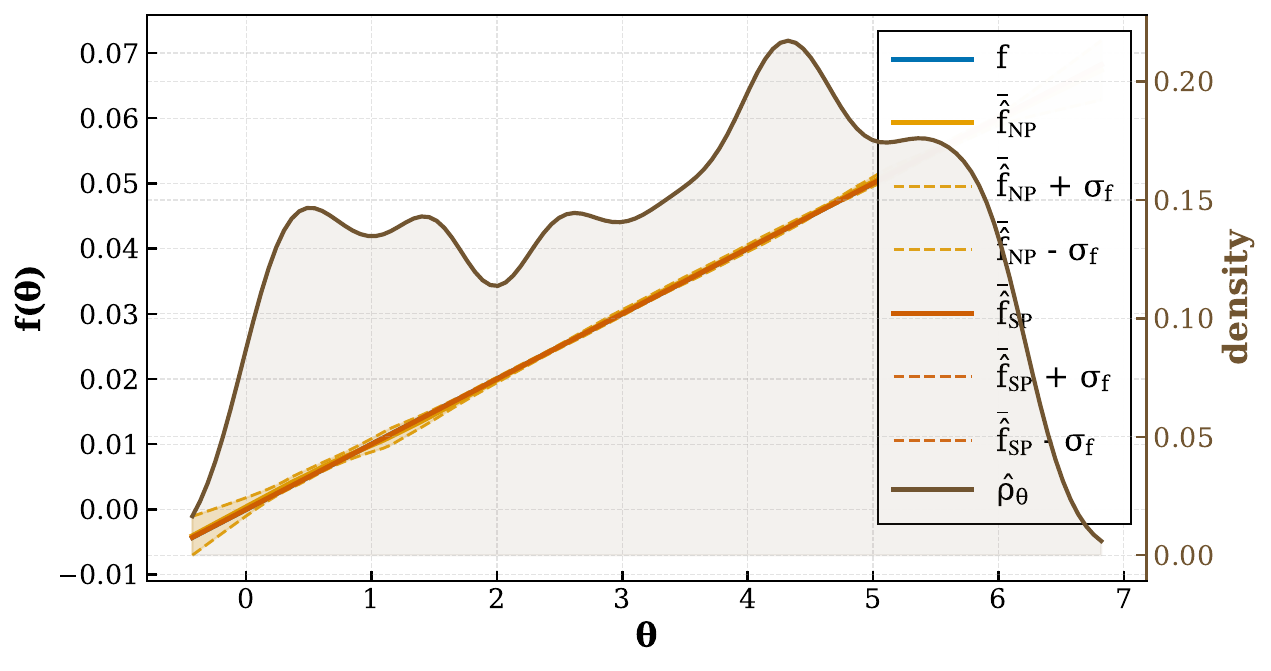}
      \caption{Estimation of environmental force $f$}
    \label{subfig:Kuramoto-environmental-force-true-vs-learned-with-KDE_10_trials}
    \end{subfigure}
    
    \caption{(Kuramoto model) Comparison of the mechanistic feature recovery ability of the fully non-parametric (NP) and semi-parametric (SP) estimation algorithms for the Kuramoto model~\eqref{eq:Kuramoto-ODE}.  In Subfigure~\ref{subfig:Kuramoto-interaction-kernel-true-vs-learned-with-KDE_10_trials} we visualize the true interaction kernel $\phi$ (blue) together with the mean recovered interaction kernels $\bar{\hat{\phi}}$ obtained via the NP (yellow) and SP (orange) approaches. Subfigure~\ref{subfig:Kuramoto-environmental-force-true-vs-learned-with-KDE_10_trials} provides the corresponding true environmental force $f$ and its recovered mean estimates $\bar{\hat{f}}$.  Mean estimates are obtained over $T_{r}=10$ independent learning trials, and the corresponding standard deviations are also plotted. The background density in Subfigure~\ref{subfig:Kuramoto-interaction-kernel-true-vs-learned-with-KDE_10_trials} (right vertical axis) corresponds to the empirical distribution $\hat{\rho}_{\Delta\theta}$ of pairwise phase differences, while in Subfigure~\ref{subfig:Kuramoto-environmental-force-true-vs-learned-with-KDE_10_trials} this corresponds to the empirical distribution $\hat{\rho}_{\theta}$ of oscillator phases observed in the training data. The corresponding relative feature recovery errors, computed via equation~\eqref{eq:feature_functions_relative_errors}, are reported in Table~\ref{tab:Kuramoto_feature_comparison}. Both formulations recover the interaction kernel with comparable accuracy, while the semi-parametric formulation achieves nearly exact recovery of the environmental force. The training data consists of $M=20$ trajectories with $N=10$ oscillators in dimension $d=1$, observed over the interval $[0,T]=[0,0.5]$ with $L=51$ time points.}
\label{fig:Kuramoto_interaction_kernel_and_fenv_P_and_NP}
\end{figure}

\begin{figure}
    \centering

    \begin{subfigure}{0.48\textwidth}
        \centering
        \includegraphics[width=\linewidth]{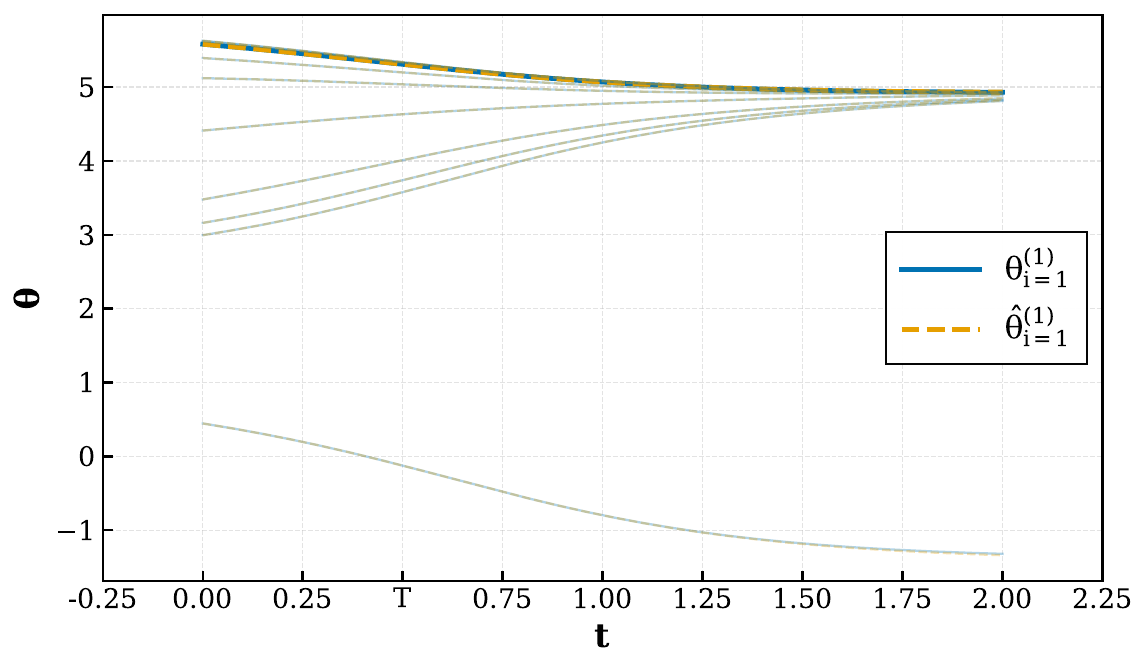}
        \caption{Semi-parametric trajectories on $M_{\mathrm{train}}$ training data}
        \label{subfig:Kuramoto_Traj_plot_P_M_train}
    \end{subfigure}
    \hfill
        \begin{subfigure}{0.48\textwidth}
        \centering
        \includegraphics[width=\linewidth]{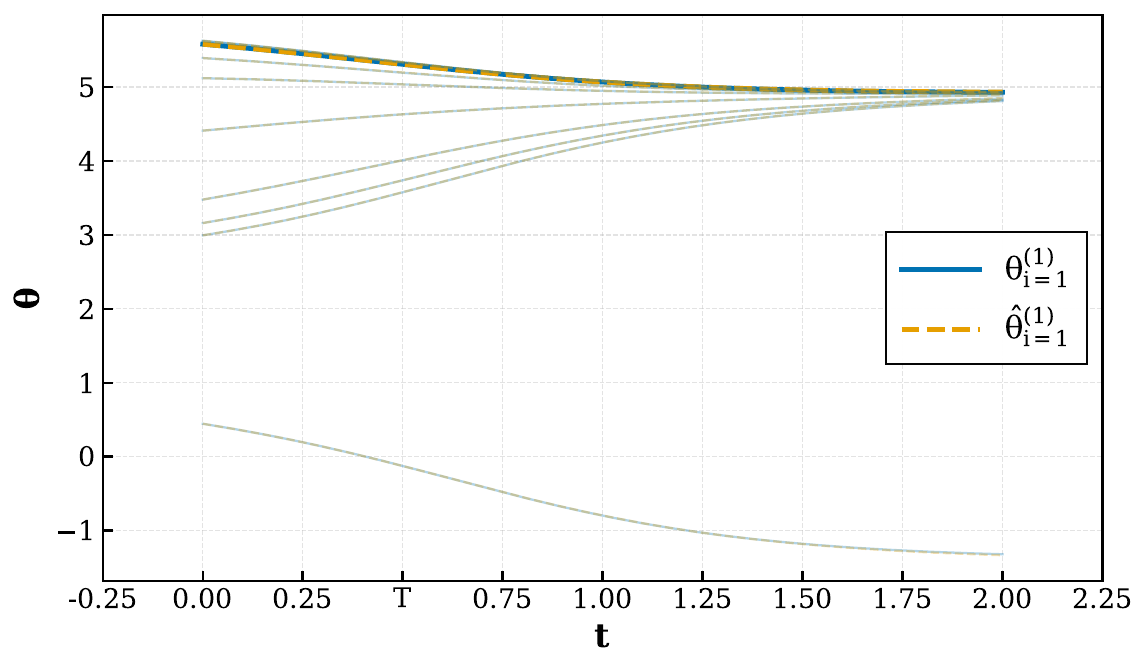}
        \caption{Fully non-parametric trajectories on $M_{\mathrm{train}}$ training data}
        \label{subfig:Kuramoto_Traj_plot_NP_M_train}
    \end{subfigure}

    \vspace{0.3cm}

    \begin{subfigure}{0.48\textwidth}
        \centering
        \includegraphics[width=\linewidth]{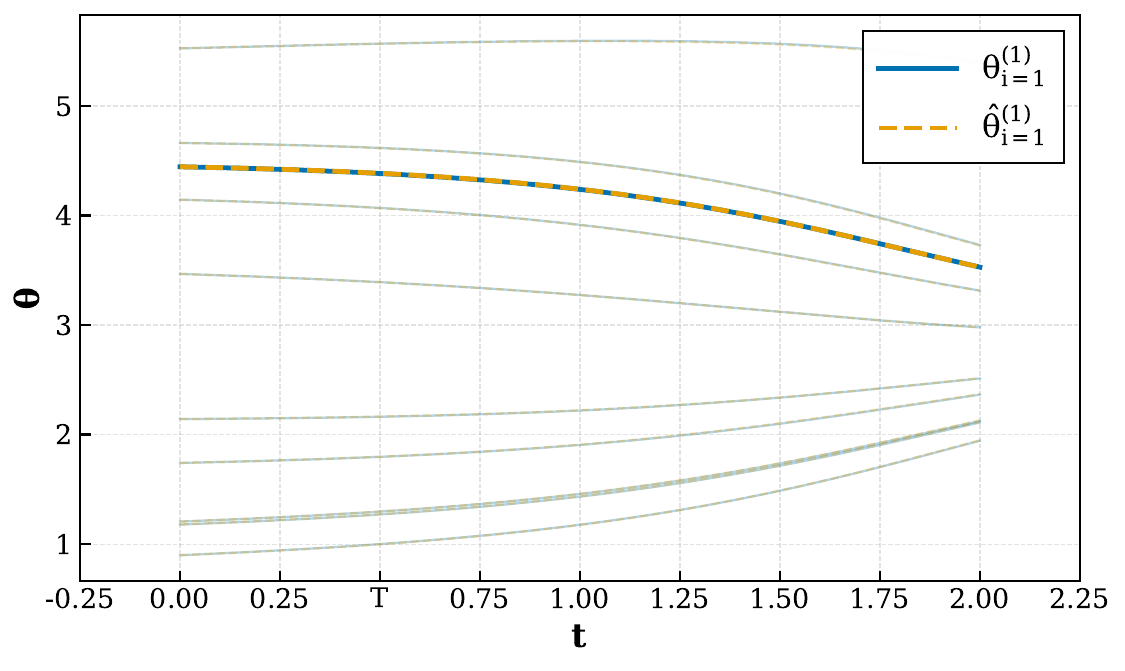}
        \caption{Semi-parametric trajectories on $M_{\mathrm{test}}$ testing data}
        \label{subfig:Kuramoto_Traj_plot_P_M_test}
    \end{subfigure}
    \hfill
    \begin{subfigure}{0.48\textwidth}
        \centering
        \includegraphics[width=\linewidth]{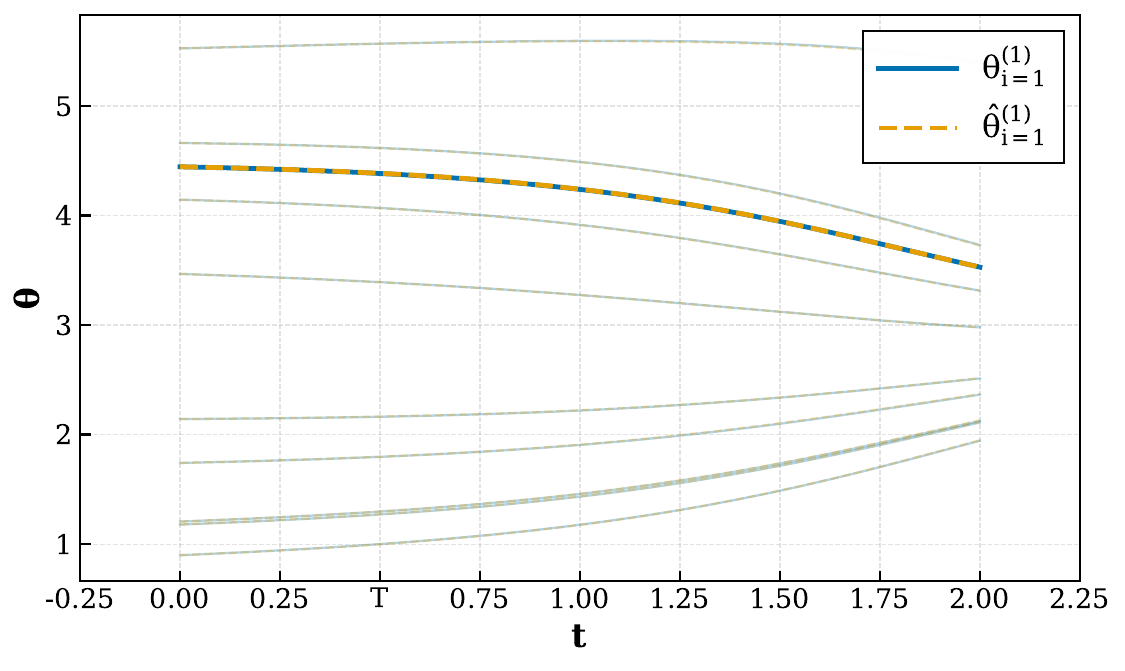}
        \caption{Fully non-parametric trajectories on $M_{\mathrm{test}}$ testing data}
        \label{subfig:Kuramoto_Traj_plot_NP_M_test}
    \end{subfigure}

    \caption{
        (Kuramoto model) Comparison of the true phase trajectories $\theta_{i}$ (blue) and the estimated phase trajectories $\hat{\theta}_{i}$ (yellow) for both the semi-parametric (left column) and non-parametric (right column) methods.  For trajectory errors, see Table~\ref{tab:Kuramoto_trajectory_comparison}.  The top row displays trajectories initialized from the $M_{\mathrm{train}}$ training trajectories, whereas the bottom rows displays trajectories generated from $M_{\mathrm{test}}$ previously unseen testing initial conditions. Note that trajectories are plotted over the full time horizon $[0,T_f]$, with the end of the training interval time $T$ indicated on the horizontal axis. Recall that models are trained using observations collected only over the interval $[0,T]=[0,0.5]$, and are evaluated beyond the fitting horizon to assess both reconstruction and predictive performance.  Both method variants closely reproduce the true dynamics on the training trajectories and generalize well to unseen initial conditions, with no visual differences between the semi-parametric and non-parametric models.  For additional details on algorithm parameters, we refer the reader to the caption of Figure~\ref{fig:Kuramoto_interaction_kernel_and_fenv_P_and_NP}.
        }
    \label{fig:Kuramoto_True_vs_Learned_Trajectories}
\end{figure}

We begin by comparing estimators for the interaction kernel $\phi$ and the environmental force $f$ for both the semi-parametric (SP) and fully non-parametric (NP) version of the algorithm; results are visualized in Figure~\ref{fig:Kuramoto_interaction_kernel_and_fenv_P_and_NP}.  Note that in the semi-parametric algorithm, we estimate $\hat{p}=0.0100000000000000141$ (see Table~\ref{tab:app:param-recovery}). We also observe that the interaction kernel $\phi$ is estimated almost identically by both approaches, with relative errors provided in Table~\ref{tab:Kuramoto_feature_comparison} on the order of $10^{-2}$, suggesting that the inference of $\phi$ is relatively insensitive to the approach, given the data utilized. For the environmental force, the semi-parametric model recovers $f$ essentially to machine precision. The fully non-parametric formulation nevertheless approximates the environmental force $f$ accurately, with a relative error $E_{f}^{\mathrm{rel}} \approx 2.0 \times 10^2$.  Indeed, despite the large difference in $E^{\mathrm{rel}}_f$ between methods, there is visually no discernible difference between the true and predicted trajectories, which are generated from both $M_{\mathrm{train}}$ and $M_{\mathrm{test}}$ on the full time horizon $[0, T_f]$, as observed in  Figure~\ref{fig:Kuramoto_True_vs_Learned_Trajectories}.  The trajectory errors are also quantified via the metrics discussed in Section~\ref{subsec:trajectory_Based_metrics}, and are provided in Table~\ref{tab:Kuramoto_trajectory_comparison}.  Note that the semi-parametric and fully non-parametric methods produce both small and similar errors with respect to trajectories (both on training and prediction data), suggesting that the two approaches provide statistically equivalent performance for the Kuramoto system on the utilized data.

\begin{table}
\centering

\begin{tabular}{lcc}
\toprule
Metric
& Non-parametric
& Semi-Parametric \\
\midrule

$E^{\mathrm{rel}}_{\mathrm{res}}$
& \num{9.13329754e-03} $\pm$ \num{1.06999932e-03}
& \num{9.21341749e-03} $\pm$ \num{1.07433881e-03} \\

$E^{\mathrm{rel}}_{\phi}$
& \num{1.15194900e-02} $\pm$ \num{6.56755290e-04}
& \num{1.15035007e-02} $\pm$ \num{6.50524925e-04} \\

$E^{\mathrm{rel}}_{f}$
& \num{1.99202020e-02} $\pm$ \num{5.68123500e-03}
& \num{1.71268822e-14} $\pm$ \num{1.18487390e-14} \\

\bottomrule
\end{tabular}
\caption{
(Kuramoto model) Comparison of relative residual error~\eqref{eq:relative_residue_error} and feature recovery metrics~\eqref{eq:feature_functions_relative_errors} for the fully non-parametric and semi-parametric estimation algorithms for the Kuramoto model~\eqref{eq:Kuramoto-ODE}. The table reports the mean error $\pm$ one standard deviation over $T_r=10$ independent learning trials.  For additional details on algorithm parameters, we refer the reader to the caption of Figure~\ref{fig:Kuramoto_interaction_kernel_and_fenv_P_and_NP}.
}
\label{tab:Kuramoto_feature_comparison}
\end{table}

\begin{table}[ht]
\centering
\begin{tabular}{llcc}
\toprule
Data set & Statistic
& Fully non-parametric
& Semi-parametric \\
\midrule

\multirow{4}{*}{$M_{\mathrm{train}}$}
& $\bar E_{\mathrm{recon}}$
& \num{1.48611521e-03} $\pm$ \num{1.29000801e-04}
& \num{1.51517472e-03} $\pm$ \num{1.32610095e-04} \\

& $\sigma_{E_{\mathrm{recon}}}$
& \num{7.05905063e-04} $\pm$ \num{1.14935014e-04}
& \num{7.25846747e-04} $\pm$ \num{1.07481546e-04} \\

& $\bar E_{\mathrm{pred}}$
& \num{5.45704659e-03} $\pm$ \num{1.45481620e-03}
& \num{5.45918478e-03} $\pm$ \num{1.40901706e-03} \\

& $\sigma_{E_{\mathrm{pred}}}$
& \num{5.66547687e-03} $\pm$ \num{2.89503932e-03}
& \num{6.02093452e-03} $\pm$ \num{2.94778488e-03} \\
\midrule

\multirow{4}{*}{$M_{\mathrm{test}}$}
& $\bar E_{\mathrm{recon}}$
& \num{1.63522403e-03} $\pm$ \num{1.99461841e-04}
& \num{1.59861149e-03} $\pm$ \num{1.29374535e-04} \\

& $\sigma_{E_{\mathrm{recon}}}$
& \num{9.23011242e-04} $\pm$ \num{2.37416998e-04}
& \num{9.09415034e-04} $\pm$ \num{2.15935277e-04} \\

& $\bar E_{\mathrm{pred}}$
& \num{5.10647667e-03} $\pm$ \num{1.43645617e-03}
& \num{4.78553671e-03} $\pm$ \num{1.23918608e-03} \\

& $\sigma_{E_{\mathrm{pred}}}$
& \num{3.90150189e-03} $\pm$ \num{1.10416966e-03}
& \num{3.71452413e-03} $\pm$ \num{8.68060498e-04} \\

\bottomrule
\end{tabular}
\caption{(Kuramoto model) Comparison of trajectory-level errors for the fully non-parametric and semi-parametric estimation algorithms for Kuramoto model~\eqref{eq:Kuramoto-ODE}.  For each learning trial, the trajectory statistics
$\bar E_{\mathrm{recon}}$,
$\bar E_{\mathrm{pred}}$,
$\sigma_{E_{\mathrm{recon}}}$,
and
$\sigma_{E_{\mathrm{pred}}}$
are computed according to~\eqref{eq:mean_trajectory_errors}~-~\eqref{eq:std_trajectory_error}.
The table reports the mean $\pm$ one standard deviation of these quantities computed on $T_r=10$ independent learning trials.
The first section corresponds to trajectories generated from $M_{\mathrm{train}}$ training initial conditions, while the second section corresponds to $M_{\mathrm{test}}$ unseen predicted initial conditions sampled from $\mu_0$.  Note that across all trajectory measures, error estimates are both small and similar for both methods.  For additional details on algorithm parameters, we refer the reader to the caption of Figure~\ref{fig:Kuramoto_interaction_kernel_and_fenv_P_and_NP}.}
\label{tab:Kuramoto_trajectory_comparison}
\end{table}

\subsubsection{Self-propelled particle (SPP) model}
\label{subsubsec:results_SPP}
Understanding how organisms move in cohesive groups has been an extensively studied mathematical modeling problem. Indeed, it is known in nature that such systems can exhibit various types of behavior, including flocking, where agents converge to a common velocity, milling, where agents rotate around a common center or axis, and swarming, which often appears as a transitional state between flocking and milling.  In this section we consider a model of self-propelled particles (SPPs), which was first introduced in~\cite{d2006self,chuang2007state} and further studied in~\cite{albi2014stability,chuang2016swarming}, and can exhibit phenomenon such as clustering, milling, and flocking.  These models are motivated by biological aggregation phenomena and have been widely used to study collective motion. For additional background on SPP models, we refer the reader to~\cite{lukeman2009conceptual,abaid2010fish,bernoff2011primer}; mathematical analysis of the dynamics of such systems may be found in~\cite{carrillo2009double,degond2011hydrodynamic,albi2014stability}.

The governing equations of the SPP system considered here take the form
    \begin{align}
    \begin{split}
        \dot{x}_i &= v_i \\
        \dot{v}_i &= (\alpha - \beta |v_i|^2)v_i - \nabla_{x_{i}} U(x_i)
        \end{split}
        \label{SPP_model_equaions}
    \end{align} 
where the generalized Morse interaction potential $U$ is given by
    \begin{align}
        U(x_i) &= \sum_{j \neq i}^N C_r e^{-|x_i -x_j|/l_r} - C_a e^{-|x_i -x_j|/l_a},
        \label{eg:Morse_potential_SPP}
    \end{align}
and combines the effects of attracting and repulsion between agents.  Here $\alpha$ scales the strength of self-propulsion, $\beta$ models a nonlinear damping effect, $l_a$ and $l_r$ are the attractive and repulsive interaction ranges, respectively, and $C_a$ and $C_r$ denote the corresponding interaction strengths.  Note that the system~\eqref{SPP_model_equaions} is second-order.

For numerical experiments, we utilize the following parameters (as reported in~\cite{chuang2007state}): $C_r =0.6, l_r=0.5, C_a=1, l_a=1, \alpha=1, \beta=0.5$ . The remaining simulation and learning parameters are reported in Tables~\ref{tab:app:data_generation_parameters},\ref{tab:app:learning_parameters}, and ~\ref{tab:app:system-initial-conditions} of Appendix~\ref{app:sec:tables_of_parameters}. Equations~\eqref{SPP_model_equaions} take the general form of a second-order collective system~\eqref{eq:second_order_system_methods}, where 
\begin{align}
\begin{split}
    f(x_i, v_i) &= (\alpha - \beta |v_i|^2)v_i \quad \text{and} \quad \phi (r) = \frac{N}{r}\left(\frac{-C_r}{l_r}e^{-r/l_r} + \frac{C_a}{l_a}e^{-r/l_a}\right)
    \end{split}
    \label{eq:SPP_f_env_term_and_interaction_kernel-terms}
\end{align}
Note that the number of agents $N$ appears in $\phi$ due to the fact that that the learning framework averages interaction forces for each agent ($1/N$ in~\eqref{eq:second_order_system_methods}), whereas the standard SPP system~\eqref{SPP_model_equaions} does not.

\begin{figure}
    \centering
    \begin{subfigure}{0.48\textwidth}
      \centering
      \includegraphics[width=\textwidth]{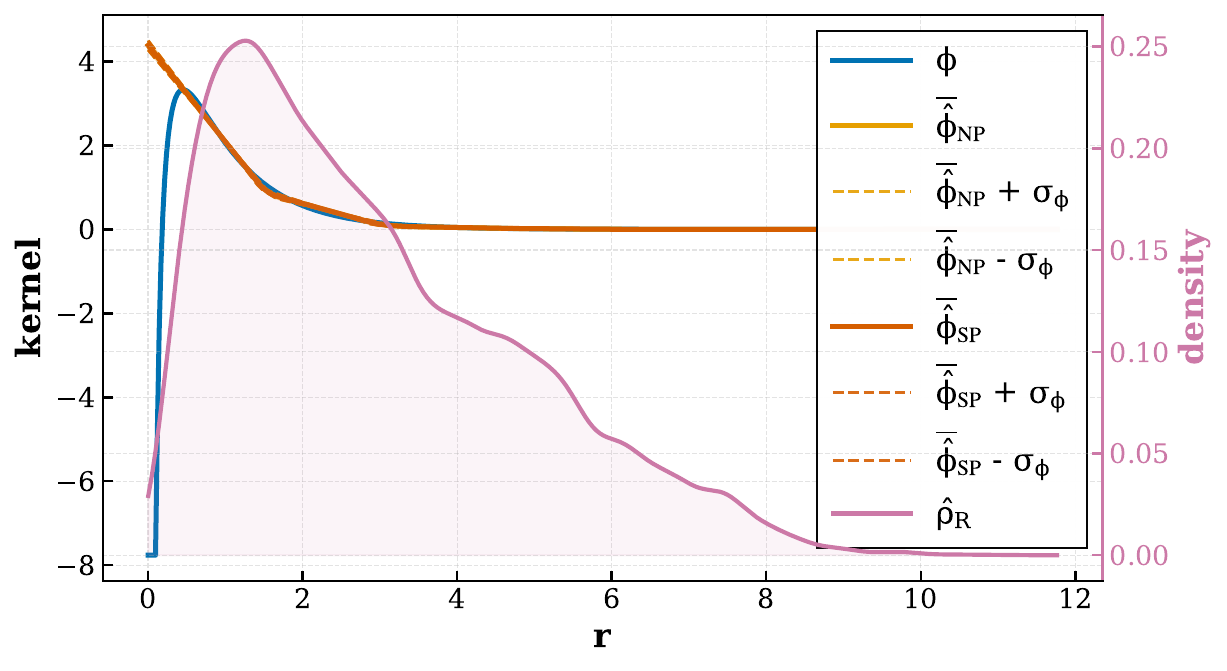}
      \caption{Estimation of interaction kernel $\phi$}
    \label{subfig:SPP_interaction_kernel_true_vs_learned_with_KDE_10_trials}
    \end{subfigure}
    \begin{subfigure}{0.48\textwidth}
      \centering
      \includegraphics[width=\textwidth]{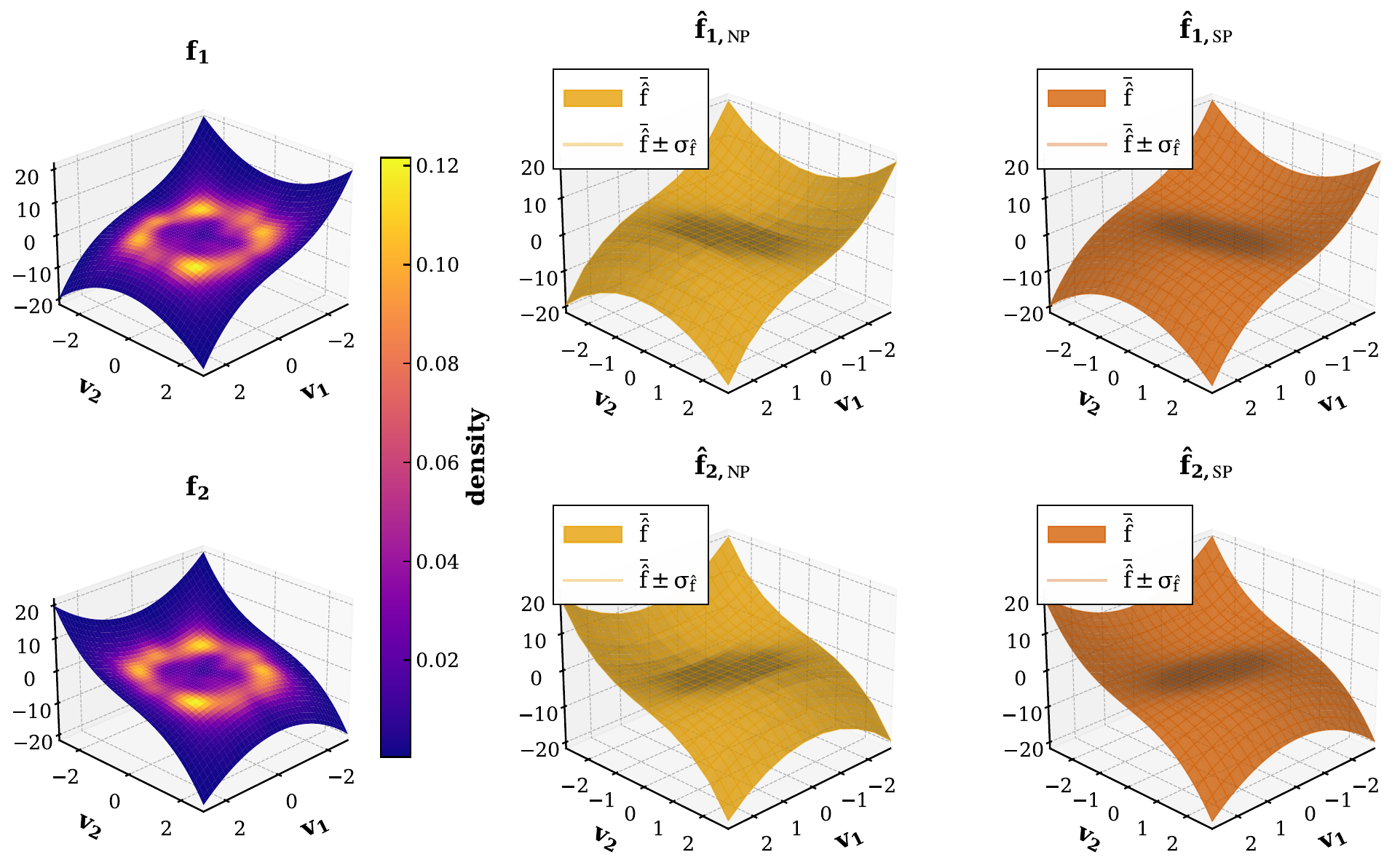}
      \caption{Estimation of environmental force $f$}
    \label{subfig:SPP_intra_agent_force_true_vs_learned_with_KDE_10_trials}
    \end{subfigure}
    
     \caption{(SPP model) Comparison of mechanistic feature recovery ability of the fully non-parametric (NP) and semi-parametric (SP) estimation algorithms for the SPP model~\eqref{SPP_model_equaions}.  In Subfigure~\ref{subfig:SPP_interaction_kernel_true_vs_learned_with_KDE_10_trials} we visualize the true interaction kernel $\phi$ (blue) with the mean recovered kernels $\bar{\hat{\phi}}$ obtained via the NP (yellow) and SP (orange) formulations. Subfigure~\ref{subfig:SPP_intra_agent_force_true_vs_learned_with_KDE_10_trials} provides the true environmental force $f$ with the corresponding mean learned estimates $\bar{\hat{ f}}$. As $f: \R^{2} \to \R^{2}$, the top and bottom rows correspond to the first and second components of $f$, respectively. Within each row, the true force (left column), the fully non-parametric mean estimate (middle column), and the semi-parametric mean estimate (right column) are shown, together with one standard deviation, which are obtained over $T_r=10$ independent learning trials. The background density in Subfigure~\ref{subfig:SPP_interaction_kernel_true_vs_learned_with_KDE_10_trials} (right vertical axis) corresponds to the empirical pairwise distance distribution $\hat{\rho}_R$, while the density visualized in Subfigure~\ref{subfig:SPP_intra_agent_force_true_vs_learned_with_KDE_10_trials} (left column) corresponds to the empirical velocity distribution $\hat{\rho}_V$ induced by the training data. The corresponding relative feature recovery errors are reported in Table~\ref{tab:SPP_feature_comparison}. The training data consists of $M=20$ trajectories with $N=10$ agents in dimension $d=2$, observed over $[0,T]=[0.0,0.5]$ with $L=51$ time points.}

\label{fig:SPP_interaction_kernel_and_fenv_P_and_NP}
\end{figure}

\begin{figure}
    \centering
    \begin{subfigure}{\textwidth}
        \centering
        \includegraphics[width=\linewidth]{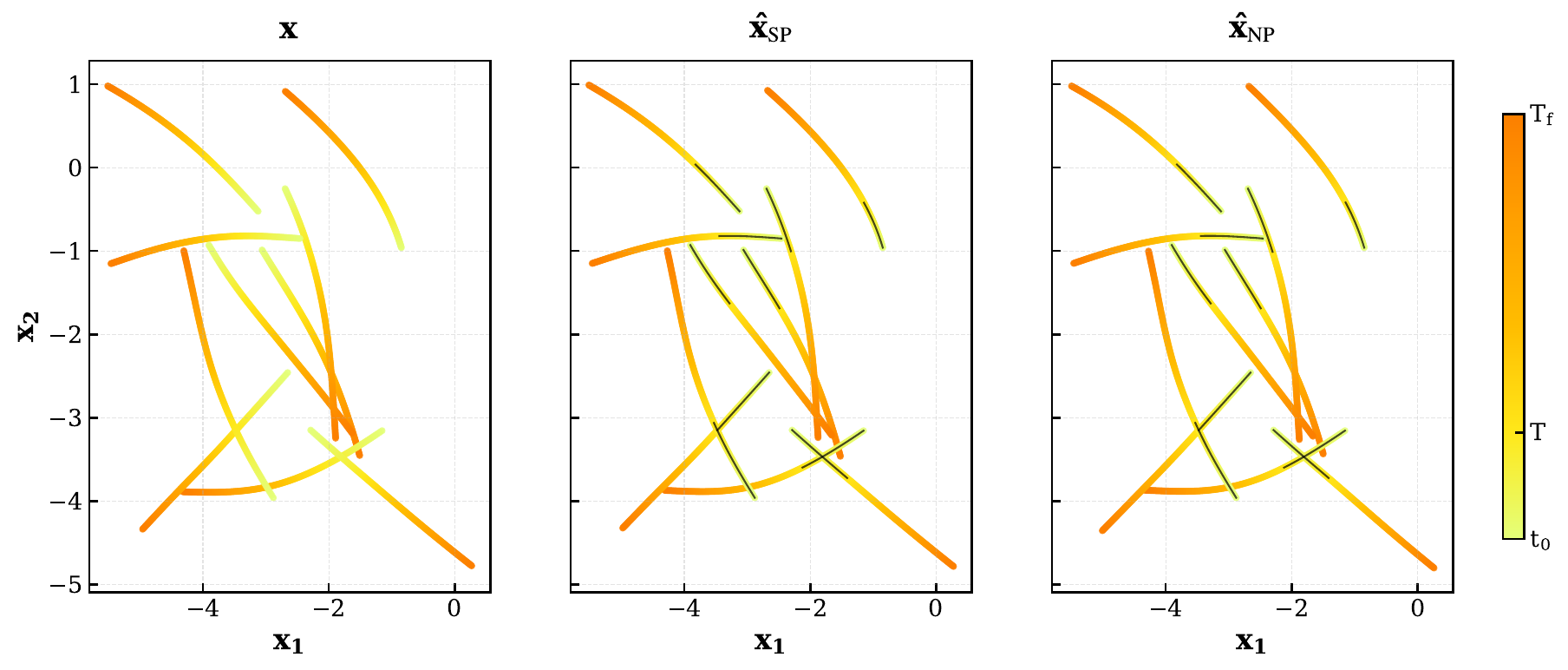}
        \caption{$M_{\mathrm{train}}$ training data utilized to obtain estimators}
        \label{subfig:SPP_Traj_plot_P_NP_M_train}
    \end{subfigure}

    \vspace{0.1cm}

    \begin{subfigure}{\textwidth}
        \centering
        \includegraphics[width=\linewidth]{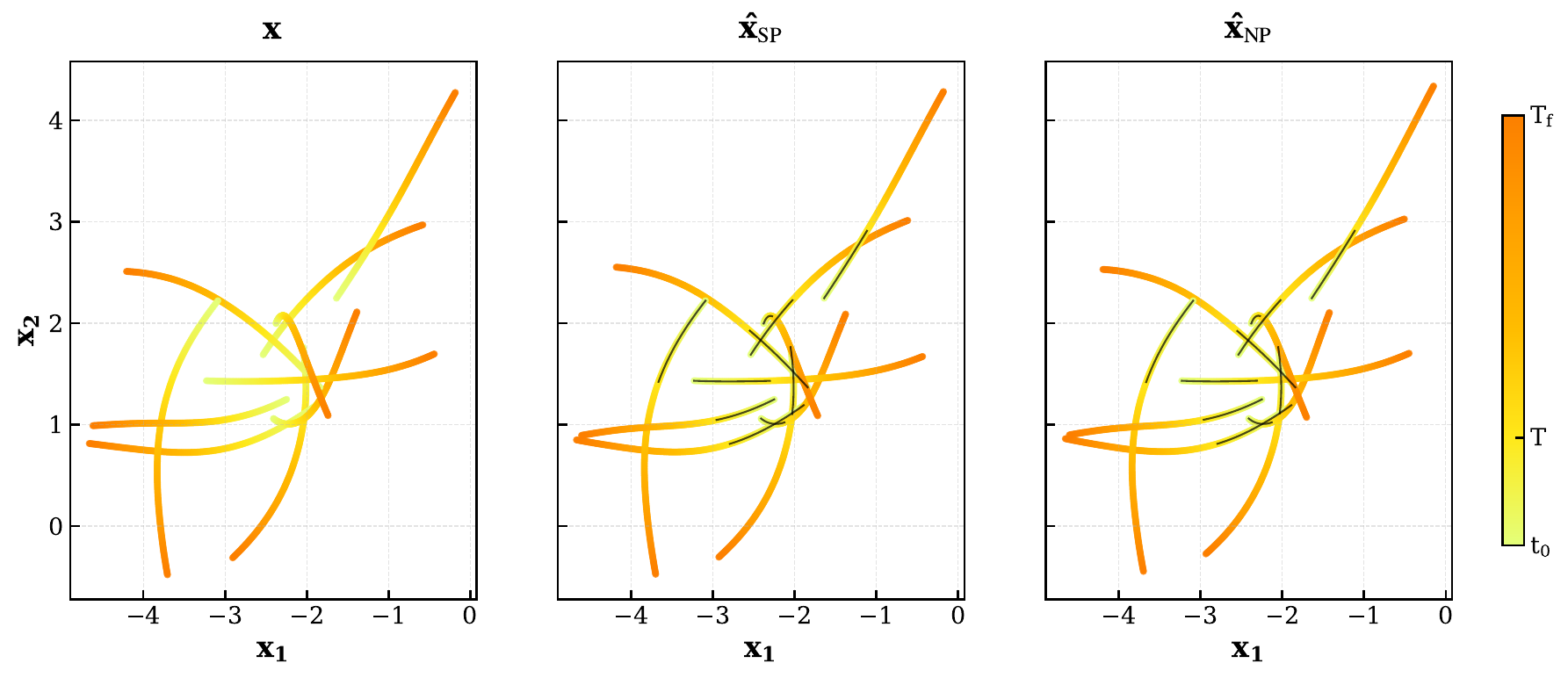}
        \caption{$M_{\mathrm{test}}$ test data utilized for evaluation}
        \label{subfig:SPP_Traj_plot_P_NP_M_test}
    \end{subfigure}

     \caption{(SPP model) Comparison of the true trajectories $x$ (left column) and the estimated trajectories $\hat{x}$ for both the semi-parametric (middle column) and fully non-parametric (right column) methods in the phase plane. For trajectory errors, see Table~\ref{tab:SPP_trajectory_comparison}.  The top row displays trajectories initialized from the $M_{\mathrm{train}}$ training initial-conditions, whereas the bottom row displays trajectories generated from $M_{\mathrm{test}}$ previously unseen testing initial conditions.  Note that trajectories are plotted over the full time horizon $[0, T_f]$, with the color gradient representing the temporal evolution of the dynamics, from yellow at $t=0$ to orange at $t=T_f$.  The black segment indicates the training interval $[0,T]$, which is utilized for model fitting; recall that no fitting is performed in the testing data in Figure~\ref{subfig:SPP_Traj_plot_P_NP_M_test}.  Here $[0,T]=[0.0,0.5]$, with testing evaluated beyond the fitting horizon $[0.0,0.5]$ until $T_f=2.0$ to assess both reconstruction and prediction in Figure~\ref{subfig:SPP_Traj_plot_P_NP_M_test}. Both method variants closely reproduce the true dynamics on the training trajectories and generalize well to unseen initial conditions.  For additional details on algorithm parameters, we refer the reader to the caption of Figure~\ref{fig:SPP_interaction_kernel_and_fenv_P_and_NP}.}
    \label{fig:SPP_True_vs_Learned_Trajectories}

\end{figure}

Figure~\ref{subfig:SPP_interaction_kernel_true_vs_learned_with_KDE_10_trials} provides a plot of the obtained estimators for the interaction kernel $\phi$ from both methods, and demonstrates that they both closely approximate $\phi$ over the region where the empirical pairwise distance distribution $\hat{\rho}_R$ is well sampled by the observed trajectory data, i.e. where the empirical pairwise distance distribution $\hat{\rho}_R$ has non-negligible mass. Note that when the interaction distance is close to $0$, the kernel $\phi$ in~\eqref{eq:SPP_f_env_term_and_interaction_kernel-terms} is singular, and as a result, there exist a limited number of trajectories with small interaction distances $r$.  As we utilize a uniformly-spaced basis of splines to approximate $\phi$, the reduced information content near $0$ leads to visible deviations between the true kernel $\phi$ and its estimators $\hat{\phi}$.  Similarly, Figure~\ref{subfig:SPP_intra_agent_force_true_vs_learned_with_KDE_10_trials} plots both the true environmental self-propelled force $f$ (left column) together with estimators from the fully non-parametric (middle column) and semi-parametric (right column) approaches; note that the top row corresponds to the first component of $f: \R^{2}\to\R^{2}$ in~\eqref{eq:SPP_f_env_term_and_interaction_kernel-terms} while the bottom row corresponds to the second component.  The approximate empirical distribution $\hat{\rho}_{V}$ is also included on the surface plot of $f$ to highlight the regions of velocity explored by the training trajectory data.  Again, in both cases, we observe qualitatively close agreement between estimators and the components of $f$.   Table~\ref{tab:SPP_feature_comparison} quantifies the accuracy the inferring mechanisms, demonstrating that the learning approaches recover $\phi$ with nearly identical accuracy, yielding $E^{\mathrm{rel}}_{\phi} \approx 0.12$. In contrast, the recovery of $f$ improves substantially under the semi-parametric formulation, with the relative error reduced by nearly two orders of magnitude. The true $x(t)$ and estimated $\hat{x}(t)$ trajectories are shown in Figure~\ref{fig:SPP_True_vs_Learned_Trajectories} on both training (Figure~\ref{subfig:SPP_Traj_plot_P_NP_M_train}) and testing (Figure~\ref{subfig:SPP_Traj_plot_P_NP_M_test}) samples. Quantitative trajectory errors reported in Table~\ref{tab:SPP_trajectory_comparison} remain small ($\approx 10^{-2}$), demonstrating that the learned models preserve the long-time collective behavior of the system.

\begin{table}
\centering
\begin{tabular}{lcc}
\toprule
Metric
& Non-parametric
& Semi-Parametric \\
\midrule

$E^{\mathrm{rel}}_{\mathrm{res}}$
& \num{3.25063615e-02} $\pm$ \num{3.80467817e-04}
& \num{1.46422590e-02} $\pm$ \num{7.41636649e-04} \\

$E^{\mathrm{rel}}_{\phi}$
& \num{1.19987060e-01} $\pm$ \num{1.99349402e-03}
& \num{1.19752915e-01} $\pm$ \num{1.84175139e-03} \\

$E^{\mathrm{rel}}_{f}$
& \num{2.97416742e-02} $\pm$ \num{3.44985201e-04}
& \num{6.48045897e-04} $\pm$ \num{1.34464077e-04} \\
\bottomrule
\end{tabular}
\caption{
(SPP model) Comparison of relative residual error~\eqref{eq:relative_residue_error} and feature recovery metrics~\eqref{eq:feature_functions_relative_errors} for the fully non-parametric and semi-parametric estimation algorithms for the SPP model~\eqref{SPP_model_equaions}. The table reports the mean error $\pm$ one standard deviation of these quantities computed over $T_r=10$ independent learning trials.  For additional details on algorithm parameters, we refer the reader to the caption of Figure~\ref{fig:SPP_interaction_kernel_and_fenv_P_and_NP}.
}
\label{tab:SPP_feature_comparison}
\end{table}

\begin{table}
\centering
\begin{tabular}{llcc}
\toprule
Data set & Statistic
& Fully non-parametric
& Semi-parametric \\
\midrule

\multirow{4}{*}{$M_{\mathrm{train}}$}
& $\bar E_{\mathrm{recon}}$
& \num{4.29165612e-03} $\pm$ \num{7.32996906e-05}
& \num{2.76263652e-03} $\pm$ \num{8.64262733e-05} \\

& $\sigma_{E_{\mathrm{recon}}}$
& \num{1.01878056e-03} $\pm$ \num{1.04784799e-04}
& \num{9.36360347e-04} $\pm$ \num{1.23291849e-04} \\

& $\bar E_{\mathrm{pred}}$
& \num{4.96923582e-02} $\pm$ \num{2.22450017e-03}
& \num{2.48460158e-02} $\pm$ \num{1.48102468e-03} \\

& $\sigma_{E_{\mathrm{pred}}}$
& \num{9.78521605e-03} $\pm$ \num{9.87781309e-04}
& \num{9.04234873e-03} $\pm$ \num{1.31457131e-03} \\
\midrule

\multirow{4}{*}{$M_{\mathrm{test}}$}
& $\bar E_{\mathrm{recon}}$
& \num{4.25656162e-03} $\pm$ \num{1.29139691e-04}
& \num{2.76701167e-03} $\pm$ \num{1.01833324e-04} \\

& $\sigma_{E_{\mathrm{recon}}}$
& \num{1.13325476e-03} $\pm$ \num{1.40514745e-04}
& \num{8.26865924e-04} $\pm$ \num{9.90060736e-05} \\

& $\bar E_{\mathrm{pred}}$
& \num{4.93943026e-02} $\pm$ \num{2.91593163e-03}
& \num{2.47350508e-02} $\pm$ \num{8.05099388e-04} \\

& $\sigma_{E_{\mathrm{pred}}}$
& \num{8.62694195e-03} $\pm$ \num{2.36222774e-03}
& \num{7.03799759e-03} $\pm$ \num{1.44251404e-03} \\

\bottomrule
\end{tabular}
\caption{(SPP model) Comparison of trajectory-level errors for the fully non-parametric and semi-parametric estimation algorithms for the SPP model~\eqref{SPP_model_equaions}.  For each learning trial, the trajectory statistics
$\bar E_{\mathrm{recon}}$,
$\bar E_{\mathrm{pred}}$,
$\sigma_{E_{\mathrm{recon}}}$,
and
$\sigma_{E_{\mathrm{pred}}}$
are computed according to~\eqref{eq:mean_trajectory_errors}~-~\eqref{eq:std_trajectory_error}.
The table reports the mean $\pm$ one standard deviation of these quantities computed over $T_r=10$ independent learning trials.
The first section corresponds to trajectories generated from $M_{\mathrm{train}}$ training initial conditions, while the second section corresponds to $M_{\mathrm{test}}$ unseen initial conditions sampled from $\mu_0$.  Note that across all trajectory measures, error estimates are both small and similar for both methods.  For additional details on algorithm parameters, we refer the reader to the caption of Figure~\ref{fig:SPP_interaction_kernel_and_fenv_P_and_NP}.
}
\label{tab:SPP_trajectory_comparison}
\end{table}

\begin{figure}[H]
\centering

\begin{minipage}[t]{0.48\textwidth}
\vspace{0pt}
    \centering

        \begin{subfigure}{\textwidth}
        \centering
        \includegraphics[width=\textwidth, height=4cm,keepaspectratio]
        {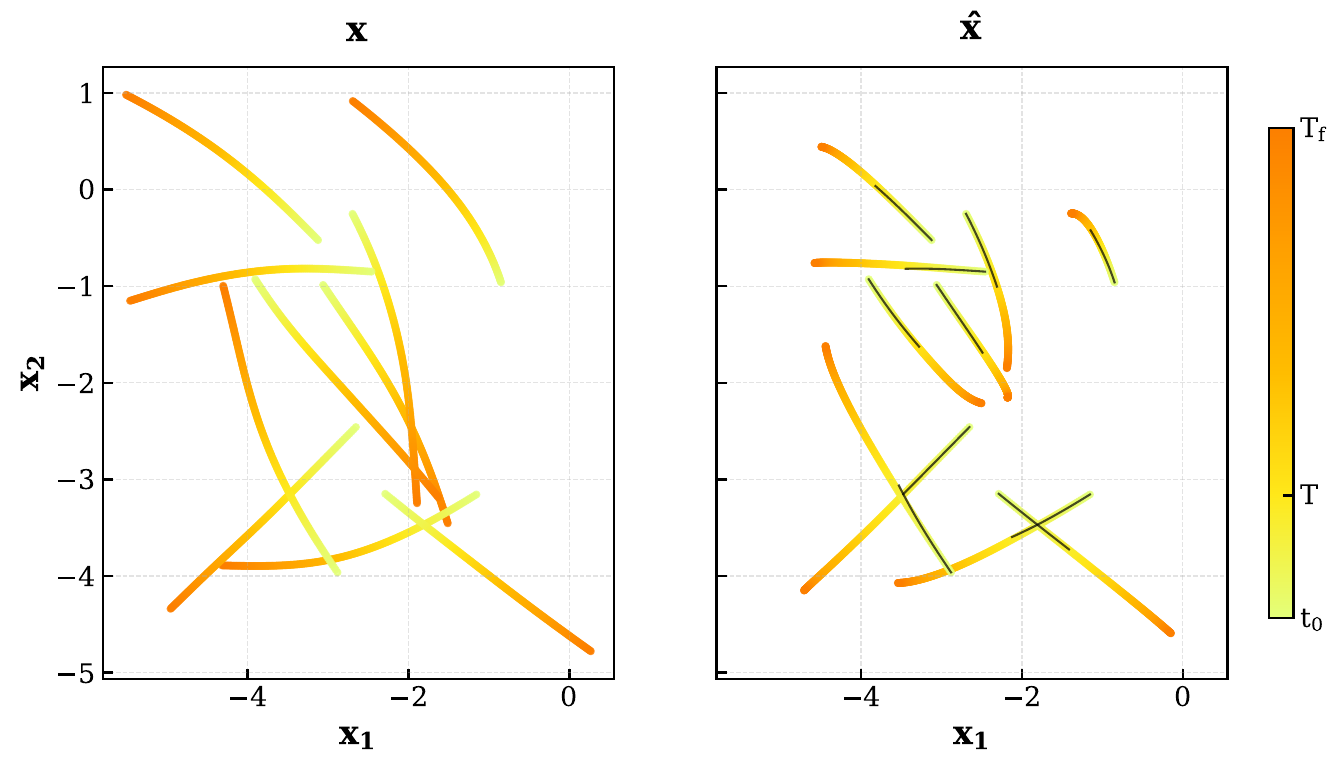}
        \caption{True (left column) and estimated (right column) trajectories $x$}
        \label{subfig:Missspecified_paramters_trajectories}
    \end{subfigure}

    \begin{subfigure}{\textwidth}
        \centering
        \includegraphics[width=\textwidth, height=3cm, keepaspectratio]
        {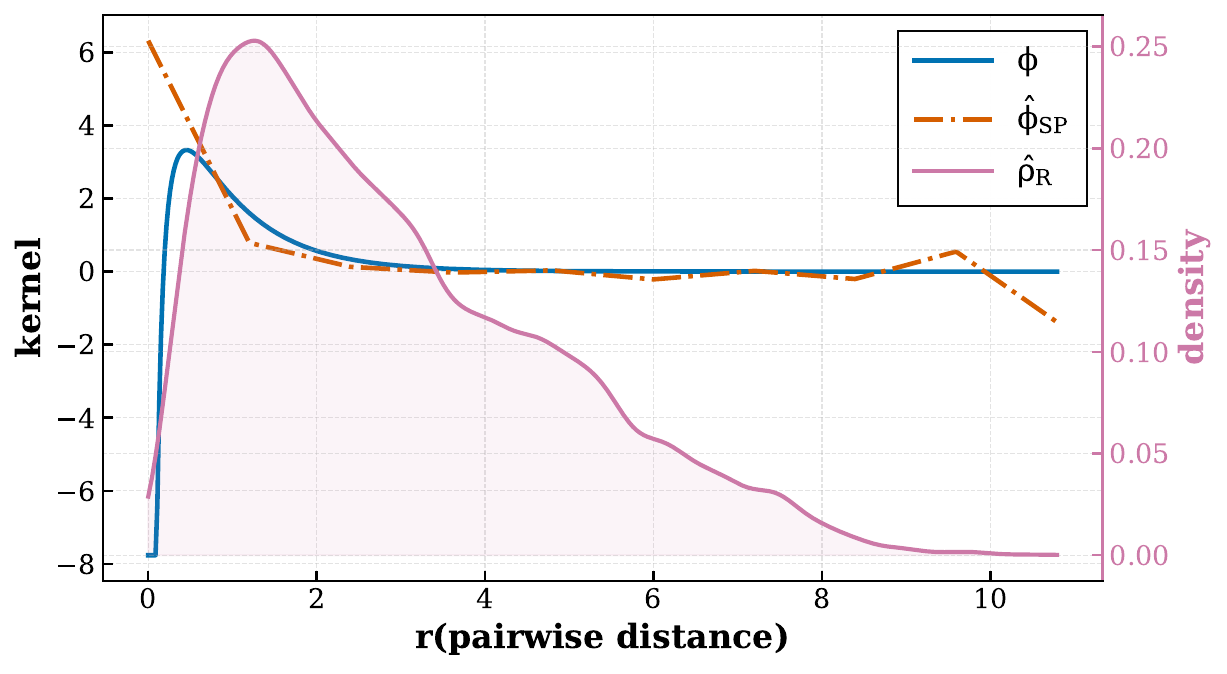}
        \caption{Interaction kernel $\phi$}
        \label{subfig:Missspecified_paramters_kernel}
    \end{subfigure}

    \vspace{0.1em}

\end{minipage}
\hfill
\begin{minipage}[t]{0.48\textwidth}
\vspace{0pt}
    \centering

    \begin{subfigure}{\textwidth}
        \centering
        \includegraphics[width=\textwidth]
        {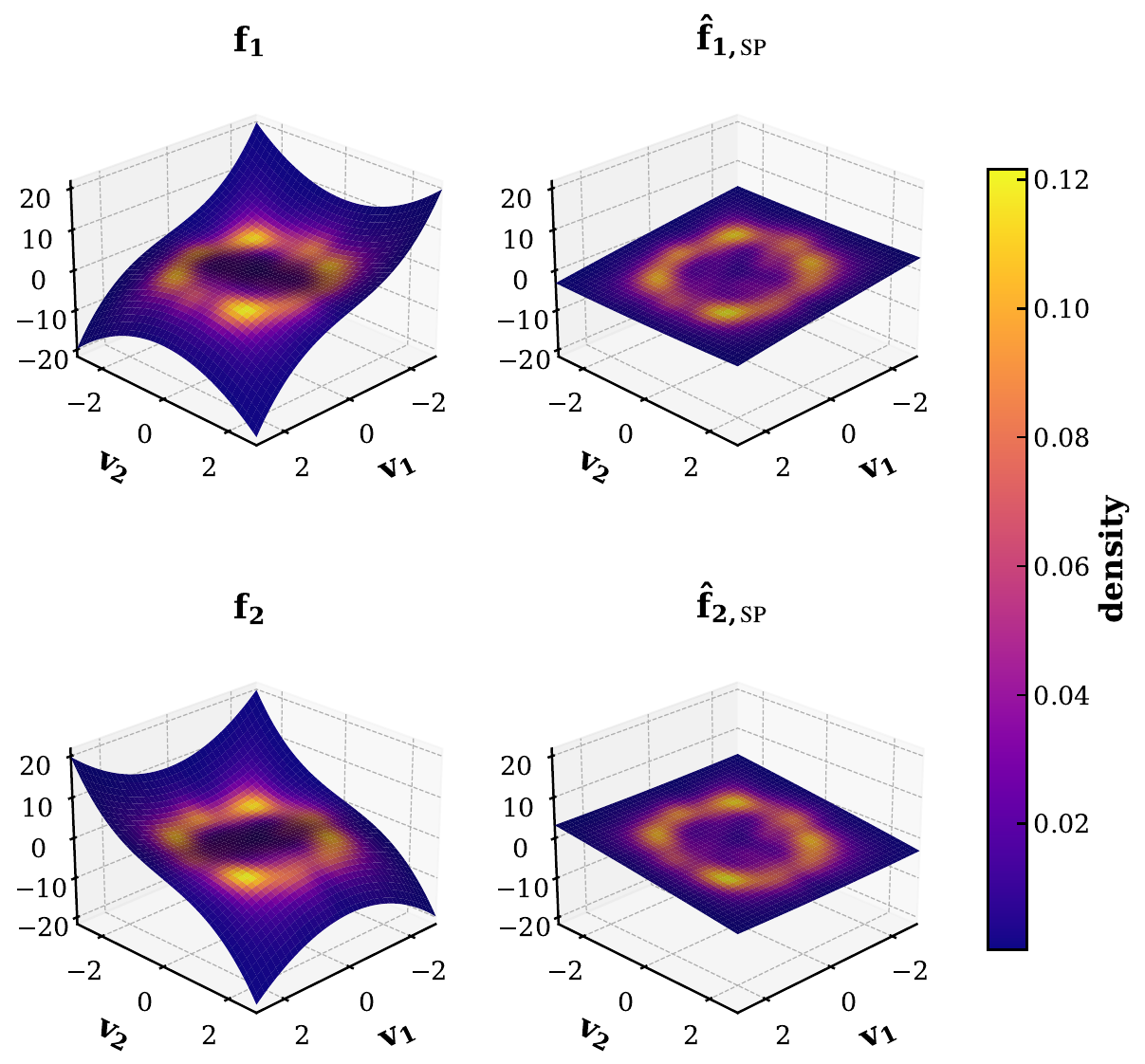}
        \caption{True (left column) and estimated (right column) environmental force $f$}
        \label{subfig:Missspecified_paramters_f}
    \end{subfigure}

\end{minipage}

\caption{(Misspecified parametric model, SPP) Results for the SPP system~\eqref{SPP_model_equaions} when the semi-parametric algorithm is supplied with an incorrect functional form for the environmental force $f$. Here $f$ is assumed of the form $f(v;p) = p_{1} v + p_{2}||v||_{1}v$, while the data is generated from the model~\eqref{SPP_model_equaions}. Note that the algorithm is unable to recover the correct mechanistic features and dynamics. Subfigure~\ref{subfig:Missspecified_paramters_trajectories} plots the resulting phase portraits for one initial condition from the $M_{\mathrm{train}}$ training data; observe that even on the training trajectories, the learned dynamics deviate noticeably from the true dynamics within the predictive interval $[T, T_f]$.  Subfigure~\ref{subfig:Missspecified_paramters_kernel} compares the true interaction kernel $\phi$ (blue solid line) with the learned kernel $\hat{\phi}$ obtained from the misspecified parametric model (red dashed line). Subfigure~\ref{subfig:Missspecified_paramters_f} compares the true (left column) and estimated (right column) environmental forces $f$, with the upper and lower panels corresponding to the first and second components, respectively. For additional details on algorithm parameters, we refer the reader to the caption of Figure~\ref{fig:SPP_interaction_kernel_and_fenv_P_and_NP}.}
\label{fig:SPP_parameter_misspecification}
\end{figure}

One limitation of the semi-parametric algorithm is that its performance depends strongly on the correctness of the prescribed functional form for the environmental force $f$. If the true $f$ which generates the data cannot be realized in the assumed form, the algorithm will generally fail to recover the underlying features and predict trajectories.  Figure~\ref{fig:SPP_parameter_misspecification} illustrates such a case for the SPP system.  Numerically, we assume the parametric form of the environmental force takes the form $f(v;p) = p_{1} v + p_{2}||v||_{1}v$, where $p=(p_{1},p_{2})$ is the vector of parameters.  Note that the true environmental force is given by~\eqref{eq:SPP_f_env_term_and_interaction_kernel-terms}, which cannot be represented by the assumed form for any $p \in \R^{2}$.  We then solve the semi-parametric estimation algorithm on training data generated from~\eqref{eq:SPP_f_env_term_and_interaction_kernel-terms}, and the resulting estimators are provided in Figure~\ref{fig:SPP_parameter_misspecification}.  Note that we obtain a marginally accurate (non-parametric) estimator for the interaction kernel $\phi$ (Figure~\ref{subfig:Missspecified_paramters_kernel}), but both the environmental force (Figure~\ref{subfig:Missspecified_paramters_f}) and trajectories (Figure~\ref{subfig:Missspecified_paramters_trajectories}) are estimated poorly.  This therefore highlights the necessity of being confident in regards to prior knowledge of parametric forms in estimation, and demonstrates the utility of a fully non-parametric approach in many experimental scenarios, when the exact structure of forces is rarely available.

\subsubsection{Phototaxis model}
\label{subsubsec:results_phototaxis}

We next consider a model of phototaxis, describing cell motion guided by an external light stimulus. Phototaxis is a fundamental behavioral mechanism in many microorganisms, allowing individuals to bias their movement toward or
away from light~\cite{bhaya2001light}.  In this work, we adapt the particle-level model introduced by~\cite{ha2009particle}, in which the motion of each bacterium is also influenced by neighboring agents through a Cucker-Smale velocity alignment term (see also Section~\ref{subsubsec:MS_Cucker_Smale}).  Specifically, the model takes the general second-order form~\eqref{eq:model_selection_general_framework-2order}, where 
\begin{align}
    \begin{split}
    f(x, v) &= I_0 (U_{\infty} e_l -  v), \quad \phi^{A} (r) = \frac{1}{(1+r^2)^{\beta}}, \quad \text{and} \quad \phi^{E}(r) \equiv 0.
    \end{split}
    \label{eq:Levi_f_env_term_and_interaction_kernel_term}
\end{align}
Here $I_0$ describe the constant intensity of the light source, $U_{\infty}$ denotes the terminal speed of the particles, and $e_{l} \in \R^{2}$ denotes the normalized direction of the light source.  For numerical simulations we fix $I_0 = 1$, $U_{\infty} =0.1$, $e_l = (-1/\sqrt{2}, 1/\sqrt{2})$, and $\beta = 0.1$.  We emphasize that this example differs significantly from the SPP model in Section~\ref{subsubsec:results_SPP}, as the SPP systems combines a nonlinear environmental force with energy-based interactions, while the model of phototaxis~\eqref{eq:Levi_f_env_term_and_interaction_kernel_term} describes velocity alignment together with a directional environmental force.

\begin{figure}
    \centering
    \begin{subfigure}{0.48\textwidth}
      \centering
      \includegraphics[width=\textwidth]{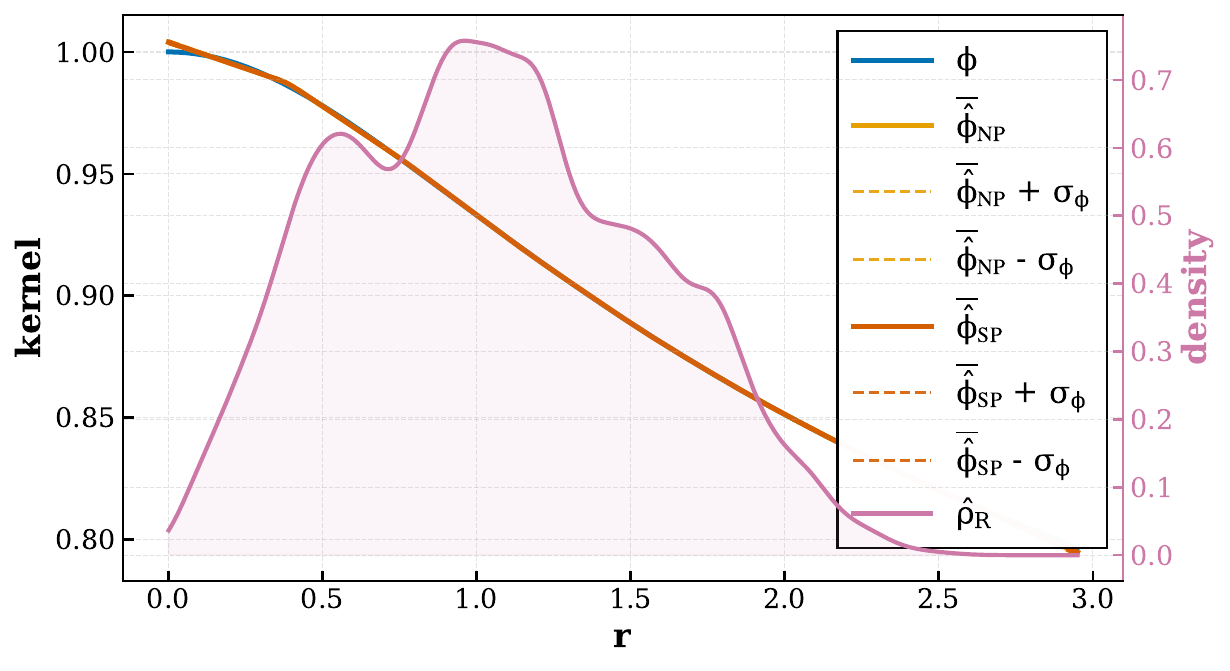}
      \caption{Estimation of interaction kernel $\phi$}
    \label{subfig:Phototaxis_interaction_kernel_true_vs_learned_with_KDE_10_trials}
    \end{subfigure}
    \begin{subfigure}{0.48\textwidth}
      \centering
      \includegraphics[width=\textwidth]{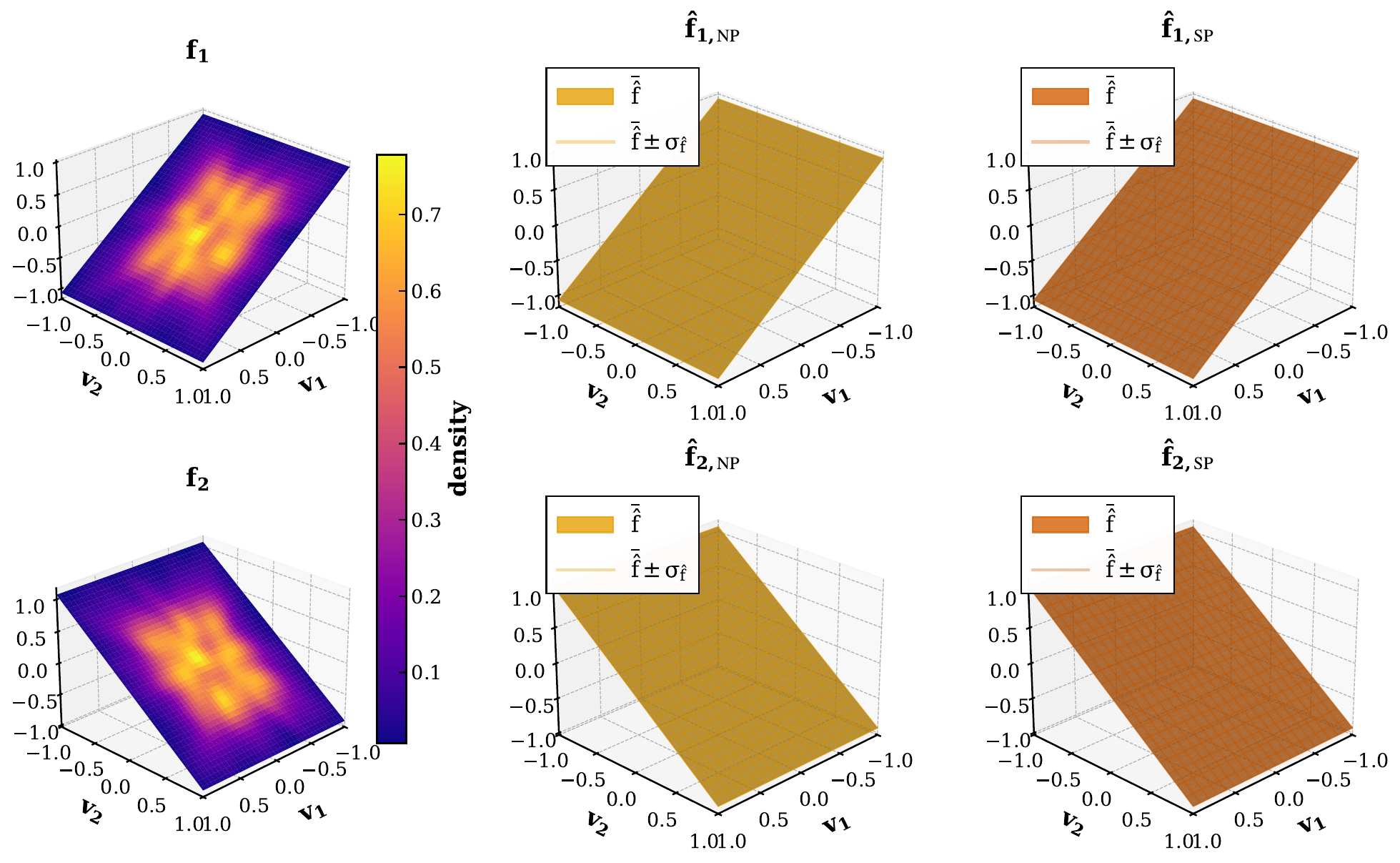}
      \caption{Estimation of environmental force $f$}
    \label{subfig:Phototaxis_intra_agent_force_true_vs_learned_with_KDE_10_trials}
    \end{subfigure}
    
    \caption{(Phototaxis model) Comparison of mechanistic feature recovery ability of the fully non-parametric (NP) and semi-parametric (SP) estimation algorithms for the phototaxis model~\eqref{eq:Levi_f_env_term_and_interaction_kernel_term}. In Subfigure~\ref{subfig:Phototaxis_interaction_kernel_true_vs_learned_with_KDE_10_trials} we visualize the true interaction kernel $\phi$ (blue) together with the mean recovered kernels $\bar{\hat{\phi}}$ obtained via the NP (yellow) and P (orange) approaches. Subfigure~\ref{subfig:Phototaxis_intra_agent_force_true_vs_learned_with_KDE_10_trials} provides the corresponding true environmental force $f$ and its recovered mean estimates $\bar{\hat{f}}$.  As $f: \R^{2} \to \R^{2}$, the top and bottom rows correspond to first and second components of $f$, respectively.  Within each row, the true force (left column), the fully non-parametric mean estimate (middle column), and semi-parametric mean estimate (right column) are shown, together with one standard deviation, which are obtained over $T_r=10$ independent learning trials.  The background density in Subfigure~\ref{subfig:Phototaxis_interaction_kernel_true_vs_learned_with_KDE_10_trials} (right vertical axis) corresponds to the empirical pairwise distance distribution $\hat{\rho}_R$, while the density visualized in Subfigure~\ref{subfig:Phototaxis_intra_agent_force_true_vs_learned_with_KDE_10_trials} (left column) corresponds to the empirical velocity distribution $\hat{\rho}_V$ induced by the training data. The corresponding relative feature recovery errors are reported in Table~\ref{tab:Phototaxis_feature_comparison}. The training data consists of $M=20$ trajectories with $N=10$ agents in dimension $d=2$, observed over $[0,T]=[0,0.5]$ with $L=51$ time points.}

\label{fig:Phototaxis_interaction_kernel_and_fenv_P_and_NP}
\end{figure}

Inferred mechanisms are plotted in Figure~\ref{fig:Phototaxis_interaction_kernel_and_fenv_P_and_NP}.  We observe a slight visible deviation between the true interaction kernel $\phi$ and the estimated kernels $\hat{\phi}$ near the boundaries of the learning interval in both approaches in Figure~\ref{subfig:Phototaxis_interaction_kernel_true_vs_learned_with_KDE_10_trials}, which is consistent with the previous observations for both the SPP and Kuramoto systems, where estimation accuracy deteriorates in regions that are poorly sampled by the trajectories, as measured by the empirical distribution $\hat{\rho}_{R}$.   In regards to the environmental force term $f$, both methods capture the correct direction of the light source, as well as the overall magnitude of influence (see Figure~\ref{subfig:Phototaxis_intra_agent_force_true_vs_learned_with_KDE_10_trials}). Table~\ref{tab:Phototaxis_feature_comparison} quantifies the estimation of mechanisms, where we note that the semi-parametric formulation achieves approximately one order of magnitude smaller relative environmental force error $E_f^{\mathrm{rel}}$ than that obtained for the non-parametric formulation. This highlights the benefit of incorporating the correct parametric structure into inference when such prior knowledge is available.

\begin{figure}
    \centering
    \begin{subfigure}{\textwidth}
        \centering
        \includegraphics[width=\linewidth]{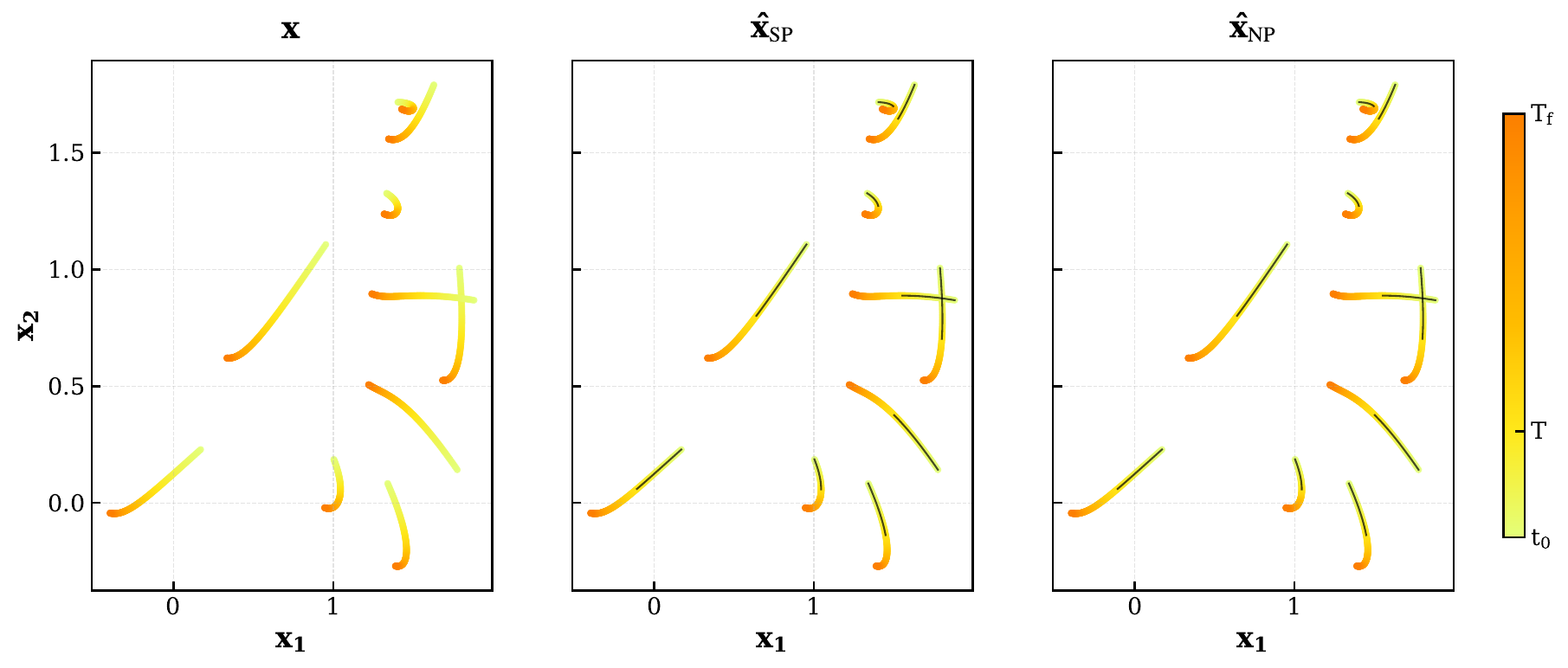}
        \caption{$M_{\mathrm{train}}$ training data utilized to obtain estimators}
        \label{subfig:Phototaxis_Traj_plot_P_NP_M_train}
    \end{subfigure}
    \vspace{0.1cm}


    \begin{subfigure}{\textwidth}
        \centering
        \includegraphics[width=\linewidth]{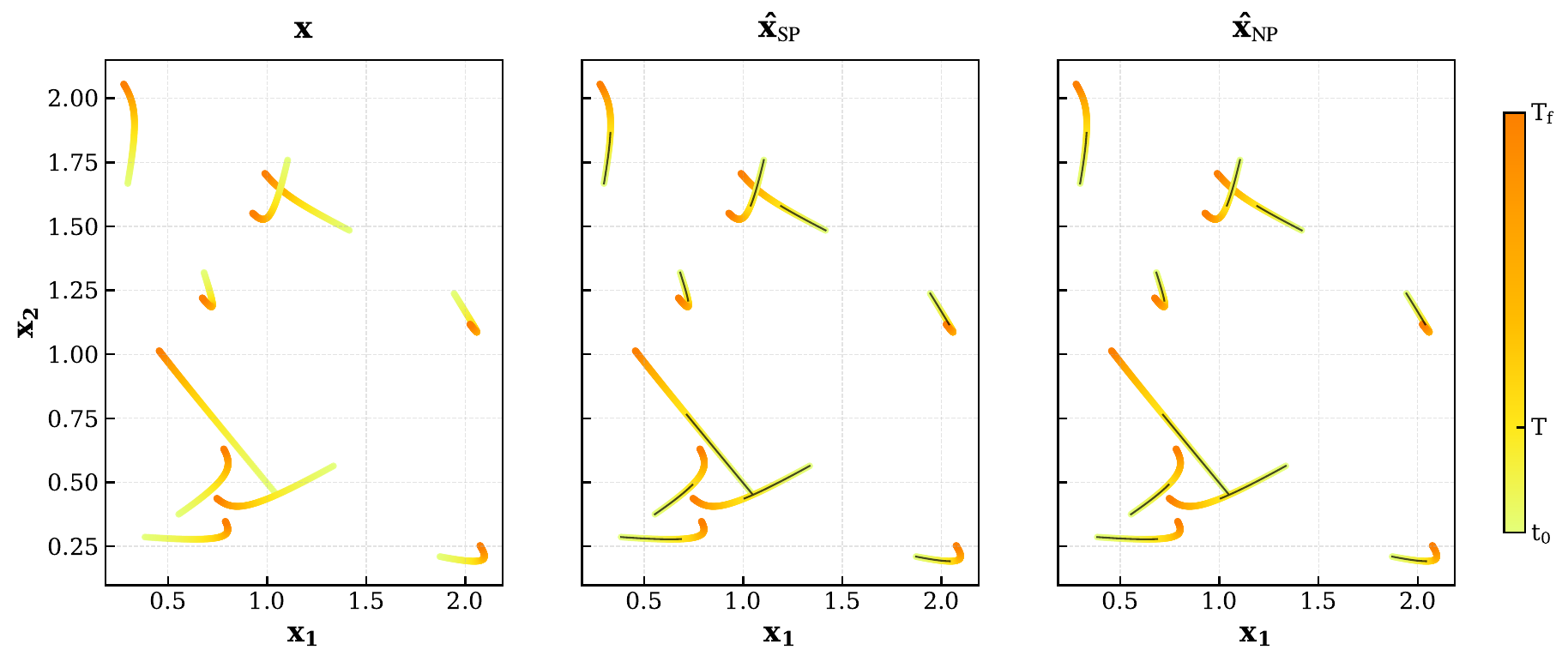}
        \caption{$M_{\mathrm{test}}$ test data utilized for evaluation}
        \label{subfig:Phototaxis_Traj_plot_P_NP_M_test}
    \end{subfigure}

    \caption{(Phototaxis model) Comparison of true trajectories $x$ (left column) and the estimated trajectories $\hat{x}$ for both the semi-parametric (middle column) and full non-parametric (right column) methods in the phase plane. For trajectory errors, see Table~\ref{tab:Phototaxis_trajectory_comparison}. The top row displays trajectories initialized from the $M_{\mathrm{train}}$ training initial-conditions, whereas the bottom row displays trajectories generated from $M_{\mathrm{test}}$ previously unseen testing initial conditions.  Note that trajectories are plotted over the full time horizon $[0, T_f]$, with the color gradient representing the temporal evolution of the dynamics, from yellow at $t=0$ to orange at $t=T_f$.  The black segment indicates the training interval $[0,T]$, which is utilized for model fitting; recall that no fitting is performed in the testing data in Figure~\ref{subfig:SPP_Traj_plot_P_NP_M_test}.  Here $[0,T]=[0,0.5]$, with testing evaluated beyond the fitting horizon $[0.0,0.5]$ until $T_f=2.0$ to assess both reconstruction and prediction in Figure~\ref{subfig:Phototaxis_Traj_plot_P_NP_M_test}. Both method variants closely reproduce the true dynamics on the training trajectories and generalize well to unseen initial conditions.  For additional details on algorithm parameters, we refer the reader to the caption of Figure~\ref{fig:Phototaxis_interaction_kernel_and_fenv_P_and_NP}.}
    \label{fig:Phototaxis_True_vs_Learned_Trajectories}

\end{figure}

Despite the differences observed in feature recovery, both formulations produce highly accurate trajectory reconstructions and predictions (Figure~\ref{fig:Phototaxis_True_vs_Learned_Trajectories}). On both the training and testing data, the reconstruction errors on the fitting interval and the prediction errors on the extrapolation interval remain on the order of $10^{-6}$ and $10^{-5}$.  Precise value are provided in Table~\ref{tab:Phototaxis_trajectory_comparison}.

\begin{table}
\centering
\begin{tabular}{lcc}
\toprule
Metric
& Non-parametric
& Semi-Parametric \\
\midrule

$E^{\mathrm{rel}}_{\mathrm{res}}$
& \num{6.72662852e-05} $\pm$ \num{6.39887610e-06}
& \num{7.29130363e-05} $\pm$ \num{7.33437682e-06} \\

$E^{\mathrm{rel}}_{\phi}$
& \num{1.98551211e-04} $\pm$ \num{2.29285528e-05}
& \num{1.98097868e-04} $\pm$ \num{2.28884560e-05} \\

$E^{\mathrm{rel}}_{f}$
& \num{5.53344049e-05} $\pm$ \num{8.75494029e-06}
& \num{3.73835494e-06} $\pm$ \num{2.41821840e-06} \\

\bottomrule
\end{tabular}
\caption{(Phototaxis model) Comparison of relative residual error~\eqref{eq:relative_residue_error} and feature recovery metrics~\eqref{eq:feature_functions_relative_errors} for the fully non-parametric and semi-parametric estimation algorithms for the phototaxis model~\eqref{eq:Levi_f_env_term_and_interaction_kernel_term}. The table reports the mean error $\pm$ one standard deviation of these quantities computed over $T_r=10$ independent learning trials.  For additional details on algorithm parameters, we refer the reader to the caption of Figure~\ref{fig:Phototaxis_interaction_kernel_and_fenv_P_and_NP}.}
\label{tab:Phototaxis_feature_comparison}
\end{table}

\begin{table}
\centering
\begin{tabular}{llcc}
\toprule
Data set & Statistic
& Non-parametric
& Semi-Parametric \\
\midrule

\multirow{4}{*}{$M_{\mathrm{train}}$}
& $\bar E_{\mathrm{recon}}$
& \num{4.02755251e-06} $\pm$ \num{6.54923071e-07}
& \num{4.49642852e-06} $\pm$ \num{7.36467099e-07} \\

& $\sigma_{E_{\mathrm{recon}}}$
& \num{1.00925916e-06} $\pm$ \num{3.05519048e-07}
& \num{1.26886962e-06} $\pm$ \num{3.89960897e-07} \\

& $\bar E_{\mathrm{pred}}$
& \num{1.36729468e-05} $\pm$ \num{2.26202000e-06}
& \num{1.17853684e-05} $\pm$ \num{1.56036745e-06} \\

& $\sigma_{E_{\mathrm{pred}}}$
& \num{2.76882280e-06} $\pm$ \num{7.38893698e-07}
& \num{3.21052055e-06} $\pm$ \num{6.86216786e-07} \\
\midrule

\multirow{4}{*}{$M_{\mathrm{test}}$}
& $\bar E_{\mathrm{recon}}$
& \num{5.06780784e-06} $\pm$ \num{6.95397944e-07}
& \num{4.52369717e-06} $\pm$ \num{5.19349962e-07} \\

& $\sigma_{E_{\mathrm{recon}}}$
& \num{1.16361256e-06} $\pm$ \num{3.30118782e-07}
& \num{1.01252441e-06} $\pm$ \num{3.03858117e-07} \\

& $\bar E_{\mathrm{pred}}$
& \num{1.54380095e-05} $\pm$ \num{2.75489736e-06}
& \num{1.21902971e-05} $\pm$ \num{1.61879705e-06} \\

& $\sigma_{E_{\mathrm{pred}}}$
& \num{2.64305127e-06} $\pm$ \num{6.20707166e-07}
& \num{2.59318984e-06} $\pm$ \num{4.52617056e-07} \\
\bottomrule
\end{tabular}
\caption{(Phototaxis model)  Comparison of trajectory-level errors for the fully non-parametric and semi-parametric estimation algorithms for the phototaxis model~\eqref{eq:Levi_f_env_term_and_interaction_kernel_term}.  For each learning trial, the trajectory statistics
$\bar E_{\mathrm{recon}}$,
$\bar E_{\mathrm{pred}}$,
$\sigma_{E_{\mathrm{recon}}}$,
and
$\sigma_{E_{\mathrm{pred}}}$
are computed according to~\eqref{eq:mean_trajectory_errors}~-~\eqref{eq:std_trajectory_error}.
The table reports the mean $\pm$ one standard deviation of these quantities computed over $T_r=10$ independent learning trials.
The first section corresponds to trajectories generated from $M_{\mathrm{train}}$ training initial conditions, while the second section corresponds to $M_{\mathrm{test}}$ unseen initial conditions sampled from $\mu_0$.  Note that across all trajectory measures, error estimates are both small and similar for both methods.  For additional details on algorithm parameters, we refer the reader to the caption of Figure~\ref{fig:Phototaxis_interaction_kernel_and_fenv_P_and_NP}.
}
\label{tab:Phototaxis_trajectory_comparison}
\end{table}

\subsection{Scalability}
\label{subsec: results_scalability}

For the remainder of the manuscript, we evaluate the non-parametric algorithm in several different scenarios, as described in Section~\ref{subsec:reliability_assessment}. In practice, prior knowledge of the exact structure of the environmental force $f$ is rarely available, making the fully non-parametric formulation more broadly applicable. The results from Section~\ref{subsec:results_nonparametric_vs_parametric_learning} have established the reliability and consistency of the fully non-parametric algorithm under repeated trials, so that for the analyses presented in this section, models are trained utilizing one training sample (i.e. $T_{r}=1$).  We will later investigate the variability of obtained estimators as a function of the amount of training data in Section~\ref{subsec:results_sample_complexity}, where repeated trials are necessary.

\begin{figure}
    \centering
    \begin{subfigure}{0.33\textwidth}
        \centering
        \includegraphics[width=\textwidth]{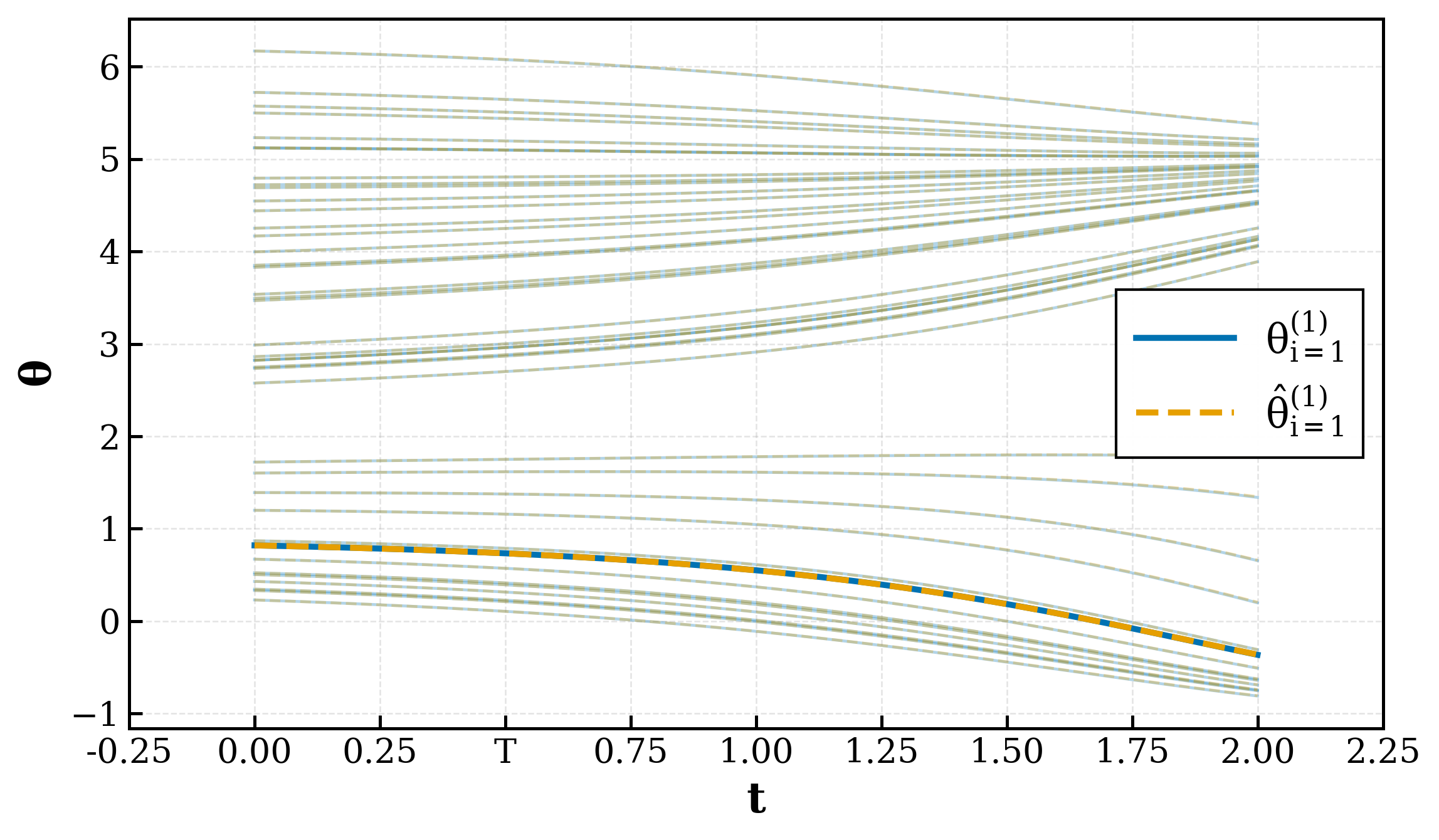}
        \caption{Kuramoto model trajectories}
        \label{subfig:Kuramoto_scalability}
    \end{subfigure}
    \hfill
    \begin{subfigure}{0.33\textwidth}
        \centering
        \includegraphics[width=\textwidth]{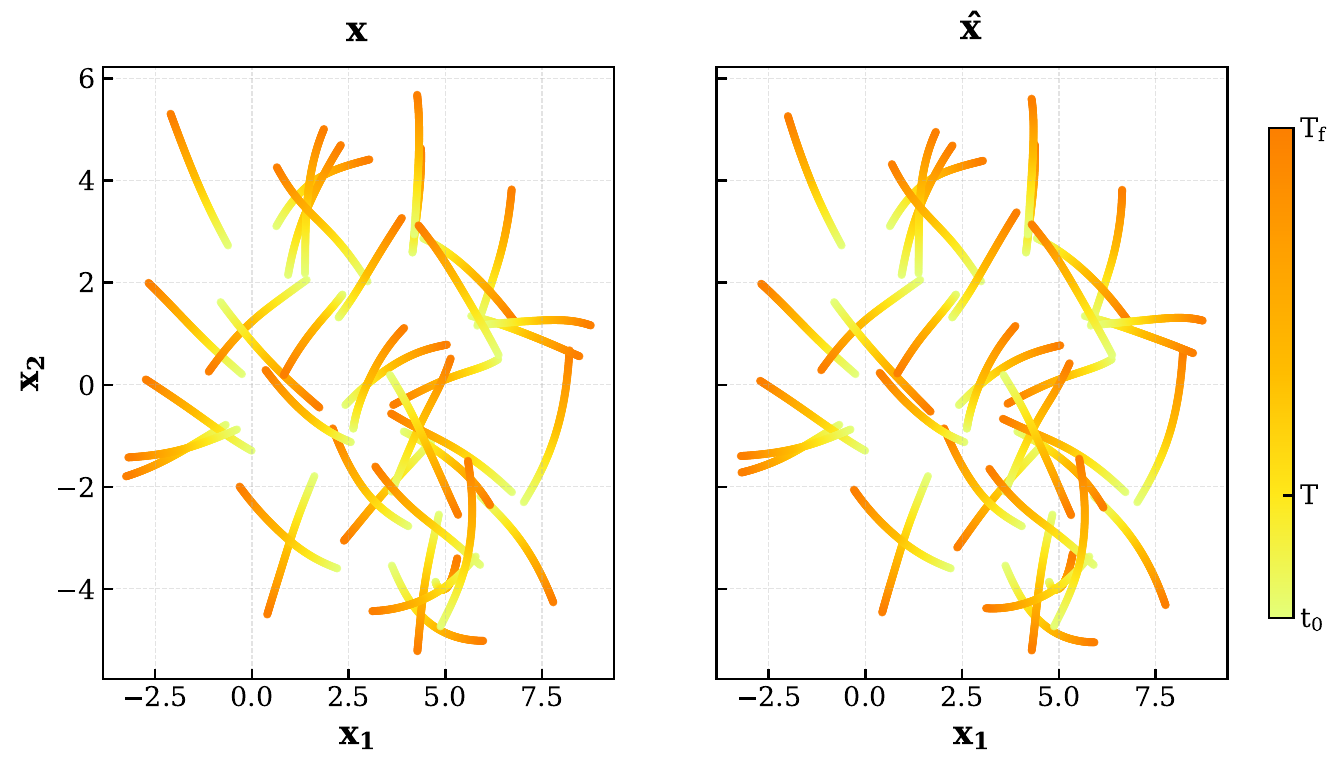}
        \caption{SPP model trajectories}
        \label{subfig:SPP_scalability}
    \end{subfigure}
    \hfill
    \begin{subfigure}{0.33\textwidth}
        \centering
        \includegraphics[width=\textwidth]{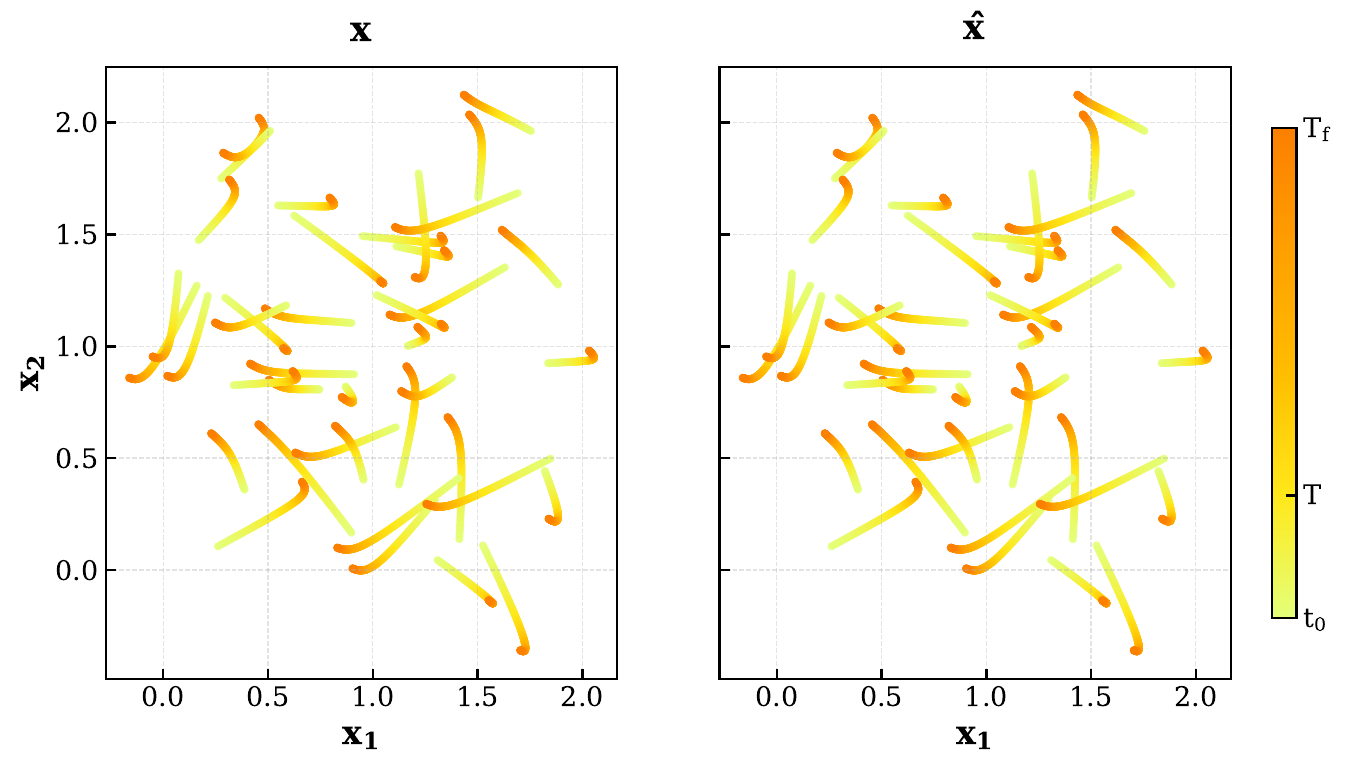}
        \caption{Phototaxis model trajectories}
        \label{subfig:Phototaxis_scalability}
    \end{subfigure}

    \caption{
    (Scaling number of agents) The true and predicted trajectories for the Kuramoto (Figure~\ref{subfig:Kuramoto_scalability}), SPP (Figure~\ref{subfig:SPP_scalability}), and phototaxis (Figure~\ref{subfig:Phototaxis_scalability}) systems obtained via the fully non-parametric method, when obtaining by the estimators by training on a smaller number of agents.  Trajectory predictions are provided on $N_{\mathrm{new}} = 4N $ agents, when $N=10$ agents were utilized in the inference. Trajectory errors are reported in Table~\ref{tab:scaling_trajectory_comparison_np_vs_p} in the semi-parametric column. Further information regarding learning and model parameters may be found in Appendix~\ref{app:sec:tables_of_parameters}.
    }
    \label{fig:Scaling_for_all_systems_4N}
\end{figure}

\begin{figure}
    \centering
    \includegraphics[width=\linewidth]{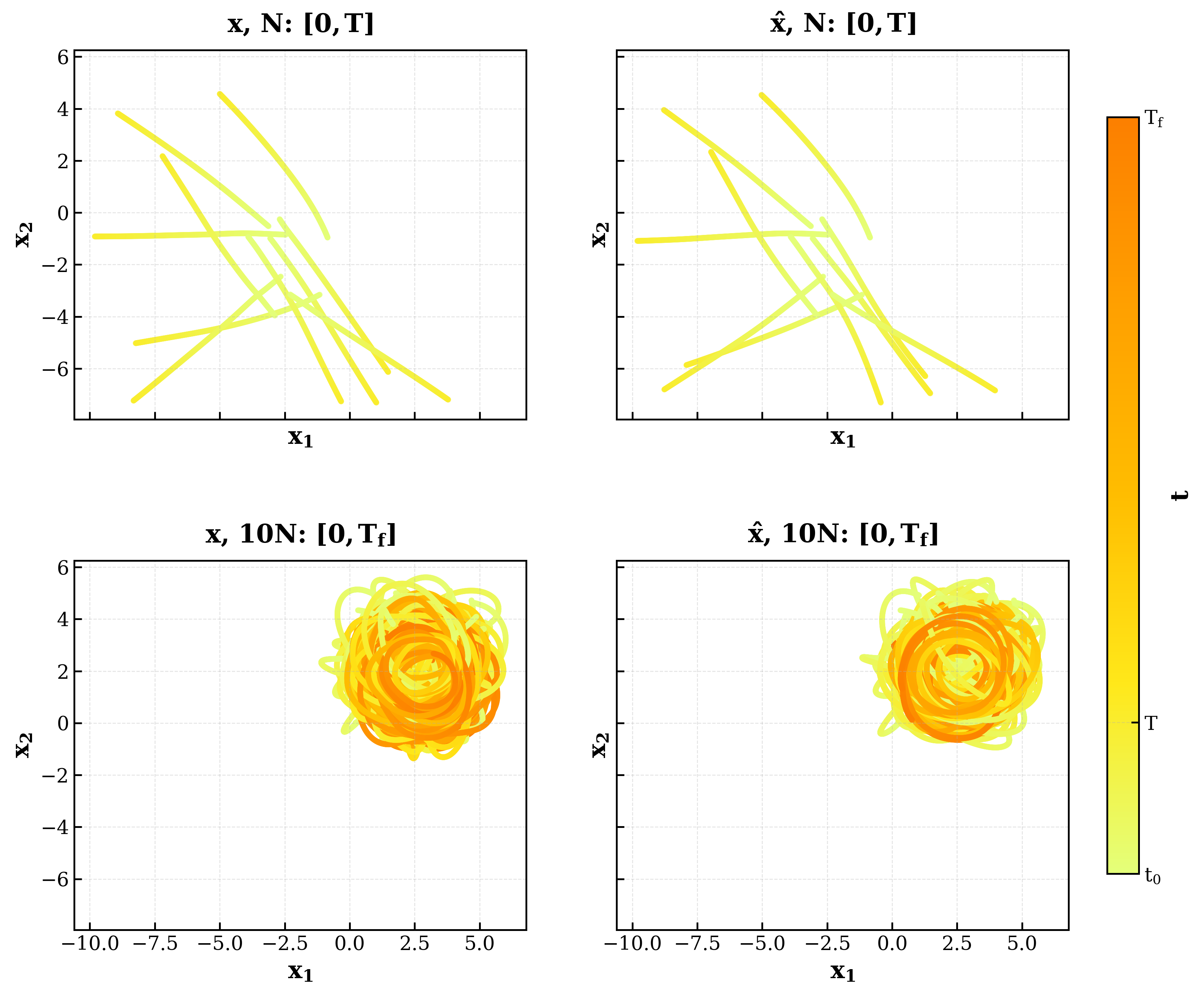}
    \caption{
        (Scaling number of agents: SPP milling model) The true and predicted trajectories for the SPP milling system obtained using the learned model, with milling parameters $C_r=1, l_r=0.5, C_a=0.5, l_a=2, \alpha=1.6$, and $\beta=0.5$. The top row shows the trajectory predictions over the training time interval $[0,T]=[0,4]$ for $N=10$ agents. The bottom row shows the trajectory predictions for a scaled system with $N_{\mathrm{new}}=100$ agents, evolved over a time horizon five times longer than the training interval, with $T_f=20$. Further information regarding the learning parameters, system parameters, and initial conditions may be found in Appendix~\ref{app:sec:tables_of_parameters}.
        }
        \label{fig:scaling_spp_swarm}

\end{figure}

To assess whether the learned mechanisms generalize across agent size, we train the algorithm using systems containing $N$ agents and evaluate the learned models on a testing data set consisting of $4N$ agents of size $M_{\mathrm{test}}^{4N}$ as described in Section~\ref{subsubsec:scalability} (for simplicity we fix $M_{\mathrm{test}}^{4N} = M_{\mathrm{test}}^{N} = M_{\mathrm{test}}$).  Figure~\ref{fig:Scaling_for_all_systems_4N} provides a qualitative comparison of the resulting trajectories for the fully non-parametric formulation, while Table~\ref{tab:scaling_trajectory_comparison_np_vs_p} summarizes the corresponding trajectory-based metrics for both the semi-parametric and fully non-parametric formulations for the Kuramoto (Figure~\ref{subfig:Kuramoto_scalability}), SPP (Figure~\ref{subfig:SPP_scalability}), and phototaxis (Figure~\ref{subfig:Phototaxis_scalability}) models.  Across all systems considered, the learned interaction mechanisms generalize consistently to larger populations, and both formulations successfully reproduce the qualitative collective behavior observed in the reference trajectories. By examining Table~\ref{tab:scaling_trajectory_comparison_np_vs_p}, we observe that the semi-parametric formulation generally yields slightly smaller mean prediction errors when compared to the fully non-parametric formulation, but the differences are modest and generally of the same order. These results thus suggest that the dominant interaction mechanisms are accurately captured by both approaches and transfer well across scales in system size. Figure~\ref{fig:scaling_spp_swarm} further demonstrates the ability of the algorithm to predict over a time horizon five times longer than the training interval while simultaneously scaling the number of agents by a factor of ten. Although the training trajectories contain relatively few agents and do not exhibit the collective milling behavior within the observed training interval, the learned governing mechanisms are sufficiently accurate to reproduce and preserve this emergent behavior when applied to the larger system and evolved into the future. The predicted agent-level trajectories may not agree exactly with the reference trajectories on an agent-by-agent basis; nevertheless, the learned model preserves the dominant macroscopic collective pattern. This is a significant strength of the proposed approach, indicating that sufficiently broad sampling of the position and velocity spaces during training enables the algorithm to recover the underlying governing laws, even when the associated emergent phenomenon is not directly observed in the training data.

\begin{table}
\centering
\begin{tabular}{llcc}
\toprule
Data set & Statistic
& Non-parametric
& Semi-parametric \\
\midrule
\multirow{4}{*}{Kuramoto}
& $\bar E_{\mathrm{recon}}$
& \num{8.63922609e-04} $\pm$ \num{1.43606753e-04}
& \num{7.93340660e-04} $\pm$ \num{9.07255942e-05} \\

& $\sigma_{E_{\mathrm{recon}}}$
& \num{2.03398640e-04} $\pm$ \num{2.37826991e-05}
& \num{2.02933207e-04} $\pm$ \num{2.68856676e-05} \\

& $\bar E_{\mathrm{pred}}$
& \num{3.36005773e-03} $\pm$ \num{1.26060389e-03}
& \num{2.72169499e-03} $\pm$ \num{9.89175778e-04} \\

& $\sigma_{E_{\mathrm{pred}}}$
& \num{1.50295418e-03} $\pm$ \num{7.71456134e-04}
& \num{1.37229820e-03} $\pm$ \num{6.66732558e-04}  \\
\midrule
\multirow{4}{*}{SPP}
& $\bar E_{\mathrm{recon}}$
& \num{7.88646353e-03} $\pm$ \num{2.47532939e-04}
& \num{0.00688492144226794} $\pm$ \num{0.0002537159047795362} \\

& $\sigma_{E_{\mathrm{recon}}}$
& \num{1.73770205e-03} $\pm$ \num{4.80451638e-04}
& \num{0.0021131679127560908} $\pm$ \num{0.0005436377108619356} \\

& $\bar E_{\mathrm{pred}}$
& \num{5.16746373e-02} $\pm$ \num{3.88665070e-03}
& \num{0.04379378751290812} $\pm$ \num{0.0022121618209391255} \\

& $\sigma_{E_{\mathrm{pred}}}$
& \num{1.66348258e-02} $\pm$ \num{3.38284586e-03}
& \num{0.012995345087749823} $\pm$ \num{0.002150118775131265}  \\
\midrule

\multirow{4}{*}{Phototaxis}
& $\bar E_{\mathrm{recon}}$
& \num{3.24022351e-06} $\pm$ \num{5.18834078e-07}
& \num{2.48279267e-06} $\pm$ \num{3.97820770e-07} \\

& $\sigma_{E_{\mathrm{recon}}}$
& \num{3.66001179e-07} $\pm$ \num{1.28189976e-07}
& \num{2.35383958e-07} $\pm$ \num{6.02447296e-08} \\

& $\bar E_{\mathrm{pred}}$
& \num{1.18822188e-05} $\pm$ \num{2.07581743e-06}
& \num{7.50772867e-06} $\pm$ \num{6.92547719e-07} \\

& $\sigma_{E_{\mathrm{pred}}}$
& \num{1.75167258e-06} $\pm$ \num{9.41635418e-07}
& \num{1.67171123e-06} $\pm$ \num{1.22320593e-06}  \\
\bottomrule
\end{tabular}
\caption{
(Scaling number of agents) Comparison of trajectory level errors for the fully non-parametric and semi-parametric estimation algorithms for the Kuramoto~\eqref{eq:Kuramoto-ODE}, SPP~\eqref{SPP_model_equaions}, and phototaxis~\eqref{eq:Levi_f_env_term_and_interaction_kernel_term} models.  For each learning trial, the trajectory statistics
$\bar E_{\mathrm{recon}}$,
$\bar E_{\mathrm{pred}}$,
$\sigma_{E_{\mathrm{recon}}}$,
and
$\sigma_{E_{\mathrm{pred}}}$
are computed according to~\eqref{eq:mean_trajectory_errors}~-~\eqref{eq:std_trajectory_error}.
The table reports the mean $\pm$ one standard deviation of these quantities computed $T_r=10$ independent learning trials.  Predicted trajectories are made with $M_{\mathrm{test}}^{4N}$ untrained initial conditions generated for $4N$ agents.  In numerical simulations, $M_{\mathrm{test}}^{4N}=M_{\mathrm{test}}$.  Further information regarding learning and model parameters may be found in Appendix~\ref{app:sec:tables_of_parameters}.
}
\label{tab:scaling_trajectory_comparison_np_vs_p}
\end{table}

\subsection{Dependence on training data}
\label{subsec:results_sample_complexity}
For the three systems previously considered (Kuramoto, SPP, and phototaxis), we are interested in the ability of the fully non-parametric framework to recover features as either the number of independent trajectory replicates $M$ or the number of temporal observations $L$ increases.  This is experimentally relevant, since in applications one would like to determine how many independent replicates are required, and how frequently each trajectory must be sampled in time, to accurately recover the governing mechanistic features. While one expects accuracy to improve as either $M$ or $L$ increases, the relative importance of these two sources of data is not obvious a priori. We therefore investigate this dependence quantitatively, as described in Section~\ref{subsubsec:sample_complexity_experiments}.

\begin{figure}
    \centering

    \begin{subfigure}{\textwidth}
        \centering
        \includegraphics[width=\textwidth]{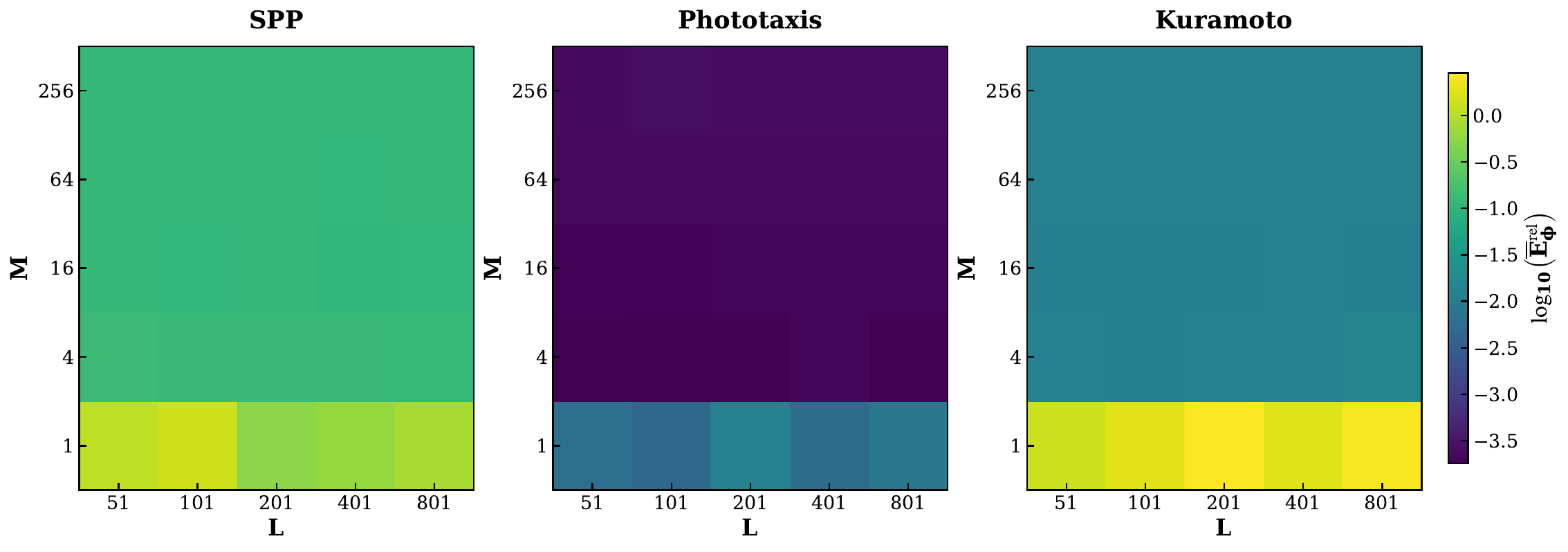}
        \caption{Mean error for interaction kernel $\phi$}
        \label{subfig:Kuramoto_E_phi_relative_ML_sweep}
    \end{subfigure}
    
    \begin{subfigure}{\textwidth}
        \centering
        \includegraphics[width=\textwidth]{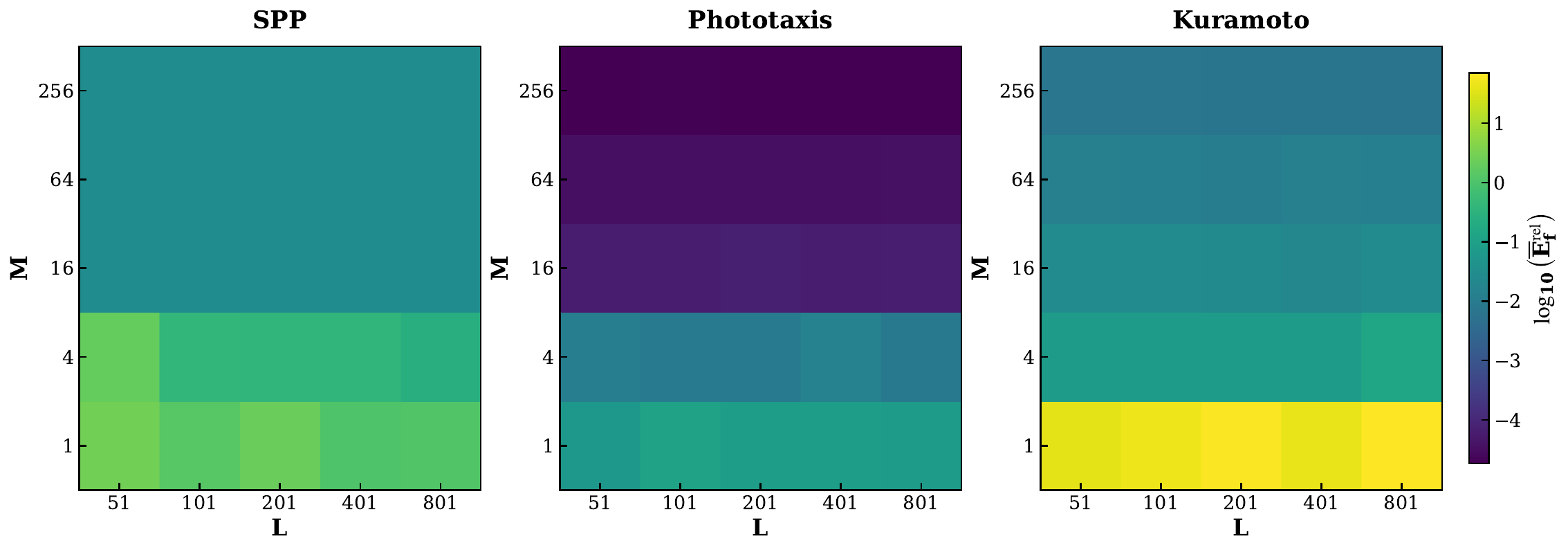}
        \caption{Mean error for environmental force $f$}
        \label{subfig:Kuramoto_E_f_relative_ML_sweep}
    \end{subfigure}

    \caption{
    Mean mechanistic relative feature recovery errors (log scale) as functions of temporal resolution $L$ and independent replicate size $M$ computed over $T_r=10$ independent learning trials. The relatively large errors observed for $M=1$ replicate, even when increasing the number of time samples $L$, suggests that a single trajectory does not contain sufficient information to accurately recover the underlying dynamics.  The above figure, for both interaction kernel $\phi$ (Figure~\ref{subfig:Kuramoto_E_phi_relative_ML_sweep}) and environmental force $f$ (Figure~\ref{subfig:Kuramoto_E_f_relative_ML_sweep}), suggests that the number of replicates $M$ is the dominant sampling parameter which controls the accuracy of the estimation algorithm.  Further information regarding learning and model parameters may be found in Appendix~\ref{app:sec:tables_of_parameters}, as well as in the figure captions of Section~\ref{subsec:results_nonparametric_vs_parametric_learning}. 
    }
    \label{fig:sample_complexity}
\end{figure}

Figure~\ref{fig:sample_complexity} demonstrates that increasing either the number of trajectories $M$ or the temporal resolution $L$ leads to more accurate recovery of the underlying interaction mechanisms.  Note that accuracy is measured with respect to the mechanistic feature recovery metrics~\eqref{eq:feature_functions_relative_errors}, and is averaged over $T_{r}=10$ learning trials; mean errors are reported with respect to these $T_{r}$ trials.  Interestingly, the results also indicate that increasing the temporal resolution $L$ alone cannot fully compensate for a lack of replicate diversity $M$. In particular, when only a single replicate ($M=1$) exists, increasing $L$ does not appear to have a significant effect on the ability to recover $\phi$ (Figure~\ref{subfig:Kuramoto_E_phi_relative_ML_sweep}) and $f$ (Figure~\ref{subfig:Kuramoto_E_f_relative_ML_sweep}), suggesting that a single trajectory does not contain sufficient information to fully identify the interaction mechanisms governing the systems.  Indeed, while increasing $L$ samples the same region of state space, increasing $M$ introduces new initial conditions and therefore explores a larger portion of state space, which is thus leveraged in the estimation algorithm. Consequently, trajectory diversity has a critical role in accurately recovering both the interaction kernels and environmental forces, with temporal resolution less crucial with regards to inference.  Recall also that the number of agents $N$ is generally large, so that each new replicate introduces a new collection of $N$ initial agent states which are utilized in estimation.

We note that if one is estimating (for example) velocities $\Vml$ from states $\Xml$ as discussed in Section~\ref{subsubsec:trajectory_data}, increasing the temporal resolution $L$ may improve accuracy in numerically differentiating $\Xml$.  As we are primarily interested in understanding the errors of the variational inference algorithms, we do not attempt to quantify the accumulation of such errors in this manuscript, where we have assumed exact velocity (and acceleration, for second--order systems) data.  One exception to this is presented in Section~\ref{subsec:results_noise_robustness}, where we consider the case of reconstruction when utilizing imperfect trajectory data (see Figure~\ref{fig:Kuramoto_noise_smoother_recovery}). 

\subsection{Robustness to noise in observation data}
\label{subsec:results_noise_robustness}
As discussed in Section~\ref{subsec:noise_robustness}, trajectory data arising in applications is typically subject to experimental noise, and we are interested in measuring the ability of the fully non-parametric variation approach to successfully infer mechanisms in the presence such noise.  We begin by assuming that the observed trajectory, velocity, and acceleration data are independency perturbed at each time sample $t_{\ell}$ by a uniform random variable of relative size $\zeta$ (see equations~\eqref{eq:noisy_data} and~\eqref{eq:noisy_dist}); we subsequently vary $\zeta$ and quantify the inference.

\begin{figure}
    \centering

    \begin{subfigure}{0.33\textwidth}
        \centering
        \includegraphics[width=\textwidth]{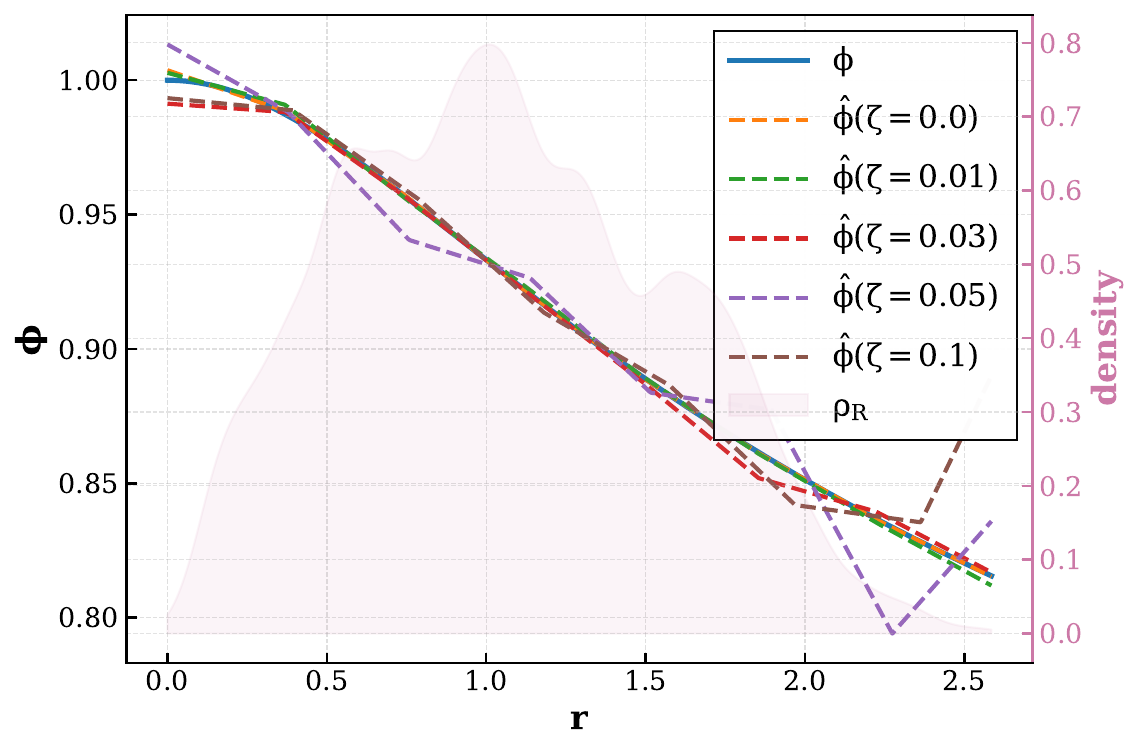}
        \caption{Interaction kernel $\phi$}
        \label{subfig:Phototaxis_phi_noise}
    \end{subfigure}
    \hfill
    \begin{subfigure}{0.33\textwidth}
        \centering
        \includegraphics[width=\textwidth]{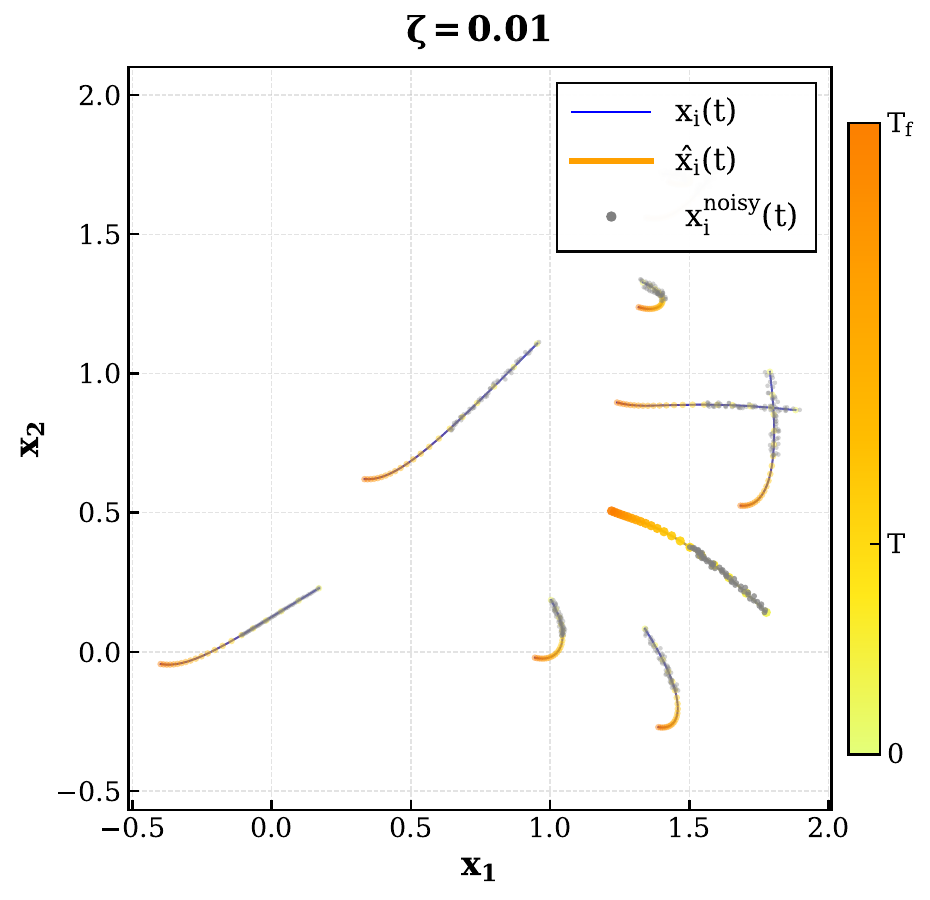}
        \caption{Trajectory data ($\zeta=0.01$) }
        \label{subfig:Phototaxis_x_zeta_001_noise}
    \end{subfigure}
    \hfill
    \begin{subfigure}{0.33\textwidth}
        \centering
        \includegraphics[width=\textwidth]{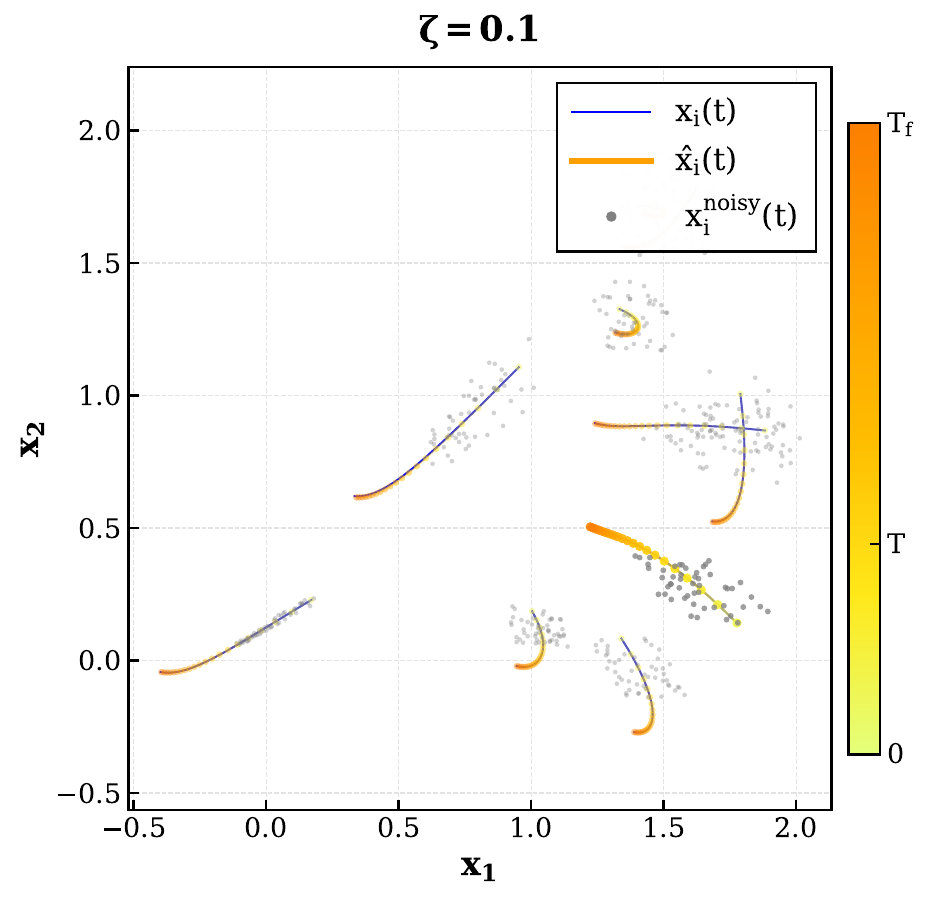}
        \caption{Trajectory data ($\zeta=0.1$)}
        \label{subfig:Phototaxis_x_zeta_010_noise}
    \end{subfigure}
    
\vspace{0.2em}
    \begin{subfigure}{\textwidth}
        \centering
        \includegraphics[width=\textwidth]{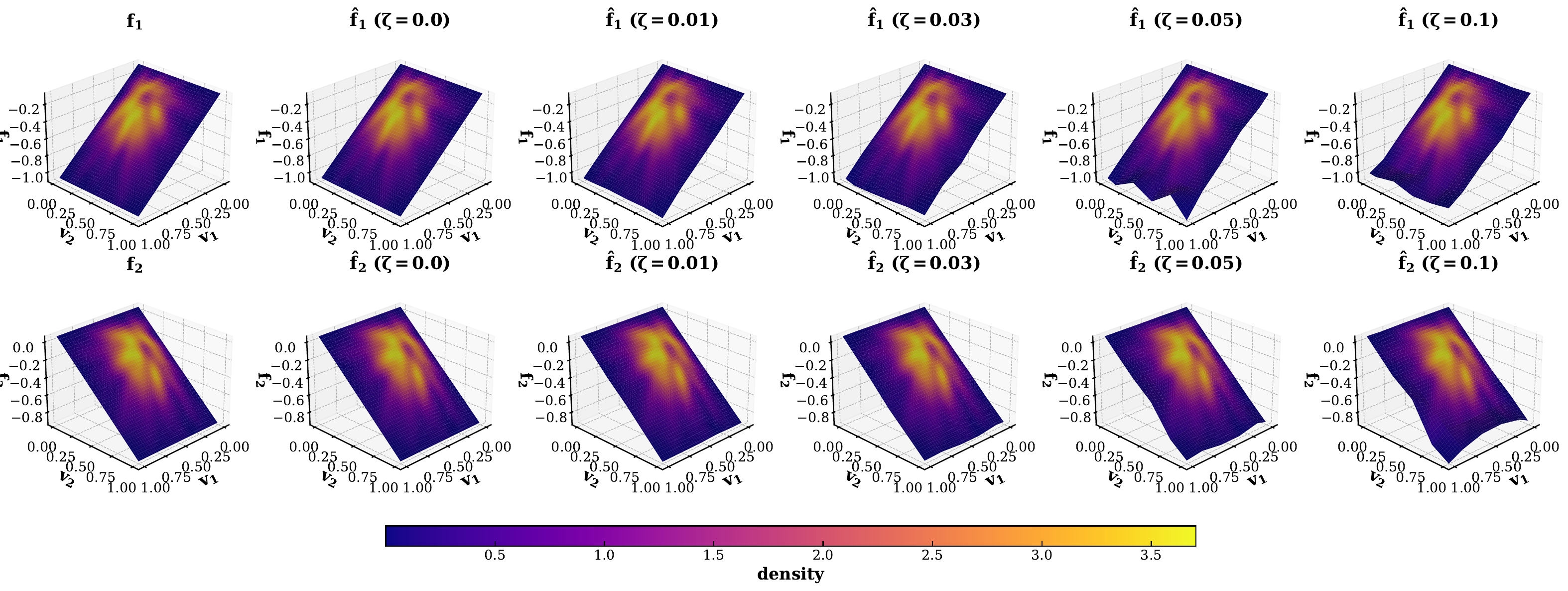}
        \caption{Environmental force $f$}
        \label{subfig:Phototaxis_f_noise}
    \end{subfigure}
    
\caption{(Noise robustness, phototaxis model) Mechanistic feature recovery under multiplicative observational noise with noise levels $\zeta$. Subfigure~\ref{subfig:Phototaxis_phi_noise} plots the estimated interaction kernels $\phi$ for different noise levels, while Subfigure~\ref{subfig:Phototaxis_f_noise} plots the corresponding estimated environmental forces $f$. The estimated empirical measures $\hat{\rho}_R$ and $\hat{\rho}_V$ induced by the training data are also included in each subfigure. Subfigures~\ref{subfig:Phototaxis_x_zeta_001_noise} and~\ref{subfig:Phototaxis_x_zeta_010_noise} compare the true, noisy, and learned trajectories for $\zeta=0.01$ and $\zeta=0.1$, respectively.  True (blue), noisy (grey), and estimated (yellow to orange, varying by time) trajectory data are all plotted; recall that training occurs on the noisy (grey) data.  Training occurs on $[0,T]$, and testing (predictions) are provided on $(T,T_{f}]$.  We observe that as $\zeta$ increases, feature recovery gradually deteriorates, primarily in regions with limited trajectory data (small measures), while the learned trajectories remain close to the true trajectories from low to moderate noise levels.  Further information regarding learning and model parameters may be found in Appendix~\ref{app:sec:tables_of_parameters}, as well as in the figure captions of Section~\ref{subsec:results_nonparametric_vs_parametric_learning}. 
}
\label{fig:photaxis_noise_recovery}
\end{figure}

Results for relative noise levels $\zeta \in [0,0.1]$ are provided in Figure~\ref{fig:photaxis_noise_recovery} for the model of phototaxis~\eqref{eq:Levi_f_env_term_and_interaction_kernel_term}, where we generally observe robustness of the algorithm with respect to observation noise. Figures~\ref{subfig:Phototaxis_phi_noise} and~\ref{subfig:Phototaxis_f_noise} illustrates feature recovery for $f$ and $\phi$, respectively, while trajectory predictions for noise levels of $\zeta=0.01$ (Figure~\ref{subfig:Phototaxis_x_zeta_001_noise}) and $\zeta=0.1$ (Figure~\ref{subfig:Phototaxis_x_zeta_010_noise}) are also provided.  We observe that as the noise level $\zeta$ increases, the learned interaction kernel and environmental force gradually deviate from their true values, with the largest deviation occurring primarily in regions where the empirical sampling measures $\hat{\rho}_{R}$ and $\hat{\rho}_{V}$ are small, i.e. in regions where the available trajectory information is limited. In contrast, regions that are well supported by the observational data remain largely unaffected by the perturbations.  Similar to previous numerical experiments, training is performed on the time interval $[0,T]$, while testing (prediction) is evaluated on $(T,T_{f}]$..  Note that even in the presence of $10\%$ relative noise levels (Figure~\ref{subfig:Phototaxis_x_zeta_010_noise}), the variational algorithm is still able to accurately predict trajectories well outside of the training interval.  Similar behavior is observed across all systems considered in Section~\ref{subsec:results_nonparametric_vs_parametric_learning}; the corresponding results are reported in Appendix~\ref{subsec:app:noise} in Figures~\ref{fig:app:noise_feature_recovery} and~\ref{fig:app:noise_trajectory_robustness}.

\begin{figure}
\centering

\begin{subfigure}{0.49\textwidth}
    \centering
    \includegraphics[width=\textwidth]{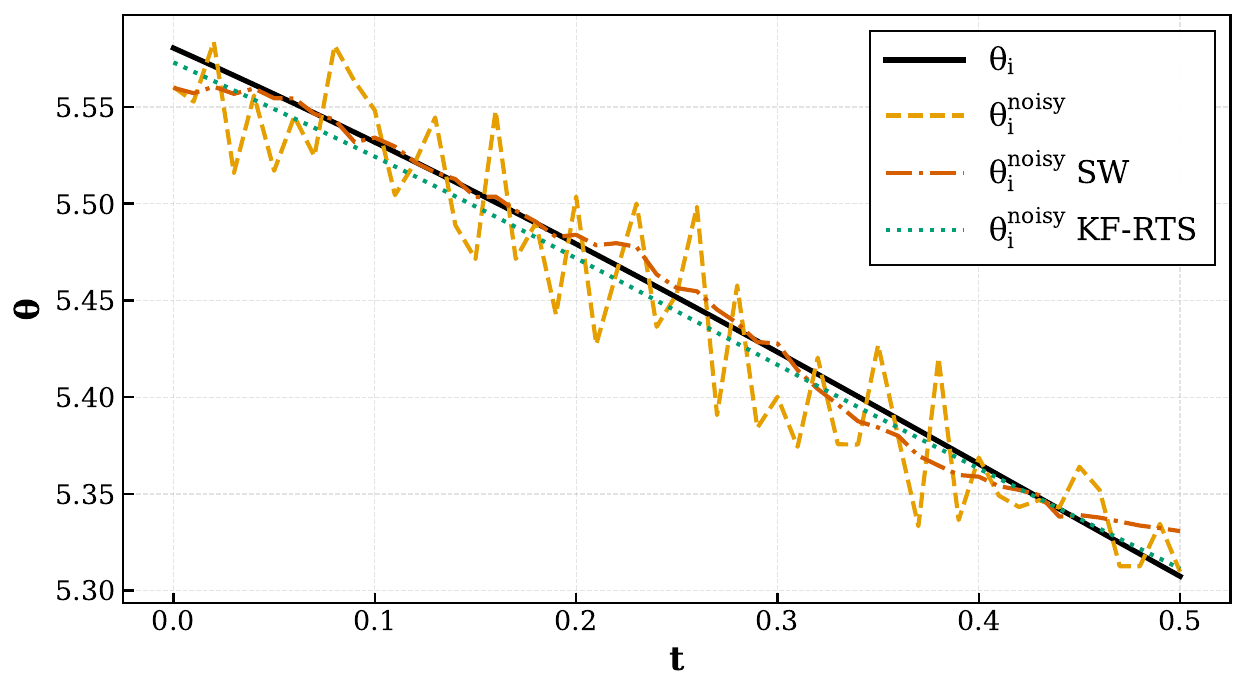}
    \caption{Trajectory $\theta$ (representative agent)}
    \label{subfig:Kuramoto_traj_X_noise_0p01_smoothers}
\end{subfigure}
\hfill
\begin{subfigure}{0.49\textwidth}
    \centering
    \includegraphics[width=\textwidth]{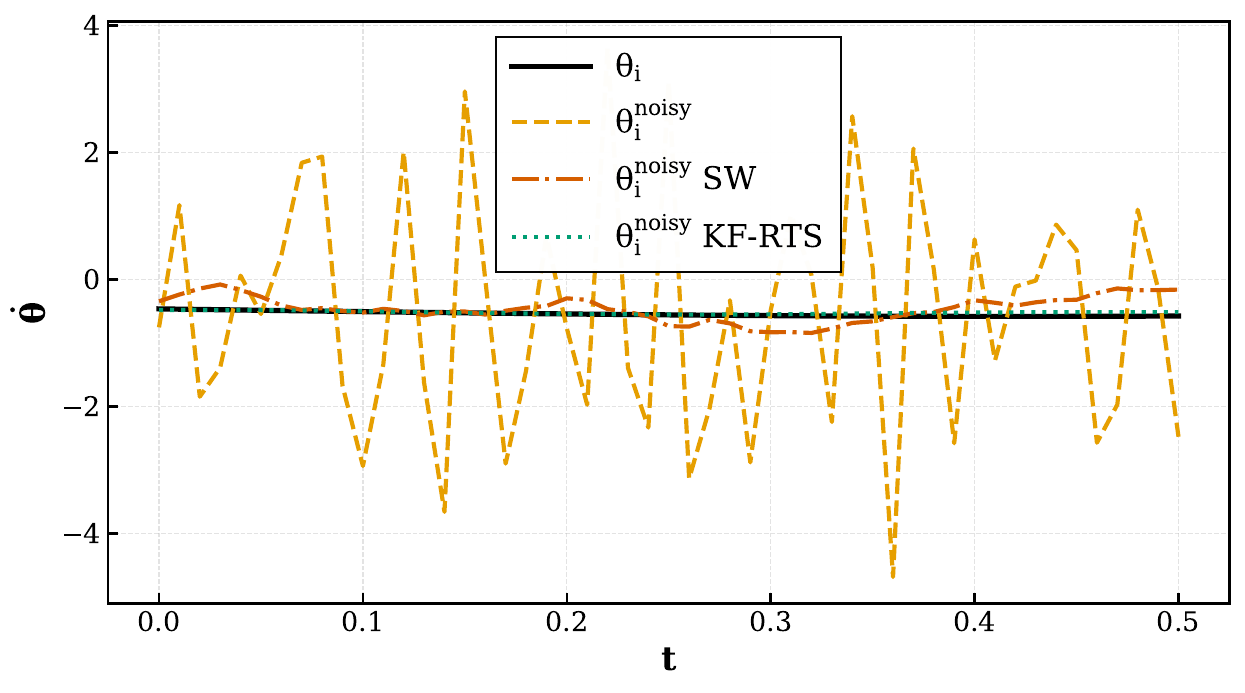}
    \caption{Velocity $\dot{\theta}$ (representative agent)}
    \label{subfig:Kuramoto_traj_LHS_noise_0p01_smoothers}
\end{subfigure}

\vspace{0.2em}

\begin{subfigure}{0.33\textwidth}
    \centering
    \includegraphics[width=\textwidth]{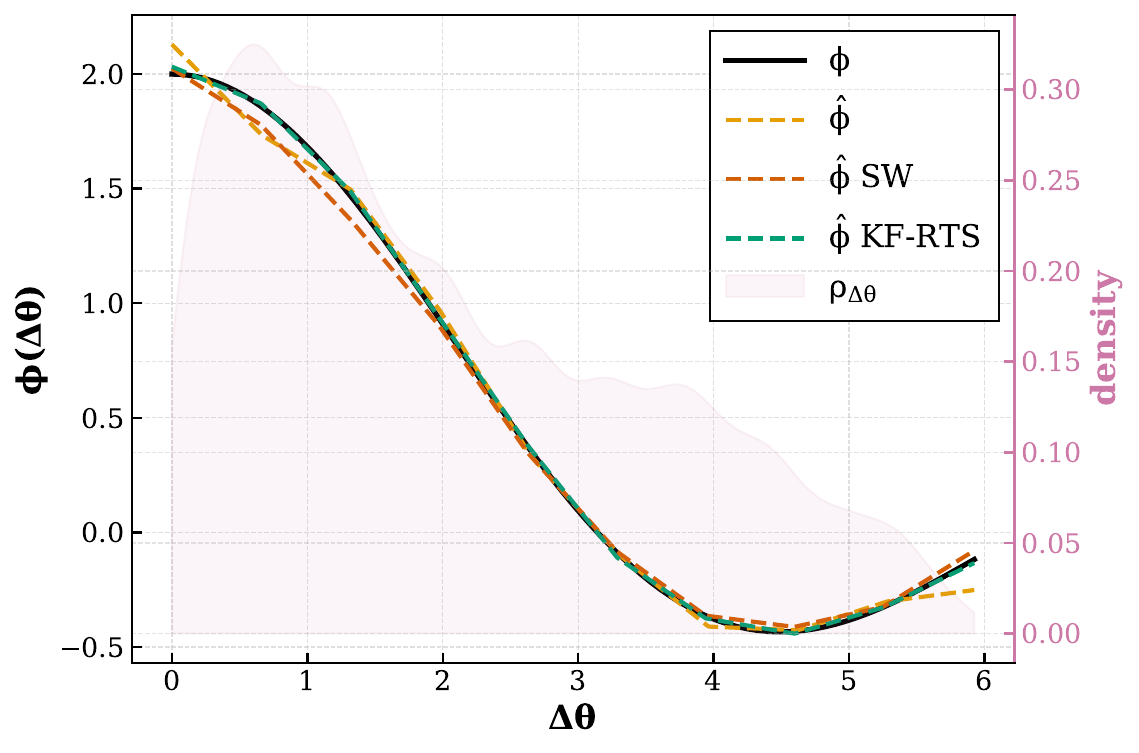}
    \caption{Interaction kernel $\phi$}
    \label{subfig:Kuramoto_phi_noise_0p01_smoothers}
\end{subfigure}
\hfill
\begin{subfigure}{0.33\textwidth}
    \centering
    \includegraphics[width=\textwidth]{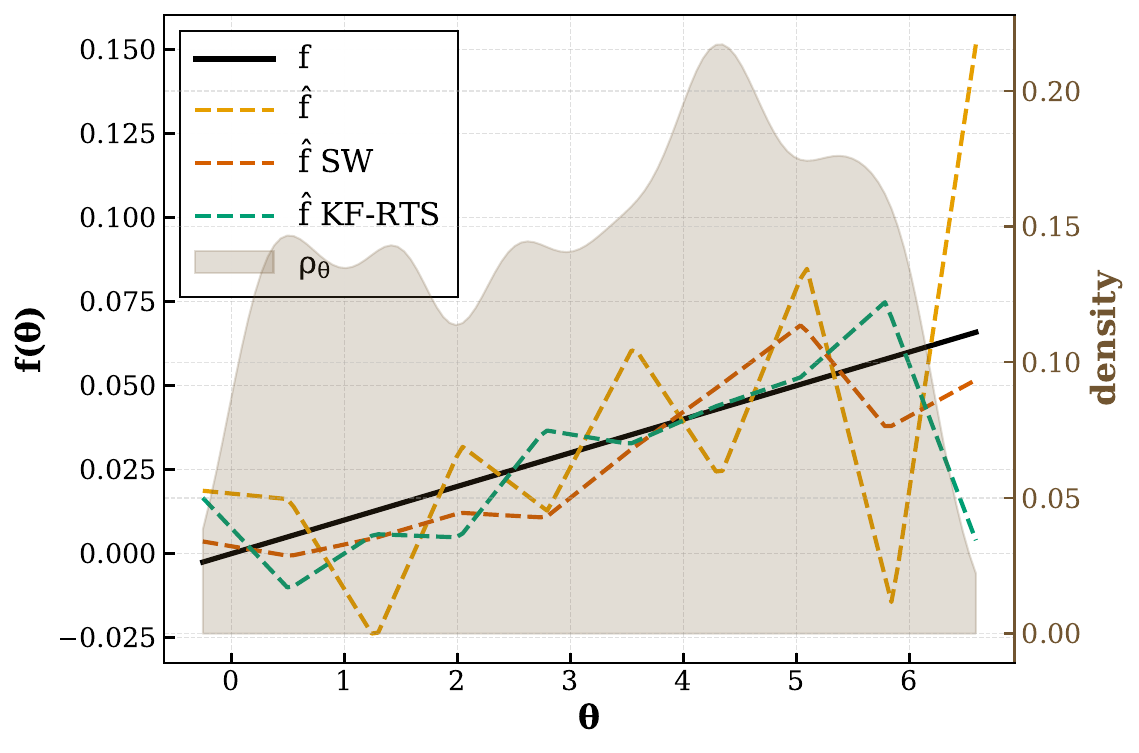}
    \caption{Environmental force $f$}
    \label{subfig:Kuramoto_f_noise_0p01_smoothers}
\end{subfigure}
\hfill
\begin{subfigure}{0.33\textwidth}
    \centering
    \includegraphics[width=\textwidth]{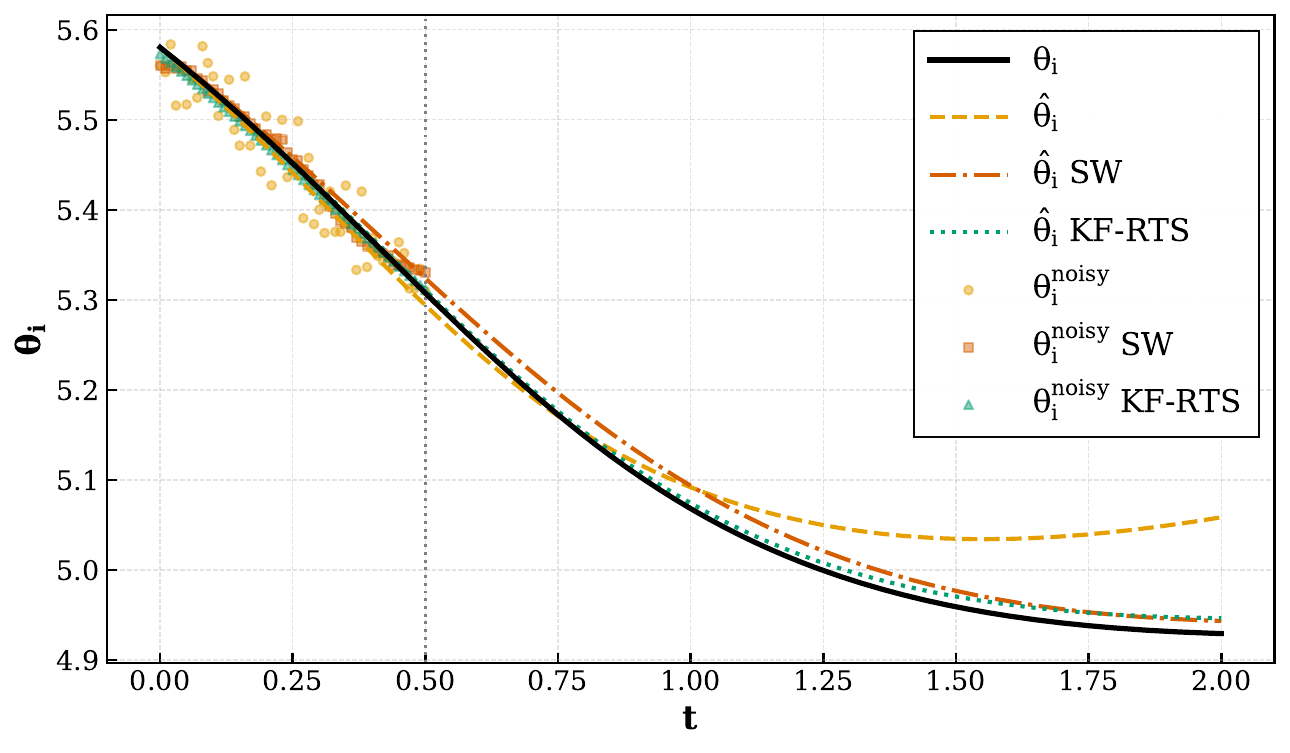}
    \caption{Trajectory data}
    \label{subfig:Kuramoto_traj_noise_0p01_smoothers}
\end{subfigure}

\caption{
(Velocity estimation with noisy data, Kuramoto model) Comparison of filtering methods for trajectory data with multiplicative observational noise in the Kuramoto model~\eqref{eq:Kuramoto-ODE}.  Here the noise level is fixed at $\zeta=0.01$ (see~\eqref{eq:noisy_data} and~\eqref{eq:noisy_dist}). Subfigures~\ref{subfig:Kuramoto_traj_X_noise_0p01_smoothers} and~\ref{subfig:Kuramoto_traj_LHS_noise_0p01_smoothers} provide plots for the noisy observations of state $\theta$ and their derivatives $\dot{\theta}$, respectively, for a representative trajectory, with true trajectories (black), noisy observations (blue), a sliding window (SW) filter (orange), and a Kalman filter with the Rauch--Tung--Striebel (KF-RTS) smoother (green) included. Subfigure~\ref{subfig:Kuramoto_phi_noise_0p01_smoothers} plots the estimated interaction kernels $\phi$, while Subfigure~\ref{subfig:Kuramoto_f_noise_0p01_smoothers} plot the recovered environmental forces $f$ obtained from the true, noisy, and filtered observations. The estimated empirical measures $\hat{\rho}_{\Delta \theta}$ and $\hat{\rho}_{\theta}$ induced by the training data are also included in each subfigure. Subfigure~\ref{subfig:Kuramoto_traj_noise_0p01_smoothers} compares the true, noisy, smoothed, and learned trajectories for a representative training trajectory (i.e. a single agent).  Further information regarding learning and model parameters may be found in Appendix~\ref{app:sec:tables_of_parameters}, as well as in the caption of Figure~\ref{fig:Kuramoto_interaction_kernel_and_fenv_P_and_NP}. 
}

\label{fig:Kuramoto_noise_smoother_recovery}
\end{figure}

We also provide results for an example where we estimate velocities from noisy trajectory data for the Kuramoto model~\eqref{eq:Kuramoto-ODE}, which are provided in Figure~\ref{fig:Kuramoto_noise_smoother_recovery}.  Figure~\ref{subfig:Kuramoto_traj_X_noise_0p01_smoothers} provides a sample trajectory (underlying and noised), together with both sliding window and Kalman filters, as discussed in Section~\ref{subsec:noise_robustness}.  In Figure~\ref{subfig:Kuramoto_traj_LHS_noise_0p01_smoothers} we plot the corresponding numerically differentiated velocities for the same representative trajectory, which are utilized as observation data $\Vml$ in the learning algorithm; results are provided in Figures~\ref{subfig:Kuramoto_phi_noise_0p01_smoothers} (interaction kernel),~\ref{subfig:Kuramoto_f_noise_0p01_smoothers} (environmental force), and~\ref{subfig:Kuramoto_traj_noise_0p01_smoothers} (estimated trajectories).  The estimation provided here is included simply as a ``proof of concept" to demonstrate that mechanism recovery is possible even when estimating velocity from noisy trajectory data, and a comprehensive investigation of such techniques is beyond the scope of this work.  We do note that the optimal choice of numerical differentiation method depends on the characteristics of the underlying dynamics and the level of observational noise, and that for the Kuramoto system presented, the Kalman filter with the Rauch-Tung-Striebel (RTS) smoother appears to provide the most accurate feature recovery and trajectory reconstruction among the methods considered.  As expected, forward differencing on noisy data yields highly inaccurate results, especially in regards to estimation of the environmental force and trajectory prediction.

\subsection{Model selection}
\label{subsec:results_model_selection}

In this section, we demonstrate the ability of the proposed model selection framework introduced in Section~\ref{sec:model_selection} to identify the mechanisms governing a collective system directly from observational data. The objective is not only to estimate the unknown interaction features but also to determine which candidate framework best explains the observed trajectory data.  We consider three second-order systems of various complexity:  1) the pure Cucker-Smale model (Section~\ref{subsubsec:MS_Cucker_Smale}), which describes velocity alignment, 2) the previously introduced phototaxis model (Section~\ref{subsubsec:MS_phototaxis}), which combines alignment with environmental forcing, and 3) a self-propelled particle model with Cucker-Smale dynamics (Section~\ref{subsubsec:MS_SPPCS}), which combines features from both the SPP and Cucker-Smale systems. We also consider a first-order model of opinion dynamics (Section~\ref{subsubsec:MS_opinion_dynamics}). Taken together, these examples allow us to assess the ability of the framework to distinguish among competing model structures and correctly identify the most likely ``active" mechanisms given the observed data.

\subsubsection{Cucker-Smale model}
\label{subsubsec:MS_Cucker_Smale}
We first consider flocking, one of the simplest forms of emergence in collective dynamics. When flocking, agents asymptotically align their velocities. The mathematical study of flocking has attracted considerable attention; see for example~\cite{cucker2007emergent,cucker2010avoiding,cucker2011general,cucker2013conditional,vicsek1995novel} and the references therein.  As a representative flocking system, we consider the Cucker-Smale model~\cite{cucker2007emergent,agueh2011analysis}, which is a second-order interacting particle system describing self-organization driven entirely by a phenomenological interaction velocity alignment kernel.  The Cucker-Smale model fits naturally within the general second-order framework~\eqref{eq:model_selection_general_framework-2order} and is therefore an ideal benchmark for testing whether the model selection procedure can correctly identify a system whose dynamics are governed solely by alignment interactions, as there are no environmental forces in the simplest form of the model. Indeed, the mechanisms defining the Cucker-Smale model are given by
\begin{align}
\begin{split}
    f(x,v) &\equiv 0 \\
    \phi^{E}(r) &\equiv 0 \\
    \phi^A (r) &= \frac{\gamma}{(1+r^2)^{\beta}}
    \end{split}
    \label{eq:Cucker–Smale f_env term and interaction kernel term}
\end{align}
where $\gamma, \beta$ are positive constants.  Depending on the choice of $\beta$, the Cucker-Smale system~\eqref{eq:Cucker–Smale f_env term and interaction kernel term} exhibits different flocking regimes~\cite{cucker2007emergent}. For example, when $\beta < 0.5$, flocking occurs for all initial conditions (\textit{unconditional flocking}), while for $\beta = 0.5$, flocking is dependent on the initial velocity configuration, and for $\beta > 0.5$, flocking depends on both the initial positions and velocities of the agents. For numerical experiments reported in this section, we select $\gamma=1.0$ and $\beta = 0.1 $, so that the dynamics exhibit unconditional flocking. As summarized in Table~\ref{tab:MS_Cucker_smale_framework_features}, the Cucker Smale model contains only alignment interactions and therefore provides a test case for evaluating whether the model selection procedure can correctly identify a pure alignment system. The remaining simulation and learning parameters used in this experiment are reported in Tables~ \ref{app:subsec:learning_algorithm_parameters} and~\ref{app:subsec:data_generation_parameters} of Appendix~\ref{app:sec:tables_of_parameters}.  

\begin{figure}
\centering

\begin{minipage}[t]{0.48\textwidth}
\vspace{0pt}
    \centering

    \begin{subfigure}{\textwidth}
        \centering
        \includegraphics[width=\textwidth, height=3cm, keepaspectratio]
        {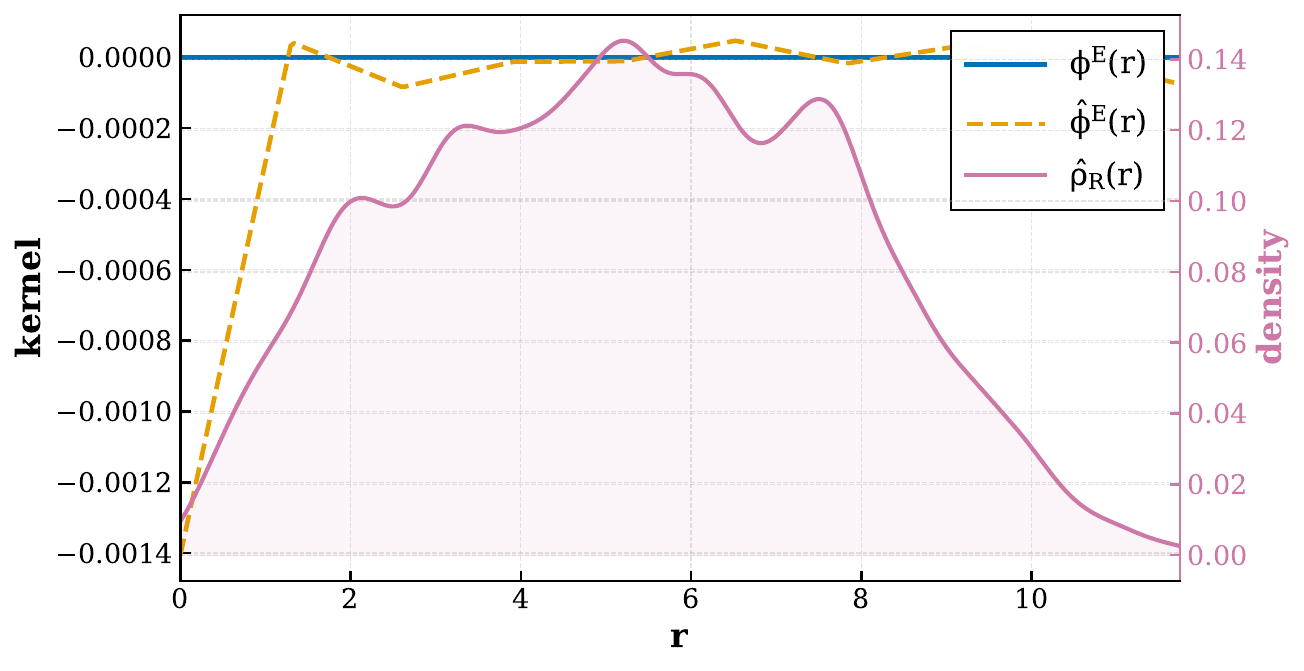}
        \caption{Energy-based interaction kernel $\phi^E$}
        \label{subfig:MS_Energy_kernel_CS}
    \end{subfigure}

    \vspace{0.1em}

    \begin{subfigure}{\textwidth}
        \centering
        \includegraphics[width=\textwidth, height=3cm,keepaspectratio]
        {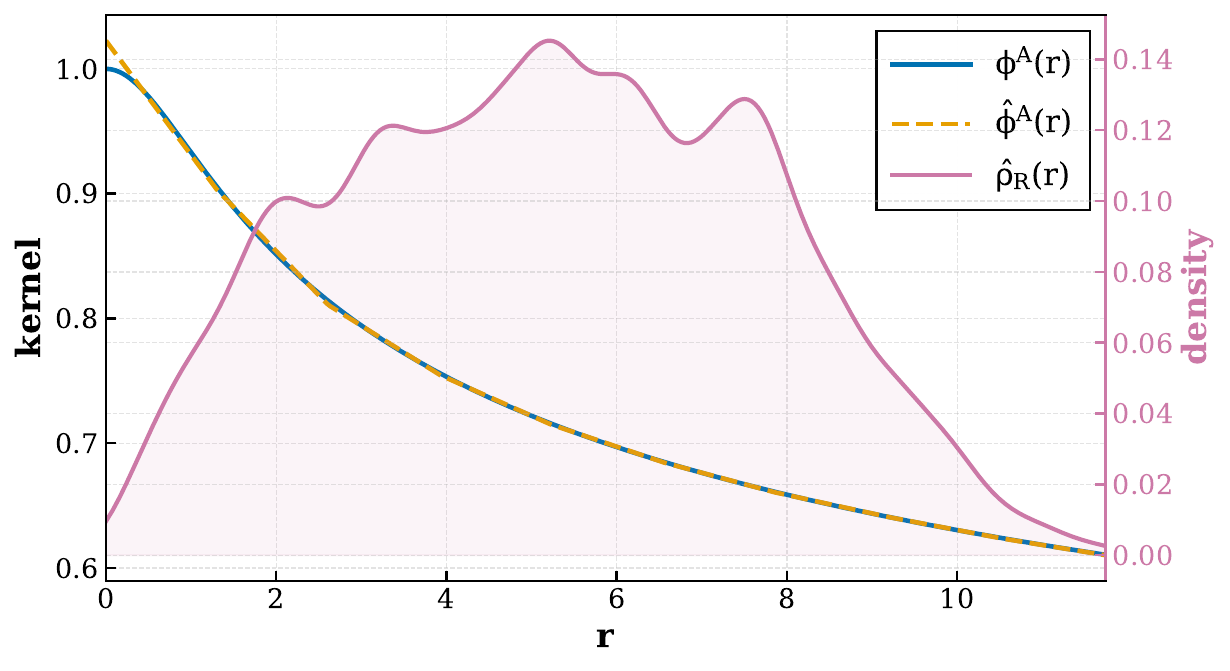}
        \caption{Alignment kernel $\phi^A$}
        \label{subfig:MS_Alignmnet_kernel_CS}
    \end{subfigure}

\end{minipage}
\hfill
\begin{minipage}[t]{0.48\textwidth}
\vspace{0pt}
    \centering

    \begin{subfigure}{\textwidth}
        \centering
        \includegraphics[width=\textwidth]{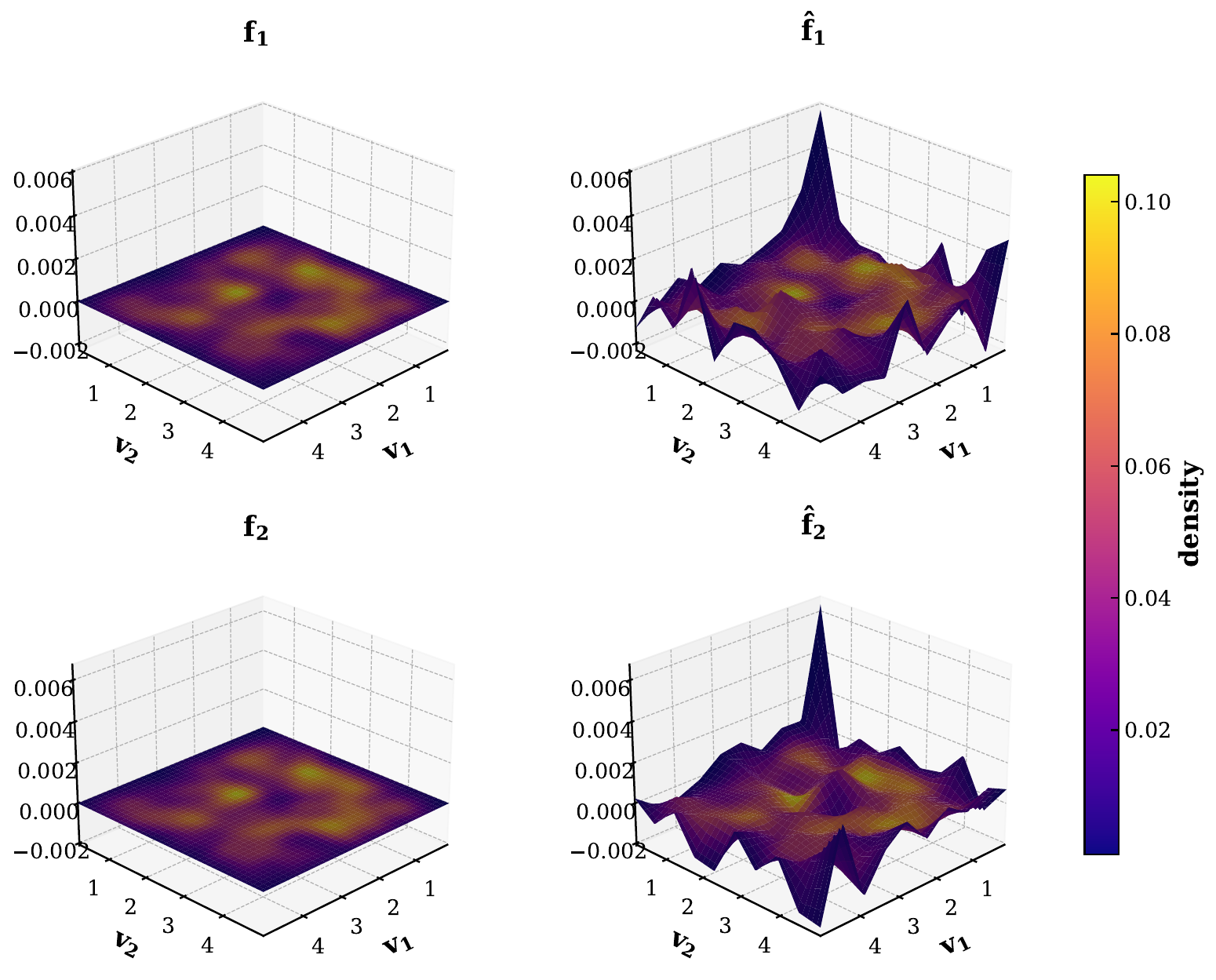}
        \caption{Environmental force $f$}
        \label{subfig:MS_f_CS}
    \end{subfigure}

\end{minipage}

    \caption{(Model selection, Cucker-Smale) Mechanisms present in the Cucker-Smale system~\eqref{eq:Cucker–Smale f_env term and interaction kernel term} within the generalized framework~\eqref{eq:model_selection_general_framework-2order}. The model selection framework estimates three quantities: an energy-based interaction kernel $\hat{\phi}^E$, an alignment interaction kernel $\hat{\phi}^A$, and an environmental force $\hat{f}$. From observations of trajectories, the algorithm correctly identify the system as alignment-dominated. Estimation is performed over $T_r=10$ independent learning trials, and the inferred alignment kernel (Figure~\ref{subfig:MS_Alignmnet_kernel_CS}) has weighted $L^{2}$ norm $\|\hat{\phi}^{A}\|_{L^2(\hat{\rho}_R)}=$\num{0.742153}$\pm$\num{0.003468}, which is several orders of magnitude larger than the inferred energy kernel $\|\hat{\phi}^{E}\|_{L^2(\hat{\rho}_R)}=$\num{1.9985e-04}$\pm$\num{1.0221e-04} (Figure~\ref{subfig:MS_Energy_kernel_CS}) and environmental force $\|\hat{f}\|_{L^2(\hat{\rho}_V)}=$\num{0.000528}$\pm$\num{0.000070} (Figure~\ref{subfig:MS_f_CS}). Note the scale of the vertical axes.  Further information regarding learning and model parameters may be found in Appendix~\ref{app:sec:tables_of_parameters}.}
    \label{fig:MS_Cucker_smale_framework_features}
    \end{figure}

We estimate all three mechanisms ($f$, $\phi^{E}$, and $\phi^{A}$) in the most general second-order framework~\eqref{eq:model_selection_general_framework-2order}; results are provided in Figure~\ref{fig:MS_Cucker_smale_framework_features}.  We observe that the fully non-parametric variational algorithm exhibits negligible training variability across repeated trials and correctly identifies the dominant alignment interaction $\phi^{A}$ while simultaneously obtaining near zero estimates for inactive energy $\phi^{E}$ and environmental force $f$ terms.  Indeed, when measuring the weighted $L^{2}$ norms~\eqref{eq:L2_weighted_def}, we obtain $\norm{\hat{f}}_{L^{2}(\hat{\rho}_{R})}, \norm{\hat{\phi}^{E}}_{L^{2}(\hat{\rho}_{V})} =O(10^{-4})$, while $\norm{\hat{\phi}^{A}}_{L^{2}(\hat{\rho}_{R})} = O(10^{-1})$.  We also provide quantitative measures of the model selection framework in Table~\ref{tab:CS_model_selection_summary} for the Cucker-Smale system.  Candidates $F_{2}-S_{1}$ are rejected by the framework compatibility gate as their relative residual errors satisfy $E_{\mathrm{res}}^{\mathrm{rel_\epsilon}} > 0.25$.  The remaining candidates $S_{2}-S_{6}$ are thus admissible and are retained for ranking. All admissible candidates satisfy the primary criterion by achieving (within tolerance $\tau_{\mathrm{traj}}$, as defined in~\eqref{eq:tie_candidate_treshhold}) the same minimum validation trajectory error $E_{\min}$. We therefore move to the secondary criterion of framework complexity. The minimum complexity is achieved by candidate $S_{2}$, corresponding to the second-order alignment-only framework, which is precisely the true underlying model~\eqref{eq:Cucker–Smale f_env term and interaction kernel term} which generated the data.  Thus, the framework selection algorithm correctly identified the pure Cucker-Smale alignment model.

\begin{table}
        \centering
        \begin{tabular}{|c|c|c|c|}
        \hline
        Order & $\phi^E$ & $\phi^A$ &  $f$ \\ \hline
        2 & - & \checkmark & - \\ \hline
        \end{tabular}
        \caption{Mechanisms of the Cucker-Smale system~\eqref{eq:Cucker–Smale f_env term and interaction kernel term}.  The symbol ``-" indicates that the mechanism is not present, while ``\checkmark" indicates that the mechanism is present.}
        \label{tab:MS_Cucker_smale_framework_features}
    \end{table}

\begin{table}
\centering
\small
\begin{tabular}{lrrrrrrrl}
\toprule
Candidate & Order & $E_{\mathrm{res}}^{\mathrm{rel}_\epsilon}$ & $E_{\mathrm{recon}}^{\mathrm{rel}}$ & $\bar E_{\mathrm{traj}}(M_{\mathrm{val}})$ & Complexity & $E_{\mathrm{res}}$ & Rank & Selected \\
\midrule
$S_2$ & 2 & 0.000626 & 0.000004 & 0.000259 & 3 & 0.000509 & 1 & Yes \\
$S_4$ & 2 & 0.000708 & 0.000004 & 0.000437 & 4 & 0.000575 & 2 & No \\
$S_5$ & 2 & 0.000624 & 0.000004 & 0.000255 & 5 & 0.000507 & 3 & No \\
$S_6$ & 2 & 0.000717 & 0.000004 & 0.000464 & 6 & 0.000583 & 4 & No \\
$F_2$ & 1 & 0.593726 & 0.003266 & 0.732061 & 2 & 0.482466 & 5 & No \\
$S_3$ & 2 & 0.593726 & 0.003266 & 0.732061 & 4 & 0.482466 & 6 & No \\
$F_1$ & 1 & 0.995070 & 0.012991 & 1.583641 & 1 & 0.808601 & 7 & No \\
$S_1$ & 2 & 0.995070 & 0.012991 & 1.583641 & 3 & 0.808601 & 8 & No \\
\bottomrule
\end{tabular}
\caption{(Model selection, Cucker-Smale) Summary of the candidate models considered during model selection.
For each candidate, we report the system order, framework compatibility diagnostics,
validation trajectory error, model complexity, residual error,
ranking, and whether the candidate was ultimately selected based on the observed trajectory data.  In this case, the framework correctly identifies the Cucker-Smale flocking model~\eqref{eq:Cucker–Smale f_env term and interaction kernel term}.
}
\label{tab:CS_model_selection_summary}
\end{table}

\subsubsection{Phototaxis model}
\label{subsubsec:MS_phototaxis}
We perform model selection on the phototaxis system introduced in Section~\ref{subsubsec:results_phototaxis}, with governing equations~\eqref{eq:Levi_f_env_term_and_interaction_kernel_term} with respect to framework~\eqref{eq:model_selection_general_framework-2order}.  Phototaxis thus combines alignment with environmental forcing, the latter of which is induced by an external light source. Therefore, this model provides a natural test case for determining whether model selection can simultaneously identify collective alignment and environmental forcing while rejecting inactive energetic interactions; these mechanisms are summarized in Table~\ref{tab:MS_Phototaxis_framework_features}.  That is, the selection should identify mechanisms $\phi^A$ and $f$ while rejecting $\phi^E$.  The simulation and learning parameters utilized in this numerical experiment are reported in Tables~\ref{app:subsec:learning_algorithm_parameters} and~\ref{app:subsec:data_generation_parameters}, with the corresponding values of $\phi^A$ and $f$ specified in Section~\ref{subsubsec:results_phototaxis}.

\begin{table}
        \centering
        \begin{tabular}{|c |c|c|c|}
        \hline
        Order & $\phi^E$ & $\phi^A$ &  $f$ \\ \hline
        2 &  - & \checkmark & \checkmark \\ \hline
        \end{tabular}
        \caption{Mechanisms of the phototaxis system~\eqref{eq:Levi_f_env_term_and_interaction_kernel_term}.  The symbol ``-" indicates that the mechanism is not present, while ``\checkmark" indicates that the mechanism is present.}
        \label{tab:MS_Phototaxis_framework_features}
    \end{table}

\begin{figure}
\centering

\begin{minipage}[t]{0.48\textwidth}
\vspace{0pt}
    \centering

    \begin{subfigure}{\textwidth}
        \centering
        \includegraphics[width=\textwidth, height=3cm, keepaspectratio]{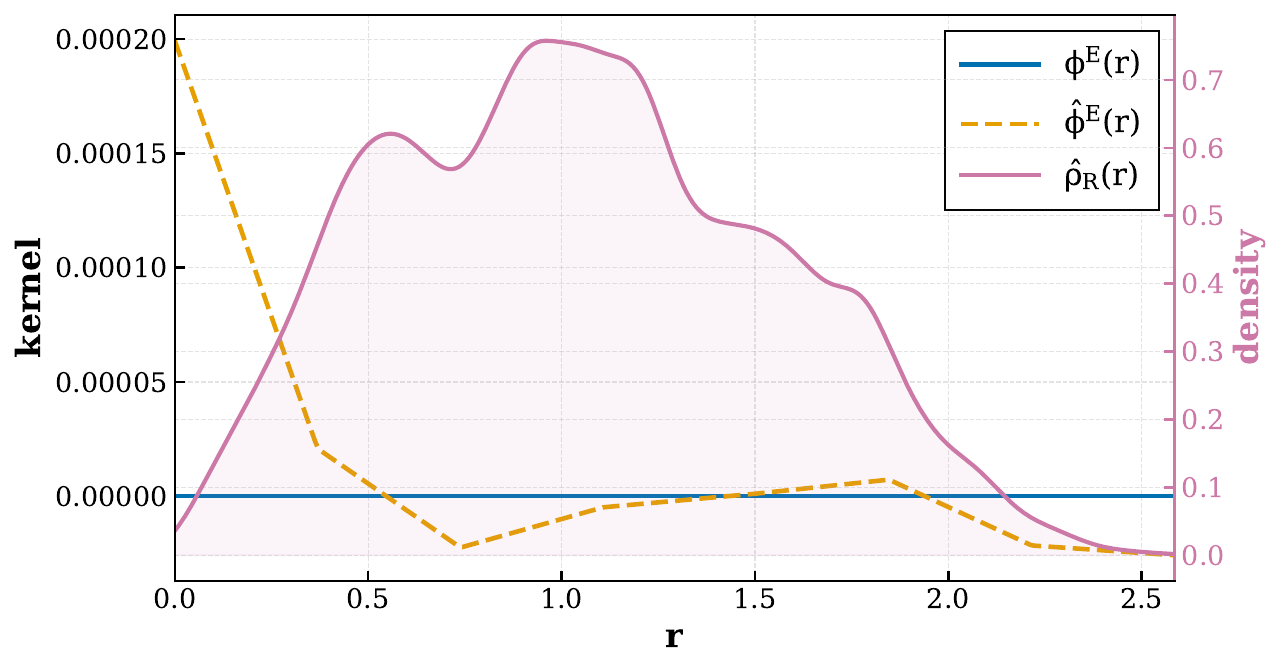}
        \caption{Energy-based interaction kernel $\phi^E$}
        \label{subfig:MS_Energy_kernel_Phototaxis}
    \end{subfigure}

    \vspace{0.1em}

    \begin{subfigure}{\textwidth}
        \centering
        \includegraphics[width=\textwidth, height=3cm,keepaspectratio]{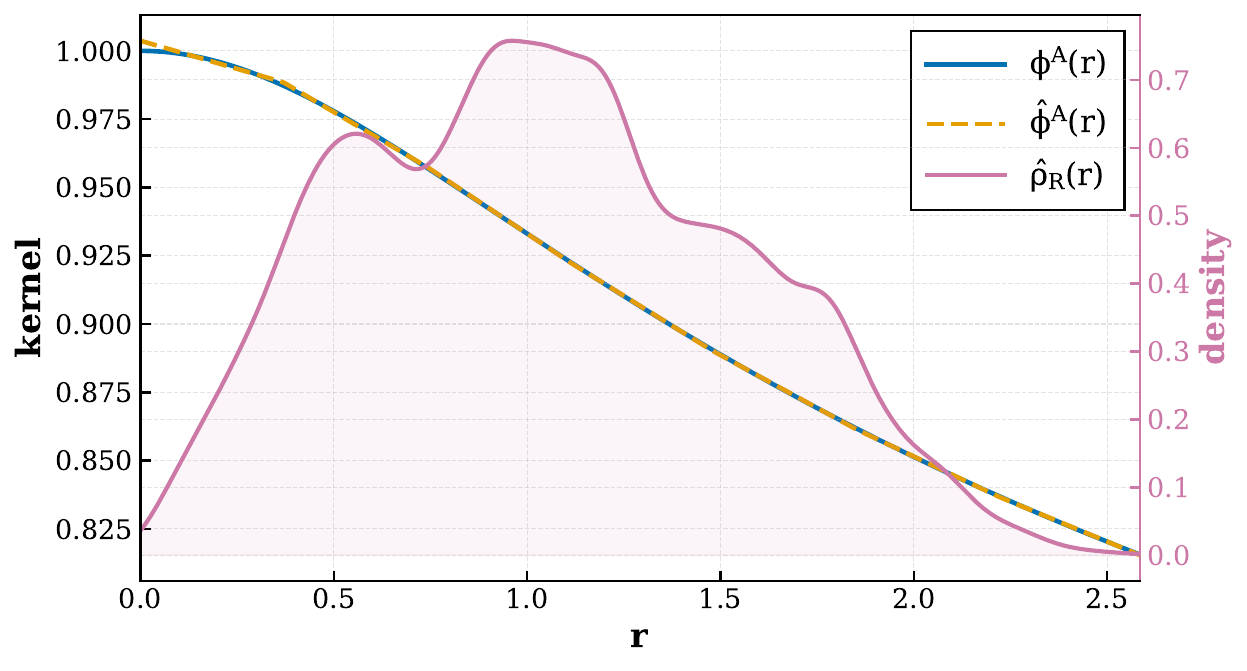}
        \caption{Alignment interaction kernel $\phi^A$}
        \label{subfig:MS_Alignmnet_kernel_Phototaxis}
    \end{subfigure}

\end{minipage}
\hfill
\begin{minipage}[t]{0.48\textwidth}
\vspace{0pt}
    \centering

    \begin{subfigure}{\textwidth}
        \centering
        \includegraphics[width=\textwidth]{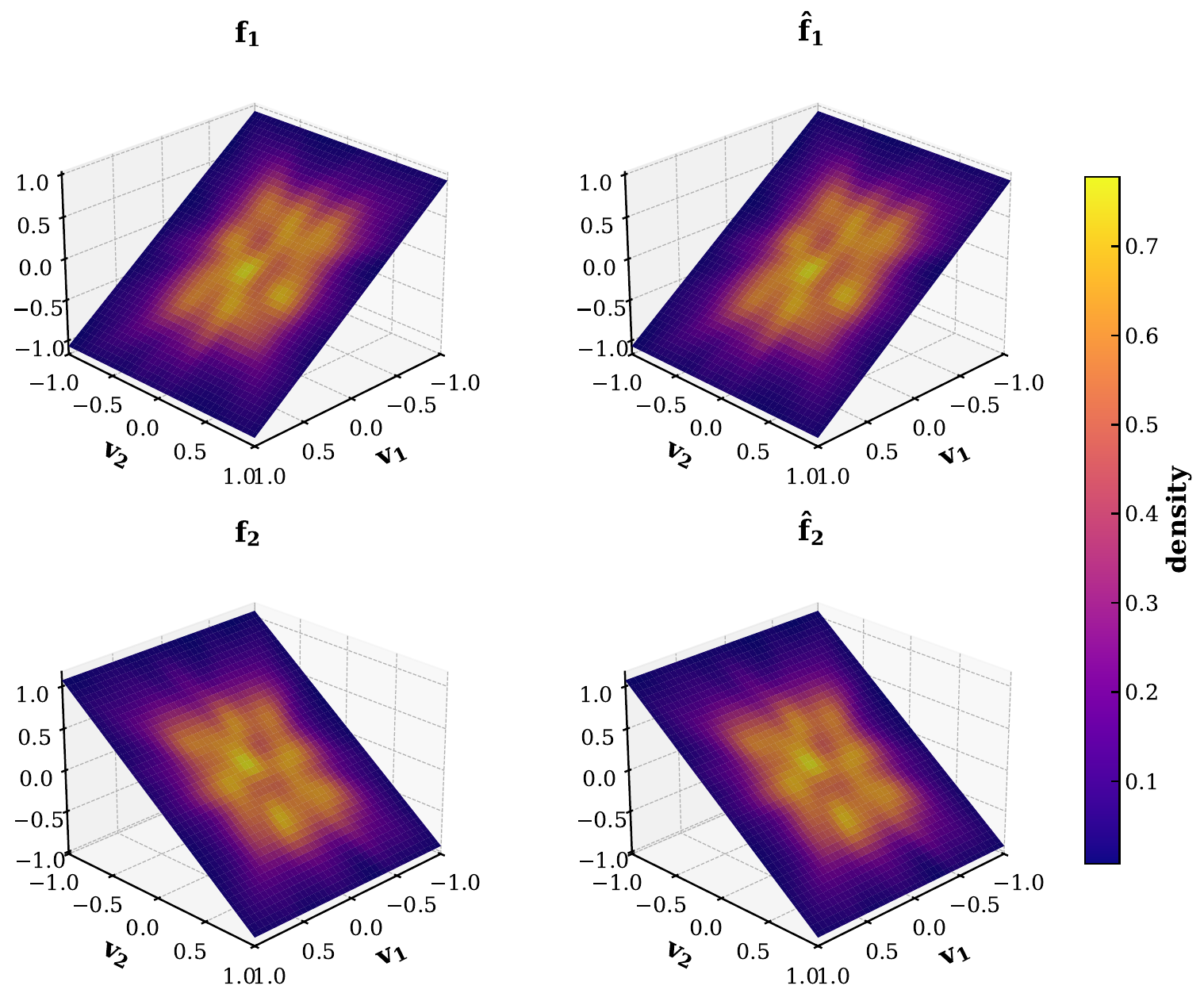}
        \caption{Environmental force $f$}
        \label{subfig:MS_f_Phototaxis}
    \end{subfigure}

\end{minipage}
    \caption{(Model selection, phototaxis) Mechanisms present in the phototaxis system~\eqref{eq:Levi_f_env_term_and_interaction_kernel_term} within the generalized framework~\eqref{eq:model_selection_general_framework-2order}. The model selection framework estimates three quantities: an energy-based interaction kernel $\hat{\phi}^E$, an alignment interaction kernel $\hat{\phi}^A$, and an environmental force $\hat{f}$. From observation of trajectories, the algorithm correctly identify the system as being governed by alignment and environmental force interactions. Estimation is performed over $T_r=10$ independent learning trials, and the inferred  learned alignment kernel (Figure~\ref{subfig:MS_Alignmnet_kernel_Phototaxis}) has weighted $L^{2}$ norm $\|\hat{\phi}^{A}\|_{L^2(\hat{\rho}_R)}=$\num{0.931035}$\pm$\num{0.002016} and the estimated environmental force has weighted $L^{2}$ norm $\|\hat{f}\|_{L^2(\hat{\rho}_V)}=$\num{0.600504}$\pm$\num{0.006211} (Figure~\ref{subfig:MS_f_Phototaxis}), both of which are several orders of magnitude larger than the inferred energy kernel $\|\hat{\phi}^{E}\|_{L^2(\hat{\rho}_R)}=$\num{7.7845e-04}$\pm$\num{6.9402e-04} (Figure~\ref{subfig:MS_Energy_kernel_Phototaxis}). Note the scale of the vertical axes.  These results indicate that the model-selection framework correctly identifies the active mechanisms while suppressing the inactive energy interaction.  Further information regarding learning and model parameters may be found in Appendix~\ref{app:sec:tables_of_parameters}.}
    \label{fig:MS_Phototaxis_framework_features}
    \end{figure}

We estimate all three mechanisms ($f$, $\phi^{E}$, and $\phi^{A}$) in the most general second-order framework~\eqref{eq:model_selection_general_framework-2order}; results are provided in Figure~\ref{fig:MS_Phototaxis_framework_features}.  As in Section~\ref{subsubsec:MS_Cucker_Smale}, the model selection algorithm correctly identifies the alignment interaction $\phi^A$ and environmental force $f$ as active features, while obtaining near zero values for the energy interaction kernel $\phi^E$. Indeed, when measuring the weighted $L^{2}$ norms~\eqref{eq:L2_weighted_def}, we obtain $\norm{\hat{\phi}^{E}}_{L^{2}(\hat{\rho}_{R})} =O(10^{-4})$, while $\norm{\hat{\phi}^{A}}_{L^{2}(\hat{\rho}_{R})},\norm{\hat{f}}_{L^{2}(\hat{\rho}_{V})},  = O(10^{-1})$.  We also provide quantitative measures on the model selection framework in Table~\ref{tab:Phototaxis_model_selection_summary}. Candidates $F_1-S_1$ are rejected by the framework compatibility gate as their relative residual errors satisfy $E_{\mathrm{res}}^{\mathrm{rel_\epsilon}} > 0.25$, while the remaining candidates $S_4-S_5$ are thus admissible and are retained for ranking.  Among these admissible candidates, the minimum validation trajectory error $E_{\min}$ is achieved by $S_4-S_6$ candidates, resulting in a tie according to the criterion defined in~\eqref{eq:tie_candidate_treshhold}. The selection procedure therefore proceeds to the secondary metric, i.e. framework complexity.  The minimum complexity is achieved by candidate $S_4$, corresponding to the second-order alignment-plus-environmental force framework, which coincides with the true data-generation model, demonstrating that the proposed model selection procedure can correctly identify both active mechanisms in the phototaxis model.  

\begin{table}
\centering
\small

\begin{tabular}{lrrrrrrrl}
\toprule
Candidate & Order & $E_{\mathrm{res}}^{\mathrm{rel}_\epsilon}$ & $E_{\mathrm{recon}}^{\mathrm{rel}}$ & $\bar E_{\mathrm{traj}}(M_{\mathrm{val}})$ & Complexity & $E_{\mathrm{res}}$ & Rank & Selected \\
\midrule
$S_4$ & 2 & 0.000067 & 0.000002 & 0.000016 & 4 & 0.000040 & 1 & Yes \\
$S_6$ & 2 & 0.000068 & 0.000002 & 0.000017 & 6 & 0.000041 & 2 & No \\
$F_2$ & 1 & 0.216497 & 0.009056 & 0.102124 & 2 & 0.130341 & 3 & No \\
$S_3$ & 2 & 0.216497 & 0.009056 & 0.102124 & 4 & 0.130341 & 4 & No \\
$S_2$ & 2 & 0.224109 & 0.015654 & 0.272981 & 3 & 0.134924 & 5 & No \\
$S_5$ & 2 & 0.224109 & 0.015654 & 0.272981 & 5 & 0.134924 & 6 & No \\
$F_1$ & 1 & 1.041332 & 0.088641 & 0.955980 & 1 & 0.626931 & 7 & No \\
$S_1$ & 2 & 1.041332 & 0.088641 & 0.955980 & 3 & 0.626931 & 8 & No \\
\bottomrule
\end{tabular}
\caption{
(Model selection, phototaxis) Summary of the candidate models considered during model selection.
For each candidate, we report the system order, framework compatibility diagnostics,
validation trajectory error, model complexity, residual error,
ranking, and whether the candidate was ultimately selected based on the observed trajectory data.  In this case, the framework correctly identifies the phototaxis model~\eqref{eq:Levi_f_env_term_and_interaction_kernel_term}.
}
\label{tab:Phototaxis_model_selection_summary}
\end{table}

\subsubsection{Self-propelled particle model with velocity alignment (SPPCS)}
\label{subsubsec:MS_SPPCS}

We next consider an extension of the self-propelled particle (SPP) model introduced in Section~\ref{subsubsec:results_SPP} that incorporates a Cucker-Smale-type velocity alignment interaction; for notational convenience, we abbreviate this model as SPPCS. In contrast to the preceding examples, this system includes two interaction kernels, corresponding to attraction-repulsion
and velocity alignment, together with an environmental/self-propulsion force. It therefore provides a useful benchmark for assessing whether the proposed framework can distinguish and recover multiple mechanistic force components from trajectory data.  The precise environmental force and energy kernel are thus given by~\eqref{eq:SPP_f_env_term_and_interaction_kernel-terms}, while the alignment kernel $\phi^{A}$ takes the form as in~\eqref{eq:Cucker–Smale f_env term and interaction kernel term}, and the ODE system is given as in the general second-order framework~\eqref{eq:model_selection_general_framework-2order}.  Note that this systems contains all three mechanisms described in the generalized framework: an environmental force, an alignment-based interaction, and an energy-based interaction.  Thus, this model provides a  more challenging selection benchmark example considered, as the correct framework requires all mechanisms to be identified simultaneously.  For trajectory generation, we utilize parameter values as in Sections~\ref{subsubsec:MS_Cucker_Smale} and~\ref{subsubsec:results_SPP}, and the remaining simulation and learning parameters are reported in Tables~\ref{app:subsec:learning_algorithm_parameters} and~\ref{app:subsec:data_generation_parameters}.  The mechanisms of the SPPCS are summarized in Table~\ref{table:SPPCS framework features-MS}, and the selection algorithm should identify non-zero mechanisms $\phi^{E}$, $\phi^{A}$, and $f$.

\begin{table}
        \centering
        \begin{tabular}{|c|c|c|c|}
        \hline
         Order & $\phi^E$ & $\phi^A$ &  $f$ \\ \hline
        2 & \checkmark & \checkmark & \checkmark \\ \hline
        \end{tabular}
         \caption{Mechanisms of the SPPCS system. The symbol ``-" indicates that the mechanism is not present, while ``\checkmark" indicates that the mechanism is present.}
        \label{table:SPPCS framework features-MS}
    \end{table}

\begin{figure}
\centering

\begin{minipage}[t]{0.48\textwidth}
\vspace{0pt}
    \centering

    \begin{subfigure}{\textwidth}
        \centering
        \includegraphics[width=\textwidth, height=3cm, keepaspectratio]{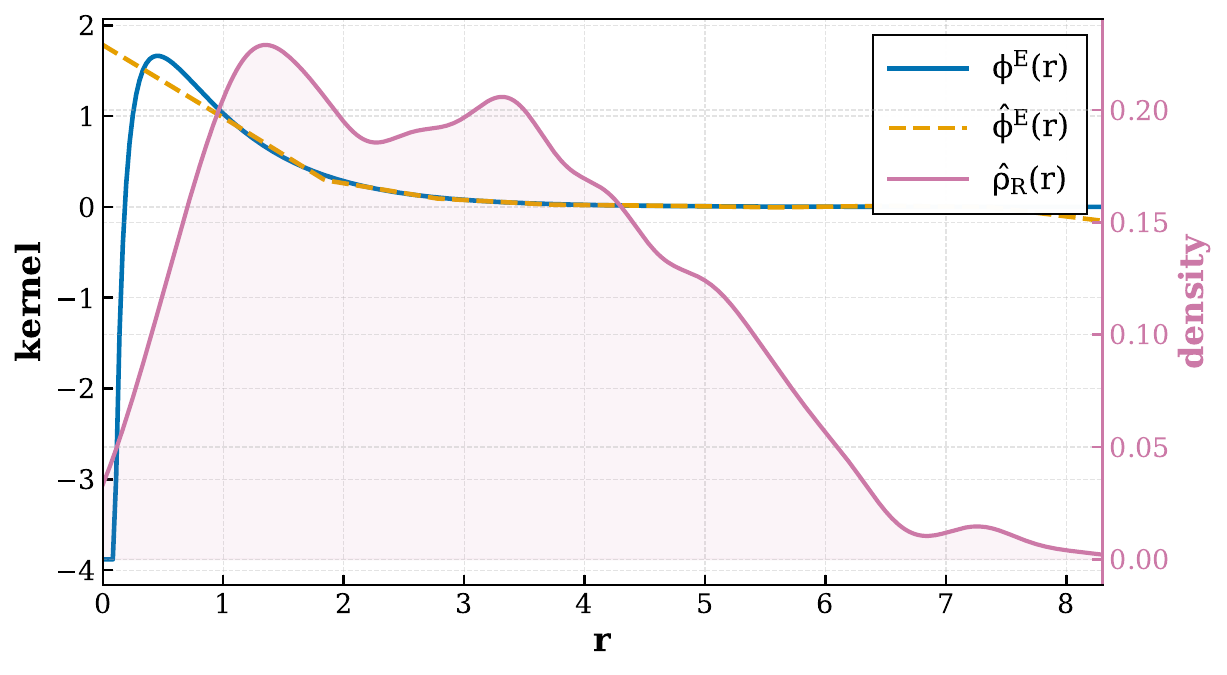}
        \caption{Energy-based interaction kernel $\phi^E$}
        \label{subfig:MS_Energy_kernel_SPPCS}
    \end{subfigure}

    \vspace{0.1em}

    \begin{subfigure}{\textwidth}
        \centering
        \includegraphics[width=\textwidth, height=3cm,keepaspectratio]{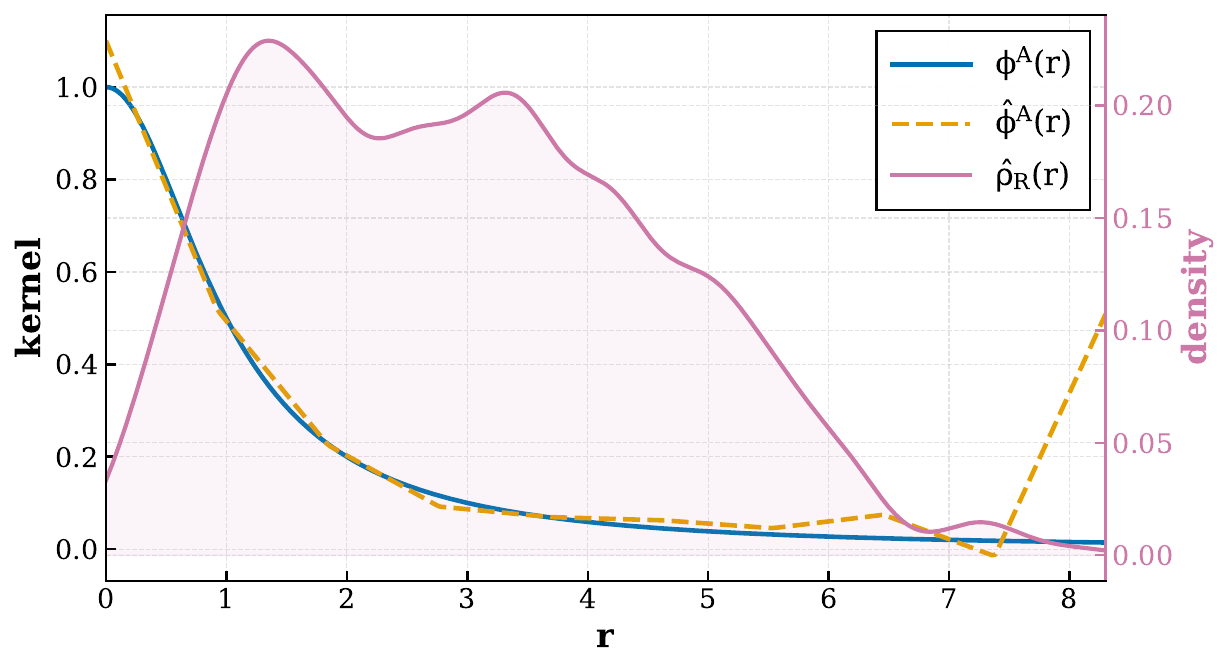}
        \caption{Alignment kernel $\phi^A$}
        \label{subfig:MS_Alignmnet_kernel_SPPCS}
    \end{subfigure}

\end{minipage}
\hfill
\begin{minipage}[t]{0.48\textwidth}
\vspace{0pt}
    \centering

    \begin{subfigure}{\textwidth}
        \centering
        \includegraphics[width=\textwidth]{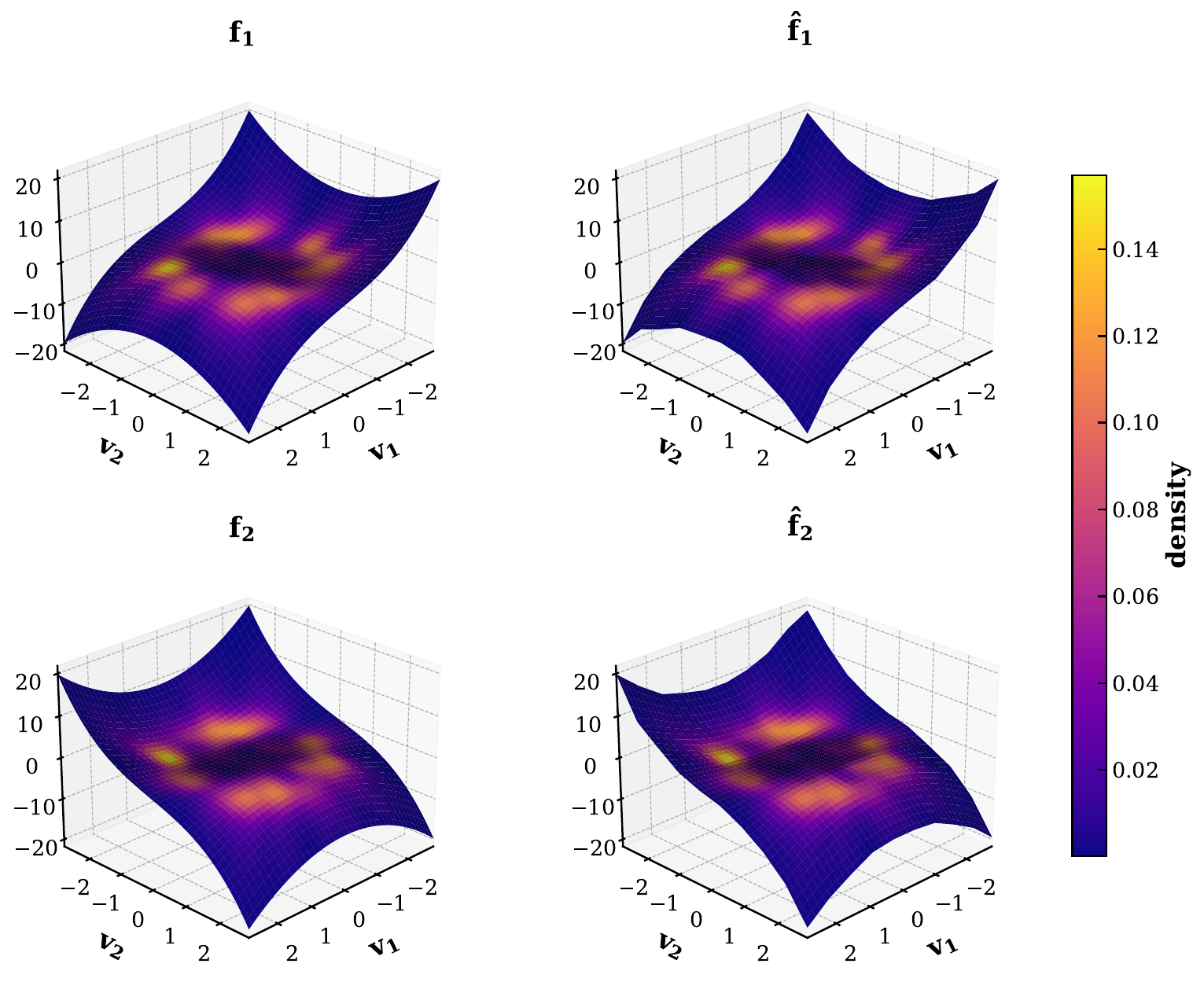}
        \caption{Environmental force $f$}
        \label{subfig:MS_f_SPPCS}
    \end{subfigure}

\end{minipage}
    
   \caption{(Model selection, SPPCS) Mechanisms present in the SPPCS system within the generalized framework~\eqref{eq:model_selection_general_framework-2order}. The model selection framework estimates three quantities: an energy-based interaction kernel $\hat{\phi}^E$, an alignment interaction kernel $\hat{\phi}^A$, and an environmental force $\hat{f}$. From observations of trajectories, the algorithm correctly identify the system as being governed by all three mechanisms. Estimation is performed over $T_r=10$ independent learning trials, and the inferred alignment kernel has weighted $L^{2}$ norm $\|\hat{\phi}^{A}\|_{L^2(\hat{\rho}_R)}=$\num{0.344557	}$\pm$\num{0.031942} (Figure~\ref{subfig:MS_Alignmnet_kernel_SPPCS}), the inferred environmental force has weighted $L^{2}$ norm $\|\hat{f}\|_{L^2(\hat{\rho}_V)}=$\num{3.062965}$\pm$\num{0.047218} (Figure~\ref{subfig:MS_f_SPPCS}), and the inferred energy kernel has weighted $L^{2}$ norm $\|\hat{\phi}^{E}\|_{L^2(\rho_R)}=$\num{6.5314e-01}$\pm$\num{6.0227e-02} (Figure~\ref{subfig:MS_Energy_kernel_SPPCS}). Note the scale of the vertical axes.  Further information regarding learning and model parameters may be found in Appendix~\ref{app:sec:tables_of_parameters}.}
    \label{fig:MS_SPPCS_framework_features}
    \end{figure}

We estimate all three mechanisms ($f$, $\phi^{E}$, and $\phi^{A}$) in the most general second-order framework~\eqref{eq:model_selection_general_framework-2order}; results are provided in Figure~\ref{fig:MS_SPPCS_framework_features}.  The algorithm correctly identifies all three model mechanisms, with all three weighted $L^{2}$ norms~\eqref{eq:L2_weighted_def} of the same order ($O(10^{-1})$).  We also provide quantitative measures on the model selection framework in Table~\ref{tab:SPPCS_model_selection_summary}. We observe that candidates $S_2-S_1$ are rejected by the framework compatibility gate as their relative residual errors satisfy $E_{\mathrm{res}}^{\mathrm{rel_\epsilon}} > 0.25$.  In addition, candidates $S_2$ and $S_5$ fail to produce vector fields that can be numerically integrated, and as candidate frameworks must generate physically meaningful trajectories before compatibility metrics can be evaluated, these candidates are rejected prior to the framework compatibility stage and removed from further consideration.  The remaining candidates $S_3-S_6$ satisfy the admissibility criteria and are retained for ranking. Among these admissible candidates, the minimum validation trajectory error $E_{\min}$ is achieved uniquely by candidate $S_6$ (within tolerance $\tau_{\mathrm{traj}}$ in~\eqref{eq:model_selection_admissible}), corresponding to the second-order energy-alignment-environmental force framework, which is precisely the true generating model.  This thus demonstrates that the proposed model selection procedure can correctly identify all active mechanisms when they are simultaneously present in the observation trajectory data.
    
\begin{table}[ht]
\centering
\small

\begin{tabular}{lrrrrrrrl}
\toprule
Candidate & Order & $E_{\mathrm{res}}^{\mathrm{rel_\epsilon}}$ & $E_{\mathrm{recon}}^{\mathrm{rel}}$ & $\bar E_{\mathrm{traj}}(M_{\mathrm{val}})$ & Complexity & $E_{\mathrm{res}}$ & Rank & Selected \\
\midrule
$S_6$ & 2 & 0.063176 & 0.000608 & 0.072231 & 6 & 0.105255 & 1 & Yes \\
$S_4$ & 2 & 0.127328 & 0.003480 & 0.354430 & 4 & 0.212138 & 2 & No \\
$F_2$ & 1 & 0.237008 & 0.003800 & 0.396505 & 2 & 0.394873 & 3 & No \\
$S_3$ & 2 & 0.237008 & 0.003800 & 0.396505 & 4 & 0.394873 & 4 & No \\
$S_2$ & 2 & N/A & N/A & N/A & 3 & N/A & 5 & No \\
$S_5$ & 2 & N/A & N/A & N/A & 5 & N/A & 6 & No \\
$F_1$ & 1 & 0.990463 & 0.080556 & 2.176686 & 1 & 1.650183 & 7 & No \\
$S_1$ & 2 & 0.990463 & 0.080556 & 2.176686 & 3 & 1.650183 & 8 & No \\
\bottomrule
\end{tabular}
\caption{
(Model selection, SPPCS) Summary of the candidate models considered during model selection.
For each candidate, we report the system order, framework compatibility diagnostics,
validation trajectory error, model complexity, residual error,
ranking, and whether the candidate was ultimately selected.  In this case, the framework correctly identifies the SPPCS model with all three mechanisms active. N/A values indicate errors in trajectory reconstruction due to numerical instability of obtained estimators.
}
\label{tab:SPPCS_model_selection_summary}
\end{table}

\subsubsection{Opinion Dynamics}
\label{subsubsec:MS_opinion_dynamics}

Here we consider an opinion dynamics model describing how individual opinions evolve through local interactions and may eventually converge toward consensus. Such models have been widely studied in sociology, control theory, and
collective behavior; see, for example,~\cite{krause2000discrete,motsch2014heterophilious} and the references therein. We study a first-order model in which each agent's state represents its opinion, and the interaction kernel determines how
strongly one agent influences another as a function of their opinion difference. The interaction kernel for this model is defined as
\begin{align}
\phi^E(r)=\begin{cases}
            1, & 0 \le r < 1/\sqrt{2},\\
            0.1, & 1/\sqrt{2} \le r \le 1,\\
            0, & r>1,
        \end{cases}
\label{eq:OP_f_env_term_and_interaction_kernel_terms}
\end{align}
with the governing equations then given by the general first-order system~\eqref{eq:model_selection_general_framework_1order}.  As summarized in Table~\ref{tab:MS_opinion_dynamics_framework_features}, the opinion dynamics model~\eqref{eq:OP_f_env_term_and_interaction_kernel_terms} contains only an energy-based interaction kernel, with no environmental force terms. Therefore, the correct framework should identify a first-order pure interaction model without $f$. The simulation and learning parameters utilized in this experiment are reported in Tables~\ref{tab:app:data_generation_parameters} and~\ref{tab:app:learning_parameters}.

\begin{table}
    \centering
    \begin{tabular}{|c|c|c|c|}
    \hline
     Order & $\phi^E$ & $\phi^A$ &  $f$ \\ \hline
     \hline
    1 & \checkmark & -- & -- \\
     \hline
    \end{tabular}
    \caption{Mechanisms of the phototaxis system~\eqref{eq:OP_f_env_term_and_interaction_kernel_terms}.  The symbol ``-" indicates that the mechanism is not present, while ``\checkmark" indicates that the mechanism is present.}
    \label{tab:MS_opinion_dynamics_framework_features}
\end{table}

\begin{figure}
    \centering

    \begin{subfigure}{0.45\textwidth}
        \centering
        \includegraphics[width=\textwidth]{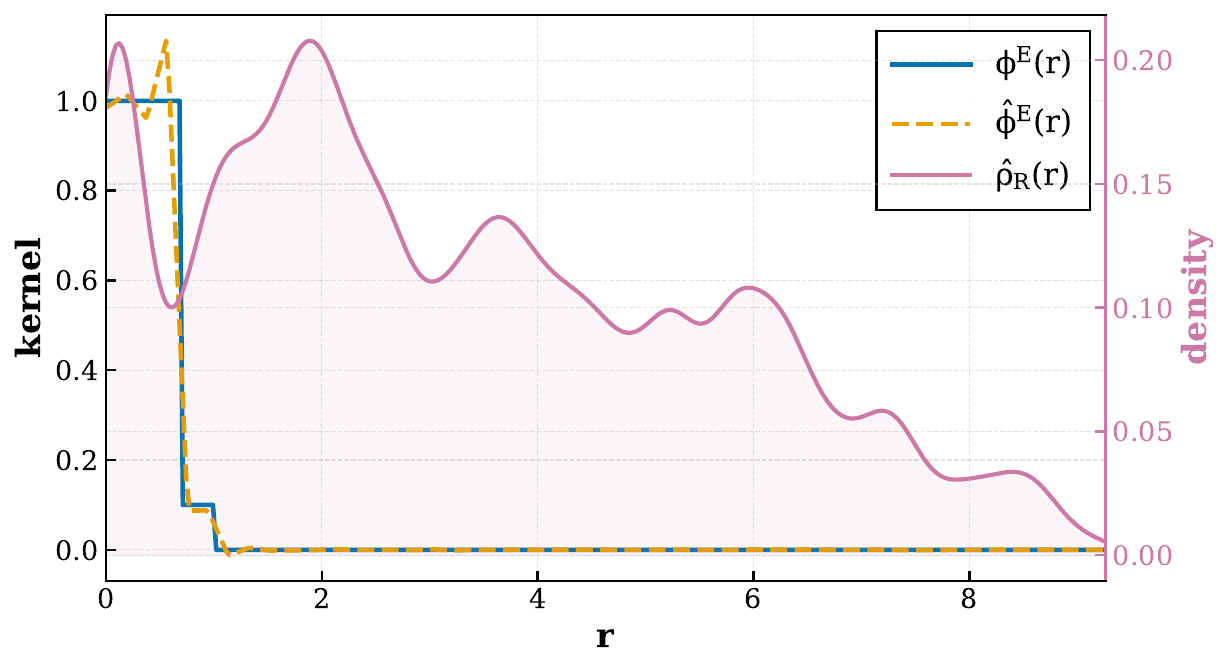}
        \caption{Interaction kernel $\phi^E$}
        \label{subfig:MS_Energy_kernel_OP}
    \end{subfigure}
    \hfill
    \begin{subfigure}{0.45\textwidth}
        \centering
        \includegraphics[width=\textwidth]{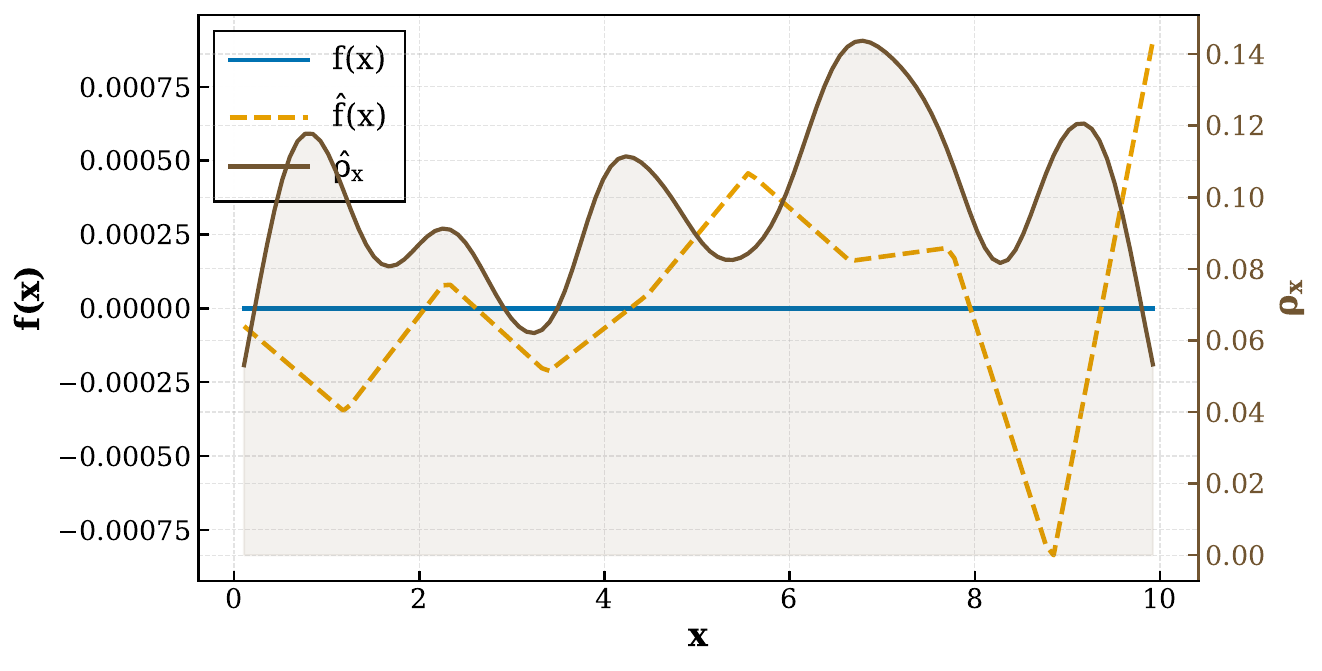}
        \caption{Environmental force $f$}
        \label{subfig:MS_f_OP}
    \end{subfigure}

    \caption{
    (Model selection, opinion dynamics) Mechanisms present for the opinion dynamics system~\eqref{eq:OP_f_env_term_and_interaction_kernel_terms} within the generalized first-order framework~\eqref{eq:model_selection_general_framework_1order}. The model selection framework estimates an energy interaction kernel $\hat{\phi}^E$ and an environmental force $\hat{f}$ from observations of trajectories, and correctly identifies a first-order model dominated by interaction forces. Estimation is performed over $T_r=10$ independent learning trials, the inferred interaction kernel has weighted $L^{2}$ $\|\hat{\phi}^{E}\|_{L^2(\hat{\rho}_R)}=$\num{3.5858e-01}$\pm$\num{1.7518e-02} (Figure~\ref{subfig:MS_Energy_kernel_OP}), while the inferred environmental force has weighted $L^{2}$ norm $\|\hat{f}\|_{L^2(\hat{\rho}_V)}=$\num{0.0000e+00}$\pm$\num{0.0000e+00} (Figure~\ref{subfig:MS_f_OP}).  Note the scale of the vertical axes.  Further information regarding learning and model parameters may be found in Appendix~\ref{app:sec:tables_of_parameters}.}
    \label{fig:MS_OP_framework_features}
\end{figure}

We estimate both mechanisms ($f$ and $\phi^{E}$) in the general first-order framework~\eqref{eq:model_selection_general_framework_1order}, with results provided in Figure~\ref{fig:MS_OP_framework_features}.  Observe that the  algorithm correctly identifies a non-zero interaction kernel $\hat{\phi}^{E}$ (Figure~\ref{subfig:MS_Energy_kernel_OP}), while obtaining a near zero estimate $\hat{f}$ (Figure~\ref{subfig:MS_f_OP}).  The algorithm also exhibits negligible variability across repeated trials, and thus consistently identifies the energy interaction as the only active feature while suppressing the inactive environmental force term (Figure~\ref{fig:MS_OP_framework_features}). The model selection framework also confirms the first-order pure interaction model as the most likely candidate; details are provided in Table~\ref{tab:OP_model_selection_summary}. Specifically, we note that all second-order ($S$) candidates are rejected by the framework compatibility gate as their relative residual errors satisfy $E_{\mathrm{res}}^{\mathrm{rel}} > 0.25$ or are undefined due to numerical instability. The remaining admissible candidates belong to the first-order ($F$) category and all satisfy the primary criterion by achieving (within tolerance $\tau_{\mathrm{traj}}$ in~\eqref{eq:model_selection_admissible}) the same minimum validation trajectory error $E_{\min}$, according to the criterion defined in~\eqref{eq:tie_candidate_treshhold}. The selection procedure therefore proceeds to the secondary criterion, i.e. framework complexity. The lowest complexity is achieved by candidate $F_1$, corresponding to the first-order energy-only framework, which is precisely the model utilized to generate the trajectory data.  Thus, even when the candidate library includes more general second-order frameworks, the procedure correctly identifies that the observed dynamics are best explained by a first-order interaction model.

\begin{table}
\centering
\small

\begin{tabular}{lrrrrrrrl}
\toprule
Candidate & Order & $E_{\mathrm{res}}^{\mathrm{rel_\epsilon}}$ & $E_{\mathrm{recon}}^{\mathrm{rel}}$ & $\bar E_{\mathrm{traj}}(M_{\mathrm{val}})$ & Complexity & $E_{\mathrm{res}}$ & Rank & Selected \\
\midrule
$F_1$ & 1 & 0.186886 & 0.006227 & 0.016430 & 1 & 0.005228 & 1 & Yes \\
$F_2$ & 1 & 0.186580 & 0.006351 & 0.015867 & 2 & 0.005220 & 2 & No \\
$S_2$ & 2 & 0.930454 & 0.020764 & 0.099662 & 3 & 0.037002 & 3 & No \\
$S_5$ & 2 & 0.927506 & 0.042013 & 0.147101 & 5 & 0.036885 & 4 & No \\
$S_1$ & 2 & 0.983127 & 0.034845 & 0.335785 & 3 & 0.039097 & 5 & No \\
$S_3$ & 2 & N/A & N/A & N/A & 4 & N/A & 6 & No \\
$S_4$ & 2 & 0.930454 & 0.020764 & 0.099662 & 4 & 0.037002 & 7 & No \\
$S_6$ & 2 & 0.927506 & 0.042069 & 0.147098 & 6 & 0.036885 & 8 & No \\
\bottomrule
\end{tabular}
\caption{
(Model selection, opinion dynamics) Summary of the candidate models considered during model selection.
For each candidate, we report the system order, framework compatibility diagnostics,
validation trajectory error, model complexity, residual error,
ranking, and whether the candidate was ultimately selected.  In this case, the framework correctly identifies the opinion dynamics model described by interactions only.  N/A values indicate errors in trajectory reconstruction due to numerical instability of obtained estimators.
}
\label{tab:OP_model_selection_summary}
\end{table}

\section{Discussion and conclusions}
\label{sec:conclusion}

In this work, we have extended variational learning methods to collective dynamical systems with both inter-agent interactions and environmental or intra-agent forces. The central idea is to exploit the structural form of collective dynamics: although the full system evolves on a high-dimensional state space, its vector field is often generated by a small number of low-dimensional functions, such as interaction kernels and environmental force laws. By learning these functions directly, rather than treating the dynamics as an arbitrary high-dimensional vector field, the proposed framework provides mechanistic estimators that are both computationally tractable and interpretable.

We introduced two extensions of the variational learning framework. The first is a semi-parametric formulation, in which the environmental force is assumed to have a known functional form depending on unknown parameters. The second is a fully non-parametric formulation, in which both the environmental force and the interaction kernels are learned from trajectory data. When prior knowledge of the environmental force is available, the semi-parametric formulation enables direct estimation of the corresponding parameters. When such information is not available, the fully non-parametric formulation provides a flexible alternative for recovering environmental forces directly from observations.

Across the benchmark systems considered in this manuscript, the proposed methods accurately recover the governing mechanisms and produce learned models that predict collective behavior beyond the training time horizon. These examples include systems with qualitatively different mechanisms, including synchronization, attraction/repulsion dynamics, velocity alignment, and
externally driven motion. The combination of low feature-recovery errors, small residual errors, accurate trajectory prediction, stability across repeated learning trials, favorable dependence on size of available training data, and robustness to observational noise suggests that the learned models capture the essential mechanistic features of the underlying dynamics.

The results also highlight the importance of the empirical sampling measures induced by the observed trajectories. Feature recovery is most accurate on regions of state space and pairwise-distance space that are well sampled by the data, while recovery may deteriorate in regions that are rarely visited. This reflects an inherent limitation of trajectory-based inference: one can only expect to learn mechanisms accurately on the portions of the state space explored by the observed dynamics. Similarly, the accuracy of the semi-parametric formulation depends on the correctness of the assumed functional form; when this structure is accurate, parameter recovery can be substantially improved, but model mis-specification can lead to biased or misleading estimates.

We also developed a model selection procedure for determining which mechanistic components are active in the observed dynamics. This procedure successfully identifies the correct governing structure in the benchmark systems considered, including models with environmental forces, energy-based interactions, alignment interactions, and combinations of these mechanisms. Thus, the framework can be used not only for force estimation, but also as a tool for mechanistic model discovery when the relevant dynamical structure is not known a priori.

More broadly, the approach developed here is not limited to the particular collective systems studied in this manuscript. The essential requirement is that the governing dynamics possess exploitable structure or symmetry, so that the high-dimensional vector field can be represented in terms of a small number of lower-dimensional functions or feature maps. Interaction kernels provide one example of such a reduction: permutation symmetry and pairwise dependence allow the dynamics of many agents to be encoded through functions of pairwise distances, relative velocities, or other low-dimensional variables. Similar ideas may be applicable to other structured dynamical systems in which physical principles, invariances, conservation laws, or known mechanistic features reduce the effective dimension of the learning problem.

Several directions remain for future work. The present study focuses primarily on homogeneous-agent systems and relatively low-dimensional interaction laws. Future work will investigate heterogeneous agents, higher-dimensional feature
dependencies, stochastic dynamics, and more complex environmental couplings. On the computational side, further study of basis construction, regularization, and hyperparameter selection may improve both accuracy and efficiency. On the
theoretical side, important questions remain concerning identifiability, convergence, approximation properties, and the relationship between the empirical sampling measures and recoverability of the underlying mechanisms.

\bibliography{env_learning_bib}{}
\bibliographystyle{unsrt}

\newpage
\appendix

\section{Variational methods for learning collective and environmental forces:  fully non-parametric approach (second-order)}
\label{sec:app:variational_learning_NP_second_order}

In this section, we describe our algorithm for inferring both the environmental and interaction forces in second-order models of collective dynamics from observed trajectory data.  Specifically, we assume that the dynamics of a system of $N$ particles are described by the following system of second-order ordinary differential equations (ODEs), for $i=1,2,\ldots , N$:
\begin{align}
\begin{split}
    \dot{x_{i}} &= v_{i} \\
    \dot{v}_{i} &= f(x_{i}, v_{i}) + \frac{1}{N}\sum_{\substack{j=1 \\ j \neq i}}^{N}\phi(|x_{j}-x_{i}|)(x_{j}-x_{i}).
\end{split}
    \label{eq:second_order_system_methods}
\end{align}
Most definitions and notations remain the same as in Section~\ref{subsec:variational_learning_NP_first_order}.
Here $x_{i}=x_{i}(t) \in \R^{d}$ denotes the state of the $i^{\text{th}}$ agent at time $t$ and $v_{i}=\dot{x}_{i}(t)$ denotes the velocity of the $i^{\text{th}}$ agent at time $t$.  The function $f:\R^{2d} \to \R^{d}$ models the environmental forces on the agents, which for simplicity we assume to be identical for all agents in the system.  For all the systems considered in this study, we found that the environmental force $f$ exhibits no dependence on the spatial variable $x$. Therefore, we simplify $f(x_i,v_i) $ to $f(v_i)$, and the corresponding mapping reduces from $f:\R^{2d}\to\R^{d} $ to $f:\R^{d}\to\R^{d}$.

\subsection{Trajectory data}
\label{subsec:app:trajectory_data_second_order}

For trajectory data, which will be utilized to construct the estimators for $f$ and $\phi$ in~\eqref{eq:second_order_system_methods}, we will now use both state space and velocity of the system.  Hence the full state of the system with velocity at time $t$ denote by
\begin{align}
    x(t) &:= \begin{pmatrix}
        x_{1}(t) \\
        x_{2}(t) \\
        \vdots \\
        x_{N}(t) 
    \end{pmatrix} \in \RdN, \qquad
    v(t):= \begin{pmatrix}
        v_{1}(t) \\
        v_{2}(t) \\
        \vdots \\
        v_{N}(t) 
    \end{pmatrix} \in \RdN.
    \label{eq:second_order_xv_def}
\end{align}
Thus, $ F_{f}$ becomes,
\begin{align}
    F_{f}(x, v) &:= \begin{pmatrix}
        f(x_{1}, v_{1}) \\
        f(x_{2}, v_{2}) \\
        \vdots \\
        f(x_{N}, , v_{N})
    \end{pmatrix} \in \RdN \quad \quad
    \label{eq:second_order_f_phi_stack}
\end{align}

 with $ F_{\phi}(x)$ remains same as in Section~\ref{subsubsec:trajectory_data}. Hence, the general form  of ~\eqref{eq:second_order_system_methods} can be denote as
\begin{align}
    \dot{v} &= F_{f}(x,v) + F_{\phi}(x).
    \label{eq:second_order_stacked}
\end{align}
We denote $x_{i} \in \R^{d}$, $v_{i} \in \R^{d}$ as the $i^{\text{th}}$ components of $x,v \in \R^{d}$, and will utilize this notation to precisely define the form of the observation data and estimation algorithm below.

Now the initial conditions denote via  
\begin{align}
    X^{(m)}_{0}, V^{(m)}_0 \in \RdN,
    \label{eq:XV_m_0}
\end{align}
for $m=1,2,\ldots , M$.  For each such $m$, denote the solution at the time $t$ of the corresponding initial-value problem (IVP)
\begin{align}
    \begin{cases}
     \dot{x} &= v\\
    \dot{v} &= F_{f}(x,v) + F_{\phi}(x) \\
    (x(0),v(0)) &=  (X^{(m)}_{0}, V^{(m)}_0)
    \end{cases}
    \label{eq:IVP_second_order}
\end{align}
as $(x^{(m)}(t), (v^{(m)}(t))$.  We assume that the experimental data consists of observations at discrete time points $0=t_{1} < t_{2} <  \cdots < t_{L}=T$, which we denote as
\begin{align}
    \Xml &:= x^{(m)}(t_{\ell})\\
    \Vml &:= \dot{x}^{(m)}(t_{\ell}),
    \label{eq:XV_m_l}
\end{align}
As before, $\Xml, \Vml \in \RdN$, and thus the set $(\Xml, \Vml)_{m,\ell=1}^{M,L}$ is generally the set of observed trajectory data which will be utilized to obtain the estimators for $f$ and $\phi$.  Note that by construction,
\begin{align}
(\Xml)_{i} &= x^{(m)}_{i}(t_{\ell}) \in \R^{d},\\
(\Vml)_{i} &= v^{(m)}_{i}(t_{\ell}) \in \R^{d},
\label{eq:i_component_Xmla_and_Vml}
\end{align}
i.e. that the $i^{\text{th}}$ component (with respect to the natural decomposition of $\Xml, \Vml$ via~\eqref{eq:second_order_xv_def}) of $\Xml, \Vml$ is the state of agent $i$ at time $t_{\ell}$ of the $m^{\text{th}}$ replicate and the velocity of agent $i$  at time $t_{\ell}$ of the $m^{\text{th}}$ replicate respectively.

The learning algorithm introduced in Section~\ref{subsec:app:estimation_algorithm_second_order} will also require observations of the $\ddot{x}=\dot{v}$ in~\eqref{eq:second_order_stacked}.  Specifically, we define
\begin{align}
    \Aml &:= \dot{v}^{(m)}(t_{\ell}),
    \label{eq:a_m_l}
\end{align}
which can be approximated from $(\Vml)_{m,\ell=1}^{M,L}$ as discussed in Section~\ref{subsubsec:trajectory_data} for the velocities of the first-order system.

\subsection{Estimation algorithm}
\label{subsec:app:estimation_algorithm_second_order}

Analogously to the algorithm presented in Section~\ref{subsubsec:estimation_algorithm}, we define an error functional of the form
\begin{align}
    \mathcal{E}_{\Hf,\Hphi}(\hat{f},\hat{\phi}) &:= \frac{1}{ML}\sum_{m,\ell=1}^{M,L}\norm*{ \Aml - \Big(F_{\hat{f}}(\Xml, \Vml)+F_{\hat{\phi}}(\Xml)\Big)}_{\RdN}^{2},
    \label{eq:second_order_error_functional}
\end{align}
where $(\Xml,\Vml,\Aml)_{m,\ell=1}^{M,L}$ denotes the (fixed) trajectory data as introduced in Section~\ref{subsec:app:trajectory_data_second_order}

For the second-order systems $h_{k}$ utilize both $(x,v)$ as in~\eqref{eq:second_order_f_phi_stack}. Hence, for each $1 \leq k \leq \nf$, writing $x=(x_{1},x_{2},\ldots , x_{N}) \in \R^{dN}$ with $x_{i} \in \R^{d}$ and $v=(v_{1},v_{2},\ldots , v_{N}) \in \R^{dN}$ with $v_{i} \in \R^{d}$, we define
\begin{align}
    h_{k}(x, v) &:= \begin{pmatrix}
        h_{k}(x_{1}, v_{1}) \\
        h_{k}(x_{2}, v_{2}) \\
        \cdots \\
        h_{k}(x_{N}, v_{N})
    \end{pmatrix}
    \label{eq:second_order_hk_stacked}
\end{align}
and the matrix $\Hml \in \R^{dN  \times \nf}$
\begin{align}
    \Hml &:= \Big( \, h_{1}(\Xml, \Vml) \quad h_{2}(\Xml, \Vml) \quad \cdots \quad h_{\nf}(\Xml, \Vml) \, \Big).
    \label{eq:Hml_2nd_order}
\end{align}

 Thus, expanding $\tilde{f} \in \Hf$ as in~\eqref{eq:basis_coefficients}, $F_{\tilde{f}}(\Xml, \Vml)$ in~\eqref{eq:second_order_error_functional} takes the form of matrix-vector multiplication:
\begin{align}
    F_{\tilde{f}}(\Xml, \Vml) &= \Hml \alpha.
    \label{eq:second_order_F_f_ml_simple}
\end{align}

Using~\eqref{eq:second_order_F_f_ml_simple} and~\eqref{eq:F_phi_ml_simple}, we observe that minimizing~\eqref{eq:second_order_error_functional} over $\Hf$ and $\Hphi$ is equivalent to minimizing
\begin{align}
    \mathcal{E}(\alpha,\beta) &:= \frac{1}{ML}\sum_{m,\ell=1}^{M,L}\norm*{\Aml - \Big( \Hml \alpha + \Psiml \beta \Big)}_{\RdN}^{2},
    \label{eq:second_order_error_functional_simple}
\end{align}
over $(\alpha,\beta) \in \R^{\nf +\nphi}$.  As~\eqref{eq:second_order_error_functional_simple} is quadratic in $(\alpha,\beta)$ (recall that the norm $\norm*{\cdot}_{\RdN}$ is defined in~\eqref{eq:inp_RdN}~-~\eqref{eq:norm_RdN} via the standard Euclidean inner product on $\R^{d}$), a solution $(\hat{\alpha},\hat{\beta})$ always exists:
\begin{align}
    (\hat{\alpha},\hat{\beta}) &:= \argmin_{\alpha \in \R^{\nf}, \, \beta \in \R^{\nphi}}\mathcal{E}(\alpha,\beta).
    \label{eq:second_order_alpha_hat_beta_hat_def}
\end{align}

The normal equations which characterize the solutions of ~\eqref{eq:second_order_alpha_hat_beta_hat_def} and the solution approach is same as in section ~\ref{subsubsec:normal}.

\subsection{Algorithm pseudocode}
\label{subsec:app:algorithm_2nd_order}

\begin{algorithm}[H]
\DontPrintSemicolon
\SetAlgoLined
\KwIn{$( \Xml, \Vml  \Aml )_{m, \ell=1}^{M,L}$ }
\KwOut{Estimators: $\hat{\phi}$, $\hat{f}$}

\BlankLine
Construct pairwise quantities and estimate the observed supports\;
\Indp Generate interaction distances $( \Rml )_{m,\ell=1}^{M,L}$ 
\BlankLine
Find the maximum and minimum interaction radii $R_{\min}, R_{\max} \in \R$\;
Find the maximum and minimum values of state variable and velocity variable $X_{\min}, X_{\max}, V_{\min}, V_{\max} \in \R^d$\;
Observed supports$ [ R_{\min}, R_{\max}], \prod_{k=1}^{d}[X_{k,\min},, X_{k,\max}]\times \prod_{k=1}^{d}[V_{k,\min}, V_{k,\max}]$\;
\Indm
\BlankLine
Construct localized basis functions \;
\Indp Construct the kernel basis $(\psi_{k})_{k=1}^{\nphi}$\;
    Construct $(h_{k})_{k=1}^{\nf}$ \;
\Indm
Create $\Cml$ by column-wise concatenating $( \,\Hml \quad \Psiml \,)$ as in ~\eqref{eq:Aml}\;
Assemble $A\hat{\theta}=\vec{b}$ (in parallel for $\ell, m$), using ~\eqref{eq:normal_eq_v1}, ~\eqref{eq:normal_eq_mat_vec}, ~\eqref{eq:normal_eq_v2}, ~\eqref{eq:theta}\;
Solve for $\hat{\theta}$ \;
\BlankLine
Assemble $\hat{\phi}$ as in ~\eqref{eq:basis_coefficients}\;
Assemble $\hat{f} = \sum_{k =1}^{\nf} \alpha_{k}h_{k}$ \;

\caption{Algorithm for learning $\phi$ (interaction) and $f$ (environmental) forces from trajectory data of second-order systems of the form~\eqref{eq:second_order_system_methods}.}
\label{alg:2nd_order_variational_algorithm_for_kernel_and_fenv_NP}
\end{algorithm}

\section{Hypothesis spaces and empirical measures}
\label{app:sec:system_profiles}

Here we provide details on both the hypothesis spaces (Section~\ref{subsubsec:hypothesis_space_selection}) and trajectory-induced empirical measures (Section~\ref{subsec:measure_on_data}) for the model systems considered in Section~\ref{subsec:results_nonparametric_vs_parametric_learning}.

\subsection{Kuramoto system}
\label{app:subsec:kuramoto}

\begin{figure}
    \centering
    \begin{subfigure}[b]{0.45\textwidth}
        \centering
        \includegraphics[width=\linewidth]{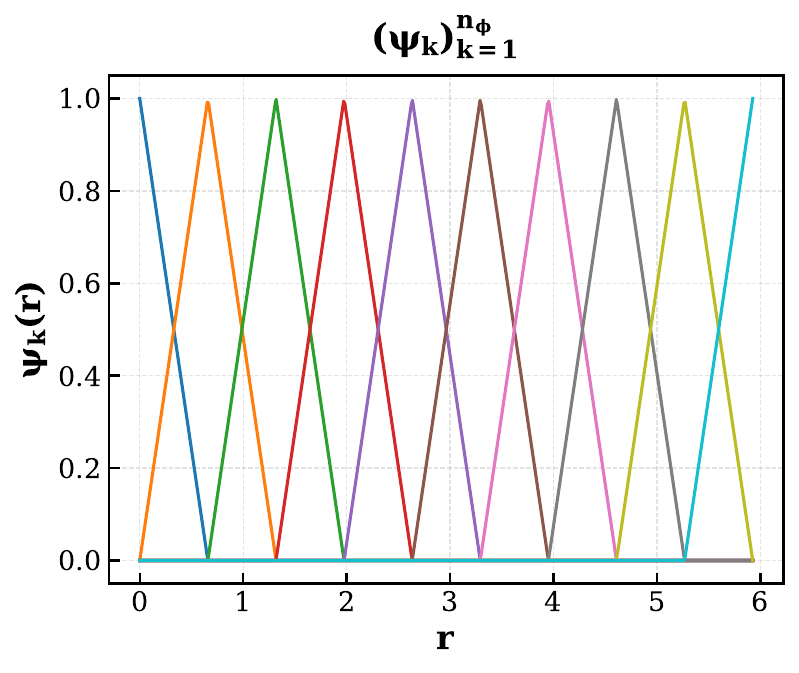}
        \caption{Basis functions defining $\mathcal{H}_{\phi}$}
        \label{fig:Kuramoto_radial_basis_functions}
    \end{subfigure}
    \hfill
    \begin{subfigure}[b]{0.45\textwidth}
        \centering
        \includegraphics[width=\linewidth]{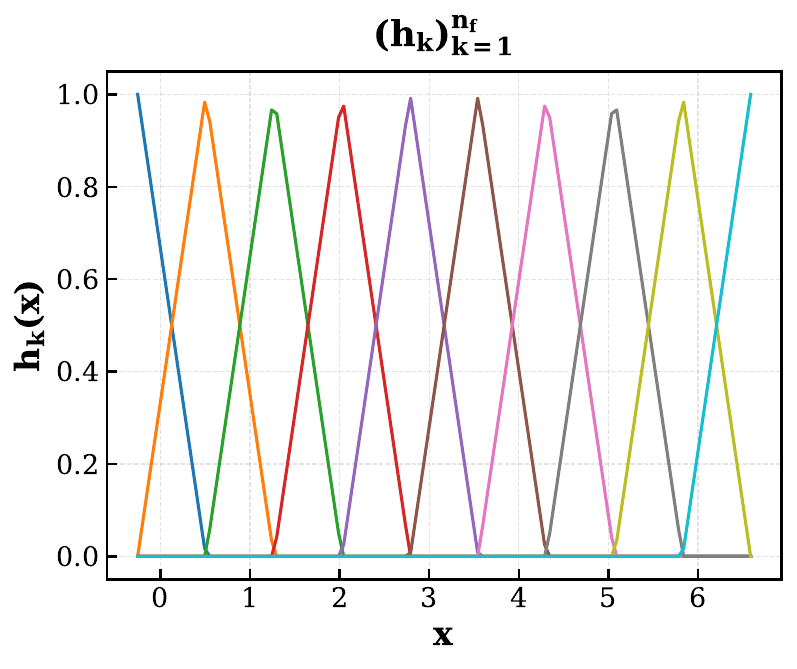}
        \caption{Basis functions defining $\mathcal{H}_{f}$}
         \label{fig:Kuramoto_f_space_basis_functions}
    \end{subfigure}
   \caption{(Kuramoto model) Linear B-spline basis functions used to construct hypothesis spaces. Subfigure~\ref{fig:Kuramoto_radial_basis_functions} plots the interaction kernel basis functions $(\psi_k)_{k=1}^{n_{\phi}}$ in $H_{\phi}$ with $n_{\phi}=10$. Subfigure~\ref{fig:Kuramoto_f_space_basis_functions} plots the environmental force basis functions $(h_k)_{k=1}^{n_f}$ in $H_f$ with $n_f=10$.}
    \label{fig:Kuramoto_basis_functions}
\end{figure}

\begin{figure}
    \centering
    \begin{subfigure}[b]{0.45\textwidth}
        \centering
        \includegraphics[width=\linewidth]{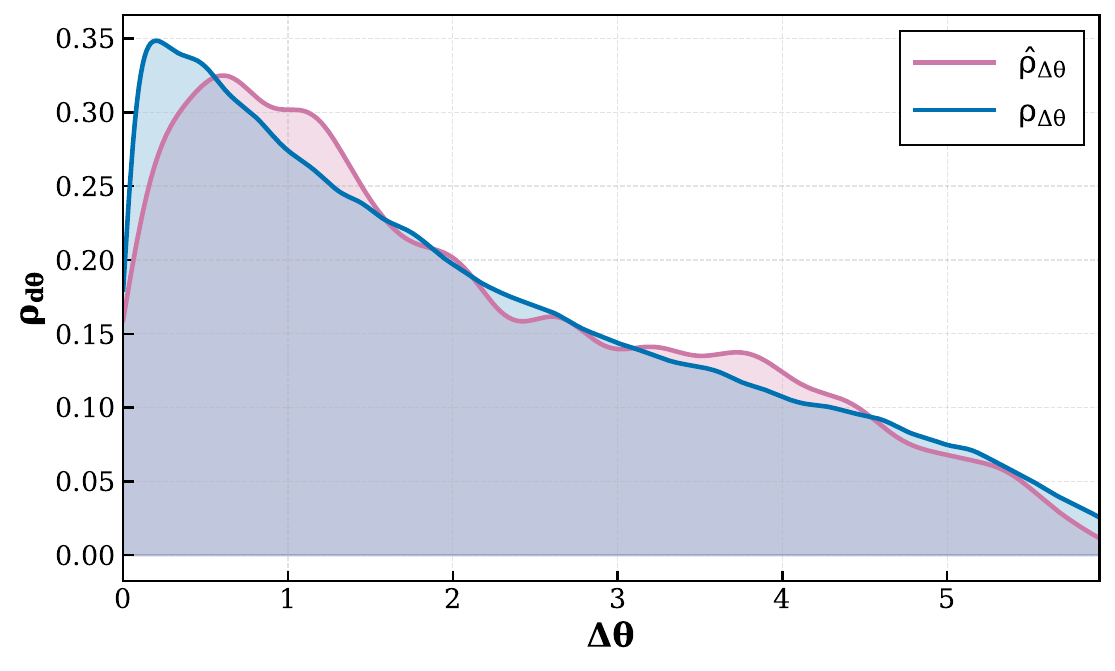}
        \caption{Induced measure on $\Delta \theta$}
        \label{fig:Kuramoto_measure_in_radial_Space}
    \end{subfigure}
    \hfill
    \begin{subfigure}[b]{0.45\textwidth}
        \centering
        \includegraphics[width=\linewidth]{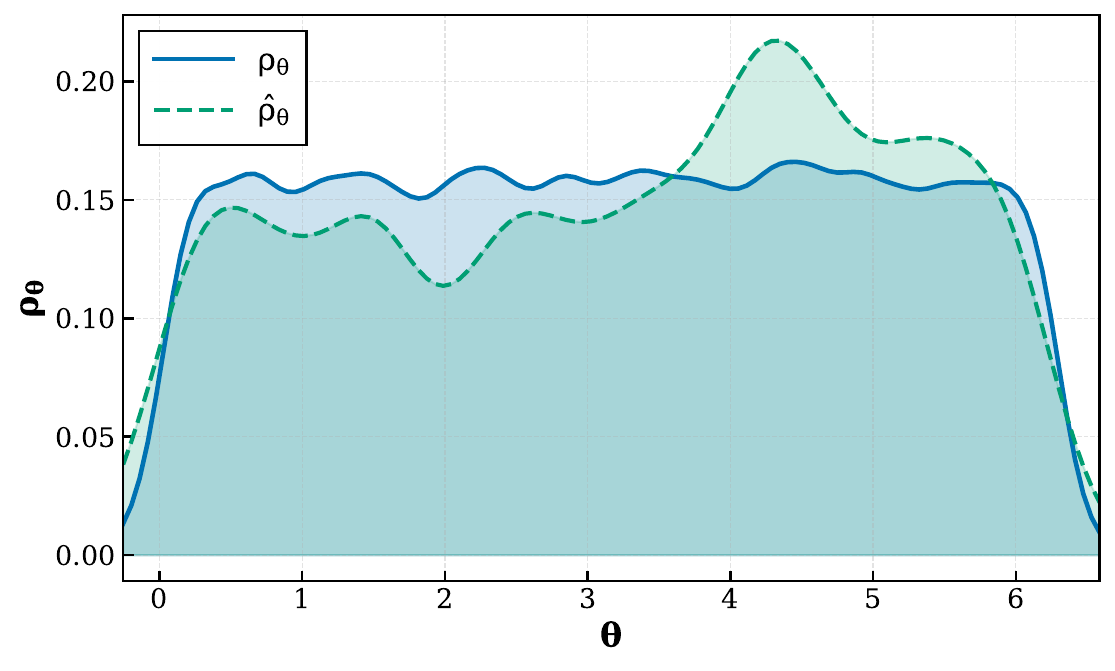}
        \caption{Induced measure on $\theta$}
        \label{fig:Kuramoto_measure_in_state_space}
    \end{subfigure}
    \caption{(Kuramoto model) Comparison of the empirical measures obtained from the $M_{\mathrm{train}}$ training data and the reference probability measures from $M_{\rho}$ samples, as discussed in Section~\ref{subsec:measure_on_data}. See Table~\ref{tab:app:data_generation_parameters} for parameters utilized.  Subfigure~\ref{fig:Kuramoto_measure_in_radial_Space} plots $\hat{\rho}_{\Delta\theta}$ versus $\rho_{\Delta\theta}$, and Subfigure~\ref{fig:Kuramoto_measure_in_state_space} plots $\hat{\rho}_{\theta}$ versus $\rho_{\theta}$. The close agreement suggests that the training trajectory data provides an accurate estimate for the regions explored by the dynamics.}
    \label{fig:Kuramoto_observation_measures}
\end{figure}

The following visualizations illustrate the basis functions for the hypothesis spaces (Figure~\ref{fig:Kuramoto_basis_functions}) and induced measures (Figure~\ref{fig:Kuramoto_observation_measures}) utilized for the Kuramoto model (Section~\ref{subsubsec:results_kuramoto}), as described in Sections~\ref{subsubsec:hypothesis_space_selection} and~\ref{subsec:measure_on_data}, respectively. These quantities define the hypothesis spaces and empirical measures employed by the estimation algorithms.

\subsection{Phototaxis system}
\label{app:subsec:phototaxis}

\begin{figure}
    \centering
    \begin{subfigure}[b]{0.35\textwidth}
        \centering
        \includegraphics[width=\linewidth]{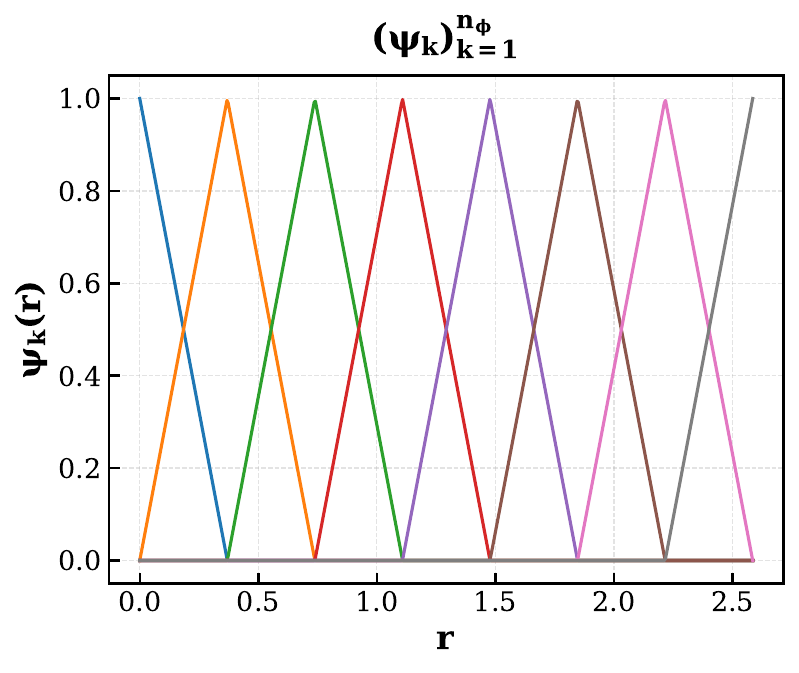}
        \caption{Basis functions defining $\mathcal{H}_{\phi}$}
         \label{fig:Phototaxis_radial_basis_functions}
    \end{subfigure}
    \hfill
    \begin{subfigure}[b]{0.64\textwidth}
        \centering
        \includegraphics[width=\linewidth]{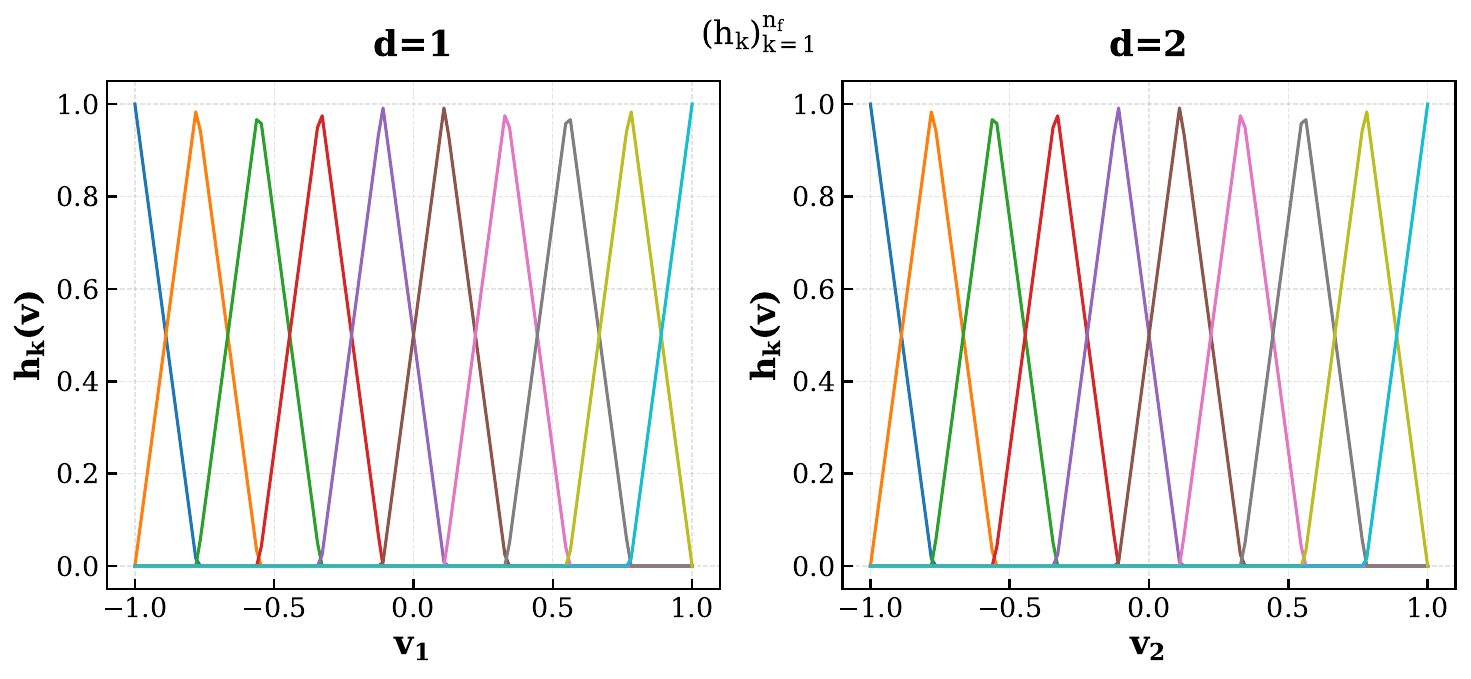}
        \caption{Basis functions defining $\mathcal{H}_{f}$}
         \label{fig:Phototaxis_f_space_basis_functions}
    \end{subfigure}
    \caption{(Phototaxis model) Linear B-spline basis functions used to construct hypothesis spaces. Subfigure~\ref{fig:Phototaxis_radial_basis_functions} plots the interaction kernel basis functions $(\psi_k)_{k=1}^{n_{\phi}}$ in $H_{\phi}$ with $n_{\phi}=8$. Subfigure~\ref{fig:Phototaxis_f_space_basis_functions} plots the environmental force basis functions $(h_k)_{k=1}^{n_f}$ in $H_f$, where the velocity space is discretized using $n_b=10$ basis functions in each dimension, resulting in $n_f=10\times10=100$ tensor product basis functions for $\mathcal{H}_{f}$.
    }
    \label{fig:Phototaxis_basis_functions}
\end{figure}

\begin{figure}
    \centering
    \begin{subfigure}[b]{0.35\textwidth}
        \centering
        \includegraphics[width=\linewidth]{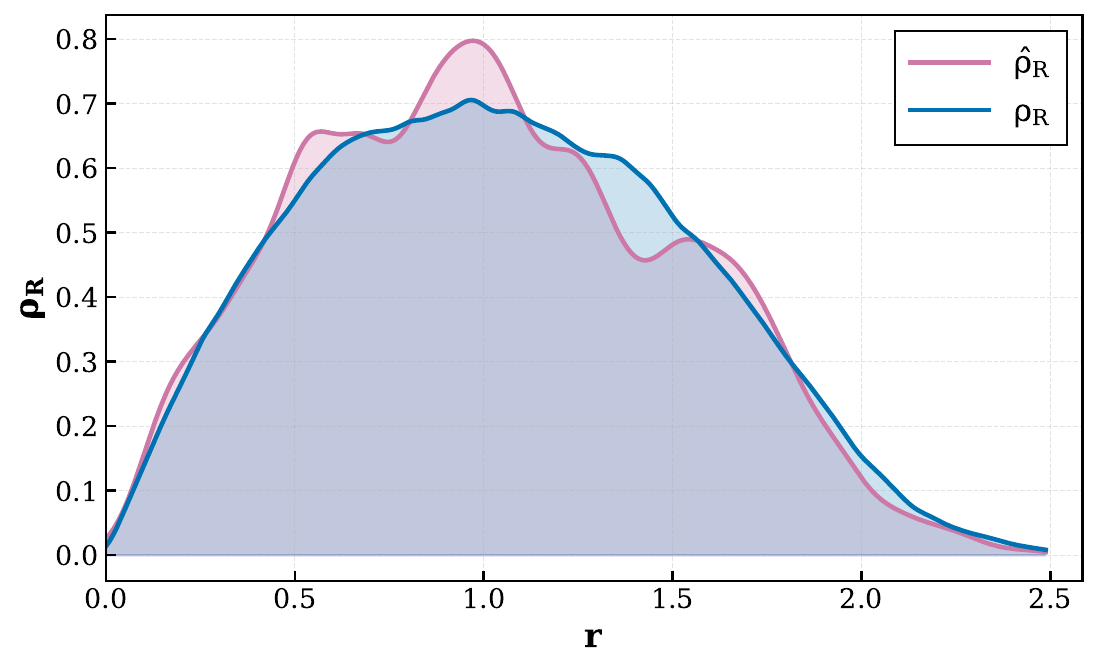}
        \caption{Induced measure on $R$}
        \label{fig:Phototaxis_measure_in_radial_Space}
    \end{subfigure}
    \hfill
    \begin{subfigure}[b]{0.64\textwidth}
        \centering
        \includegraphics[width=\linewidth]{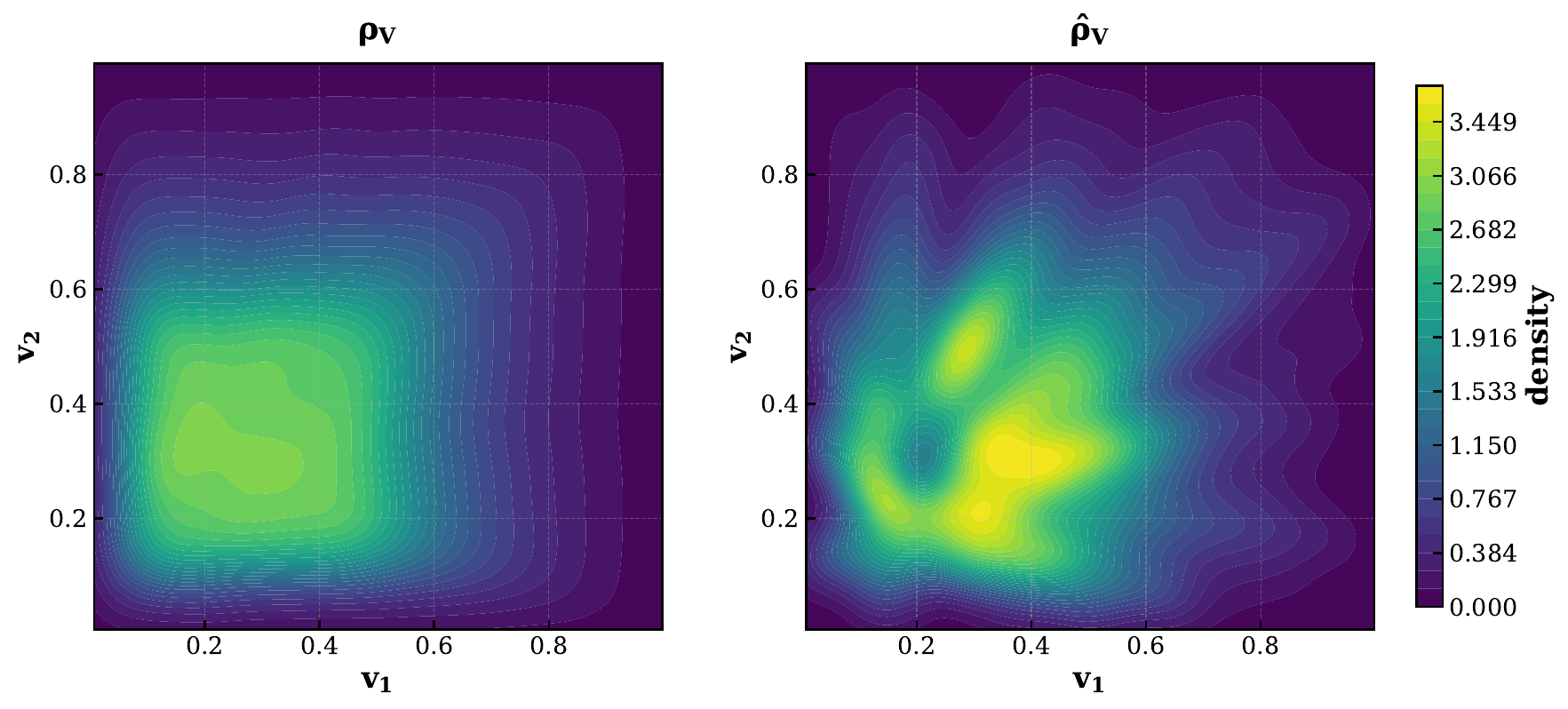}
        \caption{Induced measure on $V$}
        \label{fig:Phototaxis_measure_in_state_space}
    \end{subfigure}
   \caption{(Phototaxis model) Comparison of the empirical measures obtained from the $M_{\mathrm{train}}$ training data and the reference measure from $M_{\rho}$ samples, as discussed in Section~\ref{subsec:measure_on_data}. See Table ~\ref{tab:app:data_generation_parameters} for parameters utilized.   Subfigure~\ref{fig:Phototaxis_measure_in_radial_Space} plots $\hat{\rho}_{R}$ versus $\rho_{R}$, and Subfigure~\ref{fig:Phototaxis_measure_in_state_space} plots $\hat{\rho}_{V}$ versus $\rho_{V}$. The close agreement suggests that the training trajectory data provides an accurate estimate for the regions explored by the dynamics.}
   \label{fig:Phototaxis_observation_measures}
\end{figure}

The following visualizations illustrate the basis functions for the hypothesis spaces (Figure~\ref{fig:Phototaxis_basis_functions}) and induced measures (Figure~\ref{fig:Phototaxis_observation_measures}) utilized for the phototaxis model (Section~\ref{subsubsec:results_phototaxis}) as described in Sections~\ref{subsubsec:hypothesis_space_selection} and~\ref{subsec:measure_on_data}, respectively. These quantities define the hypothesis spaces and empirical measures employed by the learning algorithm.

\subsection{Self-propelled particle (SPP) system}
\label{app:subsec:SPP}

\begin{figure}
    \centering
    \begin{subfigure}[b]{0.35\textwidth}
        \centering
        \includegraphics[width=\linewidth]{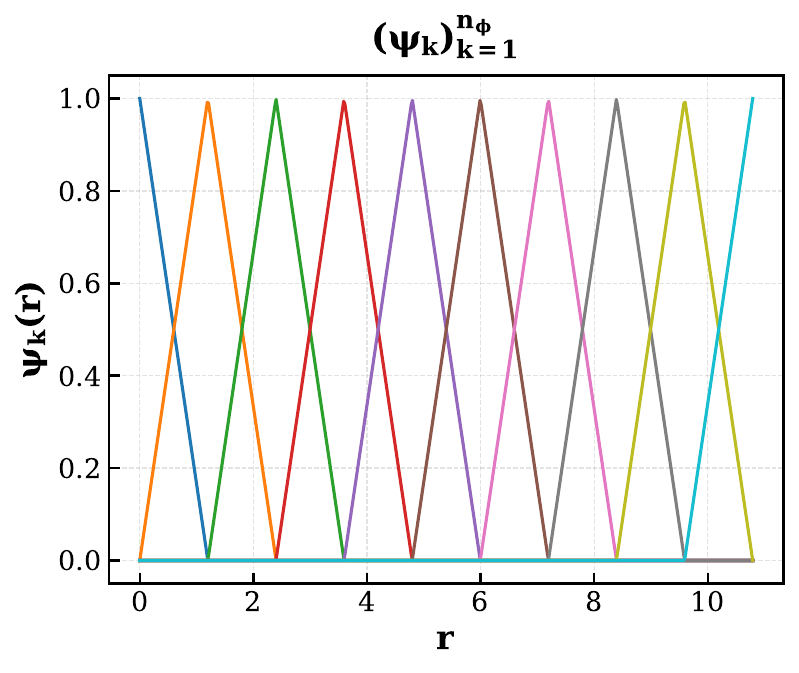}
        \caption{Basis functions defining $\mathcal{H}_{\phi}$}
         \label{fig:SPP_radial_basis_functions}
    \end{subfigure}
    \hfill
    \begin{subfigure}[b]{0.64\textwidth}
        \centering
        \includegraphics[width=\linewidth]{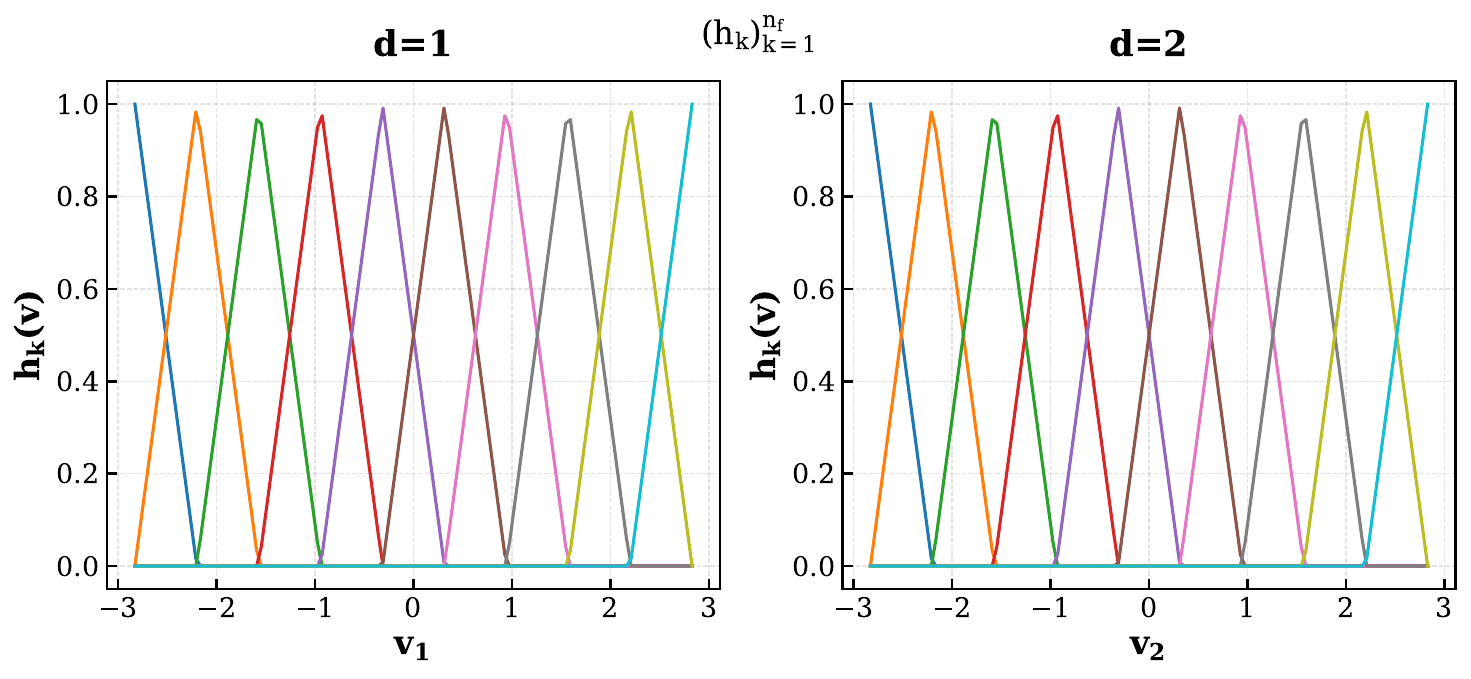}
        \caption{Basis functions defining $\mathcal{H}_{f}$}
         \label{fig:SPP_f_space_basis_functions}
    \end{subfigure}
    \caption{(SPP model) Linear B-spline basis functions used to construct the hypothesis spaces. Subfigure~\ref{fig:SPP_radial_basis_functions} plots the interaction kernel basis functions $(\psi_k)_{k=1}^{n_{\phi}}$ in $H_{\phi}$ with $n_{\phi}=8$. Subfigure~\ref{fig:Phototaxis_f_space_basis_functions} plots the environmental force basis functions $(h_k)_{k=1}^{n_f}$ in $H_f$. The velocity space is discretized using $n_b=10$ basis functions in each dimension, resulting in $n_f=10\times10=100$ tensor product basis functions for $\mathcal{H}_{f}$.
    }
    \label{fig:SPP_basis_functions}
\end{figure}

\begin{figure}
    \centering
    \begin{subfigure}[b]{0.35\textwidth}
        \centering
        \includegraphics[width=\linewidth]{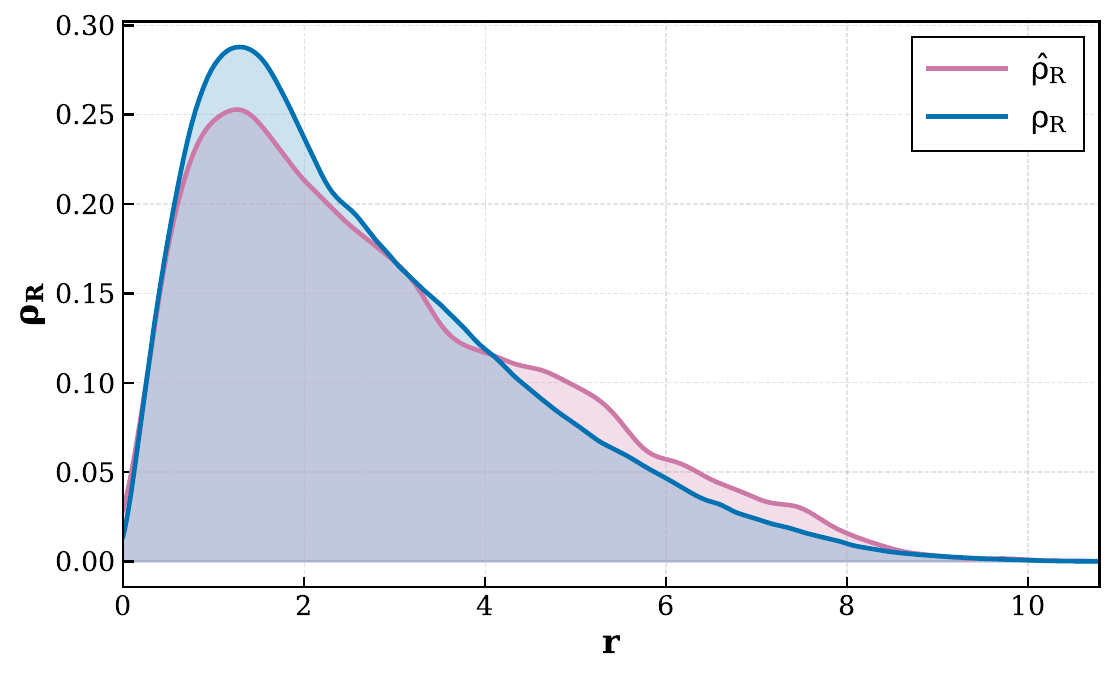}
        \caption{Induced measure on $R$}
        \label{fig:SPP_measure_in_radial_Space}
    \end{subfigure}
    \hfill
    \begin{subfigure}[b]{0.64\textwidth}
        \centering
        \includegraphics[width=\linewidth]{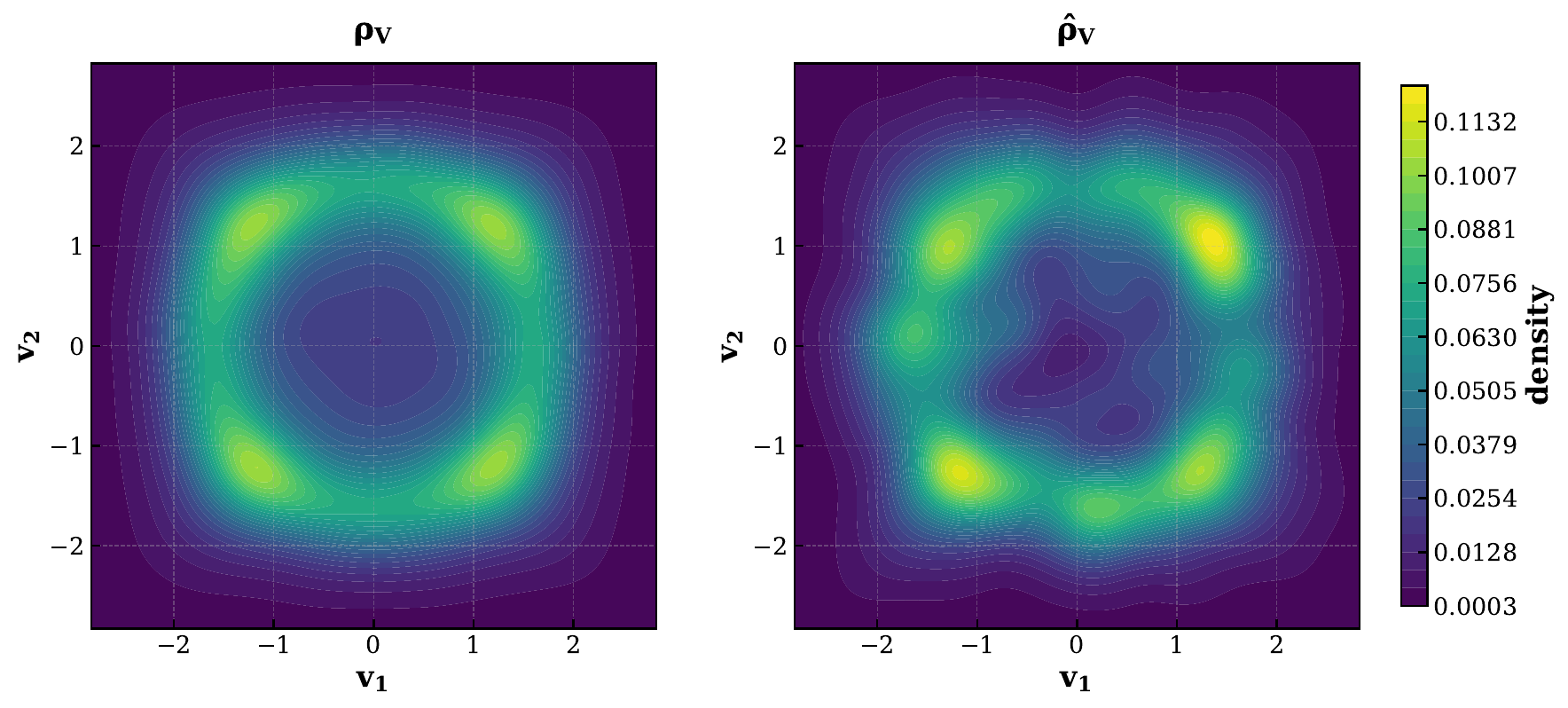}
        \caption{Induced measure on $V$}
        \label{fig:SPP_measure_in_state_space}
    \end{subfigure}
   \caption{(SPP) Comparison of the empirical measures obtained from the $M_{\mathrm{train}}$ training data and the reference  measure from $M_{\rho}$ samples, as discussed in Section~\ref{subsec:measure_on_data}. See Table ~\ref{tab:app:data_generation_parameters} for parameters utilized.   Subfigure~\ref{fig:SPP_measure_in_radial_Space} plots $\hat{\rho}_{R}$ versus $\rho_{R}$, and Subfigure~\ref{fig:SPP_measure_in_state_space} plots $\hat{\rho}_{V}$ versus $\rho_{V}$. The close agreement suggests that the training trajectory data provides an accurate estimate for the regions explored by the dynamics.}
   \label{fig:SPP_observation_measures}
\end{figure}

The following visualizations illustrate the basis functions for the hypothesis spaces (Figure~\ref{fig:SPP_basis_functions}) and induced measures (Figure~\ref{fig:SPP_observation_measures}) utilized for the self-propelled particle (SPP) model (Section~\ref{subsubsec:results_SPP}) as described in Sections~\ref{subsubsec:hypothesis_space_selection} and~\ref{subsec:measure_on_data}, respectively. These quantities define the hypothesis spaces and empirical measures employed by the learning algorithm.

\section{Metric evaluations}
\label{app:sec:metric_evaluation}
We provide tables quantifying estimation accuracy as described in Section~\ref{sec:evaluation_methodology}, as well as the effect of measurement noise on reconstruction.

\subsection{Dependence on training data}

\begin{longtable}{rrcc}
\toprule
M & L & $E_{\phi}^{\mathrm{rel}}$ & $E_{f}^{\mathrm{rel}}$ \\
\midrule
\endfirsthead
\toprule
M & L & $E_{\phi}^{\mathrm{rel}}$ & $E_{f}^{\mathrm{rel}}$ \\
\midrule
\endhead
\midrule
\multicolumn{4}{r}{Continued on next page} \\
\midrule
\endfoot
\bottomrule
\endlastfoot
1 & 51 & $\num{1.3881e+00}\pm\num{2.2499e+00}$ & $\num{3.6814e+01}\pm\num{6.4836e+01}$ \\
1 & 101 & $\num{1.8696e+00}\pm\num{2.0156e+00}$ & $\num{4.8435e+01}\pm\num{5.6091e+01}$ \\
1 & 201 & $\num{2.8650e+00}\pm\num{3.9872e+00}$ & $\num{6.4122e+01}\pm\num{8.9776e+01}$ \\
1 & 401 & $\num{1.7684e+00}\pm\num{2.3126e+00}$ & $\num{4.2537e+01}\pm\num{5.9298e+01}$ \\
1 & 801 & $\num{2.6533e+00}\pm\num{3.9945e+00}$ & $\num{7.0362e+01}\pm\num{1.1683e+02}$ \\
4 & 51 & $\num{1.2803e-02}\pm\num{2.1729e-03}$ & $\num{8.1026e-02}\pm\num{6.6889e-02}$ \\
4 & 101 & $\num{1.1935e-02}\pm\num{2.5299e-03}$ & $\num{7.9669e-02}\pm\num{5.4188e-02}$ \\
4 & 201 & $\num{1.2520e-02}\pm\num{1.2240e-03}$ & $\num{7.9739e-02}\pm\num{2.9116e-02}$ \\
4 & 401 & $\num{1.2743e-02}\pm\num{2.0112e-03}$ & $\num{7.7073e-02}\pm\num{4.9390e-02}$ \\
4 & 801 & $\num{1.4557e-02}\pm\num{3.2833e-03}$ & $\num{1.4730e-01}\pm\num{1.3601e-01}$ \\
16 & 51 & $\num{1.1944e-02}\pm\num{6.2502e-04}$ & $\num{2.6885e-02}\pm\num{7.5355e-03}$ \\
16 & 101 & $\num{1.1872e-02}\pm\num{7.0650e-04}$ & $\num{2.7195e-02}\pm\num{6.0489e-03}$ \\
16 & 201 & $\num{1.1717e-02}\pm\num{5.9480e-04}$ & $\num{2.4590e-02}\pm\num{1.0248e-02}$ \\
16 & 401 & $\num{1.1988e-02}\pm\num{3.4998e-04}$ & $\num{2.0456e-02}\pm\num{7.1757e-03}$ \\
16 & 801 & $\num{1.1814e-02}\pm\num{9.0989e-04}$ & $\num{2.7961e-02}\pm\num{9.1794e-03}$ \\
64 & 51 & $\num{1.1989e-02}\pm\num{3.8711e-04}$ & $\num{1.3728e-02}\pm\num{5.0689e-03}$ \\
64 & 101 & $\num{1.2009e-02}\pm\num{5.0137e-04}$ & $\num{1.2694e-02}\pm\num{3.9811e-03}$ \\
64 & 201 & $\num{1.2031e-02}\pm\num{4.0502e-04}$ & $\num{1.1495e-02}\pm\num{4.8004e-03}$ \\
64 & 401 & $\num{1.2195e-02}\pm\num{3.7997e-04}$ & $\num{1.3889e-02}\pm\num{5.5731e-03}$ \\
64 & 801 & $\num{1.2176e-02}\pm\num{3.7529e-04}$ & $\num{1.3101e-02}\pm\num{2.5753e-03}$ \\
256 & 51 & $\num{1.2236e-02}\pm\num{1.8152e-04}$ & $\num{7.1149e-03}\pm\num{1.6387e-03}$ \\
256 & 101 & $\num{1.2238e-02}\pm\num{2.0142e-04}$ & $\num{6.9159e-03}\pm\num{2.1138e-03}$ \\
256 & 201 & $\num{1.2295e-02}\pm\num{1.7811e-04}$ & $\num{6.6019e-03}\pm\num{2.4540e-03}$ \\
256 & 401 & $\num{1.2287e-02}\pm\num{1.3753e-04}$ & $\num{6.6174e-03}\pm\num{2.2207e-03}$ \\
256 & 801 & $\num{1.2283e-02}\pm\num{1.3878e-04}$ & $\num{6.1803e-03}\pm\num{1.9047e-03}$ \\

\caption{(Kuramoto model) Kuramoto relative interaction kernel and environmental force errors for different $M$ (number of replicates) and $L$ (number of time samples). Each entry reports the mean $\pm$ sample standard deviation over $T_{r}=10$ independent learning trials.}
\label{tab:kuramoto_ML_sweep_relative_errors_mean_std}
\end{longtable}

\begin{longtable}{rrcc}

\toprule
M & L & $E_{\phi}^{\mathrm{rel}}$ & $E_{f}^{\mathrm{rel}}$ \\
\midrule
\endfirsthead
\toprule
M & L & $E_{\phi}^{\mathrm{rel}}$ & $E_{f}^{\mathrm{rel}}$ \\
\midrule
\endhead
\midrule
\multicolumn{4}{r}{Continued on next page} \\
\midrule
\endfoot
\bottomrule
\endlastfoot
1 & 51 & $\num{6.1893e-03}\pm\num{6.2809e-03}$ & $\num{6.1180e-02}\pm\num{1.3356e-02}$ \\
1 & 101 & $\num{4.4905e-03}\pm\num{4.3087e-03}$ & $\num{1.1115e-01}\pm\num{8.8252e-02}$ \\
1 & 201 & $\num{1.3141e-02}\pm\num{2.9193e-02}$ & $\num{8.5239e-02}\pm\num{3.5008e-02}$ \\
1 & 401 & $\num{5.4909e-03}\pm\num{3.4038e-03}$ & $\num{8.5608e-02}\pm\num{3.6644e-02}$ \\
1 & 801 & $\num{8.6348e-03}\pm\num{1.2248e-02}$ & $\num{7.7069e-02}\pm\num{1.3662e-02}$ \\
4 & 51 & $\num{1.9382e-04}\pm\num{3.1394e-05}$ & $\num{1.2214e-02}\pm\num{9.3570e-03}$ \\
4 & 101 & $\num{1.8060e-04}\pm\num{2.3614e-05}$ & $\num{1.0234e-02}\pm\num{6.3196e-03}$ \\
4 & 201 & $\num{1.8843e-04}\pm\num{3.3100e-05}$ & $\num{9.0533e-03}\pm\num{7.2027e-03}$ \\
4 & 401 & $\num{2.1438e-04}\pm\num{3.3710e-05}$ & $\num{1.5331e-02}\pm\num{9.6510e-03}$ \\
4 & 801 & $\num{1.9470e-04}\pm\num{4.7980e-05}$ & $\num{8.2751e-03}\pm\num{6.2136e-03}$ \\
16 & 51 & $\num{1.9942e-04}\pm\num{3.0756e-05}$ & $\num{6.0096e-05}\pm\num{1.0329e-05}$ \\
16 & 101 & $\num{1.9766e-04}\pm\num{2.4678e-05}$ & $\num{6.0569e-05}\pm\num{8.5808e-06}$ \\
16 & 201 & $\num{2.1672e-04}\pm\num{3.2822e-05}$ & $\num{6.8149e-05}\pm\num{1.3025e-05}$ \\
16 & 401 & $\num{2.0376e-04}\pm\num{3.0521e-05}$ & $\num{5.9792e-05}\pm\num{1.2887e-05}$ \\
16 & 801 & $\num{2.0988e-04}\pm\num{3.5117e-05}$ & $\num{6.7264e-05}\pm\num{1.6062e-05}$ \\
64 & 51 & $\num{2.2848e-04}\pm\num{1.9651e-05}$ & $\num{3.3921e-05}\pm\num{3.6466e-06}$ \\
64 & 101 & $\num{2.2739e-04}\pm\num{2.5986e-05}$ & $\num{3.5154e-05}\pm\num{4.1384e-06}$ \\
64 & 201 & $\num{2.3161e-04}\pm\num{2.3412e-05}$ & $\num{3.5122e-05}\pm\num{4.0520e-06}$ \\
64 & 401 & $\num{2.2763e-04}\pm\num{1.5265e-05}$ & $\num{3.4044e-05}\pm\num{4.3860e-06}$ \\
64 & 801 & $\num{2.3308e-04}\pm\num{2.2930e-05}$ & $\num{3.6565e-05}\pm\num{5.4556e-06}$ \\
256 & 51 & $\num{2.4231e-04}\pm\num{1.2736e-05}$ & $\num{1.8466e-05}\pm\num{1.5881e-06}$ \\
256 & 101 & $\num{2.5664e-04}\pm\num{1.7316e-05}$ & $\num{2.0089e-05}\pm\num{1.9047e-06}$ \\
256 & 201 & $\num{2.4479e-04}\pm\num{2.5320e-05}$ & $\num{1.9124e-05}\pm\num{2.1326e-06}$ \\
256 & 401 & $\num{2.4944e-04}\pm\num{1.7857e-05}$ & $\num{1.9511e-05}\pm\num{2.1490e-06}$ \\
256 & 801 & $\num{2.5230e-04}\pm\num{2.0580e-05}$ & $\num{1.9472e-05}\pm\num{2.4619e-06}$ \\
\caption{(Phototaxis model) Phototaxis model relative interaction kernel and environmental force errors for different $M$ (number of replicates) and $L$ (number of time samples). Each entry reports the mean $\pm$ sample standard deviation over $T_{r}=10$ independent learning trials.} 
\label{tab:levi_ML_sweep_relative_errors_mean_std}
\end{longtable}

\begin{longtable}{rrcc}

\toprule
M & L & $E_{\phi}^{\mathrm{rel}}$ & $E_{f}^{\mathrm{rel}}$ \\
\midrule
\endfirsthead
\toprule
M & L & $E_{\phi}^{\mathrm{rel}}$ & $E_{f}^{\mathrm{rel}}$ \\
\midrule
\endhead
\midrule
\multicolumn{4}{r}{Continued on next page} \\
\midrule
\endfoot
\bottomrule
\endlastfoot
1 & 51 & $\num{1.1381e+00}\pm\num{1.2043e+00}$ & $\num{2.7564e+00}\pm\num{3.3192e+00}$ \\
1 & 101 & $\num{1.4278e+00}\pm\num{1.7454e+00}$ & $\num{1.3453e+00}\pm\num{1.0674e+00}$ \\
1 & 201 & $\num{5.3246e-01}\pm\num{5.6657e-01}$ & $\num{2.1987e+00}\pm\num{2.1506e+00}$ \\
1 & 401 & $\num{6.1445e-01}\pm\num{5.5824e-01}$ & $\num{1.0916e+00}\pm\num{1.5755e+00}$ \\
1 & 801 & $\num{8.0262e-01}\pm\num{3.6660e-01}$ & $\num{1.1428e+00}\pm\num{5.8255e-01}$ \\
4 & 51 & $\num{1.3722e-01}\pm\num{3.5001e-02}$ & $\num{1.9606e+00}\pm\num{4.5055e+00}$ \\
4 & 101 & $\num{1.2154e-01}\pm\num{2.2052e-02}$ & $\num{4.1079e-01}\pm\num{7.3962e-01}$ \\
4 & 201 & $\num{1.2518e-01}\pm\num{2.2447e-02}$ & $\num{4.0368e-01}\pm\num{7.0996e-01}$ \\
4 & 401 & $\num{1.2693e-01}\pm\num{3.5854e-02}$ & $\num{3.9452e-01}\pm\num{7.8274e-01}$ \\
4 & 801 & $\num{1.1670e-01}\pm\num{2.5842e-02}$ & $\num{2.4197e-01}\pm\num{2.3849e-01}$ \\
16 & 51 & $\num{1.1221e-01}\pm\num{8.3761e-03}$ & $\num{2.9490e-02}\pm\num{5.8072e-04}$ \\
16 & 101 & $\num{1.1085e-01}\pm\num{7.1729e-03}$ & $\num{2.9833e-02}\pm\num{1.3988e-03}$ \\
16 & 201 & $\num{1.1122e-01}\pm\num{6.9548e-03}$ & $\num{2.9854e-02}\pm\num{6.3740e-04}$ \\
16 & 401 & $\num{1.0592e-01}\pm\num{1.1698e-02}$ & $\num{2.9147e-02}\pm\num{5.4161e-04}$ \\
16 & 801 & $\num{1.0950e-01}\pm\num{7.2616e-03}$ & $\num{2.9614e-02}\pm\num{9.9649e-04}$ \\
64 & 51 & $\num{1.1433e-01}\pm\num{3.7577e-03}$ & $\num{2.8864e-02}\pm\num{2.7296e-04}$ \\
64 & 101 & $\num{1.1151e-01}\pm\num{4.6203e-03}$ & $\num{2.9080e-02}\pm\num{2.3488e-04}$ \\
64 & 201 & $\num{1.1438e-01}\pm\num{3.0293e-03}$ & $\num{2.9111e-02}\pm\num{2.0774e-04}$ \\
64 & 401 & $\num{1.1024e-01}\pm\num{4.2098e-03}$ & $\num{2.9117e-02}\pm\num{2.4098e-04}$ \\
64 & 801 & $\num{1.1347e-01}\pm\num{5.2101e-03}$ & $\num{2.9282e-02}\pm\num{2.7405e-04}$ \\
256 & 51 & $\num{1.1533e-01}\pm\num{2.2326e-03}$ & $\num{2.9606e-02}\pm\num{1.3604e-04}$ \\
256 & 101 & $\num{1.1210e-01}\pm\num{2.4077e-03}$ & $\num{2.9824e-02}\pm\num{1.2639e-04}$ \\
256 & 201 & $\num{1.1398e-01}\pm\num{2.2106e-03}$ & $\num{2.9924e-02}\pm\num{1.2837e-04}$ \\
256 & 401 & $\num{1.1318e-01}\pm\num{1.9296e-03}$ & $\num{2.9917e-02}\pm\num{1.1030e-04}$ \\
256 & 801 & $\num{1.1401e-01}\pm\num{1.4986e-03}$ & $\num{2.9998e-02}\pm\num{1.7770e-04}$ \\
\caption{(SPP model) SPP model relative interaction kernel and environmental force errors for different $M$ (number of replicates) and $L$ (number of time samples). Each entry reports the mean $\pm$ sample standard deviation over $T_{r}=10$ independent learning trials.} 
\end{longtable}
\label{tab:spp_ML_sweep_relative_errors_mean_std}

\subsection{Noise robustness}
\label{subsec:app:noise}

\begin{figure}
    \centering

    \begin{subfigure}{0.24\textwidth}
        \centering
        \includegraphics[width=\textwidth]{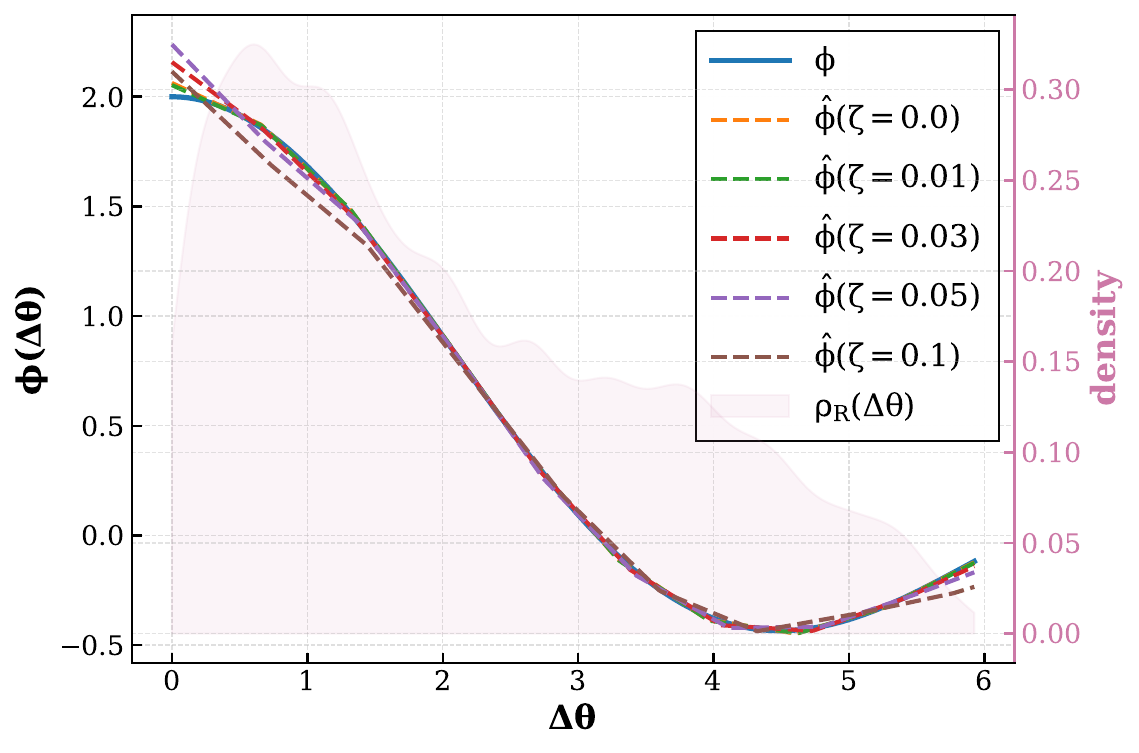}
        \caption{$\phi:$ Kuramoto}
        \label{subfig:app:Kuramoto_phi_noise}
    \end{subfigure}
    \hfill
    \begin{subfigure}{0.24\textwidth}
        \centering
        \includegraphics[width=\textwidth]{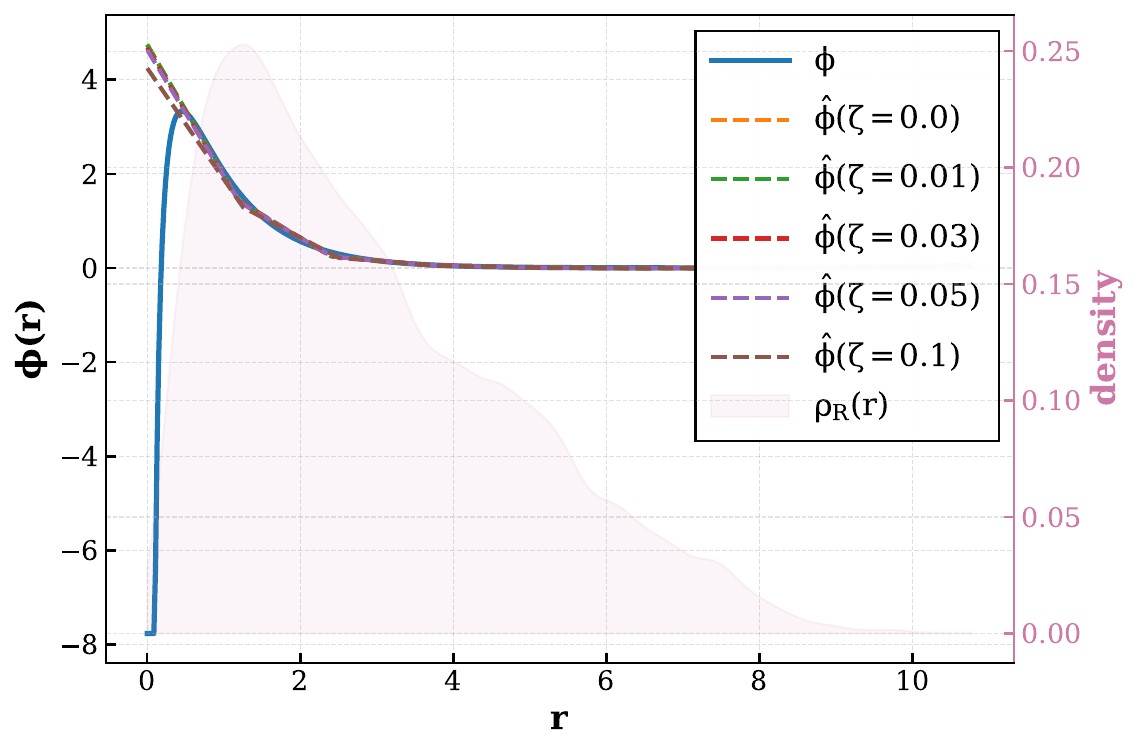}
        \caption{$\phi:$ SPP}
        \label{subfig:app:SPP_phi_noise}
    \end{subfigure}
    \hfill
    \begin{subfigure}{0.24\textwidth}
        \centering
        \includegraphics[width=\textwidth]{Images/Levy/step_6D/noise_sweep_phi_by_noise.pdf}
        \caption{$\phi:$ phototaxis}
        \label{subfig:app:Phototaxis_phi_noise}
    \end{subfigure}
    \hfill
    \begin{subfigure}{0.24\textwidth}
        \centering
        \includegraphics[width=\textwidth]{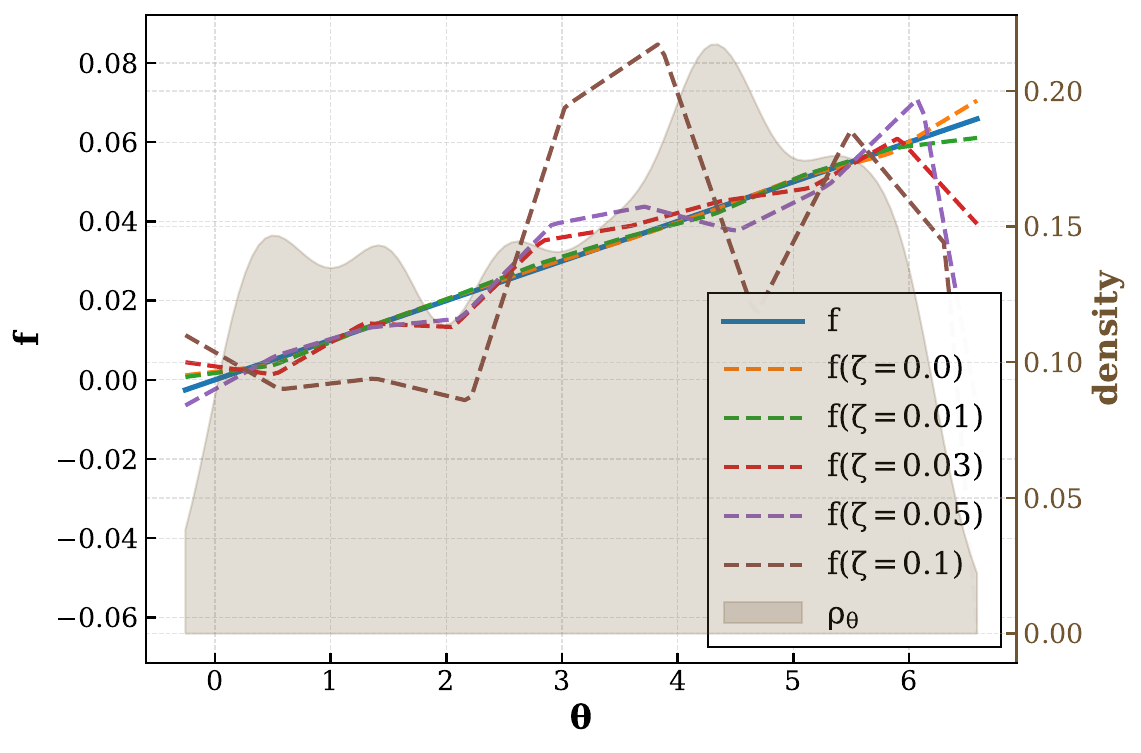}
        \caption{$f:$ Kuramoto}
        \label{subfig:app:Kuramoto_f_noise}
    \end{subfigure}
\vspace{0.2em}
    \begin{subfigure}{\textwidth}
        \centering
        \includegraphics[width=\textwidth]{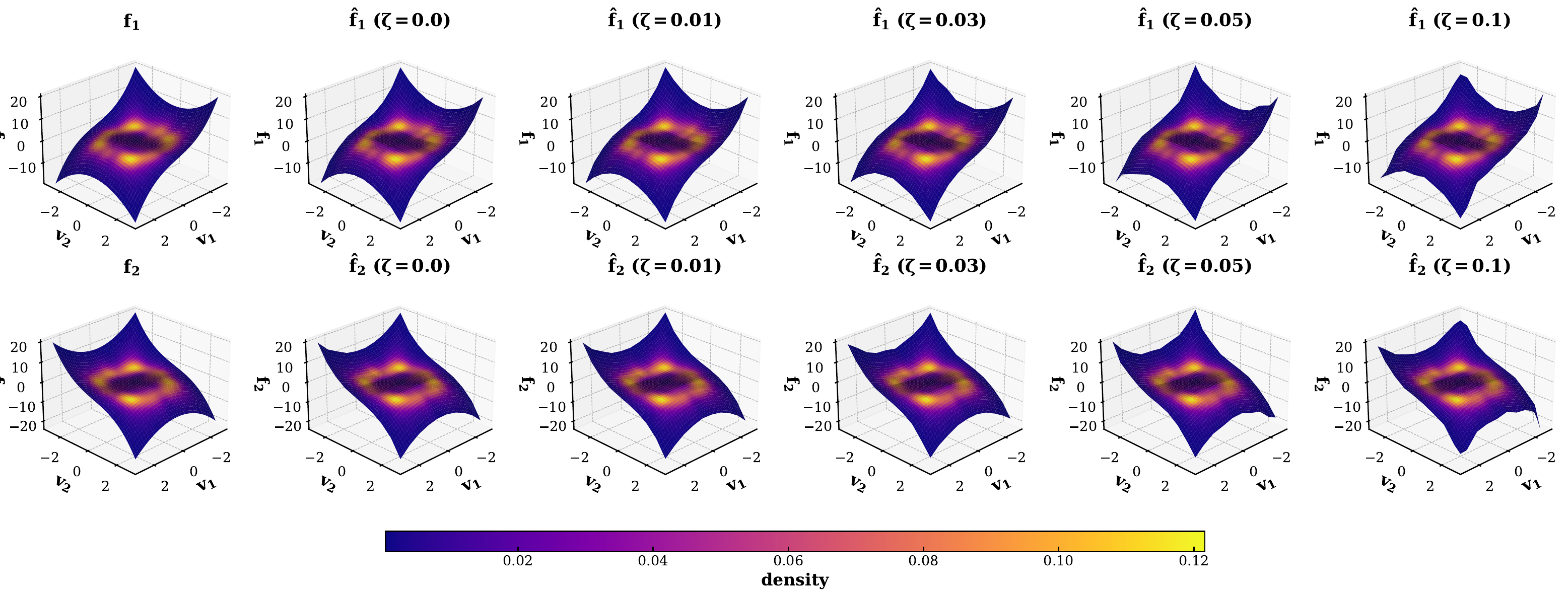}
        \caption{$f:$ SPP}
        \label{subfig:app:SPP_f_noise}
    \end{subfigure}
    \vspace{0.2em}
    \begin{subfigure}{\textwidth}
        \centering
        \includegraphics[width=\textwidth]{Images/Levy/step_6D/noise_sweep_env_surfaces_by_noise.pdf}
        \caption{$f$:phototaxis}
        \label{subfig:app:Phototaxis_f_noise}
    \end{subfigure}

    \caption{(Noise robustness, mechanism recovery) Each subfigure illustrates feature learning under multiplicative observational noise added to the training data at different noise levels $\zeta$, as described in Section~\ref{subsec:noise_robustness}. Subfigures ~\ref{subfig:app:Kuramoto_phi_noise}--~\ref{subfig:app:Phototaxis_phi_noise} show the learned interaction kernels $\phi$ for increasing values of $\zeta$. Subfigures~\ref{subfig:app:Kuramoto_f_noise}--~\ref{subfig:app:Phototaxis_f_noise} show the corresponding learned environmental force $f$ under the same noise levels. For the SPP and phototaxis systems, the recovery of $f$ is shown component-wise, where the upper and lower rows correspond to the first and second components, respectively.The background shading indicates the empirical training measure associated with each feature: $\hat{\rho}_R$ for interaction kernels in all systems except Kuramoto, where $\hat{\rho}_{\Delta\theta}$ is used; $\hat{\rho}_V$ for the environmental force in second-order systems; and $\hat{\rho}_{\theta}$ for the environmental variable in the Kuramoto system. We notice that, as the noise level $\zeta$ increases, the learned features gradually deviate from the true features, yet overall recovery is gnerally accurate and estimated features are minimally affected, primarily in regions with small induced measure (i.e. a small amount of trajectory data).
    }
    \label{fig:app:noise_feature_recovery}
\end{figure}

\begin{figure}
    \centering

    \begin{subfigure}{0.19\textwidth}
        \centering
        \includegraphics[width=\textwidth]{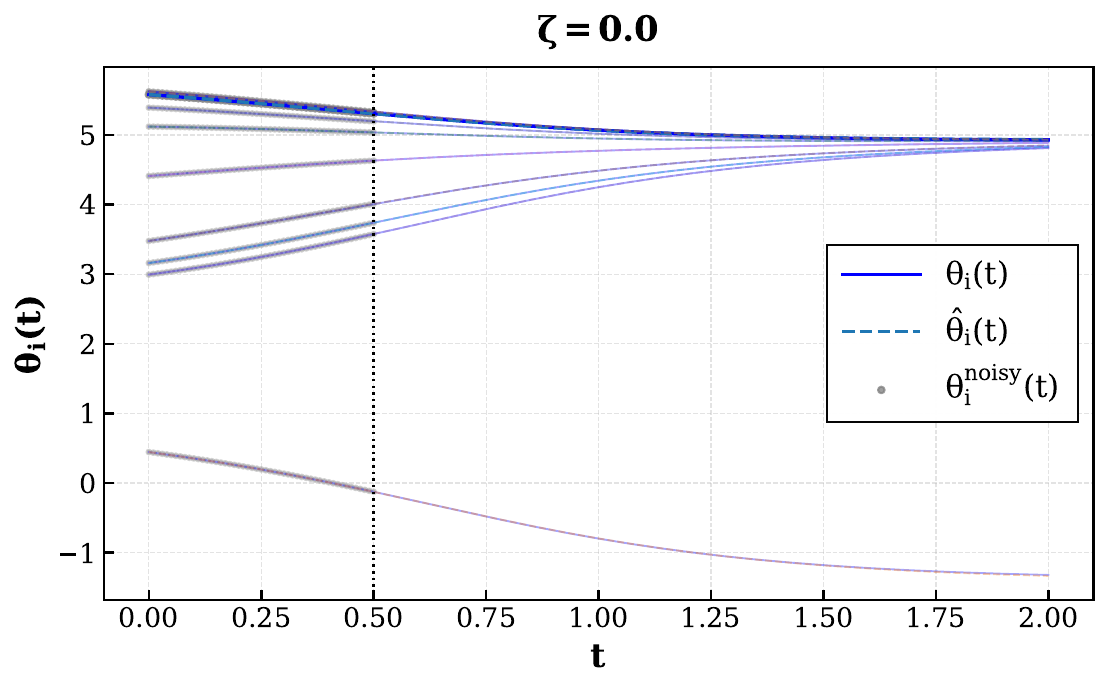}
    \end{subfigure}
    \hfill
    \begin{subfigure}{0.19\textwidth}
        \centering
        \includegraphics[width=\textwidth]{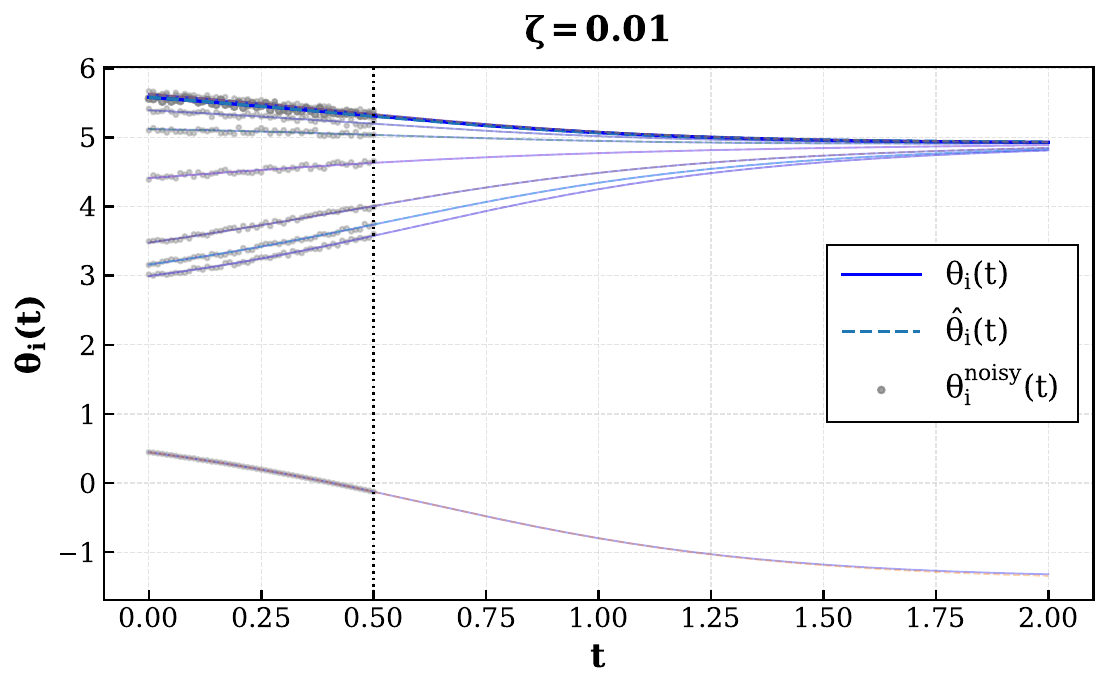}
    \end{subfigure}
    \hfill
    \begin{subfigure}{0.19\textwidth}
        \centering
        \includegraphics[width=\textwidth]{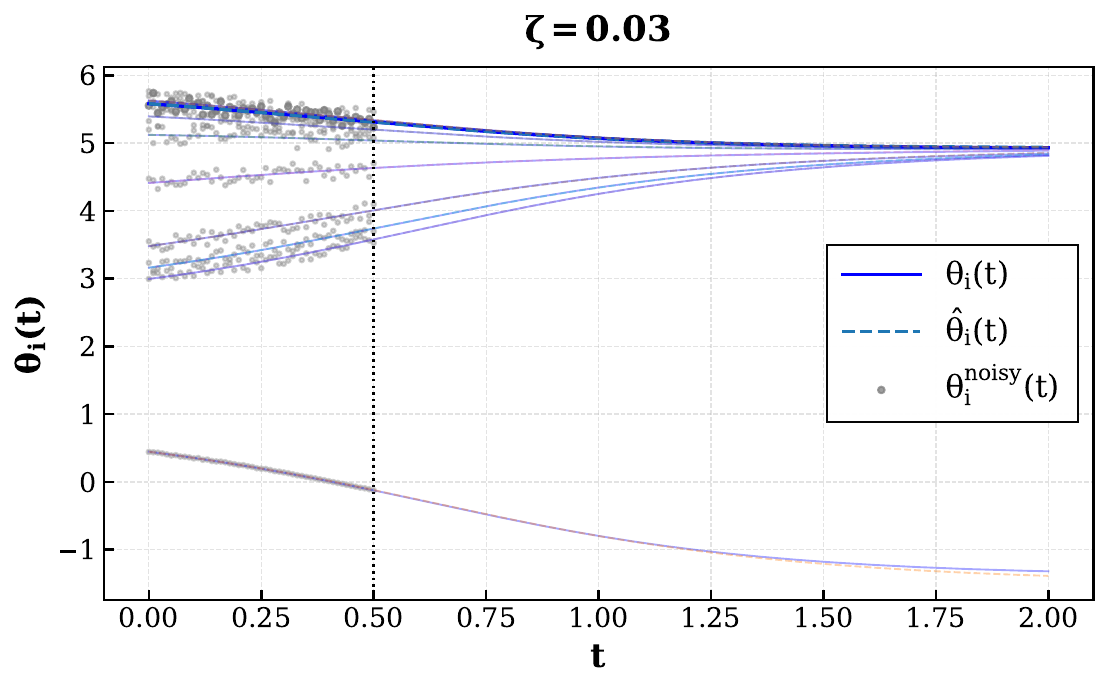}
    \end{subfigure}
     \hfill
    \begin{subfigure}{0.19\textwidth}
        \centering
        \includegraphics[width=\textwidth]{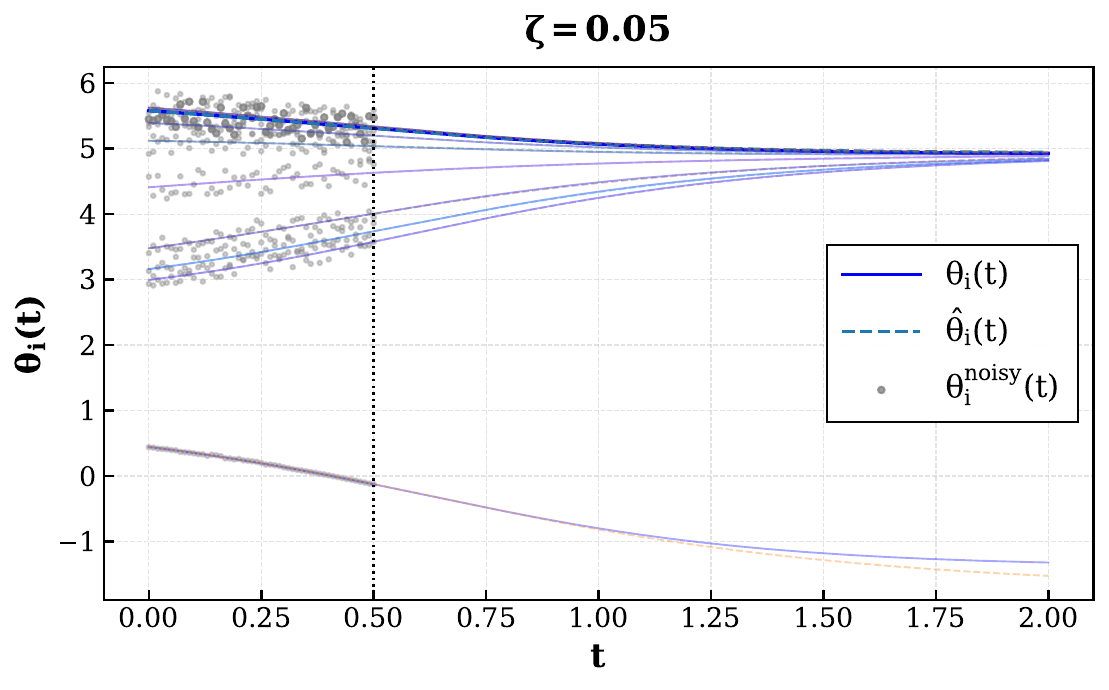}
    \end{subfigure}
    \hfill
    \begin{subfigure}{0.19\textwidth}
        \centering
        \includegraphics[width=\textwidth]{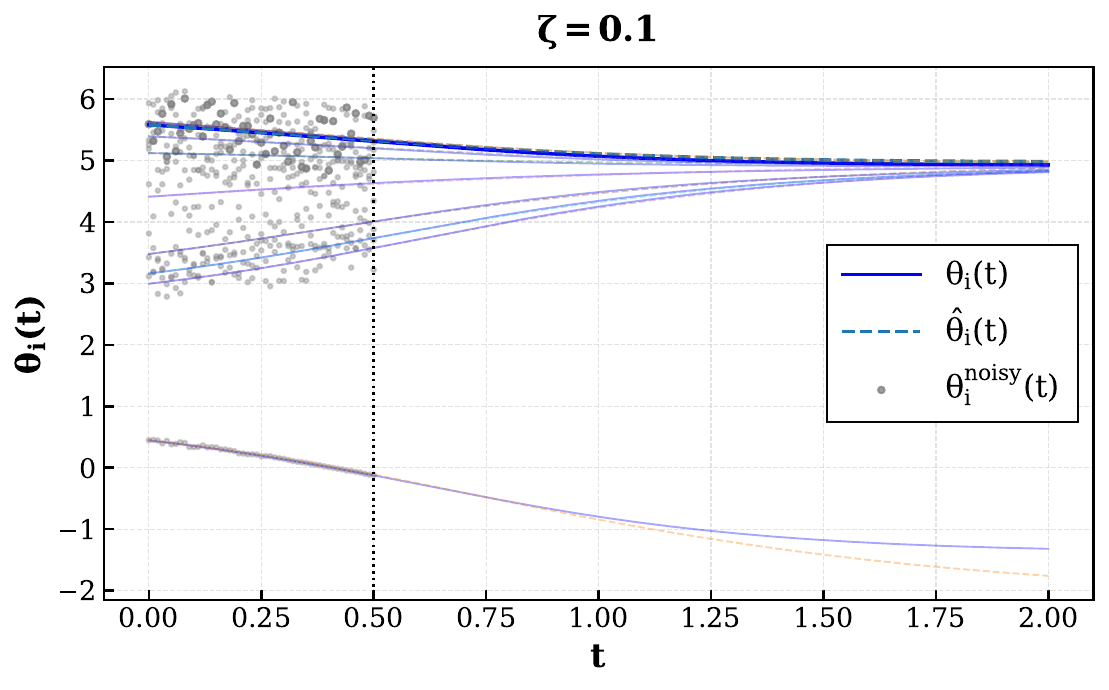}
    \end{subfigure}

    \vspace{0.4em}

    \begin{subfigure}{0.19\textwidth}
        \centering
        \includegraphics[width=\textwidth]{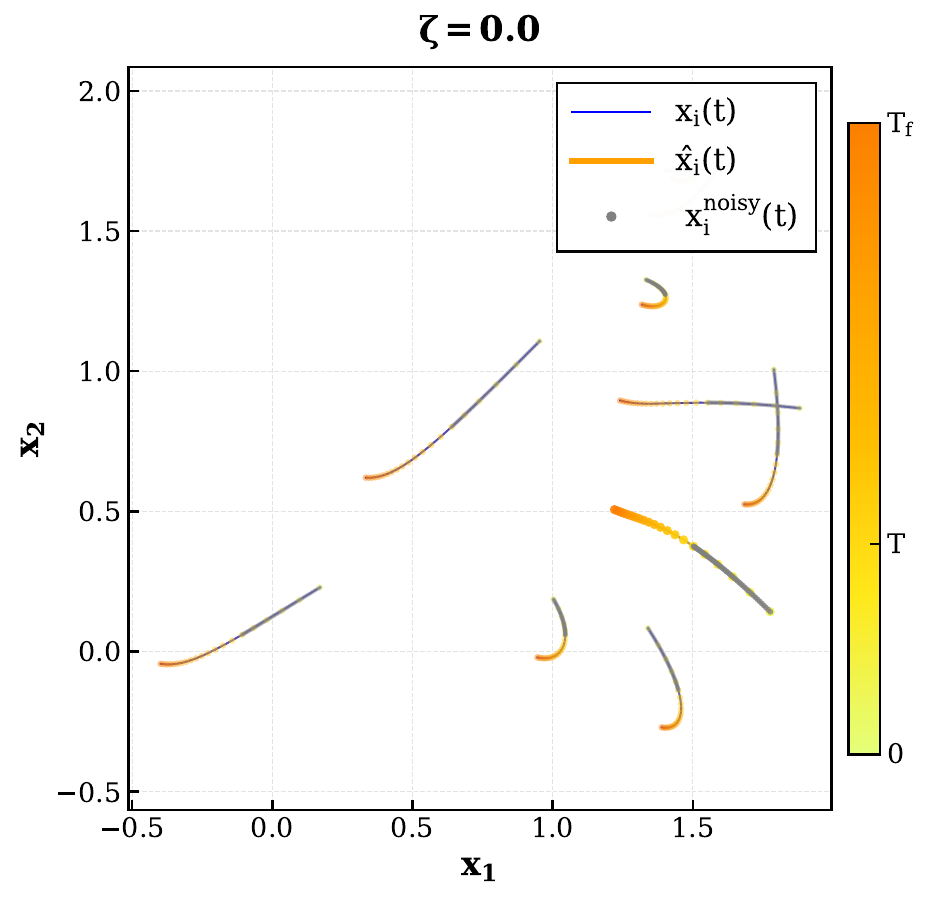}
    \end{subfigure}
    \hfill
    \begin{subfigure}{0.19\textwidth}
        \centering
        \includegraphics[width=\textwidth]{Images/Levy/step_6D/noise_sweep_seen_ic_phase_noise_0p01.pdf}
    \end{subfigure}
    \hfill
    \begin{subfigure}{0.19\textwidth}
        \centering
        \includegraphics[width=\textwidth]{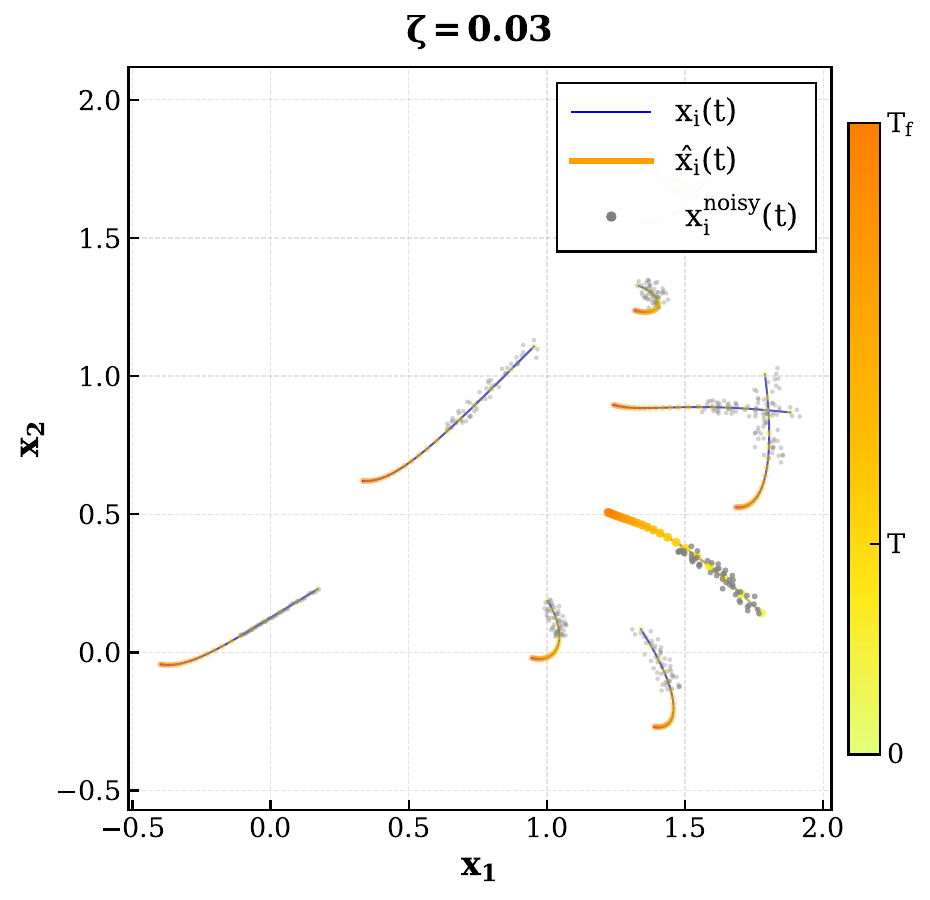}
    \end{subfigure}
     \hfill
    \begin{subfigure}{0.19\textwidth}
        \centering
        \includegraphics[width=\textwidth]{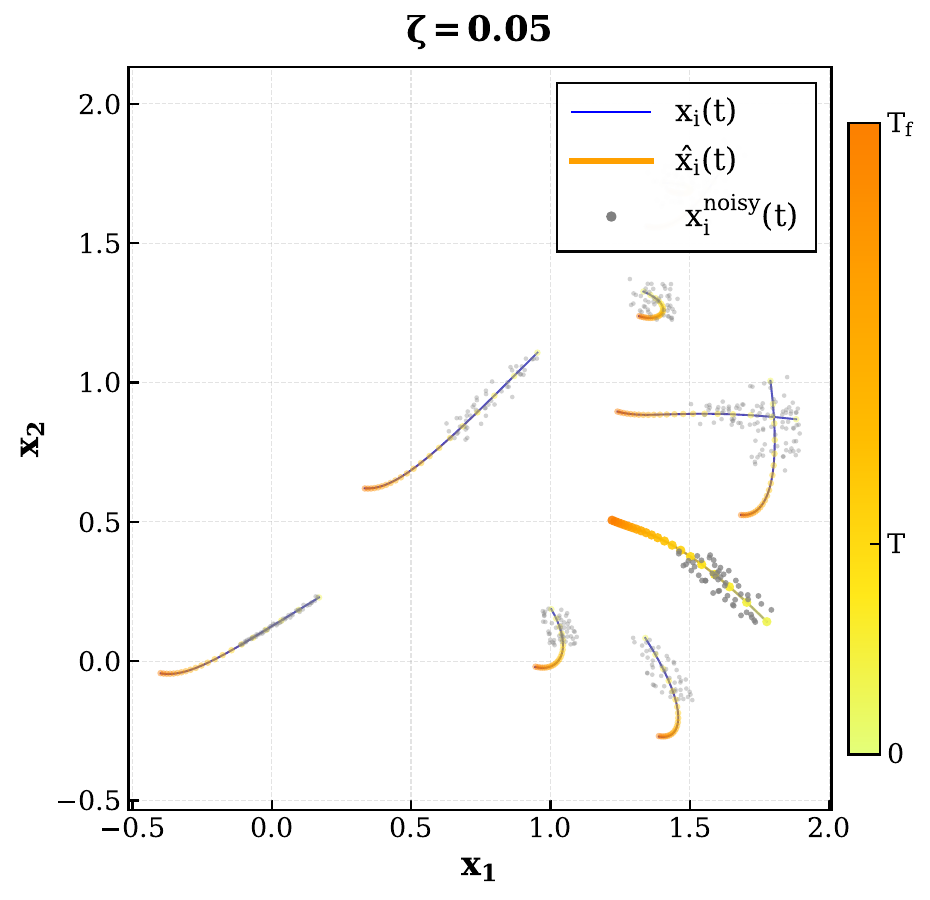}
    \end{subfigure}
    \hfill
    \begin{subfigure}{0.19\textwidth}
        \centering
        \includegraphics[width=\textwidth]{Images/Levy/step_6D/noise_sweep_seen_ic_phase_noise_0p1.pdf}
    \end{subfigure}

    \vspace{0.4em}

    \begin{subfigure}{0.19\textwidth}
        \centering
        \includegraphics[width=\textwidth]{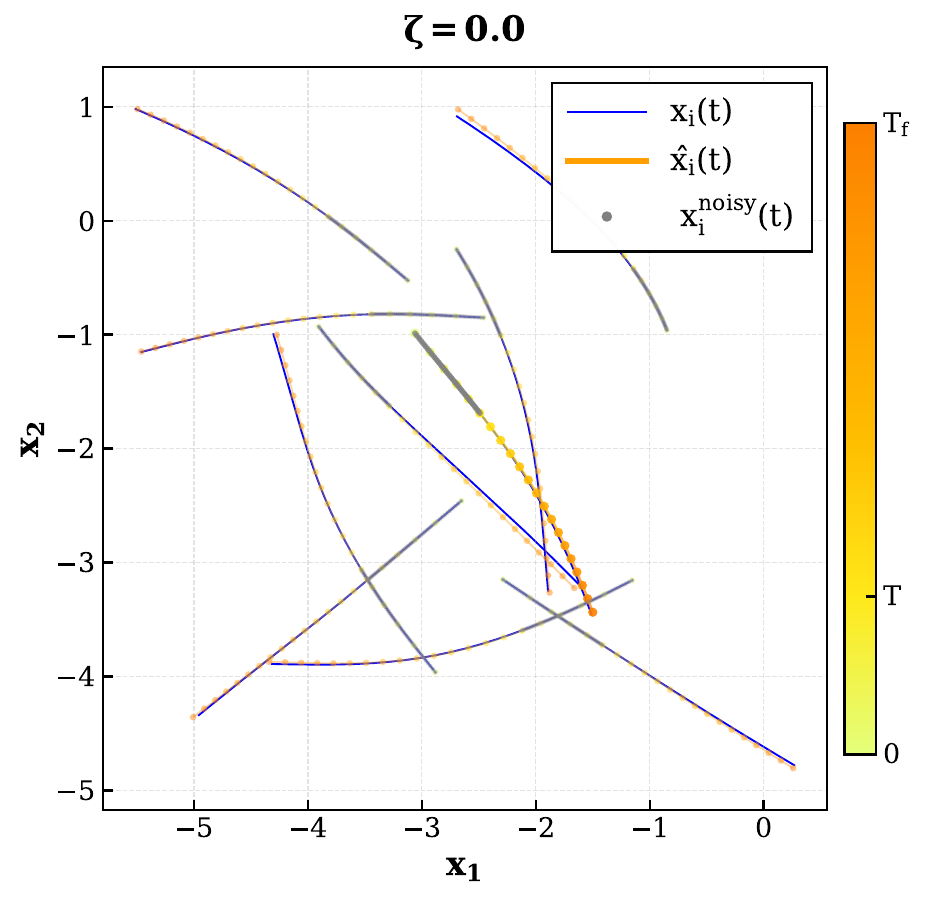}
    \end{subfigure}
    \hfill
    \begin{subfigure}{0.19\textwidth}
        \centering
        \includegraphics[width=\textwidth]{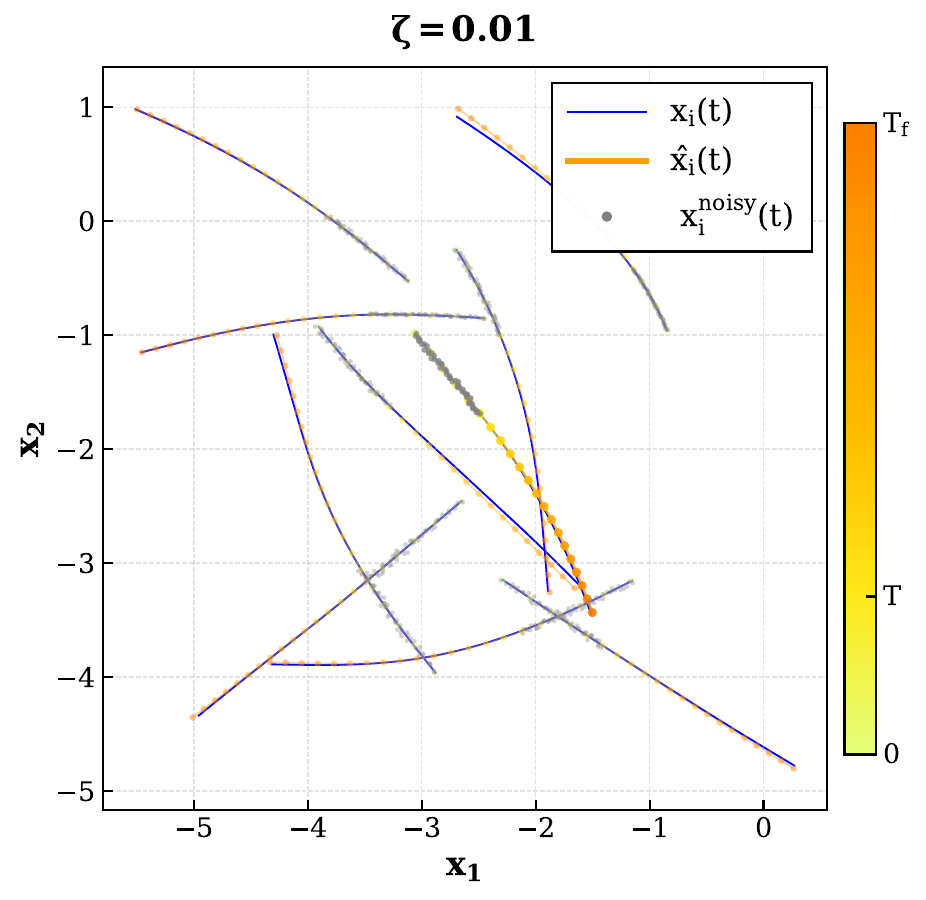}
    \end{subfigure}
    \hfill
    \begin{subfigure}{0.19\textwidth}
        \centering
        \includegraphics[width=\textwidth]{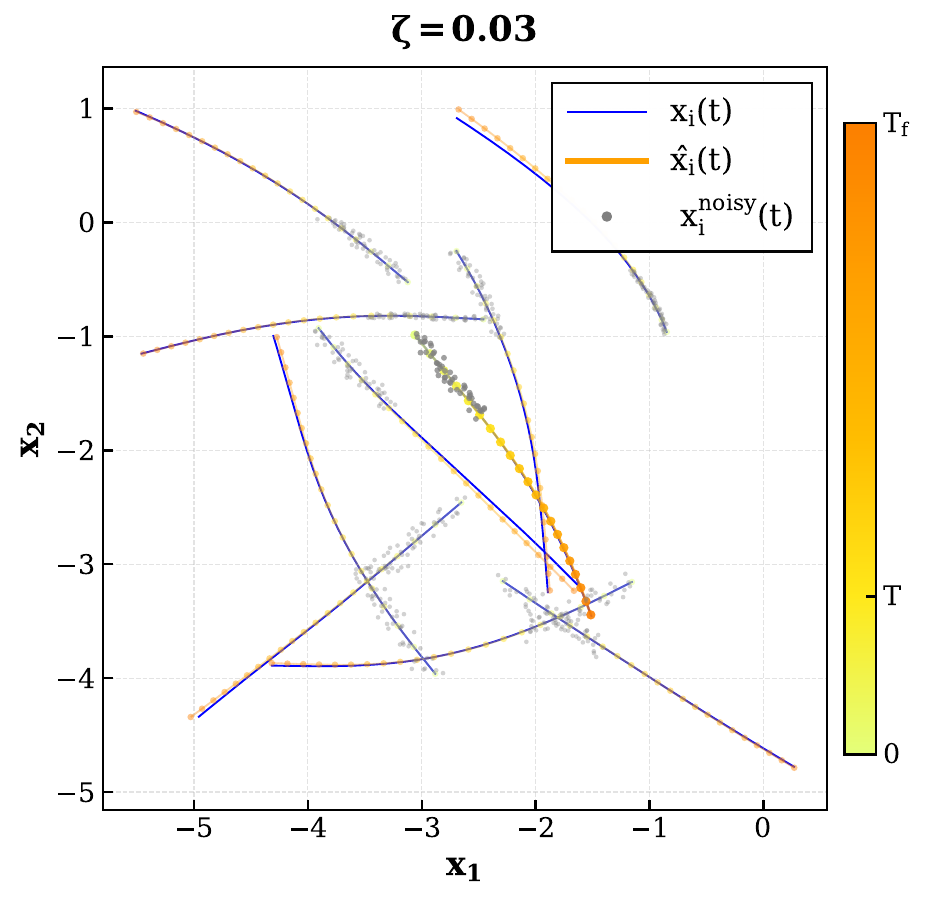}
    \end{subfigure}
        \hfill
    \begin{subfigure}{0.19\textwidth}
        \centering
        \includegraphics[width=\textwidth]{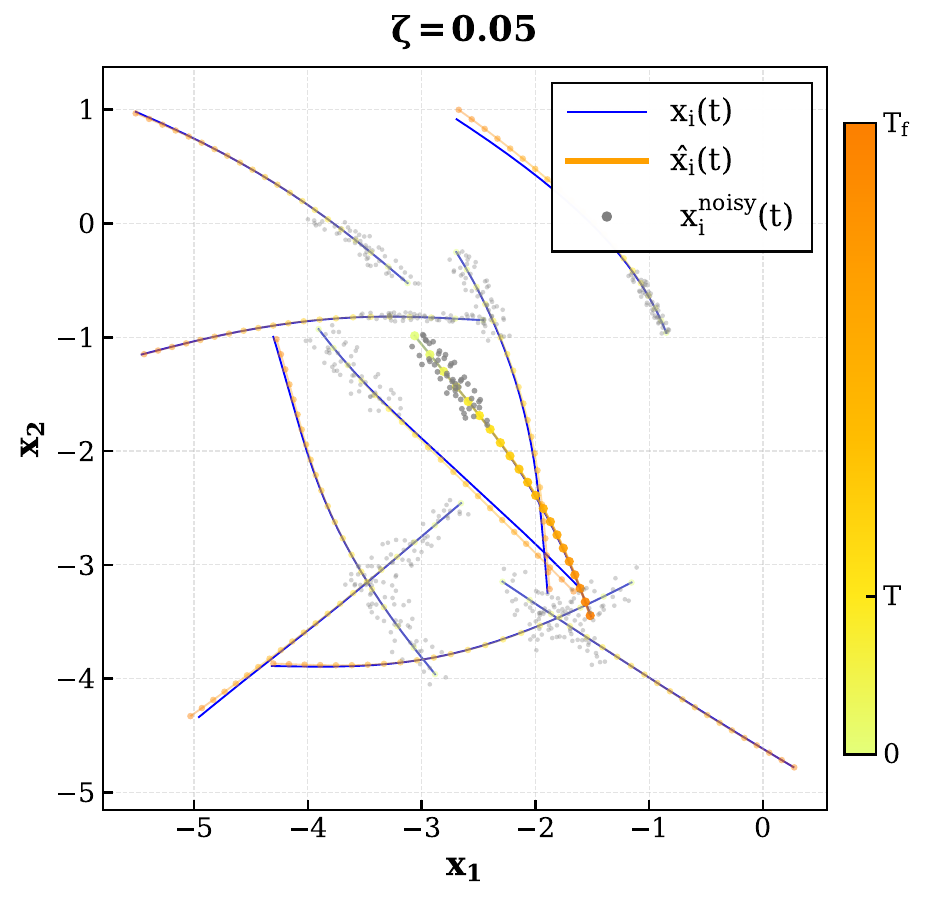}
    \end{subfigure}
    \hfill
    \begin{subfigure}{0.19\textwidth}
        \centering
        \includegraphics[width=\textwidth]{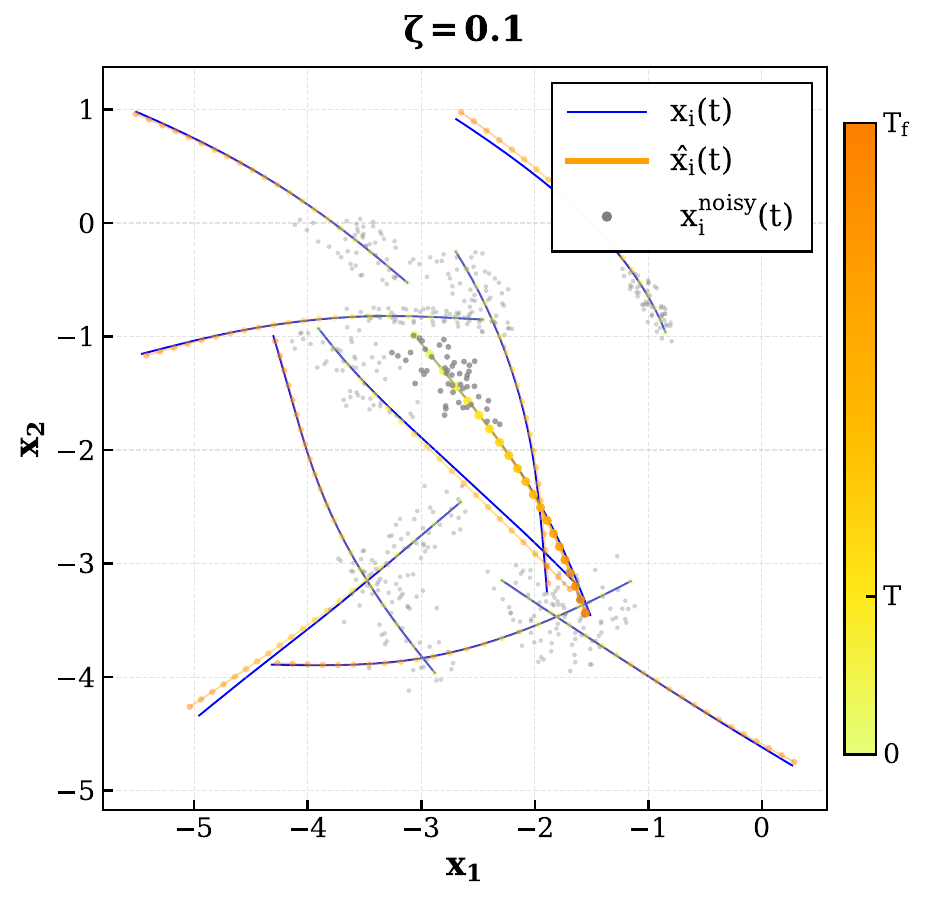}
    \end{subfigure}

    \caption{
    (Noise robustness, trajectory recovery and prediction) Each row represents the Kuramoto, Phototaxis, and SPP systems, respectively. The subfigures show one of the observed trajectories before and after being perturbed by each noise level $\zeta$ described in Section~\ref{subsec:noise_robustness}, over the full time horizon $[0,T_f]$.
    For the Kuramoto system, the solid blue line represents the true trajectory, the dashed line represents the predicted trajectory learned from the noisy observations, and the semitransparent gray dots denote the noisy trajectory used as the training data (together with noisy observations of the derivative).
    For the SPP and phototaxis systems, the solid blue lines represent the true trajectories, the semitransparent gray dotted lines represent the noisy trajectories used as the training data (together with noisy observations of the derivative), and the color-gradient lines represent the predicted trajectories learned from the noisy observations, with yellow corresponding to $t_0$ and orange corresponding to $T_f$.We observe that the learned trajectories remain close to the true trajectories at lower noise levels. As the noise level increases, gradual deviations in the predicted dynamics become apparent, particularly for the SPP system. However, the overall trajectory recovery remains stable, indicating a degree of robustness of the learned models to observational noise.
    }
\label{fig:app:noise_trajectory_robustness}
\end{figure}

The following figures illustrate the robustness of the variational framework with respect to different observational noise levels for the models discussed in Section~\ref{subsec:results_nonparametric_vs_parametric_learning}. Figure~\ref{fig:app:noise_feature_recovery} visualizes the effect of noise on recovery of mechanisms, while Figure~\ref{fig:app:noise_trajectory_robustness} shows the corresponding impact on trajectory reconstruction and prediction.

\section{Limitations}
\label{app:sec:limitations}

\subsection{Effect of initial conditions on inferrence}
\label{app:subsec:ic_variation}

\begin{figure}
    \centering
    \begin{subfigure}{0.6\textwidth}
        \centering
        \includegraphics[width=\textwidth]{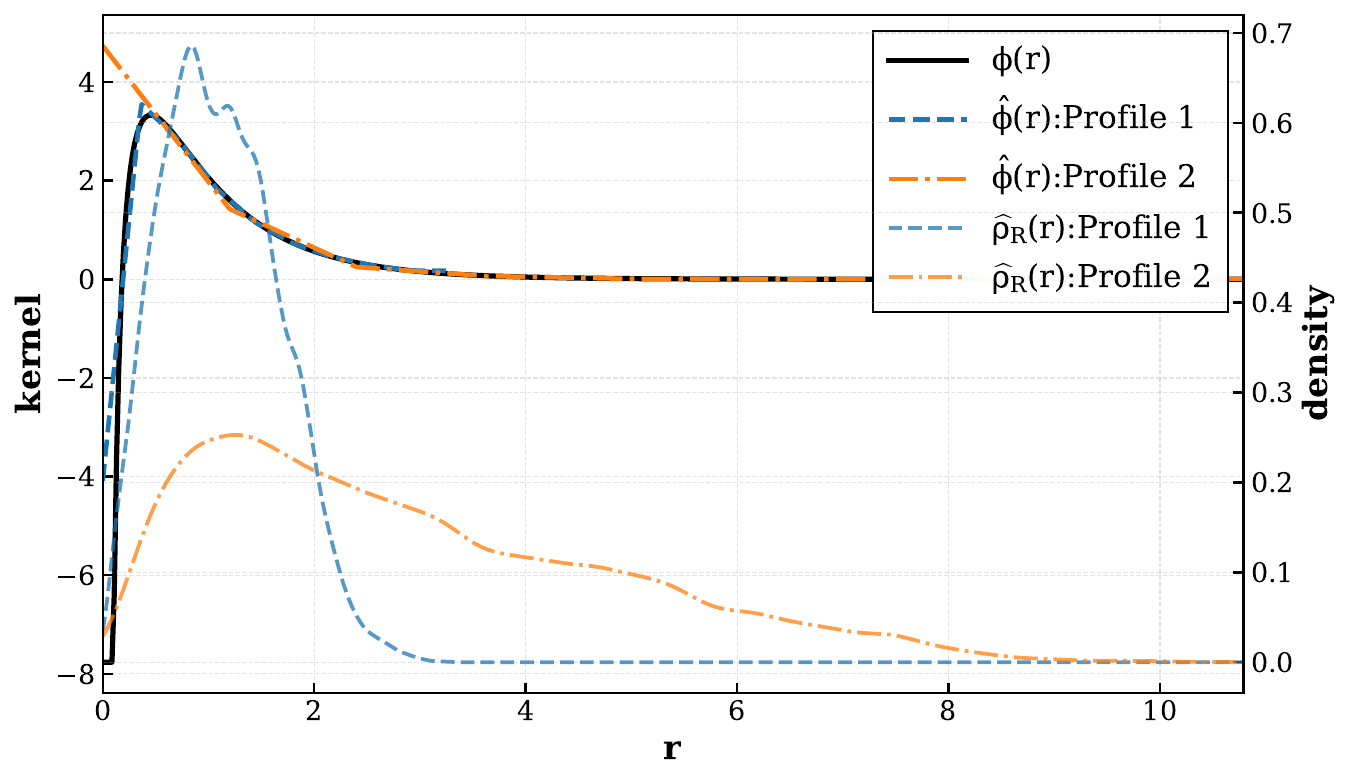}
        \caption{$\phi$}
        \label{subfig:IC_kernel}
    \end{subfigure}

    \begin{subfigure}{0.6\textwidth}
        \centering
        \includegraphics[width=\textwidth]{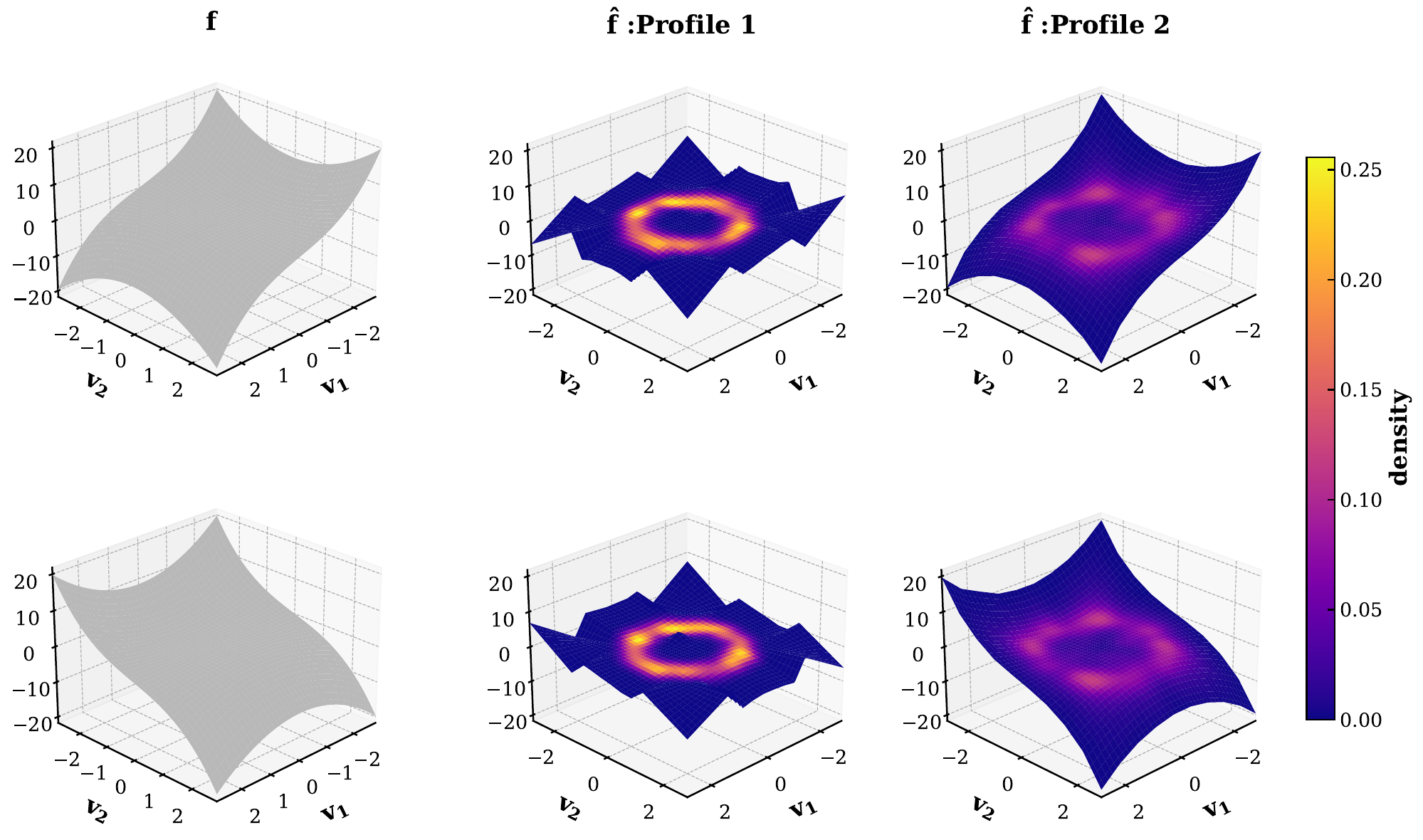}
        \caption{$f$}
        \label{subfig:IC_f}
    \end{subfigure}

    \caption{
(Initial condition distribution, SPP system) Comparison of the variational learning algorithm under two different initial condition distributions $\mu_0$. Profile~1 (blue) is trained using a narrower initial-condition distribution, while Profile~2 (orange) is trained using a distribution with larger variance. The true kernel is shown in black. Subfigure~\ref{subfig:IC_kernel} compares the recovered interaction kernel $\phi$. Both profiles recover the overall kernel accurately; however, Profile~2 fails to capture the narrow, steep peak near the origin. Subfigure~\ref{subfig:IC_f} compares the recovered environmental force $f$. Profile~1 provides a poor approximation, whereas Profile~2 accurately recovers the true environmental force, demonstrating that a broader initial condition distribution improves recovery of the environmental dynamics.
}
\label{fig:randomness_in_IC}
\end{figure}

The distribution of the initial conditions, $\mu_0$, has a significant impact on the performance of the learning algorithm. In particular, training datasets whose initial conditions span a larger region of the state space provide more informative observations, resulting in more accurate recovery of the environmental force. Conversely, when the initial-condition distribution is narrowly concentrated, portions of the state space remain poorly sampled, leading to reduced learning accuracy. Figure~\ref{fig:randomness_in_IC} illustrates this effect by comparing two training datasets generated from different initial-condition distributions.

\subsection{Basis functions}
\label{app:subsec:basis_functions}

\begin{figure}
    \centering
    \begin{subfigure}{0.48\textwidth}
        \centering
        \includegraphics[width=\textwidth]{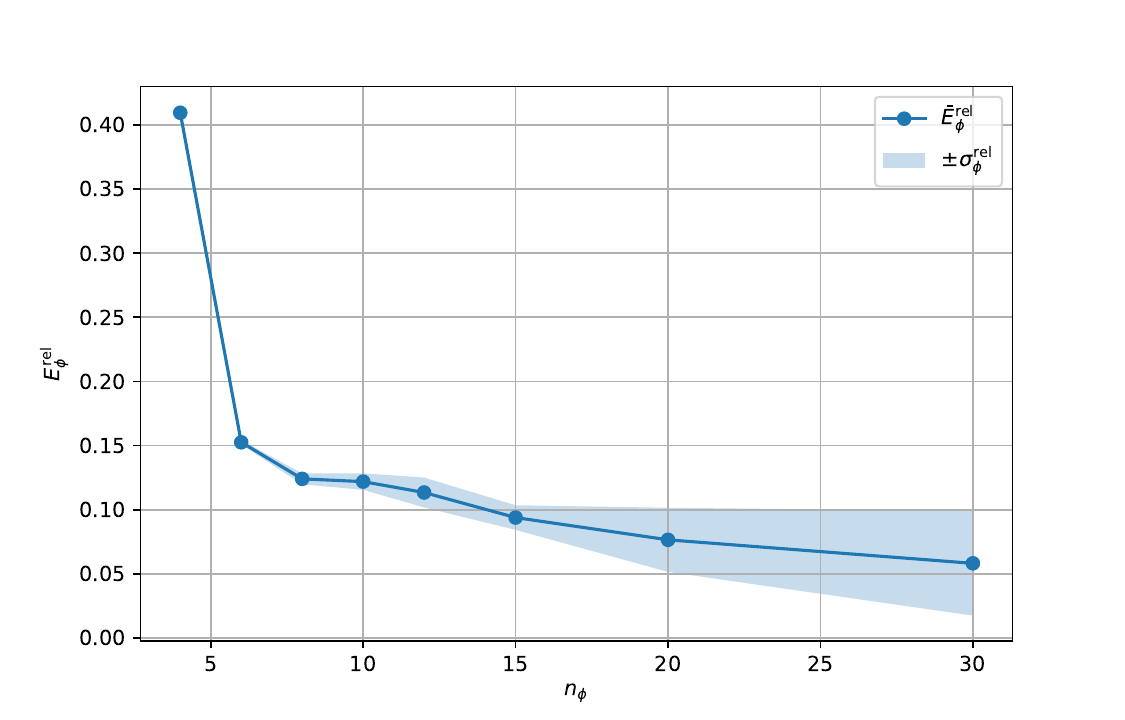}
        \caption{$\phi$}
        \label{subfig:basis_kernel}
    \end{subfigure}
    \hfill
    \begin{subfigure}{0.48\textwidth}
        \centering
        \includegraphics[width=\textwidth]{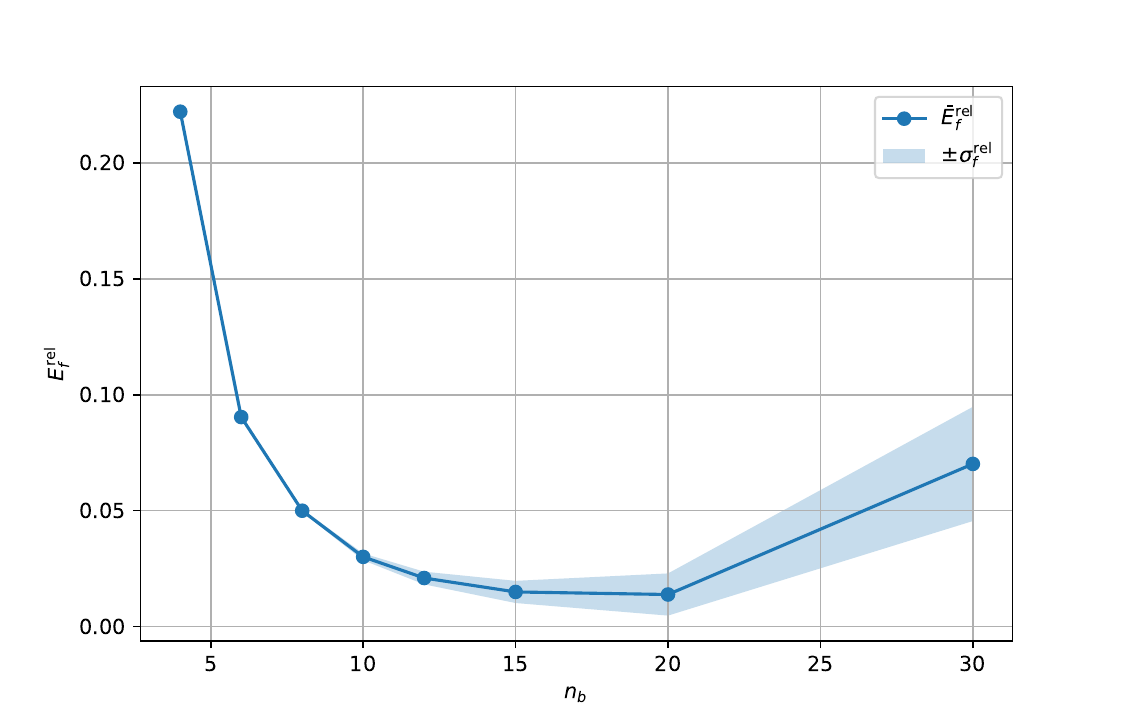}
        \caption{$f$}
        \label{subfig:basis_f}
    \end{subfigure}

    \caption{
    Effect of the number of B-spline basis functions on the recovery accuracy of the interaction kernel and environmental force for the SPP system. In each panel, one basis dimension is varied while the recovery error is averaged over all tested values of the other basis dimension. The solid curves represent the mean relative weighted $L^2$ error, and the shaded regions indicate one standard deviation across the corresponding configurations. Subfigure~\ref{subfig:basis_kernel} shows the interaction kernel error, $E_{\phi}^{\mathrm{rel}}$, as the number of kernel basis functions, $n_{\phi}$, varies. Subfigure~\ref{subfig:basis_f} shows the environmental-force error, $E_{f}^{\mathrm{rel}}$, as the number of environmental force basis functions, $n_f$, varies. In both cases, the lowest average recovery error is achieved with approximately ten basis functions.
    }
    \label{fig:basis_functions}
\end{figure}

The variational algorithm is generally sensitive to the number of basis functions used to represent both the interaction kernel and the environmental force. Figure~\ref{fig:basis_functions} illustrates the effect of varying the number of B-spline basis functions while fixing all other hyperparameters. For each value of the basis dimension being varied, the reported error is averaged with resect to all tested values of the other basis dimension.

As expected, using too few basis functions leads to underfitting and consequently larger feature recovery errors. Increasing the basis dimension improves the approximation quality and reduces the recovery error until the performance saturates. Beyond this, additional basis functions provides minimal improvement and may even degrade performance, as the least-squares equation~\eqref{eq:normal_eq_v2} becomes increasingly ill-conditioned for a fixed amount of training data. For the SPP system considered here, both the interaction kernel and the environmental force achieve their lowest average recovery error when approximately $10$ basis functions are used. This choice therefore provides a good balance between approximation accuracy and numerical stability.

These results indicate that the choice of basis dimension is an important hyperparameter in the nonparametric learning framework. Consequently, adaptive basis construction and automatic hyperparameter selection may further improve both learning accuracy and computational efficiency.

\section{Tables of parameters}
\label{app:sec:tables_of_parameters}

\subsection{Data generation parameters}
\label{app:subsec:data_generation_parameters}

\begin{table}[H]
\centering
\begin{tabular}{lccccccc}
\toprule
\textbf{Parameter}
& \textbf{Kuramoto}
& \textbf{Phototaxis}
& \textbf{SPP}
& \textbf{SPP}
& \textbf{CS2D}
& \textbf{SPPCS} 
& \textbf{OP}\\
\midrule
profile           & - & - & clumps & milling & - & clumps & -\\
System order $s$      & 1 & 2 & 2 & 2 & 2 & 2 & 1\\
State dimension $d$   & 1 & 2 & 2 & 2 & 2 & 2 & 1\\
Number of agents $N$  & 10 & 10 & 10 & 10 & 10 & 10 & 10\\
training trajectories $M_{\mathrm{train}}$ & 20 & 20 &50 & 50 & 20 & 50 & 20\\
testing trajectories $M_{\mathrm{test}}$ & 10 & 10 & 10 & 10 & 10 & 10 & 10\\
Reference trajectories $M_{\rho}$ & 2000 & 2000 & 2000 & 2000 &2000 & 2000 & 2000\\
Initial time $t_0$ & $0$ & $0.25$ & $0$ & $0$ & $0$ & $0$ & 0\\
training horizon $T$ & $0.5$ & $0.75$  & $0.5$  & $4$ & $0.5$  & $0.5$  & $10$\\
Prediction horizon $T_{f}$ & $2$ & $2.25$ & $2$ & $20$ & $2$ & $2$ & $20$\\
Time step $dt$ & $0.01$ & $0.01$ & $0.01$ & $0.01$ & $0.01$ & $0.01$ & $0.05$\\
training snapshots $L_{\mathrm{fit}}$ & $51$& $51$ & $51$ & $401$ & $51$ & $51$ & $201$\\
Full snapshots $L$ & $201$ & $201$ & $201$ & $2001$ & $201$ & $201$ & $401$\\
\bottomrule
\end{tabular}
\caption{
Data-generation parameters used in the numerical experiments.
Training data are used to estimate the interaction kernels and environmental forces,
testing data are used for trajectory validation, and the $M_{\rho}$ data set is used to
construct reference probability densities.
}
\label{tab:app:data_generation_parameters}
\end{table}

\subsection{Learning algorithm parameters}
\label{app:subsec:learning_algorithm_parameters}

\begin{table}[H]
\centering
\small
\begin{tabular}{lcccccc}
\toprule
\textbf{Parameter}
& \textbf{Kuramoto}
& \textbf{Phototaxis}
& \textbf{SPP}
& \textbf{CS2D}
& \textbf{SPPCS} 
& \textbf{OP}\\
\midrule
Interaction type
& Energy
& Alignment
& Energy
& Alignment
& Energy $+$ Alignment
& Energy \\
Radial basis functions $n_{\phi}$ & 10 & 8 & 10 & 10 & 10 & 50\\
Energy basis functions $n_{\phi^E}$ & -- & 10 & -- & 10 & 10 & 50\\
Alignment basis functions $n_{\phi^A}$ & -- & 10 & -- & 10 & 10 & 50\\
Environment basis parameter $n_b$ & 10 & 10 & 10 & 10 & 10 & 10\\
Environment basis size per state $n_{f^d}=(n_b)^d$
& 10
& 100
& 100
& 100
& 100 
& 10 \\
Environment basis $n_{f}=d*n_{f^d}$
& 10
& 200
& 200
& 200
& 200 
& 10\\
\bottomrule
\end{tabular}
\caption{
Algorithm parameters utilized in the variational least-squares framework.
The interaction kernels are approximated using radial B-spline bases, while the
environment force is represented using a tensor product of B-splines.
For first-order systems the environment variable is the state,
whereas for second-order systems it is generally the velocity.
}
\label{tab:app:learning_parameters}
\end{table}

\subsection{System initial conditions}
\label{app:subsec:system_initial_conditions}


\begin{table}[H]
\centering
\renewcommand{\arraystretch}{1.45}
\begin{tabular}{>{\raggedright\arraybackslash}p{0.18\textwidth}
                >{\raggedright\arraybackslash}p{0.72\textwidth}}
\toprule
\textbf{System} & \textbf{Initial condition $u_0$} \\
\midrule
Kuramoto &
$\theta_0^{(m,i)}{\sim} \mathcal{U}(0,2\pi)$ \\

Phototaxis &
$\Big[x_0^{(m,i)} {\sim} \mathcal U([0,2]^2),\;
v_0^{(m,i)}\sim \mu_v^{\mathrm{Phottaxis}}\Big]$ \\

SPP &
$\Big[x_0^{(m,i)}\sim\mu_x^{\mathrm{SPP}},\;
v_0^{(m,i)}\sim\mu_v^{\mathrm{SPP}}\Big]$ \\

CS2D &
$\Big[\,x_0^{(m,i)} \sim \mathcal{U}(0.1, 10), v_0^{(m,i)} \sim \mathcal{U}(0.1, 5)\Big]$\\

SPPCS &
$\Big[x_0^{(m,i)} \sim \mu_x^{\mathrm{SPP}},\;
v_0^{(m,i)}\sim\mu_v^{\mathrm{SPP}}\Big]$
\\

OP &
$x_0^{(m,i)}{\sim} \mathcal{U}(0,10)$ \\

\bottomrule
\end{tabular}
\caption{
System-specific initial conditions used in numerical experiments.
For the first-order Kuramoto system, $u_0=\theta_0$.
For second-order systems, $u_0=(x_0,v_0)$.
$\mu_v^{\mathrm{Phototaxis}}$ denotes the perturbed tensor-grid measure described in \ref{app:subsec:system_initial_conditions}
The spatial measures $\mu_x^{\mathrm{SPP}}$ and $\mu_x^{\mathrm{SPPCS}}$ are cluster-based sampling distributions.
The velocity measures $\mu_v^{\mathrm{SPP}}$ and $\mu_v^{\mathrm{SPPCS}}$ are perturbed tensor-grid distributions on $[-2\sqrt{2},2\sqrt{2}]^2$.
}
\label{tab:app:system-initial-conditions}
\end{table}

\paragraph{Phototaxis} 
For the Phototaxis system, the velocity initial conditions are generated from a perturbed tensor grid on $[-1,1]^2$. Specifically, a uniform tensor grid is constructed on the square $[-1,1]^2$, its points are randomly assigned to agents, and independent Gaussian perturbations are added. This procedure produces approximately uniform coverage of the velocity domain and improves the conditioning of the tensor-product basis used to learn the environmental force $f$.

\paragraph{SPP, SPPCS}
The spatial measure $\mu_x^{\mathrm{SPP}}$ is a cluster-based sampling distribution obtained from a mixture of uniformly populated spatial clumps. The velocity measure $\mu_v^{\mathrm{SPP}}$ is generated from a perturbed tensor grid on the square $[-2v_{\mathrm{eq}},2v_{\mathrm{eq}}]^2$, where $v_{\mathrm{eq}}=\sqrt{\alpha/\beta}$ is the equilibrium speed of
the self-propulsion model. This construction provides broad coverage of the velocity domain used for learning the environment force $f$.

\subsection{Estimates from semi-parametric approach}

\begin{table}[H]
\centering
\small
\resizebox{\linewidth}{!}{
\begin{tabular}{c c c c c}
\hline
\textbf{System}
&
\textbf{Component}
&
\textbf{Term}
&
$coeff_{\mathrm{true}}$
&
$\mathbb{E}(\hat{coeff}) \pm \sigma(\hat{coeff})$
\\
\hline

\multirow{1}{*}{Kuramoto}
& \multirow{1}{*}{$1$}
& $\theta$
& $0.01$
& $\num{0.0099999999999999} \pm \num{2.0877228669312424e-16}$
\\
\hline

\multirow{6}{*}{Phototaxis}
& \multirow{3}{*}{$1$}
& $1$
& $I_0U_\infty e_\ell^{(1)}=-\frac{0.1}{\sqrt{2}}$
& $\num{-0.07071068149235388} \pm \num{1.4237309422862539e-07}$
\\
&
& $v_1$
& $I_0=1.0$
& $\num{1.0000003194539608} \pm \num{3.538386027474474e-06}$
\\
&
& $v_2$
& $0$
& $\num{-2.967810805171508e-07} \pm \num{2.676347740402615e-06}$
\\
\cline{2-5}

&
\multirow{3}{*}{$2$}
& $1$
& $I_0U_\infty e_\ell^{(2)}=\frac{0.1}{\sqrt{2}}$
& $\num{0.07071066365646213} \pm \num{5.863941042682641e-08}$
\\
&
& $v_1$
& $0$
& $\num{-4.931084525950616e-07} \pm \num{2.709497747258749e-06}$
\\
&
& $v_2$
& $I_0=1.0$
& $\num{0.999999968739424} \pm \num{4.0266806548277446e-06}$
\\
\hline

\multirow{8}{*}{SPP}
& \multirow{4}{*}{$1$}
& $v_1$
& $\alpha=1$
& $\num{1.0002327393850916} \pm \num{0.0007946947927821555}$
\\
&
& $v_2$
& $0$
& $\num{0.000568144895427883} \pm \num{0.001043046722979779}$
\\
&
& $\|v\|^2 v_1$
& $\beta=0.5$
& $\num{0.49999639428741754} \pm \num{0.0001135084458270978}$
\\
&
& $\|v\|^2 v_2$
& $0$
& $\num{-8.231290447634012e-05} \pm \num{0.0001774521751358293}$
\\
\cline{2-5}

&
\multirow{4}{*}{$2$}
& $v_1$
& $0$
& $\num{6.237543358736741e-05} \pm \num{0.0016195088504306514}$
\\
&
& $v_2$
& $\alpha=1$
& $\num{0.9997459229867008} \pm \num{0.001191278938192407}$
\\
&
& $\|v\|^2 v_1$
& $0$
& $\num{4.2274534338444424e-05} \pm \num{0.00030534338761672563}$
\\
&
& $\|v\|^2 v_2$
& $\beta=0.5$
& $\num{0.49993885800632165} \pm \num{0.00019757261343423876}$
\\
\hline
\end{tabular}
}
\caption{
Recovery of the parametric coefficients for the candidate basis functions.
For each coefficient, we report the mean and standard deviation over
$T_r=10$ independent trials.
Coefficients corresponding to non-present mechanisms are expected
to be close to zero.
}
\label{tab:app:param-recovery}
\end{table}

\newpage

\subsection{Index of notation}

\subsection{Index of notation}
\renewcommand{\arraystretch}{1.5}
\begin{longtable}{>{\centering\arraybackslash}m{4.2cm} m{9cm}}
\caption{Parameters and their definitions.} \label{tab:notation} \\
\toprule
\textbf{Parameter} & \textbf{Definition} \\
\midrule
\endfirsthead

\multicolumn{2}{l}{\small\itshape continued from previous page} \\
\toprule
\textbf{Parameter} & \textbf{Definition} \\
\midrule
\endhead

\midrule
\multicolumn{2}{r}{\small\itshape continued on next page} \\
\endfoot

\bottomrule
\endlastfoot

\multicolumn{2}{l}{\textit{Dynamical system}} \\
\midrule
$N$ & Number of agents in the system \\
$d$ & Dimension of the state space of each agent \\
$x_i(t) \in \R^d$ & State of the $i$-th agent at time $t$ \\
$x(t) \in \RdN$ & Full stacked state of the system at time $t$ \\
$f : \R^d \to \R^d$ & Environmental force function (assumed identical for all agents) \\
$\Omega \subseteq \R^d$ & Proper subset of $\R^d$ serving as the domain of $f$ in applications where $f$ is not defined on all of $\R^d$ \\
$\phi : \R_{\geq 0} \to \R$ & Interaction kernel encoding pairwise interaction strength as a function of distance \\
$F_f(x) \in \RdN$ & Stacked environmental force: $(f(x_1),\ldots,f(x_N))$ \\
$F_\phi(x) \in \RdN$ & Stacked interaction force: $\frac{1}{N}\sum_{j\neq i}\phi(|x_j-x_i|)(x_j-x_i)$ for each agent $i$ \\
$v_i(t) \in \R^d$ & Velocity of the $i$-th agent at time $t$ (second-order systems) \\
$v(t) \in \RdN$ & Full stacked velocity of the system at time $t$: $(v_1(t),\ldots,v_N(t))$ \\
$F_f(x,v) \in \RdN$ & Stacked environmental force for second-order systems: $(f(x_1,v_1),\ldots,f(x_N,v_N))$, reducing to $F_f(v)$ when $f$ has no spatial dependence \\
$\mathcal{F}(X,Y)$ & Set of all functions from set $X$ to set $Y$ (used to specify that $\Hf,\Hphi$ are finite-dimensional subspaces of function spaces) \\

\midrule
\multicolumn{2}{l}{\textit{Trajectory data}} \\
\midrule
$\mu_0$ & Distribution on $\RdN$ from which i.i.d.\ initial conditions are sampled \\
$M$ & Number of trajectory replicates (initial condition samples) \\
$L$ & Number of discrete observation time points \\
$\{t_\ell\}_{\ell=1}^{L}$, $0=t_1<\cdots<t_L=T$ & Discrete observation times over the training interval $[0,T]$ \\
$\Delta t_\ell := t_{\ell+1}-t_\ell$ & Time step between consecutive observations \\
$X^{(m)}_0 \in \RdN$ & Initial condition of the $m$-th replicate \\
$V^{(m)}_0 \in \RdN$ & Initial velocity condition of the $m$-th replicate (second-order systems) \\
$\Xml \in \RdN$ & Observed state of the $m$-th replicate at time $t_\ell$ \\
$\Vml \in \RdN$ & Observed (or approximated) velocity of the $m$-th replicate at time $t_\ell$ \\
$\Aml \in \RdN$ & Observed (or approximated) acceleration of the $m$-th replicate at time $t_\ell$ (second-order systems only) \\
$(\Rml)_{ij} \in \R$ & Pairwise distance $|(\Xml)_i - (\Xml)_j|$ between agents $i$ and $j$ at observation $(m,\ell)$; $\Rml \in \R^{N \times N}$ \\
$\mathcal{H}$ & Low-pass filter applied to trajectory data for velocity (or acceleration) estimation \\

\midrule
\multicolumn{2}{l}{\textit{Inner product and norm on $\RdN$}} \\
\midrule
$\langle X, Y \rangle_{\RdN}$ & $\frac{1}{N}\sum_{i=1}^N \langle X_i, Y_i \rangle$, the $N$-agent-normalized inner product on $\RdN$ \\
$\|\cdot\|_{\RdN}$ & Norm on $\RdN$ induced by $\langle\cdot,\cdot\rangle_{\RdN}$ \\

\midrule
\multicolumn{2}{l}{\textit{Hypothesis spaces and bases}} \\
\midrule
$\Hf \subseteq \mathcal{F}(\R^d, \R^d)$ & Finite-dimensional hypothesis space for the environmental force $f$ \\
$\Hphi \subseteq \mathcal{F}(\R_{\geq 0}, \R)$ & Finite-dimensional hypothesis space for the interaction kernel $\phi$ \\
$\nf$ & Dimension of $\Hf$ \\
$\nphi$ & Dimension of $\Hphi$ \\
$(h_k)_{k=1}^{\nf}$ & Ordered basis for $\Hf$; each $h_k=(h_k^1,\ldots,h_k^d)$ is $\R^d$-valued \\
$(\psi_k)_{k=1}^{\nphi}$ & Ordered basis for $\Hphi$ \\
$\alpha \in \R^{\nf}$ & Coefficient vector for $\hat{f}$ in the basis $(h_k)$ \\
$\beta \in \R^{\nphi}$ & Coefficient vector for $\hat{\phi}$ in the basis $(\psi_k)$ \\
$\theta = (\alpha,\beta) \in \R^{\nf+\nphi}$ & Combined coefficient vector for both $\hat{f}$ and $\hat{\phi}$ \\
$\tilde f \in \Hf,\ \tilde\phi \in \Hphi$ & An arbitrary (not necessarily optimal) candidate pair of environmental force and interaction kernel used to define the error functional $\mathcal{E}_{\Hf,\Hphi}$ and the evaluation metrics of Section~\ref{sec:evaluation_methodology}; contrasted with the fitted estimators $\hat f,\hat\phi$ \\
$\hat{f} \in \Hf$ & Estimator of the environmental force $f$, defined as $\sum_k \hat\alpha_k h_k$ \\
$\hat{\phi} \in \Hphi$ & Estimator of the interaction kernel $\phi$, defined as $\sum_k \hat\beta_k \psi_k$ \\

\midrule
\multicolumn{2}{l}{\textit{Matrices and normal equations}} \\
\midrule
$\Hml \in \R^{dN \times \nf}$ & Matrix $\bigl(h_1(\Xml)\;\cdots\;h_{\nf}(\Xml)\bigr)$ of stacked $f$-basis terms \\
$\Hml$ (second-order case) $\in \R^{dN \times \nf}$ & Matrix $\bigl(h_1(\Xml,\Vml)\;\cdots\;h_{\nf}(\Xml,\Vml)\bigr)$ of stacked $f$-basis terms evaluated on both state and velocity \\
$\Psiml \in \R^{dN \times \nphi}$ & Matrix $\bigl(F_{\psi_1}(\Xml)\;\cdots\;F_{\psi_{\nphi}}(\Xml)\bigr)$ of stacked interaction-basis terms \\
$\Cml \in \R^{dN \times (\nf+\nphi)}$ & Concatenated matrix $(\Hml \;\; \Psiml)$ \\
$\mathcal{E}(\theta)$ & Least-squares error functional $\frac{1}{ML}\sum_{m,\ell}\|\Vml - \Cml\theta\|_{\RdN}^2$ \\
$\mathcal{E}_{\Hf,\Hphi}(\hat f,\hat \phi)$ & Second-order error functional over hypothesis spaces, equal to $\mathcal{E}(\alpha,\beta)$ once $\hat f,\hat\phi$ are expanded in their bases (same object as $\mathcal{E}(\theta)$, with $\theta=(\alpha,\beta)$) \\
$\hat{\theta} = (\hat{\alpha},\hat{\beta})$ & Minimizer of $\mathcal{E}(\theta)$; estimated coefficient vector \\
$A \in \R^{(\nf+\nphi)\times(\nf+\nphi)}$ & Normal equation matrix $\displaystyle\sum_{m,\ell}(\Cml)^T\Cml$ \\
$b \in \R^{\nf+\nphi}$ & Normal equation right-hand side $\displaystyle\sum_{m,\ell}(\Cml)^T\Vml$ \\

\midrule
\multicolumn{2}{l}{\textit{Hypothesis space construction}} \\
\midrule
$R_{\min},\,R_{\max}$ & Min/max observed pairwise distances; define the effective domain $[R_{\min},R_{\max}]$ of $\Hphi$ \\
$P$ & Number of sub-intervals in the data-dependent partition of $[R_{\min},R_{\max}]$ used to build $\Hphi$ \\
$I_p := [r_p,r_{p+1})$ & $p$-th partition element (sub-interval) of $[R_{\min},R_{\max}]$, $p=1,\ldots,P$ \\
$n_P$ & Number of localized basis functions (e.g.\ degree-$\leq 2$ polynomials or B-splines) supported on each partition element; $\nphi = P \times n_P$ \\
$Z_{\min},\,Z_{\max} \in \R^d$ & Component-wise min/max observed state values; define the hyper-rectangular partition domain of $\Hf$ \\
$X_{\min},\,X_{\max} \in \R^d$ & Component-wise min/max observed state values (second-order notation for $Z_{\min},Z_{\max}$ restricted to the state variable) \\
$V_{\min},\,V_{\max} \in \R^d$ & Component-wise min/max observed velocity values; together with $X_{\min},X_{\max}$ define the domain of $\Hf$ for second-order systems \\
$X_{k,\min},\,X_{k,\max} \in \R$ & Min/max of the $k$-th component of the state variable, $k=1,\ldots,d$ \\
$V_{k,\min},\,V_{k,\max} \in \R$ & Min/max of the $k$-th component of the velocity variable, $k=1,\ldots,d$ \\
$h_k^j : \R^d \to \R$ & Scalar-valued localized component function ($j=1,\ldots,d$) used to build the $\R^d$-valued basis function $h_k=(h_k^1,\ldots,h_k^d)$ of $\Hf$, via tensor product with the standard basis of $\R^d$ \\
$\psi_k^E,\ \psi_k^A$ & Localized basis functions for the energy-based kernel $\phi^E$ and alignment-based kernel $\phi^A$, respectively, used in the model-selection algorithm (playing the role of the generic $\psi_k$ for each kernel type) \\

\midrule
\multicolumn{2}{l}{\textit{Computational complexity}} \\
\midrule
$D$ & Dimension of the stacked state space: $D=dN$ (first-order systems), $D=2dN$ (second-order systems) \\
$n := \nf + \nphi$ & Total number of basis functions across $\Hf$ and $\Hphi$ \\

\midrule
\multicolumn{2}{l}{\textit{Semi-parametric approach}} \\
\midrule
$p \in \R^{n_p}$ & Parameter vector fully specifying the prescribed functional form of $f(x;p)$ \\
$n_p$ & Number of unknown parameters in the parametric environmental force \\
$\eta_k:\R^d\to\R^d$, $k=1,\dots,n_p$ & Known feature functions when $f$ is linear in $p$, i.e.\ $f(x;p)=\sum_k p_k\,\eta_k(x)$ \\
$\Hml$ (semi-parametric case) $\in \R^{dN \times n_p}$ & Matrix of stacked parametric feature terms $\eta_k(\Xml)$, constructed analogously to the non-parametric $\Hml$ with basis $(\eta_k)_{k=1}^{n_p}$ in place of $(h_k)_{k=1}^{\nf}$ \\

\midrule
\multicolumn{2}{l}{\textit{Induced measures on data}} \\
\midrule
$Z$ & Input variable of $f$: $Z=X$ (first-order systems), $Z=(X,V)$ (second-order systems) \\
$\omega$ & Sample outcome indexing a random draw $x_0(\omega)\sim\mu_0$ of the initial condition \\
$r_{ij}(t,\omega)$ & Pairwise distance trajectory $|x_j(t,x_0(\omega))-x_i(t,x_0(\omega))|$ between agents $i,j$ \\
$A \subseteq \R_{\geq 0}$ & Borel set used to define the pairwise-distance measure $\rho_R(A)$ \\
$B \subseteq \R^d$ & Borel set used to define the state measure $\rho_X(B)$ \\
$\rho_R$ & Continuous-time expected measure of observed pairwise distances over $[0,T]$ and $\mu_0$ \\
$\hat{\rho}_R$ & Empirical approximation of $\rho_R$ from the observed distances $(\Rml)_{m,\ell}$ \\
$\rho_Z$ & Continuous-time expected measure of the state variable $Z$ over $[0,T]$ and $\mu_0$ \\
$\hat{\rho}_Z$ & Empirical approximation of $\rho_Z$ from the observed data $(\Xml)_{m,\ell}$ \\
$\Delta t := T/L$ & Uniform discretization step of the training interval $[0,T]$ used when forming the empirical measures $\hat\rho_R,\hat\rho_X$ (distinct from the per-step $\Delta t_\ell$ used elsewhere) \\
$M_\rho$ & Number of replicates used to construct accurate empirical approximations of $\rho_R$ and $\rho_Z$ \\
$\|\psi\|_{L^2(\rho_R)}$, $\|h\|_{L^2(\rho_X)}$ & Weighted $L^2$ norms (with weight $r^2$ for $\rho_R$) used to measure feature-recovery error \\

\midrule
\multicolumn{2}{l}{\textit{Feature recovery metrics}} \\
\midrule
$E_\phi(\tilde\phi)$ & Interaction kernel error $\|\tilde\phi - \phi\|_{L^2(\rho_R)}$ (weighted by $r^2$) \\
$E_f(\tilde f)$ & Environmental force error $\|\tilde f - f\|_{L^2(\rho_Z)}$ \\
$E_\phi^{\mathrm{rel}},\,E_f^{\mathrm{rel}}$ & Relative versions of $E_\phi$ and $E_f$, normalized by $\|\phi\|_{L^2(\rho_R)}$ and $\|f\|_{L^2(\rho_Z)}$ respectively \\

\midrule
\multicolumn{2}{l}{\textit{Residual error}} \\
\midrule
$S_{\mathrm{res}}$ & RMS of the observed velocity (first-order) or acceleration (second-order) data; normalization scale \\
$E_{\mathrm{res}}(\tilde f,\tilde\phi)$ & Residual error: RMS discrepancy between observed and predicted velocities (or accelerations) \\
$E_{\mathrm{res}}^{\mathrm{rel}}$ & Relative residual error $E_{\mathrm{res}}/S_{\mathrm{res}}$ \\
$E_{\mathrm{res},\epsilon}^{\mathrm{rel}}$ & Regularized relative residual error $E_{\mathrm{res}}/(S_{\mathrm{res}}+\epsilon)$, used in the model-selection compatibility gate \\

\midrule
\multicolumn{2}{l}{\textit{Trajectory metrics}} \\
\midrule
$[0,T]$ & Training (fitting) time horizon \\
$(T,T_f]$ & Prediction (extrapolation) time horizon beyond the training window \\
$\tilde x(t) \in \RdN$ & Continuous-time solution of the IVP driven by estimators $\tilde f,\tilde\phi$ (i.e.\ $\dot{\tilde x}=F_{\tilde f}(\tilde x)+F_{\tilde\phi}(\tilde x)$, $\tilde x(0)=X_0^{(m)}$), of which $\tXml$ is the value at time $t_\ell$ \\
$\tXml \in \RdN$ & State of replicate $m$ at time $t_\ell$ simulated from the learned model $(\hat f,\hat\phi)$ \\
$e^{(m)}_\ell(\tilde f,\tilde\phi)$ & Pointwise trajectory error $\|\Xml - \tXml\|_{\RdN}$ of replicate $m$ at time $t_\ell$ \\
$E_{\mathrm{recon}}^{(m)}$ & Reconstruction error: $\max_{t_\ell\in[0,T]}e^{(m)}_\ell$ \\
$E_{\mathrm{pred}}^{(m)}$ & Prediction error: $\max_{t_\ell\in(T,T_f]}e^{(m)}_\ell$ \\
$E_{\mathrm{traj}}^{(m)}$ & Full trajectory error: $\max_{t_\ell\in[0,T_f]}e^{(m)}_\ell$ \\
$\bar{E}_{\mathrm{recon}},\bar{E}_{\mathrm{pred}},\bar{E}_{\mathrm{traj}}$ & Sample-mean reconstruction, prediction, and full trajectory errors over $M$ replicates \\
$\sigma_{E_{\mathrm{recon}}},\sigma_{E_{\mathrm{pred}}},\sigma_{E_{\mathrm{traj}}}$ & Sample standard deviations of the corresponding trajectory-error metrics \\
$E_{*}^{\mathrm{RMS}}$ & RMS trajectory error over the time interval indexed by ${*}\in\{\mathrm{recon},\mathrm{pred},\mathrm{traj}\}$ \\
$S_X$ & RMS of observed trajectory data; normalization scale for relative trajectory errors \\
$E_{*}^{\mathrm{rel}}$ & Relative RMS trajectory error $E_{*}^{\mathrm{RMS}}/(S_X+\epsilon)$ \\

\midrule
\multicolumn{2}{l}{\textit{Consistency and robustness assessment}} \\
\midrule
$T_r$ & Number of independent repeated trials used to assess reliability \\
$\eta_{i,\ell}^{(m)},\bar\eta_{i,\ell}^{(m)}\overset{\mathrm{i.i.d.}}{\sim}\mathrm{Unif}[-\zeta,\zeta]$ & Multiplicative noise applied independently to observed positions and velocities \\
$x_i^{(m),\mathrm{noisy}},\ v_i^{(m),\mathrm{noisy}} \in \R^d$ & Multiplicatively perturbed state and velocity of agent $i$, replicate $m$, at time $t_\ell$, used to assess robustness to observational noise \\
$\zeta$ & Noise level parameter controlling the magnitude of multiplicative perturbation \\
$\epsilon$ & Small regularization constant ($10^{-12}$) preventing division by zero in normalized error ratios \\

\midrule
\multicolumn{2}{l}{\textit{Model selection}} \\
\midrule
$\phi^E$ & Energy (position-based) interaction kernel in the generalized framework \\
$\phi^A$ & Alignment (velocity-based) interaction kernel (second-order systems) \\
$c$ & A candidate dynamical framework (combination of dynamical order and active force terms) \\
$F_1,F_2$ & First-order candidate frameworks (interaction-only, and interaction+environmental) in Table~\ref{tab:model_selection_candidates} \\
$S_1,\ldots,S_6$ & Second-order candidate frameworks, distinguished by which of $f,\phi^E,\phi^A$ are active, in Table~\ref{tab:model_selection_candidates} \\
$A_c,\ \hat\theta_c,\ b_c$ & Normal-equation matrix, coefficient vector, and right-hand side ($A,\hat\theta,b$ as in~\eqref{eq:normal_eq_mat_vec}) constructed for a specific candidate model $c$ in the model-selection algorithm \\
$\mathcal{C}$ & Set of admissible candidate frameworks passing the compatibility gate \\
$\mathcal{C}_{\mathrm{val}}$ & Subset of admissible candidates whose mean validation trajectory error is below $\tau_{\mathrm{traj}}$ \\
$M_{\mathrm{train}},M_{\mathrm{val}},M_{\mathrm{test}}$ & Training, validation, and test trajectory splits; $M=M_{\mathrm{train}}+M_{\mathrm{val}}+M_{\mathrm{test}}$ \\
$\tau_{\mathrm{recon}}$ & Admissibility threshold for the relative reconstruction error (set to $0.25$) \\
$\tau_{\mathrm{res}}$ & Admissibility threshold for the relative residual error (set to $0.25$) \\
$E_{\min}$ & Minimum mean validation trajectory error among admissible candidates: $\min_{c\in\mathcal{C}}\bar{E}_{\mathrm{traj}}(c)$ \\
$\tau_{\mathrm{traj}}$ & Validation trajectory threshold: $\max\!\bigl\{E_{\min}+\varepsilon_{\mathrm{abs}},\,E_{\min}(1+\varepsilon_{\mathrm{rel}})\bigr\}$ \\
$\varepsilon_{\mathrm{abs}},\varepsilon_{\mathrm{rel}}$ & Absolute and relative tolerances defining $\mathcal{C}_{\mathrm{val}}$ (set to $10^{-3}$ and $0.05$) \\
$c^*$ & Selected (highest-ranked) candidate framework \\

\end{longtable}
\renewcommand{\arraystretch}{1}

\end{document}